%% file: main.tex
\documentclass[10pt,twocolumn,letterpaper]{article}

\usepackage[pagenumbers]{cvpr}
\input{preamble}
\definecolor{cvprblue}{rgb}{0.21,0.49,0.74}
\usepackage[pagebackref,breaklinks,colorlinks,allcolors=cvprblue]{hyperref}

\def\confName{CVPR}
\def\confYear{2027}

\hypersetup{
    pdftitle={When 3DGS Meets Narrow Baselines},
    pdfauthor={Xiao Zhang, Yulun Guo, Junze Zhu, Yanlin Qian, Hua Wu, Yue Ma, Chen Guoyin}
}

\title{When 3DGS Meets Narrow Baselines}
\author{%
    Xiao Zhang\textsuperscript{1},\enspace Yulun Guo\textsuperscript{2},\enspace Junze Zhu\textsuperscript{3,4},\enspace Yanlin Qian\textsuperscript{5},\enspace Hua Wu\textsuperscript{1},\enspace Yue Ma\textsuperscript{1},\enspace Chen Guoyin\textsuperscript{1}\\[3pt]
    \normalsize\textsuperscript{1}\href{https://cixtech.com}{\textcolor{black}{CIX Technology Co.,Ltd.}}\qquad
    \textsuperscript{2}\href{https://bfjg.norincogroup.com.cn}{\textcolor{black}{Southwest Institute of Technical Physics}}\\
    \normalsize\textsuperscript{3}\href{https://nju.edu.cn}{\textcolor{black}{National Key Laboratory for Novel Software Technology, Nanjing University}}\\
    \normalsize\textsuperscript{4}\href{https://nju.edu.cn}{\textcolor{black}{School of Artificial Intelligence, Nanjing University}}\qquad
    \textsuperscript{5}\href{https://dji.com}{\textcolor{black}{DJI Technology Co.,Ltd.}}\\
    \small\ttfamily\{\href{mailto:Xiao.Zhang@cixtech.com}{\textcolor{black}{Xiao.Zhang}}, \href{mailto:David.Wu@cixtech.com}{\textcolor{black}{David.Wu}}, \href{mailto:Aaron.Ma@cixtech.com}{\textcolor{black}{Aaron.Ma}}, \href{mailto:guoyin.chen@cixtech.com}{\textcolor{black}{guoyin.chen}}\}@cixtech.com%
}

\begin{document}
\begingroup
\setcounter{dbltopnumber}{2}
\setlength{\textfloatsep}{9.5pt plus 2pt minus 2pt}
\setlength{\dbltextfloatsep}{9.5pt plus 2pt minus 2pt}
\setlength{\floatsep}{6pt plus 1pt minus 1pt}
\setlength{\dblfloatsep}{6pt plus 1pt minus 1pt}
\setlength{\intextsep}{8pt plus 2pt minus 2pt}
\captionsetup{skip=3pt}
\makeatletter
\setlength{\@fptop}{0pt}
\setlength{\@fpsep}{8pt}
\setlength{\@fpbot}{0pt plus 1fil}
\setlength{\@dblfptop}{0pt}
\setlength{\@dblfpsep}{8pt}
\setlength{\@dblfpbot}{0pt plus 1fil}
\makeatother
\twocolumn[{%
\renewcommand\twocolumn[1][]{#1}%
\vspace*{-49pt}
\maketitle
\vspace{-3em}
\begin{center}
    \includegraphics[width=\textwidth]{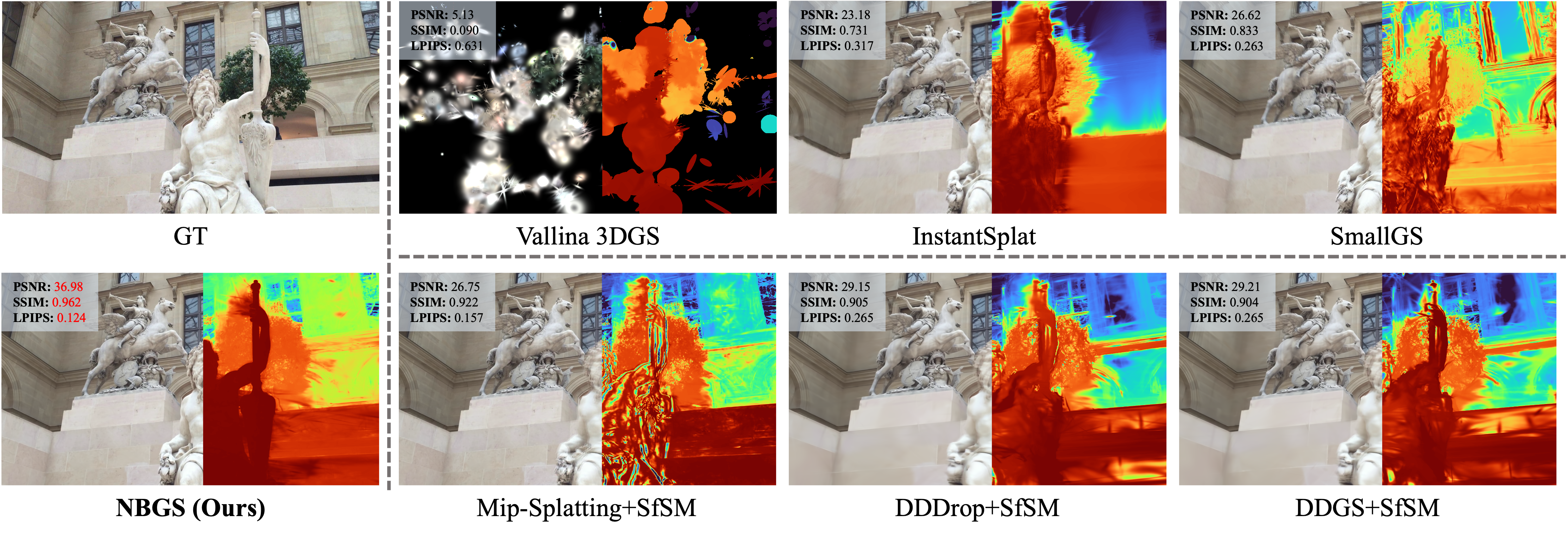}
    \captionof{figure}{\textbf{3DGS under a narrow baseline.} Rendered RGB and depth on a narrow-baseline clip. Default Vallina 3DGS removes almost all Gaussians and collapses. NBGS recovers both the rendering appearance and the scene depth.}
    \label{fig:teaser}
\end{center}
\vspace{0.5em}
}]

\noindent\begin{minipage}[t]{\columnwidth}
\input{sections/00_abstract}
\end{minipage}\par

\input{sections/compact/01_introduction}

\input{sections/compact/02_related_work}

\input{sections/compact/05_results_evidence_figures}

\input{sections/compact/03_method}
\input{sections/compact/04_experiments}

\raggedbottom
\input{sections/compact/05_results}
\input{sections/06_conclusion}
\input{sections/llm_usage_statement}

{\small
\bibliographystyle{ieeenat_fullname_unsorted}
\bibliography{references}
}

\clearpage
\endgroup
\input{sections/appendix}

\end{document}

%% file: preamble.tex
\usepackage{tikz}
\usepackage{microtype}
\usepackage{stfloats}
\usepackage{float}

\input{sections/review_macros}

\makeatletter
\@ifpackageloaded{lineno}{%
  \IfFormatAtLeastTF{2025-06-01}{%
    \IfPackageAtLeastTF{lineno}{2026/02/16}{%
      \RemoveFromHook{build/column/before}[lineno]
      \AddToHook{build/column/after}{%
        \setbox\@outputbox\vbox{%
          \boxmaxdepth\@maxdepth
          \protected@write\@auxout{}{%
            \string\def\string\@LN@column{\if@firstcolumn1\else2\fi}}%
          \box\@outputbox}}%
    }{}%
  }{}%
}{}
\makeatother

%% file: sections/review_macros.tex
\newif\ifreviewchanges
\ifdefined\showreviewchanges\reviewchangestrue\fi

%% file: sections/00_abstract.tex
\begin{abstract}
3D Gaussian Splatting (3DGS) achieves high-quality, real-time novel-view synthesis, but narrow baselines cannot satisfy the conditions of standard 3DGS. Even when poses and sparse depth can be recovered from such captures, weak parallax leaves the geometry highly uncertain. Previous depth-regularized methods may treat uncertain estimates as exact targets and guide Gaussians to erroneous geometry. More critically, the camera extents, which can be 1\%-0.1\% of the scene scale, are used to prune the scenes and collapse reconstructions. Beyond this scale mismatch,  absolute near cutoffs ignoring reconstruction scale will blur or alias rendering results. To address these challenges, we propose the Narrow-Baseline 3DGS (NBGS). We first represent the scene scale by the depths instead of camera extents to stop the collapse. Then we model each sparse inverse-depth estimate as a normal distribution, and these distributions guide ray supervision. Finally, the near-side bounds of these distributions determine each camera's near plane to remove blurring and aliasing. Across small motion clips, our framework reaches 32.01 dB PSNR, exceeding the SOTA Mip-Splating by 1.38 dB. On narrow-baseline iPhone clips with LiDAR references, it also attains the highest $\delta_{1.25}$ accuracy ($72.5\%$), exceeding the SOTA Mip-Splating by $ 1.1\%$.
\end{abstract}

%% file: sections/compact/01_introduction.tex
\section{Introduction}
\label{sec:introduction}

\looseness=-1
3D Gaussian Splatting (3DGS) has emerged as a powerful representation for photorealistic 3D reconstruction and novel-view synthesis~\cite{kerbl2023gaussians}. Despite its success, 3DGS remains vulnerable to a fundamental geometric limit: narrow-baseline reconstruction. When camera translations become small relative to scene depth, multi-view observations provide increasingly weak parallax, making depth estimation severely ill-conditioned. This difficulty is not simply a consequence of having too few images. Even densely sampled views can provide little additional geometric information when their camera centers are tightly clustered. As the baseline-to-depth ratio approaches zero, small correspondence or camera-pose errors can induce large depth errors, challenging the geometric foundations on which 3D reconstruction depends.

More critically, narrow-baseline geometry exposes fundamental limitations in the standard 3DGS backend, potentially leading to catastrophic reconstruction failure. As camera baselines shrink, camera-extent-based density control becomes unreliable, sparse depth estimates grow uncertain, and fixed near-plane thresholds become incompatible with the scene scale. These issues can cause excessive Gaussian pruning, erroneous geometry, and unstable rendering.

Recent approaches have extended Gaussian reconstruction to uncalibrated cameras and limited-view settings~\cite{fu2024cf3dgs,fan2024instantsplat,ni2026g4splat}. However, reducing the number of required views or recovering unknown cameras does not, by itself, resolve the degeneracy caused by vanishing parallax. Structure-from-small-motion (SfSM)~\cite{ha2016smallmotion} offers a means of recovering camera poses and sparse geometry from small camera motions, but these estimates remain uncertain, and the downstream 3DGS backend still inherits conditions that are unsuitable for narrow-baseline observations. This motivates a complementary question: how can Gaussian reconstruction remain stable and geometrically meaningful when the available camera motion provides only weak depth constraints?

To address these challenges, we propose the NBGS reconstruction framework that explicitly accounts for scene depth and geometric uncertainty under narrow baselines. Our contributions are threefold:

\begin{itemize}
    \item Depth-based Scene Rescaling. The scene scale determined by camera extent becomes undersized and leads 3DGS to collapse for narrow-baseline conditions. We use the median depth, instead of camera extent, to stabilize 3DGS by avoiding pruning too many Gaussians.
    \item Depth-guided Adaptive Near Plane Filtering. Filtering and rasterization by fixed near planes can cause blurring and aliasing. We set the filter bounds adaptively by depth, preventing unstable Gaussian projections and unintended removal of valid geometry.
    \item Probabilistic Depth Modeling and Supervision. Due to the uncertain geometry caused by weak parallax, we remodel sparse inverse-depth estimates as reliability-conditioned distributions and introduce probabilistic ray supervision to refine uncertain geometry during Gaussian optimization.
\end{itemize}

%% file: sections/compact/02_related_work.tex
\section{Related Work}
\label{sec:related}

\subsection{Narrow-Baseline Geometry Estimation}
Incremental and global structure-from-motion (SfM) recover cameras and geometry from multi-view correspondences, relying on informative parallax between views~\cite{schonberger2016sfm,pan2024glomap}. Under narrow baselines, however, translation, focal length, and depth become poorly constrained, so small correspondence errors amplify, making structure recovery considerably harder~\cite{tao2024globaltranslation}. Ha et al. proposed structure-from-small-motion(SfSM) to address this problem~\cite{ha2016smallmotion} by jointly estimating shared intrinsics, poses, and sparse depths from narrow-baseline scenes. Though this method improves conditioning, it cannot recover parallax, and the resulting depths remain sensitive to correspondence, pose, and calibration errors.

Learning-based geometry offers a complementary way by injecting priors rather than adding measured parallax. VGGSfM learns tracking and differentiable bundle adjustment~\cite{wang2024vggsfm}, and MP-SfM propagates uncertainty from monocular surface priors into SfM~\cite{pataki2025mpsfm}. DUSt3R and MASt3R estimate point maps and geometry-aware matches~\cite{wang2024dust3r,leroy2024mast3r}, while VGGT jointly predicts cameras and dense geometry~\cite{wang2025vggt}. For video sequences, MegaSaM combines learned priors with global optimization~\cite{li2025megasam}, MonST3R handles scene motion~\cite{zhang2025monst3r}, CUT3R accumulates geometry through a persistent state~\cite{wang2025cut3r}, and MASt3R-SLAM integrates reconstruction priors into tracking and fusion~\cite{murai2025mast3rslam}. Closest to narrow-baseline scenes, Small GS refines narrow-baseline camera poses through Gaussian rendering~\cite{yao2025smallgs}. These approaches substantially improve robustness, yet the depth remains ambiguous under weak parallax and should be treated as uncertain. 

\subsection{Depth Guidance for Neural Rendering}
A number of works regularize radiance fields and Gaussians by depth when geometry is underconstrained. RegNeRF regularizes patches rendered from unseen views, FreeNeRF constrains frequencies and near-camera density, and CoR-GS exploits disagreement between reconstructions~\cite{niemeyer2022regnerf,yang2023freenerf,zhang2024corgs}. SPARF jointly refines radiance fields and noisy camera poses~\cite{truong2023sparf}. Depth supervision provides a more direct geometric signal. DS-NeRF supervises with sparse SfM depths, and SparseNeRF distills depth rankings~\cite{deng2022dsnerf,wang2023sparsenerf}. In the 3D Gaussian, FSGS and DNGaussian combine sparse-view optimization with depth regularization~\cite{zhu2024fsgs,li2024dngaussian}, DepthSplat couples depth prediction with splatting~\cite{xu2025depthsplat}, and D$^2$GS uses depth-aware dropout and far-region enhancement~\cite{song2026d2gs}. NexusGS densifies initialization with epipolar depth priors~\cite{zheng2025nexusgs}, and FrugalNeRF selects depth targets by reprojection error~\cite{lin2025frugalnerf}.

When depth itself is unreliable, several methods have explored how to model its reliability. Roessle et al. complete sparse SfM depths into dense priors with uncertainty~\cite{roessle2022densepriors}, CDGS confidence-weights depth regularization~\cite{zhang2024cdgs}, CoMapGS weights supervision by covisibility~\cite{jang2025comapgs}, and UGOT combines uncertainty with distributional optimal transport~\cite{sun2024ugot}. These methods mostly represent uncertainty as weighting, ignoring the fact that depth uncertainty can affect the 3DGS backend.

\begin{figure*}[!t]
    \captionsetup{skip=3pt}
    \centering
    \includegraphics[width=\textwidth,trim=0 4.36bp 0 6.98bp,clip]{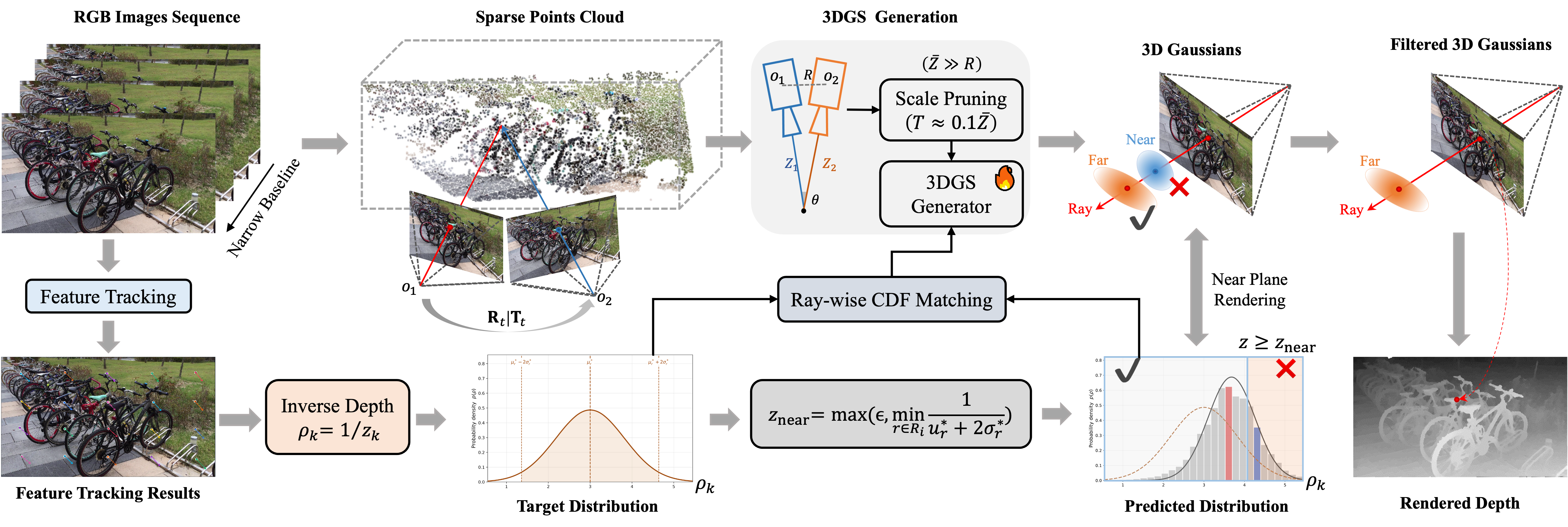}
    \caption{\textbf{Overall framework.} SfSM initialize 3DGS with the median depth $\overline{z}$ as scene scale. Sparse-depth distributions replace exact depth targets for supervision, and the near plane is set by the depth distribution for filtering and rendering.}
    \label{fig:method_overview}
\end{figure*}

\subsection{Scale-Dependent Components in Gaussian Splatting}
3DGS scales its size-pruning and clone/split thresholds, as well as position learning rates, by the extent of camera centers~\cite{kerbl2023gaussians}. Grubert et al. note that this extent reflects the capture trajectory rather than scene size~\cite{grubert2025improving}. They instead use the mean distance from the camera centroid to SfM points and remove size pruning. Under narrow baselines, the camera extent can be much smaller than scene depth, making this distinction especially important for density control.

Rendering also depends on depth and sampling scale. Mip-NeRF and Zip-NeRF integrate spatial footprints to reduce aliasing~\cite{barron2021mipnerf,barron2023zipnerf}, while Mip-Splatting combines 3D covariance smoothing with 2D anti-aliasing~\cite{yu2024mipsplatting}. Extensions improve pixel integration and multi-scale rendering~\cite{liang2024analyticsplatting,yan2024multiscalegs,franke2024trips}, high-frequency detail and view-space bounds~\cite{huang2025spectralgs,steiner2025aaags}, and depth ordering~\cite{radl2024stopthepop}. These advances primarily address sampling and rendering consistency. With uncertain SfSM geometry at an arbitrary scale, however, fixed near cutoffs can reject valid foreground or observations needed for filter estimation. Our method derives a shared boundary for covariance filtering and rasterization from sparse-depth uncertainty, while retaining 2D anti-aliasing.

%% file: sections/compact/05_results_evidence_figures.tex
\input{sections/compact/system_quant_table}

\newcommand{\RGBComparisonFigure}{%
\begin{figure*}[t]
    \centering
    \includegraphics[width=\textwidth]{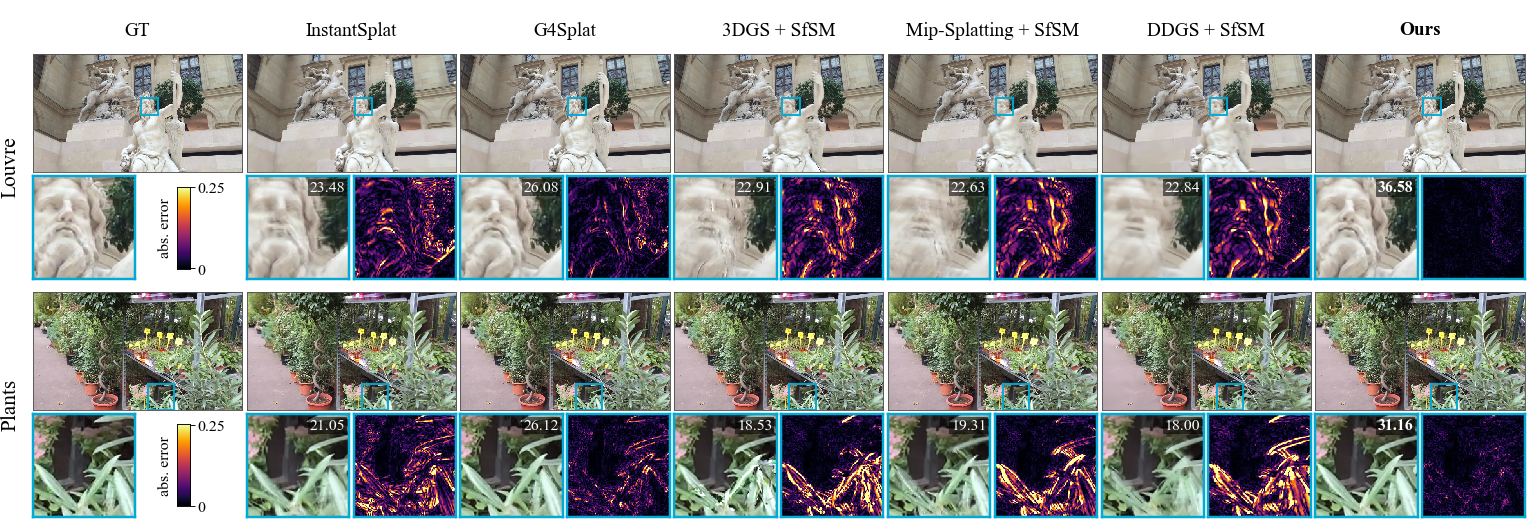}
    \caption{\textbf{Rendering results on Louvre and Plants.} Insets enlarge foreground detail (left) and show their corresponding absolute RGB error maps (right).}
    \label{fig:louvre_render}
\end{figure*}
}
\newcommand{\PhoneDepthFigure}{%
\begin{figure*}[t]
    \centering
    \includegraphics[width=\textwidth]{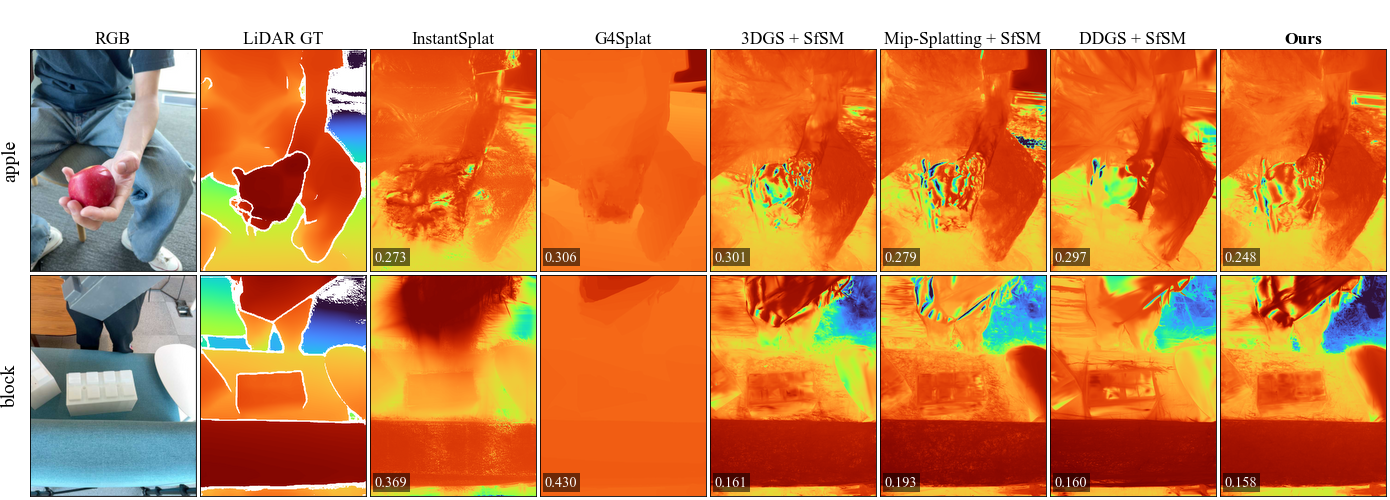}
    \caption{\textbf{Rendered depth against iPhone LiDAR.} Red is near; white indicates missing depth.}
    \label{fig:iphone_depth}
\end{figure*}
}

%% file: sections/compact/system_quant_table.tex
\newcommand{\SystemQuantTable}{%
\begin{table*}[t]
    \centering
    \caption{\textbf{Ten-scene comparison.} PSNR$\uparrow$/SSIM$\uparrow$/LPIPS$\downarrow$;  Bold/underlined: best/second.}
    \label{tab:system_quant}
    \fontsize{6}{7.2}\selectfont
    \setlength{\tabcolsep}{0.65pt}
    \renewcommand{\arraystretch}{1.10}
    \begin{tabular*}{\textwidth}{@{\extracolsep{\fill}}l*{11}{c}@{}}
        \toprule
        Method & Bikes & Dinos & Flowers & Frogs & Louvre & Notre-Dame & Plants & Stairs & Stones & Trees & Avg. \\
        \midrule
        \multicolumn{12}{@{}l}{\itshape External Baselines} \\
        Vanilla 3DGS~\cite{kerbl2023gaussians} & 7.36/0.00/0.71 & 9.82/0.12/0.57 & 5.36/0.00/0.81 & 8.54/0.08/0.65 & 3.82/0.00/0.65 & 5.49/0.00/0.73 & 5.83/0.01/0.75 & 6.69/0.00/0.71 & 5.49/0.03/0.72 & 5.87/0.08/0.69 & 6.43/0.03/0.70 \\
        InstantSplat~\cite{fan2024instantsplat} & 20.98/0.53/0.25 & \underline{27.09}/0.76/0.25 & 21.60/0.70/0.17 & 21.10/0.65/0.25 & 23.18/0.73/0.32 & 21.98/0.67/0.21 & 22.37/0.74/0.19 & 22.41/0.65/0.21 & 21.59/0.53/0.27 & 18.98/0.55/0.26 & 22.13/0.65/0.24 \\
        G4Splat~\cite{ni2026g4splat} & 29.72/0.89/0.10 & \textbf{30.71}/0.88/0.19 & 27.50/0.90/0.10 & 25.97/0.84/0.13 & \underline{30.24}/0.90/0.19 & 27.61/0.89/\textbf{0.12} & 25.61/0.85/0.13 & 26.58/0.84/0.13 & 31.10/0.91/0.11 & 25.75/0.86/0.11 & 28.08/0.87/0.13 \\
        Small GS~\cite{yao2025smallgs} & 20.53/0.52/0.27 & 21.29/0.51/0.42 & 19.07/0.50/0.31 & 21.28/0.69/0.26 & 26.62/0.83/0.26 & 16.07/0.23/0.39 & 16.97/0.42/0.36 & 19.96/0.42/0.39 & 19.37/0.31/0.44 & 17.90/0.48/0.35 & 19.91/0.49/0.35 \\
        \addlinespace[1.5pt]
        \multicolumn{12}{@{}l}{\itshape Initial solution and variants} \\
        DDGS~\cite{song2026d2gs} & 31.22/0.90/0.16 & 25.41/0.86/0.21 & 28.74/0.91/0.13 & 29.59/0.89/0.19 & 29.21/0.90/0.26 & 27.04/0.86/0.16 & 22.90/0.81/0.20 & 31.24/0.92/0.14 & 32.26/0.92/0.15 & 27.99/0.91/0.14 & 28.56/0.89/0.18 \\
        DDDrop~\cite{song2026d2gs} & 31.21/0.90/0.16 & 25.31/0.86/0.21 & 28.73/0.91/0.13 & 29.57/0.89/0.19 & 29.15/0.90/0.26 & 27.03/0.86/0.16 & 22.88/0.82/0.20 & 31.23/0.92/0.14 & 32.22/0.92/0.15 & 27.98/0.91/0.14 & 28.53/0.89/0.18 \\
        Mip-Splatting$^{\ddagger}$~\cite{yu2024mipsplatting} & \textbf{35.02}/\textbf{0.96}/\textbf{0.06} & 26.37/\underline{0.91}/0.14 & \textbf{30.99}/\textbf{0.94}/\textbf{0.08} & \underline{33.04}/\underline{0.94}/\underline{0.10} & 29.54/0.92/\underline{0.17} & \underline{28.41}/\underline{0.89}/0.13 & 25.09/0.86/0.13 & \underline{33.16}/\textbf{0.95}/\underline{0.08} & 34.44/0.95/\underline{0.09} & \textbf{30.28}/\textbf{0.94}/\textbf{0.08} & \underline{30.63}/0.93/\underline{0.11} \\
        \addlinespace[1.5pt]
        \multicolumn{12}{@{}l}{\itshape Our method} \\
        Vanilla 3DGS$^{\dagger}$& 34.62/0.96/\underline{0.06} & 26.19/\textbf{0.91}/\textbf{0.13} & \underline{30.98}/\underline{0.94}/0.08 & 31.76/0.93/0.11 & 29.61/\underline{0.92}/0.18 & 28.26/0.89/0.13 & \underline{25.86}/\underline{0.88}/\underline{0.13} & 33.14/0.95/0.08 & \underline{34.46}/\underline{0.95}/0.09 & 30.04/0.94/0.08 & 30.49/\underline{0.93}/0.11 \\
        \textbf{NBGS (Ours)} & \underline{34.87}/\underline{0.96}/0.06 & 25.49/0.90/\underline{0.14} & 30.96/\underline{0.94}/\underline{0.08} & \textbf{34.77}/\textbf{0.95}/\textbf{0.09} & \textbf{37.00}/\textbf{0.96}/\textbf{0.12} & \textbf{28.45}/\textbf{0.89}/\underline{0.12} & \textbf{30.29}/\textbf{0.94}/\textbf{0.09} & \textbf{33.17}/\underline{0.95}/\textbf{0.08} & \textbf{34.85}/\textbf{0.95}/\textbf{0.08} & \underline{30.21}/\underline{0.94}/\underline{0.08} & \textbf{32.01}/\textbf{0.94}/\textbf{0.09} \\
        \bottomrule
    \end{tabular*}
\par\vspace{2pt}\begin{minipage}{\textwidth}\fontsize{7}{8}\selectfont $^{\ddagger}$Opacity reset disabled, which also disables size pruning.\\
$^{\dagger}$With $R$ replaced by $\overline Z$ in size pruning, clone/split threshold and position learning rate (Sec.~\ref{sec:depth_scaled_pruning}).
 \end{minipage}
\end{table*}
}

%% file: sections/compact/03_method.tex
\section{Method}
\label{sec:method}

\subsection{Why 3DGS Fails under Narrow Baselines}

A narrow-baseline scene is recorded by a camera that moves only slightly, typically within a fraction of a second. We first characterize narrow-baseline geometry and then show how it fails to meet three implicit conditions of 3DGS.

\begin{table}[!b]
    \centering
    \footnotesize
    \caption{\textbf{Capture geometry.}
Camera-extent-to-depth ratio $\beta$ and angular proxy
$\theta_{\mathrm{proxy}}$ for the evaluated clips.}
    \label{tab:beta}
    \setlength{\tabcolsep}{1.2pt}
    \begin{tabular*}{\columnwidth}{@{\extracolsep{\fill}}lccccc@{}}
        \toprule
        \multicolumn{6}{@{}l}{\textit{Small-motion clips~\cite{ha2016smallmotion}}} \\
        Scene & Bikes & Dinos & Flowers & Frogs & Louvre \\
        $\beta(\%)$\,/\,$\theta_{\max}(^\circ)$ & 0.29\,/\,0.34 & 0.88\,/\,1.00 & 0.74\,/\,0.85 & 1.19\,/\,1.36 & 0.44\,/\,0.51 \\
        \addlinespace[2pt]
        Scene & Notre-Dame & Plants & Stairs & Stones & Trees \\
        $\beta(\%)$\,/\,$\theta_{\max}(^\circ)$ & 0.11\,/\,0.12 & 0.92\,/\,1.05 & 0.60\,/\,0.69 & 0.69\,/\,0.79 & 0.83\,/\,0.95 \\
        \midrule
        \multicolumn{6}{@{}l}{\textit{iPhone clips~\cite{gao2022dynamic}}} \\
        Scene & apple & backpack & block & mochi & \\
        $\beta(\%)$\,/\,$\theta_{\max}(^\circ)$ & 0.51\,/\,0.58 & 0.75\,/\,0.85 & 1.23\,/\,1.41 & 0.42\,/\,0.48 & \\
        \bottomrule
    \end{tabular*}
\end{table}

The base-to-height ratio classically characterizes multi-view geometry in previous research, especially for defining narrow baseline scenes ~\cite{delon2007small,yu2014accidental,ha2016smallmotion}. Given images $\mathcal{I}=\{I_t\}_{t=1}^{T}$, let $R=r_{\max}$ be the camera extent, with $r_{\max}$ representing the largest distance of the camera centers from their mean, and let $\overline Z$ be the median sparse-point depth.  We characterize the capture geometry by comparing the camera
extent $R$ with the median sparse-point depth $\overline Z$:
\begin{equation}
    \beta=\frac{R}{\overline Z},
    \qquad
    \theta_{\mathrm{proxy}}=2\arctan(\beta).
    \label{eq:baseline_ratio}
\end{equation}
The angular proxy $\theta_{\mathrm{proxy}}$ corresponds to an idealized symmetric lateral baseline of length $2R$ viewing a point at depth $\overline Z$. It provides an angular interpretation of the baseline-to-depth ratio. Small values of $\beta$ indicate limited camera extent relative to scene depth.

Table~\ref{tab:beta} shows narrow-baseline clips reach only about 0.11\%--1.23\% on $\beta$ and $0.12^\circ-1.41^\circ$ on $\theta_{proxy}$, while captures orbiting an object have $\beta$ of order one and $\theta_{proxy}$ above $1.5^\circ$ that COLMAP requires for a stable triangulation~\cite{delon2007small,schonberger2016sfm}. Such narrow-baseline geometry has two characteristics. First, $\beta$ is small and implies that the camera extent greatly underestimates the scene size. Second, the global scale is uncertain, since SfSM fixes it arbitrarily under weak parallax. These characteristics violate the 3DGS conditions.

\paragraph{Condition 1: Camera extent can reflect scene scale and prune Gaussians.} 3DGS scales its clone/split threshold, size-pruning threshold and position learning rate by the camera extent $R$, implicitly assuming that the cameras span the whole scene~\cite{grubert2025improving}. Under a narrow baseline, however, $R$ is only a fraction $\beta$ of the real scene scale, so the pruning threshold $0.1R$ projects to pixels even smaller than 1 pixel at most points. Ordinary Gaussians are therefore judged oversized, and size pruning removes most of the model after the first opacity reset and keeps doing so at every densification step, leading to a catastrophic collapse.

\begin{figure*}[!t]
    \centering
    \captionsetup{skip=3pt}
    \includegraphics[width=\textwidth]{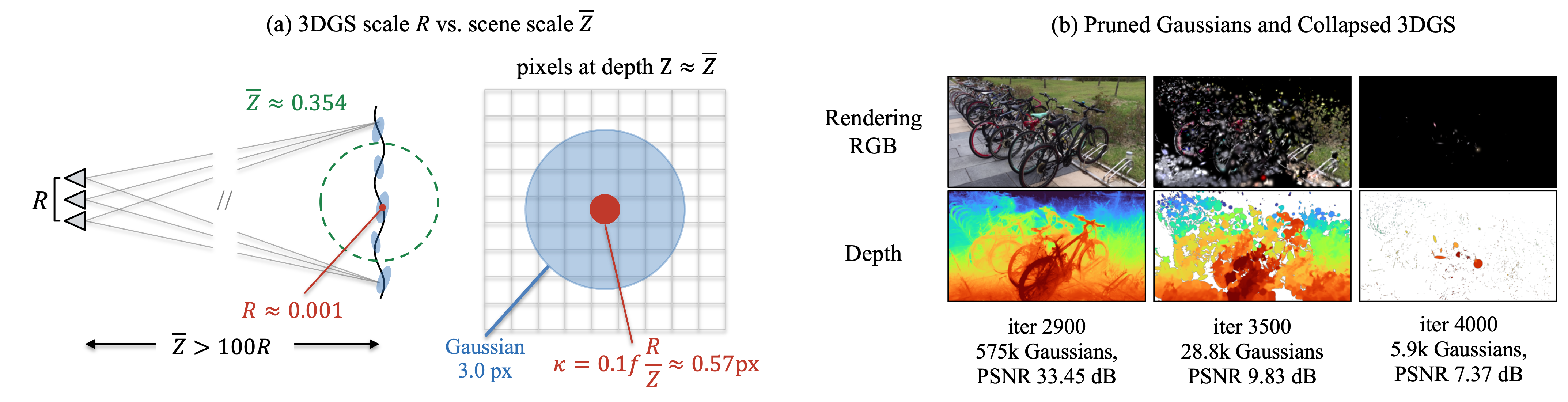}
    \caption{\textbf{How incorrect scene scale influences 3DGS pruning.} (a) Camera extent $R$ versus median depth $\overline Z$, and pruning radius $\kappa$ versus a typical Gaussian in pixels at depth $\overline Z$  (b) The pruning whose size is determined by scene scale makes renders and depths collapse.}
    \label{fig:pruning_collapse}
\end{figure*}

\paragraph{Condition 2: Triangulation is well-conditioned and sparse depths are reliable.} Previous depth-supervised 3DGS treat sparse depths as exact targets and use them for supervision. Under weak parallax, a matching error in depth estimation can be magnified many times as the depth $Z$ and baseline length $B$ change (detailed in Appendix~\ref{app:compact_uncertainty}), so even subpixel errors become a large fraction of depth. Treating these depths as exact targets pulls Gaussians towards biased surfaces and discards the uncertainty that separates reliable tracks from unreliable ones.

\paragraph{Condition 3: A fixed absolute near plane separates valid from invalid Gaussians.} Rasterization and covariance filtering discard Gaussians closer than a fixed absolute depth, such as $z=0.2$, which presumes a scene scale at which this value lies well before the nearest surface. SfSM fixes the scale arbitrarily, so the same constant may fall in front of the nearest surface in one reconstruction and behind it in another, either removing valid foreground Gaussians or retaining unstable near-camera projections.

To address the unmet conditions, we remodel the narrow-baseline geometry. Fig.~\ref{fig:method_overview} summarizes our method NBGS. We first track features through the clip and recover camera poses, a shared focal length and sparse points, which initialize 3DGS. Scale-adjusted pruning replaces $R$ by $\overline Z$ in density control. The near-side bounds of the depth distributions set a per-camera near plane, which covariance filtering and rasterization share. During optimization, uncertainty-aware ray supervision matches the distribution of Gaussian contributions along each ray through a CRPS loss.

\subsection{Depth-based Scene Rescaling}
\label{sec:depth_scaled_pruning}

Condition 1 fails most visibly by size pruning. In 3DGS, size pruning removes every Gaussian whose largest scale exceeds $0.1R$. Fig.~\ref{fig:pruning_collapse} shows that, on Bikes, at the median sparse depth $\overline Z\approx340R$, this threshold corresponds to only $\kappa=0.57$ pixels, whereas the median Gaussian of a normally converging model spans about 3.0 pixels. Size pruning therefore judges 93.9\% of the Gaussians that the scene needs as oversized and removes them, collapsing the reconstruction.

\begin{figure*}[!t]
    \captionsetup{skip=3pt}
    \centering
    \includegraphics[width=\textwidth,trim=0 14.84bp 0 7.85bp,clip]{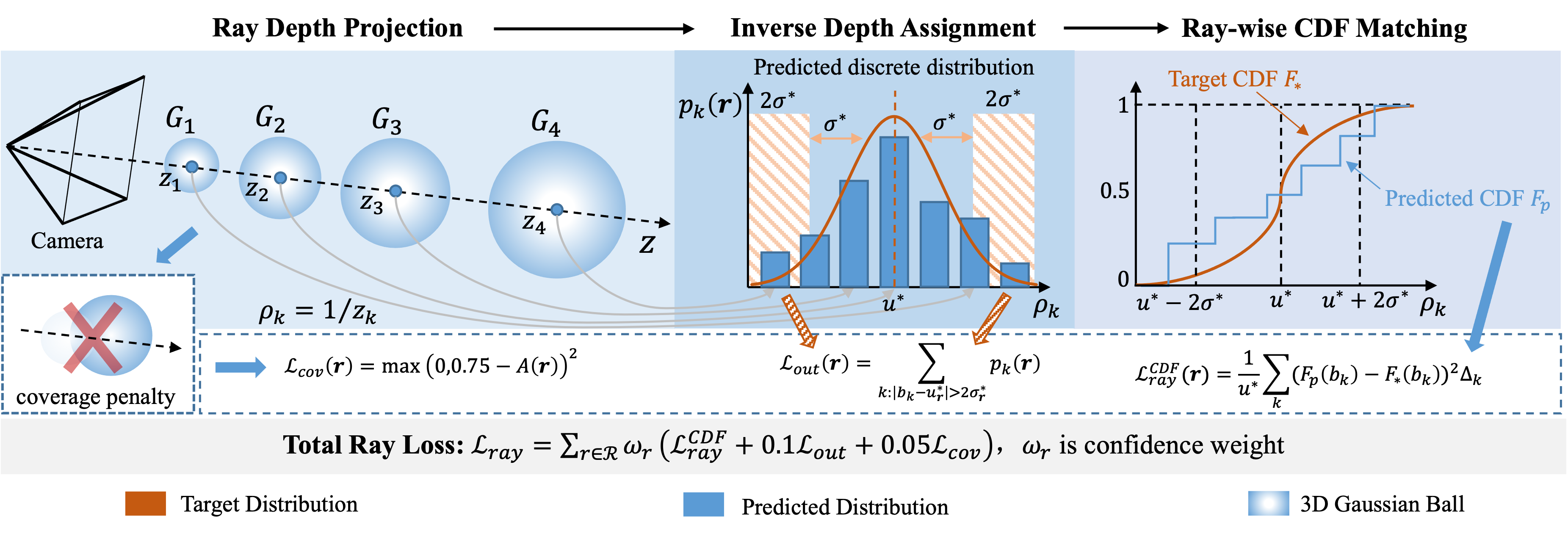}
    \caption{\textbf{Uncertainty-aware ray supervision.} Gaussian contributions form a predicted distribution, matched to the confidence-conditioned target.}
    \label{fig:ray_loss}
\end{figure*}

\begin{table}[b]\centering
\caption{\textbf{Pruning study in Vanilla 3DGS.} All metrics are averaged over ten narrow-baseline scenes and five seeds at 15k iterations.}
\label{tab:pruning_ablation}
\fontsize{7}{8.2}\selectfont\setlength{\tabcolsep}{3pt}
\begin{tabular*}{\columnwidth}{@{\extracolsep{\fill}}ccccccc@{}}\toprule
\multicolumn{4}{c}{Configuration} & \multicolumn{3}{c}{Ten scenes} \\
\cmidrule(lr){1-4}\cmidrule(l){5-7}
Reset & Prune & Split & LR & PSNR$\uparrow$ & SSIM$\uparrow$ & LPIPS$\downarrow$ \\\midrule
on & $R$ & $R$ & $R$ & 6.43 & 0.033 & 0.699 \\
off & off & $R$ & $R$ & 30.06 & 0.919 & 0.116 \\
on & $\overline Z$ & $R$ & $R$ & 30.37 & 0.922 & 0.113 \\
\midrule
on & $\overline Z$ & $\overline Z$ & $R$ & 29.75 & 0.916 & 0.120 \\
off & off & $\overline Z$ & $\overline Z$ & 30.30 & 0.925 & 0.108 \\
on & $\overline Z$ & $\overline Z$ & $\overline Z$ & \textbf{30.49} & \textbf{0.927} & \textbf{0.106} \\
\bottomrule\end{tabular*}\end{table}

The pruning starts after the first opacity reset. For Vanilla 3DGS, the model reaches 33.45 dB with 575k Gaussians at iteration 2,900. The first pruning removes 95.0\% of them, starting with the large Gaussians that cover flat regions such as the ground. Gaussians at the border of a hole must grow to cover it, but any of them that exceeds $\kappa$ is pruned again at the next step. By iteration 4,000, only 1\% Gaussians remain. The removal is irreversible, and the collapse happens in all experiments on the narrow-baseline scenes, which end between 3.8 and 9.8 dB PSNR.

Table~\ref{tab:pruning_ablation} shows that correcting the pruning scale avoids collapse while retaining size pruning. Disabling opacity reset also disables size pruning in this implementation, reaching 30.06 dB but bypassing size-based removal of large floaters. Instead, replacing $R$ with $\overline Z$ in size pruning alone preserves the default schedule and reaches 30.37 dB.

The same scale also controls the clone/split threshold and the position learning rate. These parameters determine how Gaussians are densified and subsequently moved during optimization. Additionally, replacing $R$ in the clone/split threshold, while leaving the learning rate unchanged, reduces PSNR to 29.75 dB. Scaling the position learning rate by $\overline Z$ as well restores a consistent scale across all three uses. This configuration achieves the highest mean PSNR of 30.49 dB and outperforms the other settings on small-motion clips.

\subsection{Probabilistic Depth Modeling and Supervision}
\label{sec:ray_supervision}

After stabilizing scene scale and density control, we account for the uncertainty in the sparse depths used for reconstruction. Under weak parallax, small image measurement errors can produce substantial depth errors. In a rectified two-view approximation, their relative uncertainty satisfies $\sigma_Z/Z \simeq \sigma_d/d$, where $d$ is the disparity (detailed in Appendix~\ref{app:compact_uncertainty}). Table~\ref{tab:feature_lidar_uncertainty} illustrates this sensitivity via tracked features and Lidar GT in iPhone clips. After aligning each scene by one mean-depth scale, though median reprojection errors are only 0.07--0.24 pixels, depth nRMSE remains 7.26\%--17.46\%.

Treating these estimates as exact targets can pull Gaussians towards incorrect surfaces. We instead construct a normal distribution whose width controls how tightly each estimate constrains reconstruction. For a sparse observation $r=(t,j)$, let $\widehat Z_{tj}$ denote its estimated camera-space depth. The track reliability $c_r$ combines tracking, reprojection, parallax, conditioning, and visibility diagnostics. We retain finite positive-depth observations with $c_r \geq 0.05$, forming the supervised ray set. Using inverse depth, we define:
\begin{equation}
    \begin{aligned}
        \mu_r^*
        &= \frac{1}{\widehat Z_{tj}}, \\
        (\sigma_r^*)^2
        &= (\mu_r^*)^2
        \frac{\sigma_{\mathrm{rel}}^2}
        {c_r}
        + (0.5\overline{\Delta}_{\rho})^2, \\
        \rho_r^*
        &\sim \mathcal N\!\left(
            \mu_r^*,(\sigma_r^*)^2
        \right).
    \end{aligned}
    \label{eq:target_sigma}
\end{equation}
Here, $\sigma_{\mathrm{rel}}$ sets the relative target width, and we add $(0.5\overline{\Delta}_{\rho})^2$ to the target variance, ensuring that its standard deviation is at least half the inverse-depth bin width. Therefore, lower reliability produces a broader target, allowing greater geometric adjustment during optimization. This distribution will guide supervision and near-plane filter settings.

\begin{table}[t]\centering
\caption{\textbf{SfSM feature-depth errors against LiDAR.} nRMSE is normalized by mean LiDAR depth.}
\label{tab:feature_lidar_uncertainty}
\fontsize{7}{8.2}\selectfont\setlength{\tabcolsep}{2pt}\renewcommand{\arraystretch}{1.06}
\begin{tabular*}{\columnwidth}{@{\extracolsep{\fill}}lrrrr@{}}\toprule
Scene & Pairs & Reproj. (px) & AbsRel (\%) & nRMSE (\%) \\\midrule
apple & 2,253 & 0.16 & 4.51 & 16.01 \\
backpack & 4,036 & 0.10 & 5.37 & 7.26 \\
block & 3,923 & 0.07 & 4.10 & 10.55 \\
mochi-high-five & 930 & 0.24 & 7.23 & 17.46 \\
\bottomrule\end{tabular*}\end{table}

Let $e_0,\ldots,e_K$ be uniformly spaced inverse-depth bin boundaries with width $h$. Each Gaussian distributes its contribution among the bins according to its inverse depth $1/z_i$. The assignment weight $q_{ik}$ specifies the fraction assigned to bin $k$, with $\sum_k q_{ik}=1$. The normalized bin masses are:
\begin{equation}
    \begin{aligned}
        p_k(r)
        &= \frac{\sum_i a_i(r)q_{ik}}
        {\max(A(r),\epsilon)}, \\
        F_r(e_k)
        &= \sum_{l=1}^{k}p_l(r),
        \qquad A(r)=\sum_i a_i(r).
    \end{aligned}
    \label{eq:ray_distribution}
\end{equation}
Thus, $p_k(r)$ measures the rendering contribution assigned to bin $k$, while $F_r(e_k)$ accumulates the contributions through that bin. We match this cumulative distribution to the target normal CDF, $F_r^*(e_k)=\Phi((e_k-\mu_r^*)/\sigma_r^*)$. Comparing cumulative values preserves the ordering of inverse depths and penalizes contributions displaced along the ray. Our ray-wise CDF matching loss is
\begin{equation}
    \mathcal L_{\mathrm{ray}}^{\mathrm{CDF}}(r)
    =
    \frac{h}{\max(\mu_r^*,\epsilon)}
    \sum_{k=1}^{K-1}
    \left[
        F_r(e_k)-F_r^*(e_k)
    \right]^2.
    \label{eq:ray_cdf}
\end{equation}
Here, $\Phi$ is the standard normal CDF. Multiplication by $h$ approximates integration over inverse depth, and division by $\mu_r^*$ normalizes the loss by the target scale. This squared-CDF objective is closely related to the continuous ranked probability score~\cite{gneiting2007proper}. Reliable estimates produce concentrated targets, whereas uncertain estimates spread the target mass over a wider inverse-depth range.

We additionally discourage outside-mass contributions using the valid interval $\mathcal I_r=[\mu_r^*-2\sigma_r^*,\,\mu_r^*+2\sigma_r^*]$. This interval contains 95.45\% of the modeled normal mass. An outside-mass loss suppresses relative contributions beyond it, which can be described as:
\begin{equation}
    \mathcal L_{\mathrm{out}}(r)= \sum_{k:\,b_k\notin\mathcal I_r}p_k(r)
\end{equation}

\SystemQuantTable

CDF matching constrains the relative distribution of rendering contributions along a ray. However, a ray with very low accumulated opacity can produce the same normalized distribution as one with substantial opacity. Matching the CDF alone may leave an observed feature insufficiently represented by the Gaussian scene. We therefore add an opacity coverage loss at supervised feature locations:
\begin{equation}
    \mathcal L_{\mathrm{cov}}(r)
    =
    \max\!\left(0,\,\tau-A(r)\right)^2,
    \qquad \tau=0.75,
    \label{eq:ray_coverage}
\end{equation}
where $A(r)=\sum_i T_i(r)\alpha_i(r)$ is the accumulated opacity. This loss encourages sufficient rendering contribution along each supervised ray. Together, CDF matching, outside mass, and opacity coverage constrain both the distribution and total magnitude of rendering contributions with confidence ratio $w_r=c_r/\sum_{s\in\mathcal R}c_s$. The ray supervision is as follows:
\begin{equation}
    \mathcal L_{\mathrm{ray}}
    =
    \sum_{r\in\mathcal R}w_r
    \left(
        \mathcal L_{\mathrm{ray}}^{\mathrm{CDF}}(r)
        +0.10\mathcal L_{\mathrm{out}}(r)
        +0.05\mathcal L_{\mathrm{cov}}(r)
    \right).
    \label{eq:ray_objective}
\end{equation}

\subsection{Depth-Guided Adaptive Near-Plane Filtering}
\label{sec:renderer_validity}

Condition3 fails when a fixed near cutoff falls inside valid foreground geometry, and this failure makes the image blurry or aliased. Common rasterization implementations used by 3DGS and Mip-Splatting reject Gaussians with camera-space depth $z\leq0.2$~\cite{kerbl2023gaussians,yu2024mipsplatting}. This absolute threshold does not account for the recovered scene scale and can therefore discard valid foreground Gaussians. Under narrow baselines, their depths remain similar across views, so the same foreground may be repeatedly excluded.

The cutoff can also affect covariance prefiltering. Filter scale is estimated from distances in views that pass the depth-validity test. If a foreground Gaussian fails this test in every view, the original implementation assigns a fallback distance that may be much larger than its actual depth. Since the filter width increases with this distance, the Gaussian becomes excessively smoothed, which can degrade object boundaries, contributing to blur and ghosting. 

We therefore derive each camera's near plane from the depth distributions introduced in Section~\ref{sec:ray_supervision}. The boundary should accommodate the near-side extent of the retained observations, including their uncertainty. We take the minimum of the valid interval as the near-plane filter scale, leaving more room for uncertain foreground geometry as follows:
\begin{equation}
    \begin{aligned}
        \underline Z_r
        &= \frac{1}{\mu_r^*+2\sigma_r^*}, \\
        z_{\mathrm{near},t}
        &= \min_{r\in\mathcal R_t}\underline Z_r.
    \end{aligned}
    \label{eq:near_clip}
\end{equation}

Covariance prefiltering and rasterization share this boundary, so both stages use the same valid depth range. The boundary follows reconstruction rescaling when the target distributions are rescaled consistently.

%% file: sections/compact/04_experiments.tex
\section{Experimental Setup}
\label{sec:experiments}

\subsection{Dataset and Evaluation}
\label{sec:data}
We evaluate 10 narrow-baseline scenes from the small-motion dataset~\cite{ha2016smallmotion}, using 30 frames each. The middle 4 are reserved for rendering evaluation; the other 26 supervise Gaussian optimization. We measure PSNR, SSIM~\cite{wang2004ssim} and LPIPS~\cite{zhang2018lpips} as metrix. We additionally extract clips of 15--20 consecutive frames from four DyCheck iPhone sequences: apple, backpack, block, and mochi-high-five~\cite{gao2022dynamic}. The selected clips satisfy our narrow-baseline criterion, $\theta_{\max}<1.5^\circ$. Their LiDAR depth serves as ground truth for depth evaluation. Their middle four frames are likewise reserved for Gaussian evaluation. We measure rendered-depth AbsRel~\cite{eigen2014depth} and $\delta_{1.25}$~\cite{ladicky2014pulling} after one median-scale alignment per scene, on common static pixels, excluding unreliable depth edges. Every scene run methods under seeds 0--4.

\subsection{Baselines and Variants}
InstantSplat~\cite{fan2024instantsplat} and G4Splat~\cite{ni2026g4splat} use their native frontends. Small GS~\cite{yao2025smallgs} supplies refined poses to an RGB-only backend. Vanilla 3DGS~\cite{kerbl2023gaussians}, DDGS~\cite{song2026d2gs}, DDDrop and Mip-Splatting~\cite{yu2024mipsplatting} share SfSM geometry. These backends all use 15k iterations for training and refinement. Our initial Vanilla 3DGS and NBGS replace the camera extent in all three uses in Sec.~\ref{sec:depth_scaled_pruning}; Mip-Splatting disables reset and coupled size pruning. NBGS-Base retains opacity reset and size pruning from NBGS and omits ray supervision. All methods retain the 7.5k densification window.

\RGBComparisonFigure
\PhoneDepthFigure
\label{sec:controlled_comparisons}
\label{sec:setup_end}

%% file: sections/compact/05_results.tex
\section{Results and Analysis}
\label{sec:results}

\subsection{Rendering Quality}
\label{sec:image_quality}

We first examine how the Gaussian backend uses the recovered SfSM geometry. With the same cameras and sparse points, NBGS reaches 32.01 dB PSNR, compared with 30.49 dB for the depth-scaled Vanilla initial solution and 30.63 dB for Mip-Splatting (Table~\ref{tab:system_quant}). It also exceeds G4Splat, the strongest external baseline evaluated, by 3.93 dB. Improvements over the initial solution occur on eight scenes, with the largest gains on Louvre and Plants. Performance remains scene-dependent: Mip-Splatting retains higher PSNR on four scenes.

Figure~\ref{fig:louvre_render} makes these differences visible in foreground detail. On Louvre, competing SfSM backends soften facial contours and fingers or introduce repeated boundaries. On Plants, near-camera leaves exhibit blurred or doubled edges. NBGS preserves more distinct contours in these regions. These examples illustrate localized improvements; they do not establish their prevalence across each image. The comparison uses seed-0 renders, except archived InstantSplat and G4Splat results on Louvre. Crop-selection details and complete scene comparisons are provided in Appendix~\ref{app:full_qualitative}.

\subsection{Depth Accuracy on Phone Captures}
\label{sec:iphone_depth}
\input{sections/compact/iphone_depth_table}

We next evaluate whether the rendering improvements are accompanied by more accurate depth. NBGS and Mip-Splatting achieve similar PSNR on the iPhone clips, yet their mean depth AbsRel differs: 0.214 versus 0.253 (Table~\ref{tab:iphone_depth}). Their corresponding $\delta_{1.25}$ accuracies are 0.725 and 0.714. Similar image quality can therefore conceal different geometric errors.

The improvement varies across scenes. Vanilla 3DGS remains slightly more accurate on block and mochi-high-five, although NBGS-Base achieves lower average error. Figure~\ref{fig:iphone_depth} compares the rendered depths with LiDAR references and shows that our method improves average depth accuracy under depth ambiguity and weak parallax provided by narrow-baseline scenes.

\subsection{Ablations}
\label{sec:ray_ablation}
\label{sec:confidence_ablation}
\label{sec:matched_near}

The ablation studies examine whether the improvements arise from depth supervision, target representation or rendering boundaries. Comparisons use the NBGS Base configuration, which retains our scene rescaling while disabling ray supervision. Geometry, initialization and training settings are held fixed within each comparison.

\paragraph{Depth supervision.}
Table~\ref{tab:ray_ablation}(a) compares different ways of using estimated depth along rendering rays. DDGS represents an existing depth-guided approach. Despite sharing the same SfSM input, DDGS performs below the RGB-only reference. Therefore, depth guidance does not automatically improve reconstruction when the input depths are uncertain. The Base variants further compare supervision choices within a shared backend. CDF matching achieves 32.01 dB and exceeds relative-L1 supervision by 1.04 dB, while the no-ray control achieves a lower metric. 
These comparisons distinguish the behavior of different depth-use strategies and show that if depth is directly used as a reliable prior, it may even lead to degradation. Using depth information as soft supervision, such as CDF loss, 3DGS can better fit the corresponding depth information and achieve better results

\paragraph{Target representation.}
Target width and bin count control different aspects of ray supervision: the former specifies the tolerance around an estimated depth, while the latter determines the resolution of distribution matching. Confidence-conditioned targets achieve lower LPIPS than fixed-width targets, supporting the use of track reliability to adapt this tolerance. We further compare $K\in\{4,6,8\}$ while keeping the continuous target widths and inverse-depth range fixed. More bins refine the approximation of the same distributional objective. Six bins achieve the best PSNR, SSIM, and LPIPS among the tested settings. Increasing the resolution to eight bins brings no further improvement. We therefore adopt $K=6$ as a compact discretization with the best observed mean performance.

\paragraph{Near-plane selection.}
The near-plane ablation examines how the rendering boundary affects foreground reconstruction. A fixed cutoff of $0.2$ can lie inside valid foreground geometry. It then removes Gaussians needed to represent the nearest surfaces. Replacing this cutoff with our adaptive near plane achieves a PSNR improvement of 1.50 dB (Table~\ref{tab:matched_near}(c)). The improvement is concentrated in the affected foreground. The adaptive boundary therefore recovers foreground quality lost under an overly restrictive cutoff.
A small fixed cutoff of $10^{-4}$ also restores reconstruction quality, but influences covariance filtering and rasterization afterward. Our policy preserves uncertainty margins around valid foreground observations while adapting the accepted depth range to the reconstructed scene.
\input{sections/compact/combined_scene_ablations}

%% file: sections/compact/iphone_depth_table.tex
\begin{table}[!t]
    \centering
    \caption{\textbf{Phone LiDAR evaluation.} AbsRel$\downarrow$, $\delta_{1.25}$$\uparrow$, PSNR$\uparrow$. Bold/underlined: best/second.}
    \label{tab:iphone_depth}
    \fontsize{7}{8.4}\selectfont
    \setlength{\tabcolsep}{0.65pt}
    \renewcommand{\arraystretch}{1.10}
    \begin{tabular*}{\columnwidth}{@{\extracolsep{\fill}}l*{7}{c}@{}}
        \toprule
        & \multicolumn{5}{c}{AbsRel$\downarrow$} & & \\
        \cmidrule(lr){2-6}
        Method & apple & backpack & block & mochi & Avg. & $\delta_{1.25}\uparrow$ & PSNR$\uparrow$ \\
        \midrule
        \multicolumn{8}{@{}l}{\itshape External Baselines} \\
        G4Splat~\cite{ni2026g4splat} & 0.306 & 0.301 & 0.430 & \underline{0.274} & 0.328 & 0.453 & 27.16 \\
        Small GS~\cite{yao2025smallgs} & 0.296 & 0.180 & 0.193 & \textbf{0.251} & 0.230 & 0.676 & 23.41 \\
        \addlinespace[1.5pt]
        \multicolumn{8}{@{}l}{\itshape With SfSM~\cite{ha2016smallmotion}} \\
        DDGS~\cite{song2026d2gs} & \underline{0.271} & 0.173 & 0.164 & 0.318 & 0.232 & 0.646 & 26.93 \\
        DDDrop~\cite{song2026d2gs} & 0.287 & 0.161 & 0.163 & 0.314 & 0.231 & 0.656 & 26.89 \\
        Mip-Splatting~\cite{yu2024mipsplatting} & 0.349 & 0.149 & 0.185 & 0.329 & 0.253 & \underline{0.714} & \underline{28.08} \\
        \addlinespace[1.5pt]
        \multicolumn{8}{@{}l}{\itshape Our method} \\
        Revised Vanilla 3DGS & 0.305 & \underline{0.133} & \textbf{0.153} & 0.292 & \underline{0.221} & 0.701 & 27.92 \\
        \textbf{NBGS (Ours)} & \textbf{0.270} & \textbf{0.132} & \underline{0.156} & 0.297 & \textbf{0.214} & \textbf{0.725} & \textbf{28.08} \\
        \bottomrule
    \end{tabular*}
\end{table}

%% file: sections/compact/combined_scene_ablations.tex
\begin{table}[!t]
\centering
\caption{\textbf{Ablation studies.}\\
NBGS: $w/ L_{ray}^{CDF},Cond. width, K=6, Z_{\mathrm{near},t}.$.}
\label{tab:ray_ablation}
\label{tab:loss_ablations}
\label{tab:confidence_ablation}
\label{tab:matched_near}

\small
\setlength{\tabcolsep}{3pt}
\renewcommand{\arraystretch}{1.04}

\begin{tabular*}{\columnwidth}
{@{\extracolsep{\fill}}lccc@{}}
\toprule
Setting & PSNR$\uparrow$ & SSIM$\uparrow$ & LPIPS$\downarrow$ \\

\midrule
\multicolumn{4}{@{}l}{\textbf{(a) Ray supervision}} \\
DDGS~\cite{song2026d2gs}
    & 28.56 & 0.89 & 0.18 \\
NBGS Base, w/o
    & 30.06 & 0.92 & 0.12 \\
NBGS Base, w/ $L_{\mathrm{ray}}^{1}$
    & \underline{30.97} & \underline{0.93} & \underline{0.09} \\

\midrule
\multicolumn{4}{@{}l}{\textbf{(b) Target width / bins}} \\
Width $c=1$, $K=6$
    & 31.99 & 0.94 & 0.09 \\
Cond. width, $K=4$
    & \underline{32.00} & 0.94 & 0.09 \\
Cond. width, $K=8$
    & 32.00 & \underline{0.94} & \underline{0.09} \\

\midrule
\multicolumn{4}{@{}l}{\textbf{(c) Near-plane policy}} \\
Fixed $0.2$
    & 30.51 & 0.92 & 0.11 \\
Fixed $10^{-4}$
    & \underline{31.58} & \underline{0.93} & \underline{0.09} \\

\midrule
\textbf{NBGS(ours)}
    & {32.01} & \textbf{0.94} & \textbf{0.09} \\
\bottomrule
\end{tabular*}
\end{table}

%% file: sections/06_conclusion.tex
\section{Conclusion}
\label{sec:conclusion}
We show how narrow-baseline scenes challenge the 3DGS backend and present our solutions NBGS. Three conditions cannot hold for narrow-baseline scenes. For the undersized scene scale, we replace the camera extent with the median depth and stop 3DGS from collapsing. For the fixed near-plane filter, we design an adaptive filtering guided by depth to fix blurriness and aliasing in images. For uncertain geometry and depth, we propose probabilistic depth supervision to fully use the uncertain depth and refine geometry. The three improvements make our framework reach 32.01 dB PSNR, exceeding the SOTA Mip-Splating by 1.38 dB on small motion clips. On narrow-baseline iPhone clips with LiDAR references, it also attains $72.5\%$ $\delta_{1.25}$ accuracy, exceeding SOTA Mip-Splating by $1.1\%$.

Though we have improved the performance of 3DGS on narrow baselines, there are still certain limitations. Our current framework's design and evaluation focus on narrow-baseline clips, and its generalization to wide-baseline captures remains unverified. Extending it to a broader range of camera motions will require a more flexible geometry frontend and corresponding uncertainty estimates. Also, we will explore efficient geometry estimation and Gaussian optimization in future work, with the goal of building a unified framework for real-time, on-device reconstruction on handheld devices.

%% file: sections/llm_usage_statement.tex
\section*{LLM Usage Statement}

The authors used ChatGPT (GPT-5.6 Sol) to assist with grammar checking and language editing of the manuscript. Codex 5.6 was used to assist with adapting and reproducing baselines whose original code was not directly compatible with small-motion clips, including DDGS and Small GS. The authors take full responsibility for the manuscript and experimental results.

%% file: sections/appendix.tex
\appendix
\twocolumn[{%
  \pdfbookmark[0]{Appendix}{appendix.start}%
  \centering\Large\bfseries Appendix\par\vspace{1em}%
}]
\setcounter{figure}{0}
\setcounter{table}{0}
\setcounter{equation}{0}
\renewcommand{\thesection}{S\arabic{section}}
\renewcommand{\thefigure}{S\arabic{figure}}
\renewcommand{\thetable}{S\arabic{table}}
\renewcommand{\theequation}{S\arabic{equation}}
\renewcommand{\theHsection}{supplement.S.\arabic{section}}
\renewcommand{\theHfigure}{supplement.S.\arabic{figure}}
\renewcommand{\theHtable}{supplement.S.\arabic{table}}
\renewcommand{\theHequation}{supplement.S.\arabic{equation}}

\input{sections/theory_lidar}

\onecolumn
\ifdefined\nolinenumbers\nolinenumbers\fi
\input{sections/comparison_intro}
\input{sections/scene_comparisons}
\input{sections/phone_comparisons}

\twocolumn
\ifdefined\linenumbers\linenumbers\fi
\input{sections/comparison_protocol}

%% file: sections/theory_lidar.tex
\section{Narrow-Baseline Clips and Depth Uncertainty}
\label{app:compact_uncertainty}
\label{app:feature_lidar_validation}

\subsection{Datasets and clip selection}
\begin{figure*}[!b]\centering
\includegraphics[width=.96\textwidth]{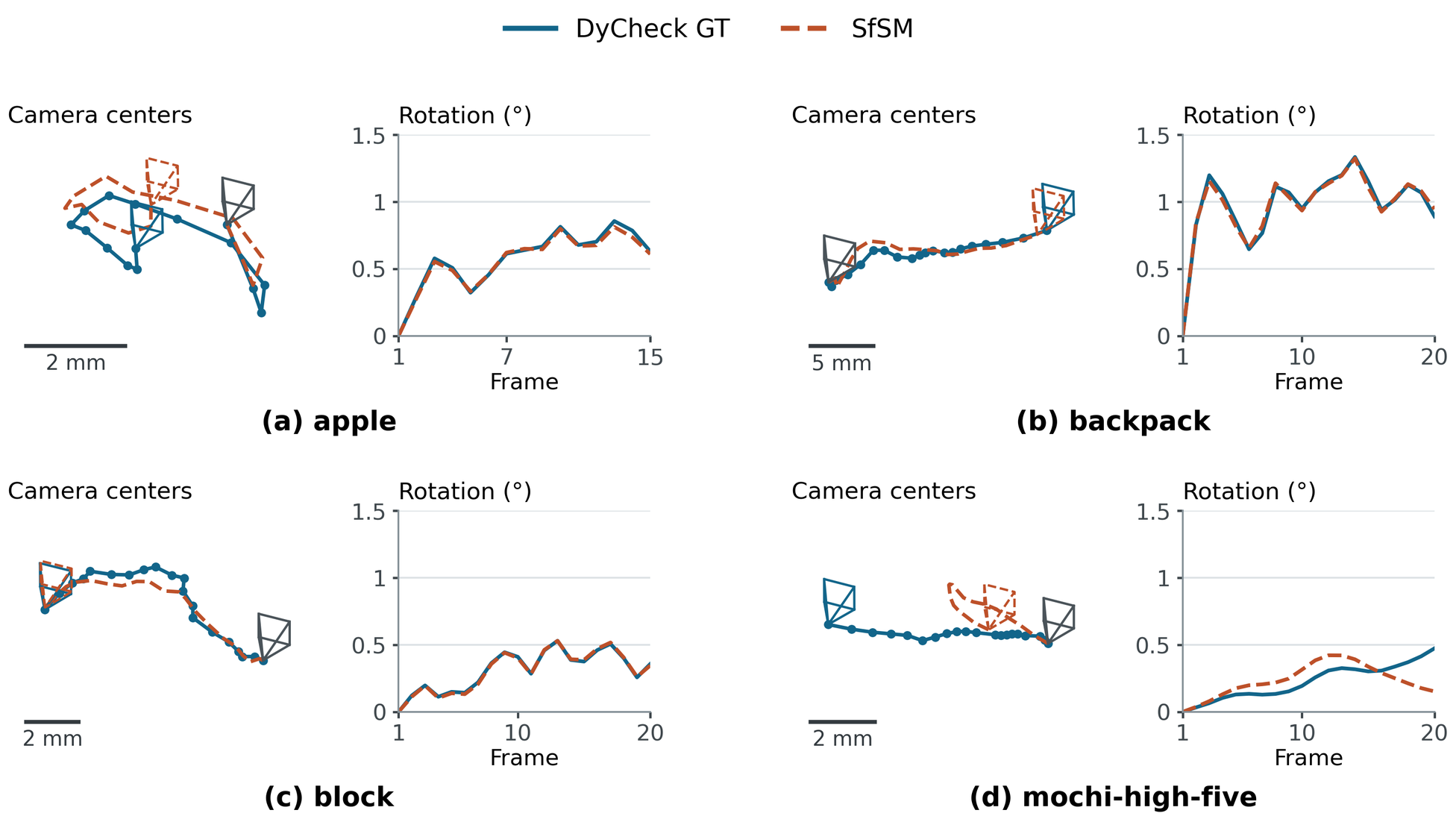}
\caption{\textbf{DyCheck and SfSM camera poses.} Left in each panel, an orthographic view of the camera-center paths. Wireframes mark the shared first pose (gray) and each final pose. Scale bars are in millimeters. Right, the angle of each camera's rotation relative to its first frame.}
\label{fig:supp_dycheck_poses}
\end{figure*}

We use the ten scenes of the Small-Motion dataset~\cite{ha2016smallmotion} to evaluate rendering quality. SfSM processes all 30 frames of each clip. The middle four frames supply evaluation views, and the other 26 supply the images used for Gaussian optimization. To examine depth as well as appearance, we also extract short windows from DyCheck iPhone sequences~\cite{gao2022dynamic}, which provide camera poses and LiDAR depth.

For DyCheck, we consider windows of 15 or 20 consecutive frames. We use the released poses and LiDAR to select windows in which the largest distance between camera centers is at most $0.017$ times the median of the frame-level LiDAR depth medians and the largest relative rotation is at most $1.4^\circ$. We rank candidates by reprojection consistency on the middle four frames, then run SfSM with the recorded focal length and zero distortion. We retain windows with at least 200 features and sparse-depth AbsRel at most $0.10$ after matching the median SfSM and LiDAR depths of the paired reference-frame features.

The retained camera-0 windows are apple (frames 3--17), backpack (20--39), block (6--25), and mochi-high-five (3--22), all at $720\times960$. The angular proxy in Eq.~\eqref{eq:baseline_ratio}, evaluated on the recovered geometry, ranges from $0.48^\circ$ to $1.41^\circ$. LiDAR thus enters clip selection and depth evaluation. Gaussian training uses the image observations and SfSM geometry.

Figure~\ref{fig:supp_dycheck_poses} pairs the GT and SfSM poses for all 75 images. For each clip, we express both trajectories in their respective first-camera frames and fit one positive scale to the SfSM center displacements by least squares. The maximum separation of the GT centers ranges from $4.61$\,mm on apple to $18.04$\,mm on backpack. The rotation angles follow similar profiles on backpack and block. On mochi-high-five the paths diverge, with a center RMSE of $2.08$\,mm after alignment.

\clearpage
\subsection{Why small disparity makes depth uncertain}
\input{sections/baseline_sensitivity_table}
\input{sections/inverse_depth_sensitivity_table}

Consider two rectified cameras with focal length $f$ and lateral separation $B$. The disparity of a point at depth $Z$ is $d=fB/Z$. Writing $\widehat Z$ for the depth obtained after a disparity perturbation $\delta d$, we have
\begin{equation}
 d=\frac{fB}{Z},\qquad
 \widehat Z=\frac{fB}{d+\delta d},\qquad
 \frac{\widehat Z-Z}{Z}
 =-\frac{\delta d/d}{1+\delta d/d}.
 \label{eq:supp_depth_exact}
\end{equation}
The same subpixel error accounts for a larger fraction of a small disparity. Holding $f$ and $B$ fixed and taking $|\delta d|\ll d$ gives
\begin{equation}
\begin{aligned}
 \delta Z&\simeq-\frac{Z^2}{fB}\delta d,\qquad
 \sigma_Z^2\simeq\frac{Z^4}{f^2B^2}\sigma_d^2,\\
 \frac{\sigma_Z}{Z}&\simeq\frac{\sigma_d}{d}
 =\frac{\sigma_d Z}{fB}.
\end{aligned}
\label{eq:supp_depth_error}
\end{equation}
For the same disparity noise, a more distant point or a shorter baseline increases the relative depth uncertainty (Table~\ref{tab:supp_sensitivity_example}). The entries use a local approximation. Equation~\eqref{eq:supp_depth_exact} gives the asymmetric response to larger disparity errors.

\newpage
Inverse depth $\rho=1/Z=d/(fB)$ is linear in disparity~\cite{civera2008inversedepth}, so
\begin{equation}
 \sigma_\rho=\frac{\sigma_d}{fB},\qquad
 \frac{\sigma_\rho}{\rho}=\frac{\sigma_d}{d}
 \simeq\frac{\sigma_Z}{Z},\qquad
 \sigma_Z\simeq Z^2\sigma_\rho.
 \label{eq:supp_inverse_depth_error}
\end{equation}
At a fixed baseline, $\sigma_\rho$ is constant under this noise model. As a point moves farther away, $\rho=1/Z$ decreases, so the same absolute uncertainty accounts for a larger fraction of its inverse depth.

For example, in the $5$\,mm row of Table~\ref{tab:supp_inverse_depth_sensitivity}, doubling $Z$ from $1$ to $2$\,m halves $\rho$ from $1$ to $0.5$\,m$^{-1}$. The absolute standard deviation stays at $0.03077$\,m$^{-1}$, and the relative standard deviation doubles from $3.08\%$ to $6.15\%$. These percentages follow exactly from the linear inverse-depth model. For depth $Z$, Eq.~\eqref{eq:supp_depth_error} gives the same percentages to first order.

\clearpage
\subsection{Sparse feature depths against LiDAR}
\label{app:phone_depth}
\begin{figure*}[!b]\centering
\includegraphics[width=.83\textwidth]{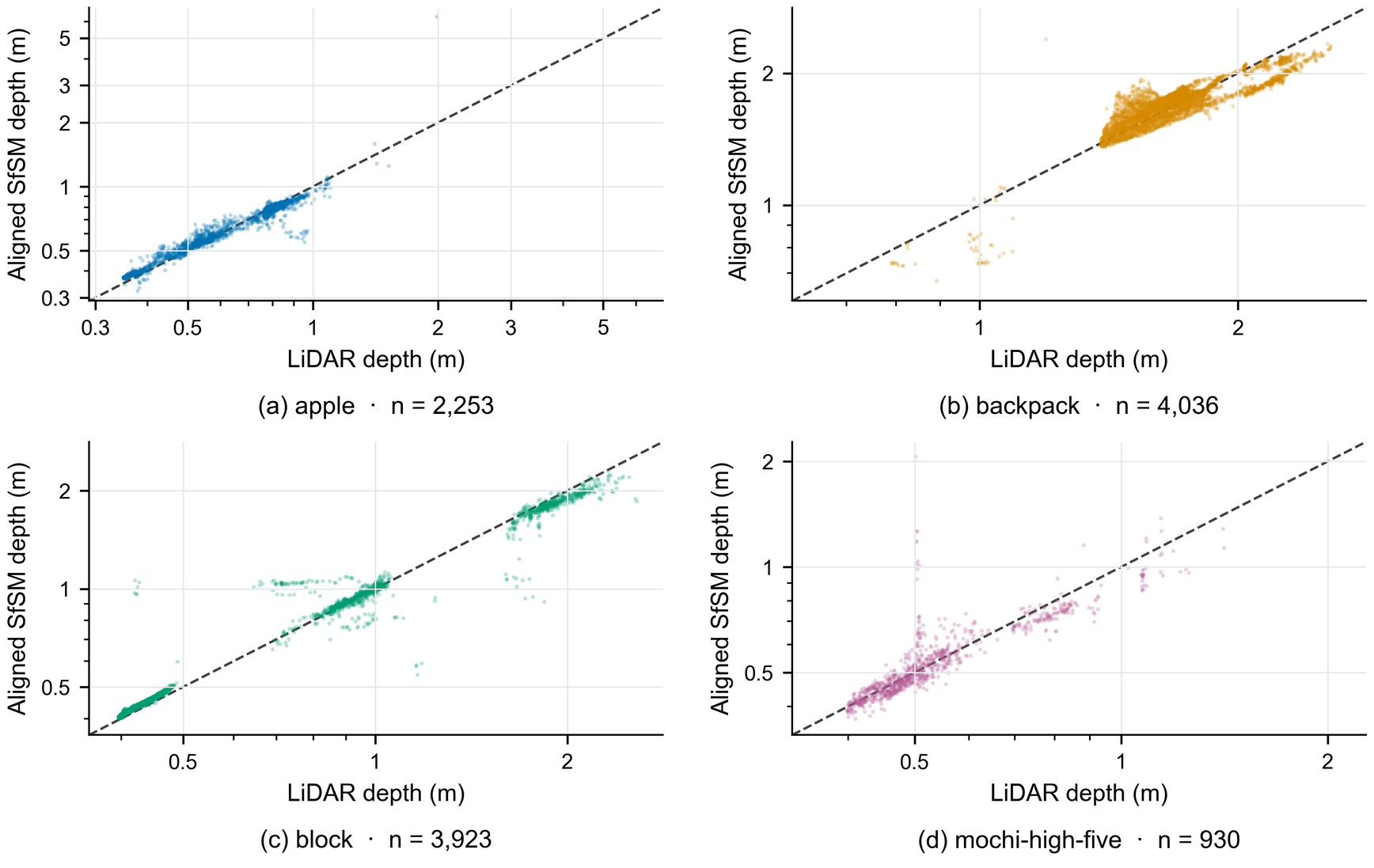}
\caption{\textbf{LiDAR depth and aligned SfSM estimates.} Every valid pair is shown. Both axes are logarithmic, in meters, with matching limits within each panel. Dashed lines indicate agreement.}
\label{fig:supp_lidar_depth_agreement}
\end{figure*}

How closely does a well-reprojected feature agree with its LiDAR depth? For Table~\ref{tab:feature_lidar_uncertainty}, we take the first frame of each retained clip and pair each SfSM feature depth $\widehat Z_j$ with the nearest-pixel LiDAR depth $D_j$. Of 11,778 features, 11,142 have positive finite depths, LiDAR in $(0,2.999)$\,m, and a valid exported depth standard deviation. We use all these pairs, with no confidence or static-region threshold.

SfSM recovers depth up to scale. Following the sparse-feature comparison in Table~\ref{tab:feature_lidar_uncertainty}, we match the average SfSM depth to the average LiDAR depth over the paired features in each scene. We then compute the signed discrepancy $e_j$, its scene-normalized value $r_j$, and the pointwise absolute relative error $E_j$ as follows.
\begin{equation}
\begin{aligned}
 a_s&=\frac{\operatorname{avg}_jD_j}{\operatorname{avg}_j\widehat Z_j},\qquad
 e_j=a_s\widehat Z_j-D_j,\\
 r_j&=\frac{e_j}{\operatorname{avg}_jD_j},\qquad
 E_j=\frac{|e_j|}{D_j}.
\end{aligned}
\label{eq:supp_lidar_residual}
\end{equation}
This alignment sets the average signed discrepancy to zero while preserving differences between individual features. Section~\ref{sec:iphone_depth} evaluates a different quantity, the rendered depth on common static pixels, using one median-scale alignment per scene.

\begin{table}[H]\centering\footnotesize
\caption{\textbf{Sparse-depth errors after scale alignment.} nRMSE uses the scene's average LiDAR depth. P90 is the 90th percentile of $E_j$ within each scene. The last column weights the four scenes equally.}
\label{tab:supp_feature_lidar_errors}
\setlength{\tabcolsep}{1.2pt}
\begin{tabular*}{\linewidth}{@{\extracolsep{\fill}}lrrrrr@{}}\toprule
 & apple & backpack & block & \shortstack{mochi-\\high-five} & \shortstack{Four-scene\\average}\\\midrule
Pairs & 2,253 & 4,036 & 3,923 & 930 & --\\
MAE (m) & 0.0291 & 0.0923 & 0.0329 & 0.0428 & --\\
AbsRel (\%) & 4.51 & 5.37 & 4.10 & 7.23 & 5.30\\
nRMSE (\%) & 16.01 & 7.26 & 10.55 & 17.46 & 12.82\\
P90 (\%) & 8.53 & 11.95 & 4.78 & 12.76 & 9.51\\
\shortstack[l]{Median reproj.\\(px)} & 0.16 & 0.10 & 0.07 & 0.24 & --\\
\bottomrule\end{tabular*}
\end{table}

For each scene we take the median of the tracks' median reprojection residuals. These scene values range from $0.07$ to $0.24$ pixels, yet depth nRMSE ranges from $7.26\%$ to $17.46\%$ (Table~\ref{tab:supp_feature_lidar_errors}). Figure~\ref{fig:supp_lidar_depth_agreement} shows which points deviate from agreement after scale alignment. A track can therefore fit its observed image location closely while its position along the ray remains uncertain. Section~\ref{sec:ray_supervision} therefore supervises the ray with a distribution around the estimated inverse depth.

\clearpage
\subsection{How errors vary across features}
\begin{figure*}[!b]\centering
\includegraphics[width=1.0\textwidth]{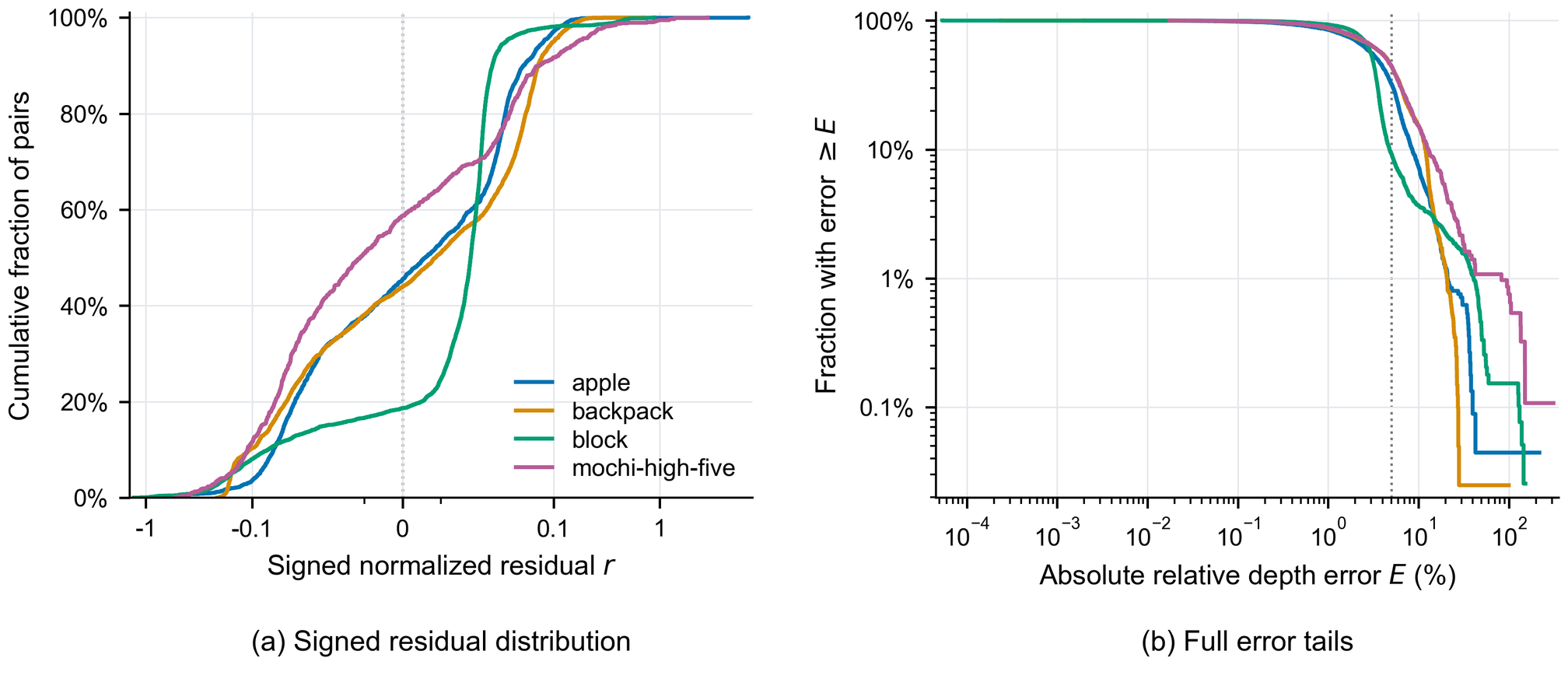}
\caption{\textbf{Full empirical depth-error distributions.} (a) Signed residual ECDFs use a symmetric-log horizontal axis, linear within $\pm0.02$. (b) Absolute-relative-error tails use logarithmic axes. The dotted line marks $5\%$ error. No residual outliers are removed.}
\label{fig:supp_lidar_error_distribution}
\end{figure*}

Depth errors vary across tracks within each scene in Table~\ref{tab:feature_lidar_uncertainty}. On mochi-high-five, the average absolute relative error is $7.23\%$, and the 90th percentile reaches $12.76\%$. Figure~\ref{fig:supp_lidar_error_distribution} shows this variation for all four clips. Panel~(a) shows whether depth is overestimated or underestimated. Panel~(b) gives the fraction of features exceeding any chosen relative error.

Table~\ref{tab:supp_feature_lidar_coverage} summarizes the spread in two ways. The empirical interval uses twice the sample standard deviation $s_s$ of its normalized residuals. It counts features with $|r_j|\leq2s_s$. The per-feature interval uses each feature's own depth standard deviation $\widehat\sigma_{Z,j}$, exported by SfSM. It counts features with $|e_j|\leq2a_s\widehat\sigma_{Z,j}$ after alignment to LiDAR.
Across all pairs, the scene-fitted intervals cover $94.79\%$ of the same observations used to estimate their widths. The per-feature intervals exported by SfSM cover $25.92\%$. Both columns describe depth discrepancies measured in the four reference frames. These can include camera and LiDAR sampling errors, particularly at depth boundaries.

In Sec.~\ref{sec:ray_supervision}, the normal distribution serves a different purpose. Its mean places supervision at the estimated inverse depth, and its width allows more movement along a less reliable ray. The stated $95.45\%$ is the mass inside two standard deviations of that target distribution. Table~\ref{tab:supp_feature_lidar_coverage} measures empirical coverage in depth space and does not estimate coverage of the training targets.

\begin{table}[H]\centering\small
\caption{\textbf{Depth interval coverage.} The four-scene average gives each scene equal weight. The last row gives each feature pair equal weight.}
\label{tab:supp_feature_lidar_coverage}
\setlength{\tabcolsep}{3pt}
\begin{tabular*}{\linewidth}{@{\extracolsep{\fill}}lrr@{}}\toprule
Scene & \shortstack{Empirical\\$\pm2s_s$ (\%)} & \shortstack{Exported\\$\pm2a_s\widehat\sigma_Z$ (\%)}\\\midrule
apple & 99.20 & 52.11\\
backpack & 90.96 & 18.53\\
block & 95.64 & 11.24\\
mochi-high-five & 97.10 & 56.45\\\midrule
Four-scene average & 95.72 & 34.58\\
All 11,142 feature pairs & 94.79 & 25.92\\
\bottomrule\end{tabular*}
\end{table}

%% file: sections/baseline_sensitivity_table.tex
\begin{table*}[!b]\centering
\caption{\textbf{First-order depth uncertainty.} We fix $f=1300$ pixels and $\sigma_d=0.2$ pixels. RSD is the relative standard deviation $\sigma_Z/Z$.}
\label{tab:supp_sensitivity_example}
\setlength{\tabcolsep}{3pt}\renewcommand{\arraystretch}{1.16}
\begin{tabular*}{\textwidth}{@{\extracolsep{\fill}}r*{4}{rr}@{}}\toprule
 & \multicolumn{2}{c}{$B=1$ mm} & \multicolumn{2}{c}{$B=5$ mm} & \multicolumn{2}{c}{$B=10$ mm} & \multicolumn{2}{c}{$B=50$ mm}\\
\cmidrule(lr){2-3}\cmidrule(lr){4-5}\cmidrule(lr){6-7}\cmidrule(lr){8-9}
$Z$ (m) & $\sigma_Z$ (m) & RSD (\%) & $\sigma_Z$ (m) & RSD (\%) & $\sigma_Z$ (m) & RSD (\%) & $\sigma_Z$ (m) & RSD (\%)\\\midrule
0.5 & 0.03846 & 7.69 & 0.00769 & 1.54 & 0.00385 & 0.77 & 0.00077 & 0.15\\
1 & 0.15385 & 15.38 & 0.03077 & 3.08 & 0.01538 & 1.54 & 0.00308 & 0.31\\
2 & 0.61538 & 30.77 & 0.12308 & 6.15 & 0.06154 & 3.08 & 0.01231 & 0.62\\
5 & 3.84615 & 76.92 & 0.76923 & 15.38 & 0.38462 & 7.69 & 0.07692 & 1.54\\
10 & 15.38462 & 153.85$^{\dagger}$ & 3.07692 & 30.77 & 1.53846 & 15.38 & 0.30769 & 3.08\\
\bottomrule\end{tabular*}
\par\vspace{3pt}\begin{minipage}{\textwidth}\footnotesize $^{\dagger}d\leq\sigma_d$: a negative one-standard-deviation disparity perturbation gives non-positive disparity.\end{minipage}
\end{table*}

%% file: sections/inverse_depth_sensitivity_table.tex
\begin{table*}[!b]\centering
\caption{\textbf{Inverse-depth uncertainty.} Each row fixes the baseline. The focal length and disparity noise are the same as in Table~\ref{tab:supp_sensitivity_example}.}
\label{tab:supp_inverse_depth_sensitivity}
\setlength{\tabcolsep}{3pt}\renewcommand{\arraystretch}{1.16}
\begin{tabular*}{\linewidth}{@{\extracolsep{\fill}}rr*{5}{r}@{}}\toprule
 & & \multicolumn{5}{c}{$\sigma_\rho/\rho$ (\%) at depth $Z$}\\
\cmidrule(lr){3-7}
$B$ (mm) & $\sigma_\rho$ ($\mathrm{m}^{-1}$) & 0.5 m & 1 m & 2 m & 5 m & 10 m\\\midrule
1 & 0.15385 & 7.69 & 15.38 & 30.77 & 76.92 & 153.85\\
5 & 0.03077 & 1.54 & 3.08 & 6.15 & 15.38 & 30.77\\
10 & 0.01538 & 0.77 & 1.54 & 3.08 & 7.69 & 15.38\\
50 & 0.00308 & 0.15 & 0.31 & 0.62 & 1.54 & 3.08\\
\bottomrule\end{tabular*}
\end{table*}

%% file: sections/comparison_intro.tex
\section{Complete Scene Comparisons}\label{app:aligned_qualitative}\label{app:full_qualitative}
\subsection{Small-Motion Benchmark}\label{app:smallmotion_visualizations}
Figure~\ref{fig:louvre_render} highlights foreground contours on Louvre and Plants. Here we show the same types of detail across all ten Small-Motion scenes~\cite{ha2016smallmotion}. RGB crops show local detail, and full-image RGB errors locate differences from the reference. Each depth map has its own color scale for comparing scene structure.

%% file: sections/scene_comparisons.tex
\definecolor{suppcyan}{HTML}{00A9D6}\definecolor{supporange}{HTML}{E45128}
\newsavebox{\SuppSceneBox}\newlength{\SuppSceneAvailable}
\newcommand{\PlaceSuppScene}{%
\setlength{\SuppSceneAvailable}{\dimexpr\textheight-\pagetotal-6pt\relax}%
\ifdim\SuppSceneAvailable<0.5\textheight\clearpage\setlength{\SuppSceneAvailable}{\dimexpr\textheight-6pt\relax}\fi%
\noindent\makebox[\textwidth][c]{\ifdim\dimexpr\ht\SuppSceneBox+\dp\SuppSceneBox\relax>\SuppSceneAvailable\resizebox*{!}{\SuppSceneAvailable}{\usebox{\SuppSceneBox}}\else\usebox{\SuppSceneBox}\fi}\par}
\par\smallskip
\pdfbookmark[3]{Bikes}{supp.scene.bikes}
\begin{lrbox}{\SuppSceneBox}\begin{minipage}{\textwidth}
\noindent{\large\bfseries Bikes}\hfill{\small\color{black!65}Small-Motion dataset}\par\vspace{3pt}
\noindent{\small\bfseries Rendered RGB}\par\vspace{2pt}
\noindent \begin{minipage}[t]{0.194\textwidth}\centering\fontsize{7.4}{8.5}\selectfont\strut Original RGB\par\vspace{1pt}\includegraphics[width=\linewidth,bb=0 0 1344 624,viewport=8 436 328 616,clip]{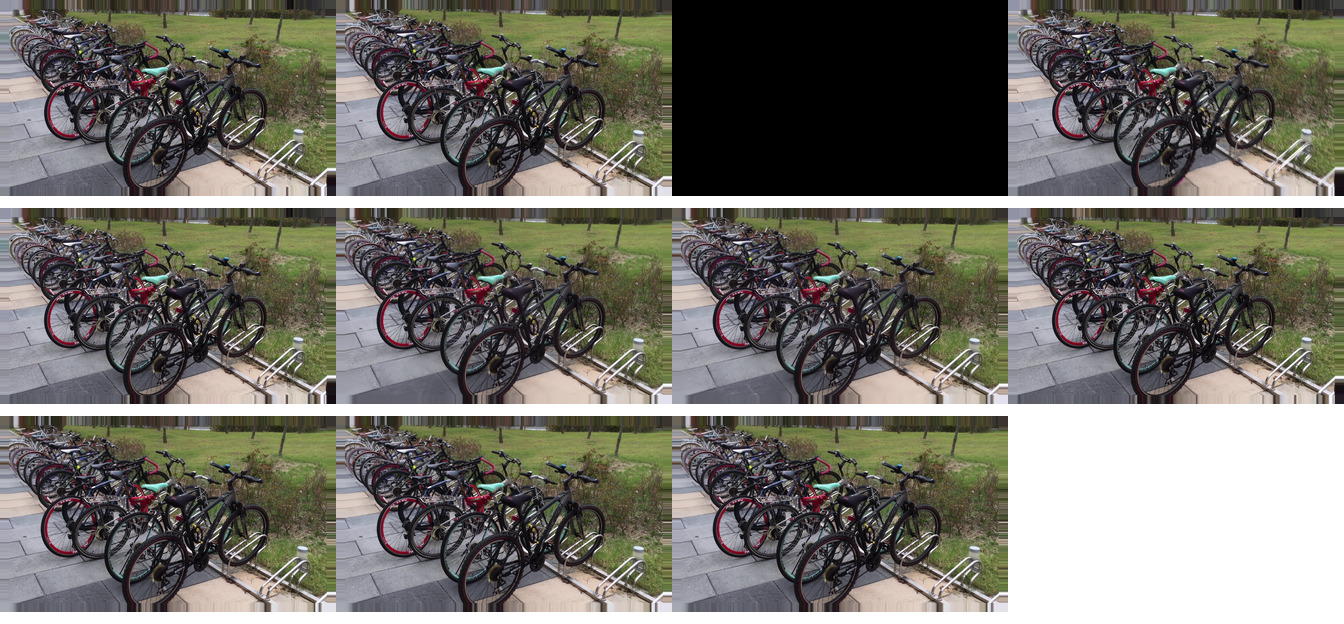}\par\vspace{2pt}\strut SfSM-undistorted RGB\par\vspace{1pt}\includegraphics[width=\linewidth,bb=0 0 1344 624,viewport=344 436 664 616,clip]{figures/small_motion/bikes/rgb.jpg}\end{minipage}\hfill\begin{minipage}[t]{0.194\textwidth}\centering\fontsize{7.4}{8.5}\selectfont\strut Vanilla original\par\vspace{1pt}\setlength{\unitlength}{\dimexpr\linewidth/1000\relax}\begin{picture}(1000,562.500)\put(0,0){\includegraphics[width=\linewidth,bb=0 0 1344 624,viewport=680 436 1000 616,clip]{figures/small_motion/bikes/rgb.jpg}}\put(572.917,208.333){\color{suppcyan}\linethickness{0.45pt}\framebox(145.833,145.833){} }\put(692.708,57.292){\color{supporange}\linethickness{0.45pt}\framebox(145.833,145.833){} }\end{picture}\par\vspace{2pt}\hbox to\linewidth{{\setlength{\fboxsep}{0pt}\color{suppcyan}\fbox{\includegraphics[width=0.478\linewidth,bb=0 0 960 1200,viewport=8 968 232 1192,clip]{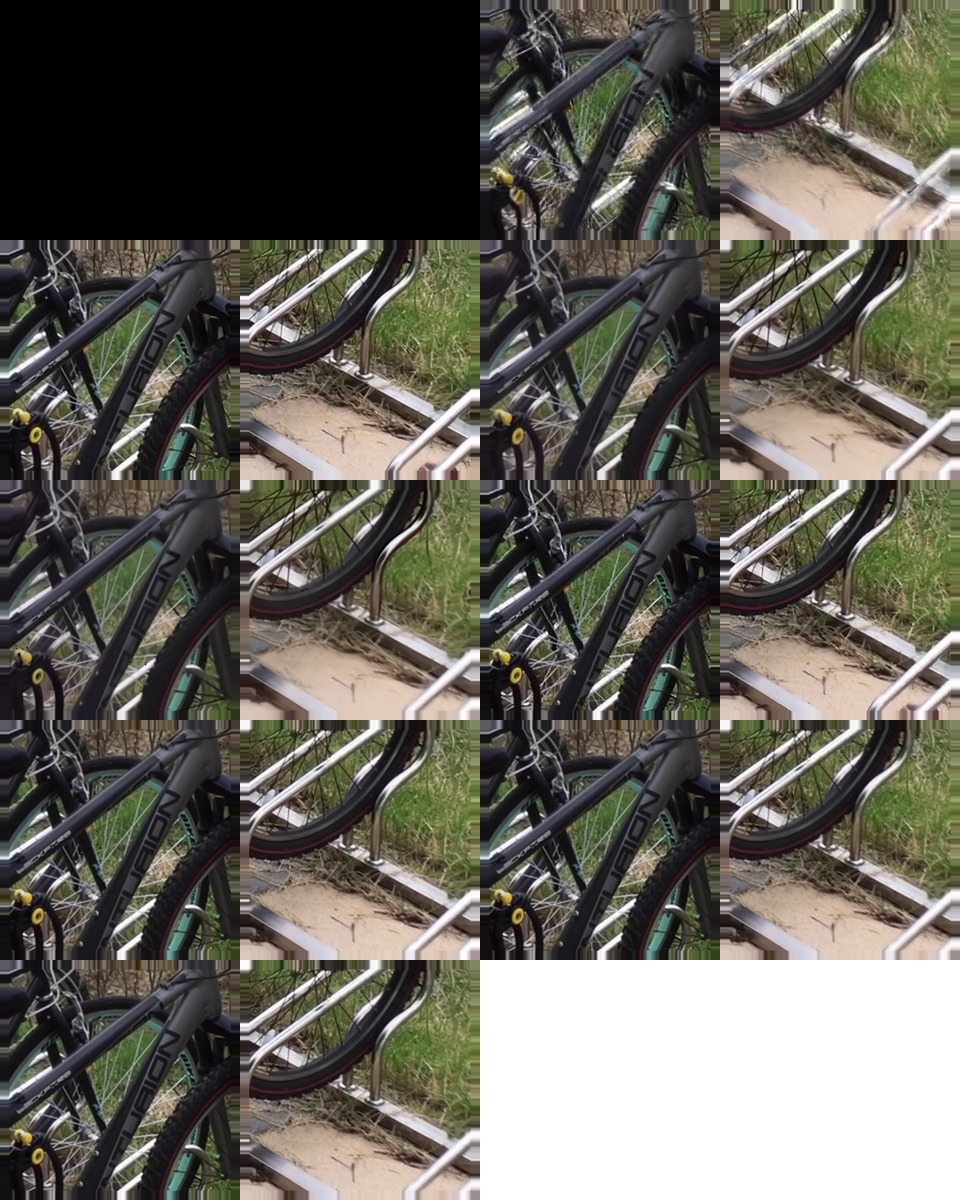}}}\hfill{\setlength{\fboxsep}{0pt}\color{supporange}\fbox{\includegraphics[width=0.478\linewidth,bb=0 0 960 1200,viewport=248 968 472 1192,clip]{figures/small_motion/bikes/rgb_details.jpg}}}}\end{minipage}\hfill\begin{minipage}[t]{0.194\textwidth}\centering\fontsize{7.4}{8.5}\selectfont\strut Small GS\par\vspace{1pt}\setlength{\unitlength}{\dimexpr\linewidth/1000\relax}\begin{picture}(1000,562.500)\put(0,0){\includegraphics[width=\linewidth,bb=0 0 1344 624,viewport=1016 436 1336 616,clip]{figures/small_motion/bikes/rgb.jpg}}\put(572.917,208.333){\color{suppcyan}\linethickness{0.45pt}\framebox(145.833,145.833){} }\put(692.708,57.292){\color{supporange}\linethickness{0.45pt}\framebox(145.833,145.833){} }\end{picture}\par\vspace{2pt}\hbox to\linewidth{{\setlength{\fboxsep}{0pt}\color{suppcyan}\fbox{\includegraphics[width=0.478\linewidth,bb=0 0 960 1200,viewport=488 968 712 1192,clip]{figures/small_motion/bikes/rgb_details.jpg}}}\hfill{\setlength{\fboxsep}{0pt}\color{supporange}\fbox{\includegraphics[width=0.478\linewidth,bb=0 0 960 1200,viewport=728 968 952 1192,clip]{figures/small_motion/bikes/rgb_details.jpg}}}}\end{minipage}\hfill\begin{minipage}[t]{0.194\textwidth}\centering\fontsize{7.4}{8.5}\selectfont\strut InstantSplat\par\vspace{1pt}\setlength{\unitlength}{\dimexpr\linewidth/1000\relax}\begin{picture}(1000,562.500)\put(0,0){\includegraphics[width=\linewidth,bb=0 0 1344 624,viewport=8 228 328 408,clip]{figures/small_motion/bikes/rgb.jpg}}\put(572.917,208.333){\color{suppcyan}\linethickness{0.45pt}\framebox(145.833,145.833){} }\put(692.708,57.292){\color{supporange}\linethickness{0.45pt}\framebox(145.833,145.833){} }\end{picture}\par\vspace{2pt}\hbox to\linewidth{{\setlength{\fboxsep}{0pt}\color{suppcyan}\fbox{\includegraphics[width=0.478\linewidth,bb=0 0 960 1200,viewport=8 728 232 952,clip]{figures/small_motion/bikes/rgb_details.jpg}}}\hfill{\setlength{\fboxsep}{0pt}\color{supporange}\fbox{\includegraphics[width=0.478\linewidth,bb=0 0 960 1200,viewport=248 728 472 952,clip]{figures/small_motion/bikes/rgb_details.jpg}}}}\end{minipage}\hfill\begin{minipage}[t]{0.194\textwidth}\centering\fontsize{7.4}{8.5}\selectfont\strut DDDrop\par\vspace{1pt}\setlength{\unitlength}{\dimexpr\linewidth/1000\relax}\begin{picture}(1000,562.500)\put(0,0){\includegraphics[width=\linewidth,bb=0 0 1344 624,viewport=344 228 664 408,clip]{figures/small_motion/bikes/rgb.jpg}}\put(572.917,208.333){\color{suppcyan}\linethickness{0.45pt}\framebox(145.833,145.833){} }\put(692.708,57.292){\color{supporange}\linethickness{0.45pt}\framebox(145.833,145.833){} }\end{picture}\par\vspace{2pt}\hbox to\linewidth{{\setlength{\fboxsep}{0pt}\color{suppcyan}\fbox{\includegraphics[width=0.478\linewidth,bb=0 0 960 1200,viewport=488 728 712 952,clip]{figures/small_motion/bikes/rgb_details.jpg}}}\hfill{\setlength{\fboxsep}{0pt}\color{supporange}\fbox{\includegraphics[width=0.478\linewidth,bb=0 0 960 1200,viewport=728 728 952 952,clip]{figures/small_motion/bikes/rgb_details.jpg}}}}\end{minipage}\par\vspace{4pt}
\noindent \begin{minipage}[t]{0.194\textwidth}\centering\fontsize{7.4}{8.5}\selectfont\strut DDGS\par\vspace{1pt}\setlength{\unitlength}{\dimexpr\linewidth/1000\relax}\begin{picture}(1000,562.500)\put(0,0){\includegraphics[width=\linewidth,bb=0 0 1344 624,viewport=680 228 1000 408,clip]{figures/small_motion/bikes/rgb.jpg}}\put(572.917,208.333){\color{suppcyan}\linethickness{0.45pt}\framebox(145.833,145.833){} }\put(692.708,57.292){\color{supporange}\linethickness{0.45pt}\framebox(145.833,145.833){} }\end{picture}\par\vspace{2pt}\hbox to\linewidth{{\setlength{\fboxsep}{0pt}\color{suppcyan}\fbox{\includegraphics[width=0.478\linewidth,bb=0 0 960 1200,viewport=8 488 232 712,clip]{figures/small_motion/bikes/rgb_details.jpg}}}\hfill{\setlength{\fboxsep}{0pt}\color{supporange}\fbox{\includegraphics[width=0.478\linewidth,bb=0 0 960 1200,viewport=248 488 472 712,clip]{figures/small_motion/bikes/rgb_details.jpg}}}}\end{minipage}\hfill\begin{minipage}[t]{0.194\textwidth}\centering\fontsize{7.4}{8.5}\selectfont\strut G4Splat\par\vspace{1pt}\setlength{\unitlength}{\dimexpr\linewidth/1000\relax}\begin{picture}(1000,562.500)\put(0,0){\includegraphics[width=\linewidth,bb=0 0 1344 624,viewport=1016 228 1336 408,clip]{figures/small_motion/bikes/rgb.jpg}}\put(572.917,208.333){\color{suppcyan}\linethickness{0.45pt}\framebox(145.833,145.833){} }\put(692.708,57.292){\color{supporange}\linethickness{0.45pt}\framebox(145.833,145.833){} }\end{picture}\par\vspace{2pt}\hbox to\linewidth{{\setlength{\fboxsep}{0pt}\color{suppcyan}\fbox{\includegraphics[width=0.478\linewidth,bb=0 0 960 1200,viewport=488 488 712 712,clip]{figures/small_motion/bikes/rgb_details.jpg}}}\hfill{\setlength{\fboxsep}{0pt}\color{supporange}\fbox{\includegraphics[width=0.478\linewidth,bb=0 0 960 1200,viewport=728 488 952 712,clip]{figures/small_motion/bikes/rgb_details.jpg}}}}\end{minipage}\hfill\begin{minipage}[t]{0.194\textwidth}\centering\fontsize{7.4}{8.5}\selectfont\strut Vanilla$^{\dagger}$\par\vspace{1pt}\setlength{\unitlength}{\dimexpr\linewidth/1000\relax}\begin{picture}(1000,562.500)\put(0,0){\includegraphics[width=\linewidth,bb=0 0 1344 624,viewport=8 20 328 200,clip]{figures/small_motion/bikes/rgb.jpg}}\put(572.917,208.333){\color{suppcyan}\linethickness{0.45pt}\framebox(145.833,145.833){} }\put(692.708,57.292){\color{supporange}\linethickness{0.45pt}\framebox(145.833,145.833){} }\end{picture}\par\vspace{2pt}\hbox to\linewidth{{\setlength{\fboxsep}{0pt}\color{suppcyan}\fbox{\includegraphics[width=0.478\linewidth,bb=0 0 960 1200,viewport=8 248 232 472,clip]{figures/small_motion/bikes/rgb_details.jpg}}}\hfill{\setlength{\fboxsep}{0pt}\color{supporange}\fbox{\includegraphics[width=0.478\linewidth,bb=0 0 960 1200,viewport=248 248 472 472,clip]{figures/small_motion/bikes/rgb_details.jpg}}}}\end{minipage}\hfill\begin{minipage}[t]{0.194\textwidth}\centering\fontsize{7.4}{8.5}\selectfont\strut Mip-Splatting\par\vspace{1pt}\setlength{\unitlength}{\dimexpr\linewidth/1000\relax}\begin{picture}(1000,562.500)\put(0,0){\includegraphics[width=\linewidth,bb=0 0 1344 624,viewport=344 20 664 200,clip]{figures/small_motion/bikes/rgb.jpg}}\put(572.917,208.333){\color{suppcyan}\linethickness{0.45pt}\framebox(145.833,145.833){} }\put(692.708,57.292){\color{supporange}\linethickness{0.45pt}\framebox(145.833,145.833){} }\end{picture}\par\vspace{2pt}\hbox to\linewidth{{\setlength{\fboxsep}{0pt}\color{suppcyan}\fbox{\includegraphics[width=0.478\linewidth,bb=0 0 960 1200,viewport=488 248 712 472,clip]{figures/small_motion/bikes/rgb_details.jpg}}}\hfill{\setlength{\fboxsep}{0pt}\color{supporange}\fbox{\includegraphics[width=0.478\linewidth,bb=0 0 960 1200,viewport=728 248 952 472,clip]{figures/small_motion/bikes/rgb_details.jpg}}}}\end{minipage}\hfill\begin{minipage}[t]{0.194\textwidth}\centering\fontsize{7.4}{8.5}\selectfont\strut \textbf{NBGS}\par\vspace{1pt}\setlength{\unitlength}{\dimexpr\linewidth/1000\relax}\begin{picture}(1000,562.500)\put(0,0){\includegraphics[width=\linewidth,bb=0 0 1344 624,viewport=680 20 1000 200,clip]{figures/small_motion/bikes/rgb.jpg}}\put(572.917,208.333){\color{suppcyan}\linethickness{0.45pt}\framebox(145.833,145.833){} }\put(692.708,57.292){\color{supporange}\linethickness{0.45pt}\framebox(145.833,145.833){} }\end{picture}\par\vspace{2pt}\hbox to\linewidth{{\setlength{\fboxsep}{0pt}\color{suppcyan}\fbox{\includegraphics[width=0.478\linewidth,bb=0 0 960 1200,viewport=8 8 232 232,clip]{figures/small_motion/bikes/rgb_details.jpg}}}\hfill{\setlength{\fboxsep}{0pt}\color{supporange}\fbox{\includegraphics[width=0.478\linewidth,bb=0 0 960 1200,viewport=248 8 472 232,clip]{figures/small_motion/bikes/rgb_details.jpg}}}}\end{minipage}\par\vspace{4pt}
\noindent{\small\bfseries Relative rendered depth}\par\vspace{2pt}
\noindent \begin{minipage}[t]{0.194\textwidth}\centering\fontsize{7.4}{8.5}\selectfont\strut Relative depth\par\vspace{4pt}\raggedright Per-method p02--p98. Black: invalid. No depth ground truth or metric units.\par\vspace{5pt}\includegraphics[width=\linewidth,height=5pt]{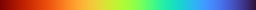}\par near\hfill far\end{minipage}\hfill\begin{minipage}[t]{0.194\textwidth}\centering\fontsize{7.4}{8.5}\selectfont\strut Vanilla original\par\vspace{1pt}\includegraphics[width=\linewidth,bb=0 0 1344 624,viewport=8 436 328 616,clip]{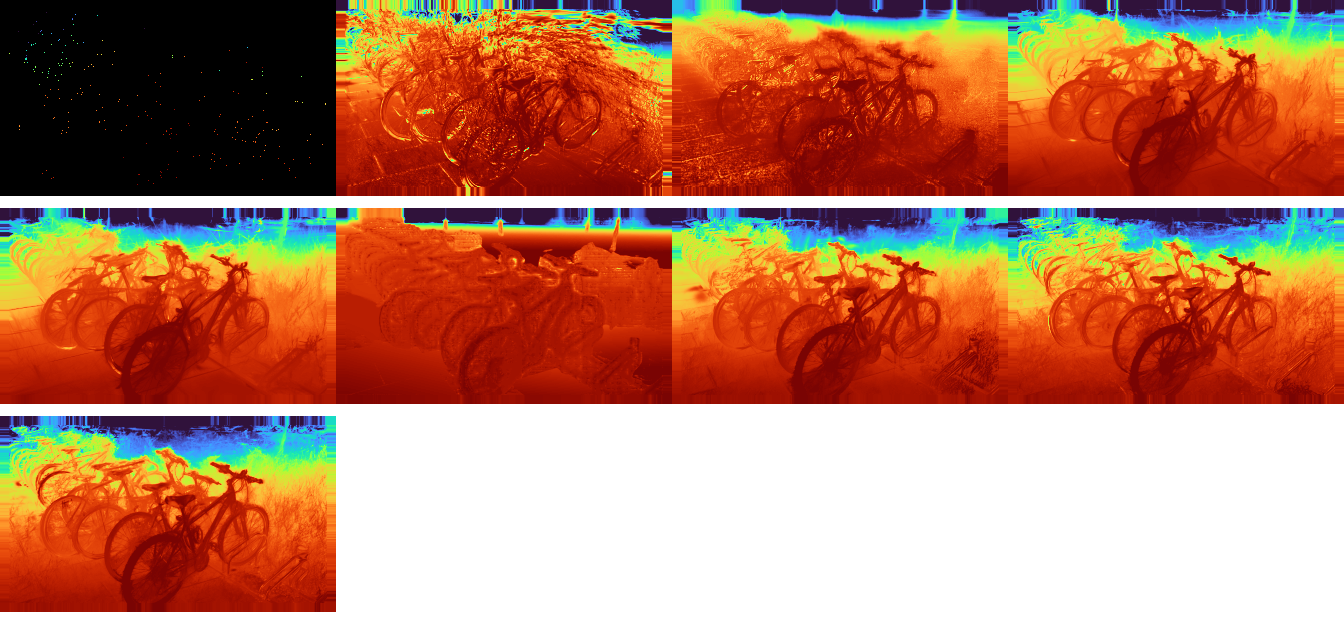}\end{minipage}\hfill\begin{minipage}[t]{0.194\textwidth}\centering\fontsize{7.4}{8.5}\selectfont\strut Small GS\par\vspace{1pt}\includegraphics[width=\linewidth,bb=0 0 1344 624,viewport=344 436 664 616,clip]{figures/small_motion/bikes/depth.png}\end{minipage}\hfill\begin{minipage}[t]{0.194\textwidth}\centering\fontsize{7.4}{8.5}\selectfont\strut InstantSplat\par\vspace{1pt}\includegraphics[width=\linewidth,bb=0 0 1344 624,viewport=680 436 1000 616,clip]{figures/small_motion/bikes/depth.png}\end{minipage}\hfill\begin{minipage}[t]{0.194\textwidth}\centering\fontsize{7.4}{8.5}\selectfont\strut DDDrop\par\vspace{1pt}\includegraphics[width=\linewidth,bb=0 0 1344 624,viewport=1016 436 1336 616,clip]{figures/small_motion/bikes/depth.png}\end{minipage}\par\vspace{4pt}
\noindent \begin{minipage}[t]{0.194\textwidth}\centering\fontsize{7.4}{8.5}\selectfont\strut DDGS\par\vspace{1pt}\includegraphics[width=\linewidth,bb=0 0 1344 624,viewport=8 228 328 408,clip]{figures/small_motion/bikes/depth.png}\end{minipage}\hfill\begin{minipage}[t]{0.194\textwidth}\centering\fontsize{7.4}{8.5}\selectfont\strut G4Splat\par\vspace{1pt}\includegraphics[width=\linewidth,bb=0 0 1344 624,viewport=344 228 664 408,clip]{figures/small_motion/bikes/depth.png}\end{minipage}\hfill\begin{minipage}[t]{0.194\textwidth}\centering\fontsize{7.4}{8.5}\selectfont\strut Vanilla$^{\dagger}$\par\vspace{1pt}\includegraphics[width=\linewidth,bb=0 0 1344 624,viewport=680 228 1000 408,clip]{figures/small_motion/bikes/depth.png}\end{minipage}\hfill\begin{minipage}[t]{0.194\textwidth}\centering\fontsize{7.4}{8.5}\selectfont\strut Mip-Splatting\par\vspace{1pt}\includegraphics[width=\linewidth,bb=0 0 1344 624,viewport=1016 228 1336 408,clip]{figures/small_motion/bikes/depth.png}\end{minipage}\hfill\begin{minipage}[t]{0.194\textwidth}\centering\fontsize{7.4}{8.5}\selectfont\strut \textbf{NBGS}\par\vspace{1pt}\includegraphics[width=\linewidth,bb=0 0 1344 624,viewport=8 20 328 200,clip]{figures/small_motion/bikes/depth.png}\end{minipage}\par\vspace{4pt}
\noindent{\small\bfseries Absolute RGB error}\par\vspace{2pt}
\noindent \begin{minipage}[t]{0.194\textwidth}\centering\fontsize{7.4}{8.5}\selectfont\strut RGB error, $[0,3]$\par\vspace{4pt}\raggedright Channel-summed absolute error. Method-matched RGB reference.\par\vspace{5pt}\includegraphics[width=\linewidth,height=5pt]{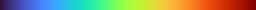}\par 0\hfill 3\end{minipage}\hfill\begin{minipage}[t]{0.194\textwidth}\centering\fontsize{7.4}{8.5}\selectfont\strut Vanilla original\par\vspace{1pt}\includegraphics[width=\linewidth,bb=0 0 1344 624,viewport=8 436 328 616,clip]{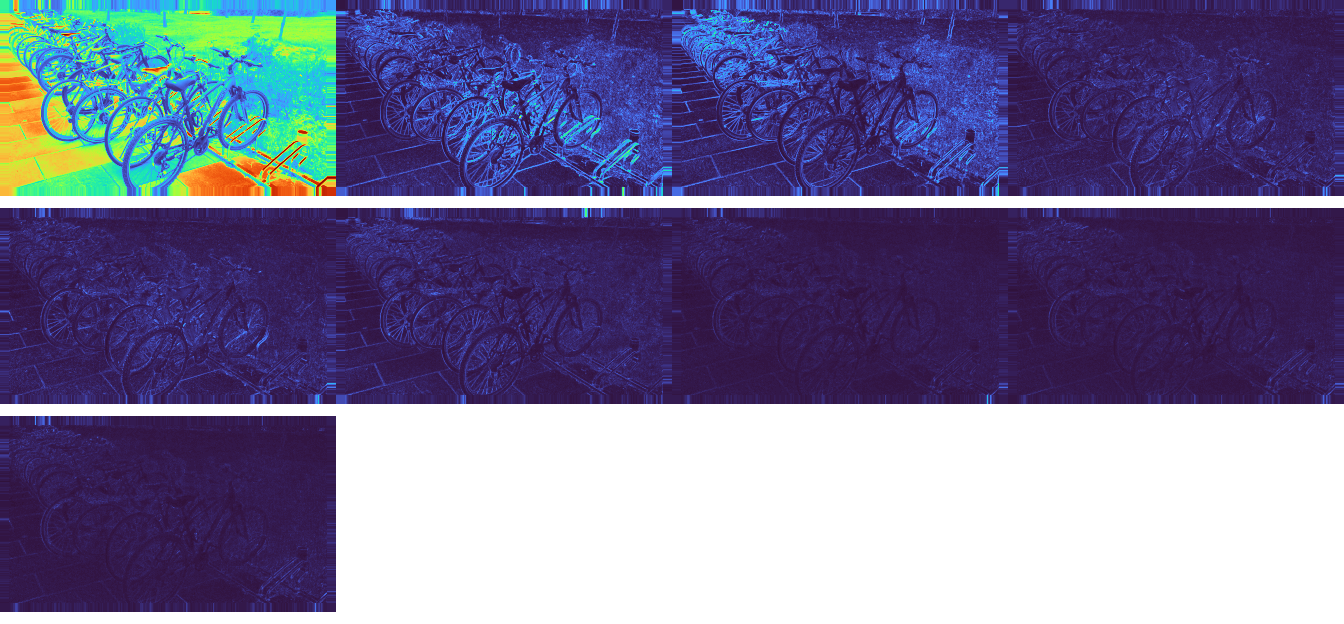}\end{minipage}\hfill\begin{minipage}[t]{0.194\textwidth}\centering\fontsize{7.4}{8.5}\selectfont\strut Small GS\par\vspace{1pt}\includegraphics[width=\linewidth,bb=0 0 1344 624,viewport=344 436 664 616,clip]{figures/small_motion/bikes/rgb_error.png}\end{minipage}\hfill\begin{minipage}[t]{0.194\textwidth}\centering\fontsize{7.4}{8.5}\selectfont\strut InstantSplat\par\vspace{1pt}\includegraphics[width=\linewidth,bb=0 0 1344 624,viewport=680 436 1000 616,clip]{figures/small_motion/bikes/rgb_error.png}\end{minipage}\hfill\begin{minipage}[t]{0.194\textwidth}\centering\fontsize{7.4}{8.5}\selectfont\strut DDDrop\par\vspace{1pt}\includegraphics[width=\linewidth,bb=0 0 1344 624,viewport=1016 436 1336 616,clip]{figures/small_motion/bikes/rgb_error.png}\end{minipage}\par\vspace{4pt}
\noindent \begin{minipage}[t]{0.194\textwidth}\centering\fontsize{7.4}{8.5}\selectfont\strut DDGS\par\vspace{1pt}\includegraphics[width=\linewidth,bb=0 0 1344 624,viewport=8 228 328 408,clip]{figures/small_motion/bikes/rgb_error.png}\end{minipage}\hfill\begin{minipage}[t]{0.194\textwidth}\centering\fontsize{7.4}{8.5}\selectfont\strut G4Splat\par\vspace{1pt}\includegraphics[width=\linewidth,bb=0 0 1344 624,viewport=344 228 664 408,clip]{figures/small_motion/bikes/rgb_error.png}\end{minipage}\hfill\begin{minipage}[t]{0.194\textwidth}\centering\fontsize{7.4}{8.5}\selectfont\strut Vanilla$^{\dagger}$\par\vspace{1pt}\includegraphics[width=\linewidth,bb=0 0 1344 624,viewport=680 228 1000 408,clip]{figures/small_motion/bikes/rgb_error.png}\end{minipage}\hfill\begin{minipage}[t]{0.194\textwidth}\centering\fontsize{7.4}{8.5}\selectfont\strut Mip-Splatting\par\vspace{1pt}\includegraphics[width=\linewidth,bb=0 0 1344 624,viewport=1016 228 1336 408,clip]{figures/small_motion/bikes/rgb_error.png}\end{minipage}\hfill\begin{minipage}[t]{0.194\textwidth}\centering\fontsize{7.4}{8.5}\selectfont\strut \textbf{NBGS}\par\vspace{1pt}\includegraphics[width=\linewidth,bb=0 0 1344 624,viewport=8 20 328 200,clip]{figures/small_motion/bikes/rgb_error.png}\end{minipage}\par\vspace{4pt}
\end{minipage}\end{lrbox}
\PlaceSuppScene
\clearpage
\pdfbookmark[3]{Dinos}{supp.scene.dinos}
\begin{lrbox}{\SuppSceneBox}\begin{minipage}{\textwidth}
\noindent{\large\bfseries Dinos}\hfill{\small\color{black!65}Small-Motion dataset}\par\vspace{3pt}
\noindent{\small\bfseries Rendered RGB}\par\vspace{2pt}
\noindent \begin{minipage}[t]{0.194\textwidth}\centering\fontsize{7.4}{8.5}\selectfont\strut Original RGB\par\vspace{1pt}\includegraphics[width=\linewidth,bb=0 0 1344 768,viewport=8 520 328 760,clip]{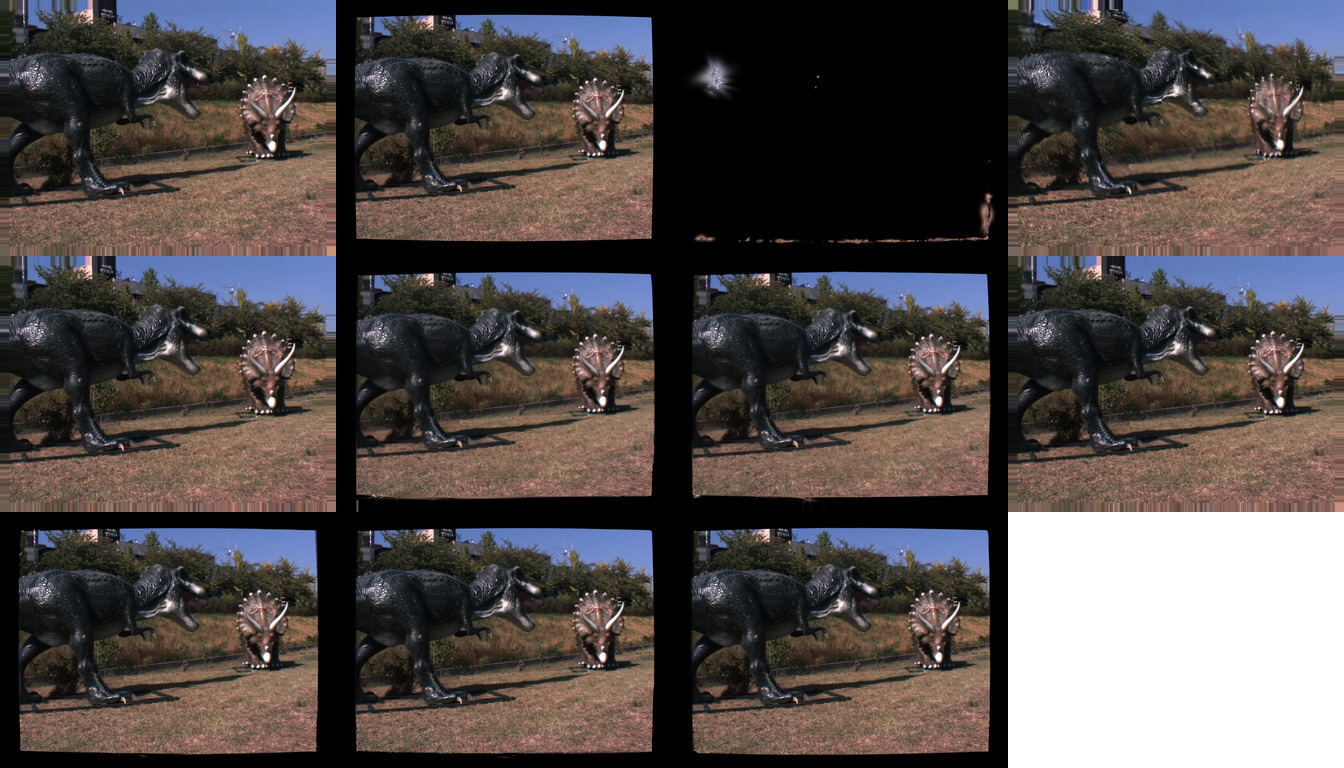}\par\vspace{2pt}\strut SfSM-undistorted RGB\par\vspace{1pt}\includegraphics[width=\linewidth,bb=0 0 1344 768,viewport=344 520 664 760,clip]{figures/small_motion/dinos/rgb.jpg}\end{minipage}\hfill\begin{minipage}[t]{0.194\textwidth}\centering\fontsize{7.4}{8.5}\selectfont\strut Vanilla original\par\vspace{1pt}\setlength{\unitlength}{\dimexpr\linewidth/1000\relax}\begin{picture}(1000,750.000)\put(0,0){\includegraphics[width=\linewidth,bb=0 0 1344 768,viewport=680 520 1000 760,clip]{figures/small_motion/dinos/rgb.jpg}}\put(351.562,343.750){\color{suppcyan}\linethickness{0.45pt}\framebox(218.750,218.750){} }\put(656.250,328.125){\color{supporange}\linethickness{0.45pt}\framebox(187.500,187.500){} }\end{picture}\par\vspace{2pt}\hbox to\linewidth{{\setlength{\fboxsep}{0pt}\color{suppcyan}\fbox{\includegraphics[width=0.478\linewidth,bb=0 0 960 1200,viewport=8 968 232 1192,clip]{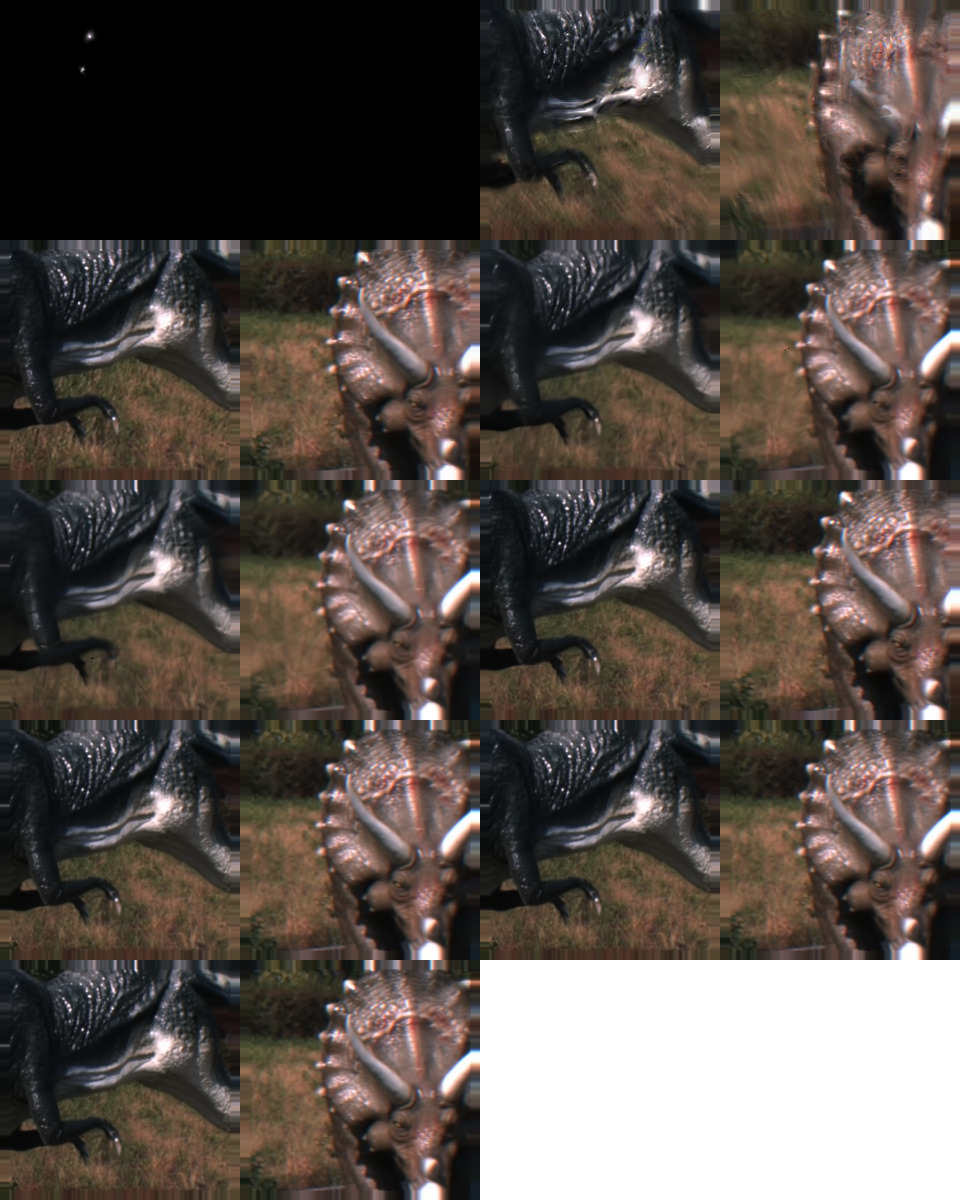}}}\hfill{\setlength{\fboxsep}{0pt}\color{supporange}\fbox{\includegraphics[width=0.478\linewidth,bb=0 0 960 1200,viewport=248 968 472 1192,clip]{figures/small_motion/dinos/rgb_details.jpg}}}}\end{minipage}\hfill\begin{minipage}[t]{0.194\textwidth}\centering\fontsize{7.4}{8.5}\selectfont\strut Small GS\par\vspace{1pt}\setlength{\unitlength}{\dimexpr\linewidth/1000\relax}\begin{picture}(1000,750.000)\put(0,0){\includegraphics[width=\linewidth,bb=0 0 1344 768,viewport=1016 520 1336 760,clip]{figures/small_motion/dinos/rgb.jpg}}\put(351.562,343.750){\color{suppcyan}\linethickness{0.45pt}\framebox(218.750,218.750){} }\put(656.250,328.125){\color{supporange}\linethickness{0.45pt}\framebox(187.500,187.500){} }\end{picture}\par\vspace{2pt}\hbox to\linewidth{{\setlength{\fboxsep}{0pt}\color{suppcyan}\fbox{\includegraphics[width=0.478\linewidth,bb=0 0 960 1200,viewport=488 968 712 1192,clip]{figures/small_motion/dinos/rgb_details.jpg}}}\hfill{\setlength{\fboxsep}{0pt}\color{supporange}\fbox{\includegraphics[width=0.478\linewidth,bb=0 0 960 1200,viewport=728 968 952 1192,clip]{figures/small_motion/dinos/rgb_details.jpg}}}}\end{minipage}\hfill\begin{minipage}[t]{0.194\textwidth}\centering\fontsize{7.4}{8.5}\selectfont\strut InstantSplat\par\vspace{1pt}\setlength{\unitlength}{\dimexpr\linewidth/1000\relax}\begin{picture}(1000,750.000)\put(0,0){\includegraphics[width=\linewidth,bb=0 0 1344 768,viewport=8 264 328 504,clip]{figures/small_motion/dinos/rgb.jpg}}\put(351.562,343.750){\color{suppcyan}\linethickness{0.45pt}\framebox(218.750,218.750){} }\put(656.250,328.125){\color{supporange}\linethickness{0.45pt}\framebox(187.500,187.500){} }\end{picture}\par\vspace{2pt}\hbox to\linewidth{{\setlength{\fboxsep}{0pt}\color{suppcyan}\fbox{\includegraphics[width=0.478\linewidth,bb=0 0 960 1200,viewport=8 728 232 952,clip]{figures/small_motion/dinos/rgb_details.jpg}}}\hfill{\setlength{\fboxsep}{0pt}\color{supporange}\fbox{\includegraphics[width=0.478\linewidth,bb=0 0 960 1200,viewport=248 728 472 952,clip]{figures/small_motion/dinos/rgb_details.jpg}}}}\end{minipage}\hfill\begin{minipage}[t]{0.194\textwidth}\centering\fontsize{7.4}{8.5}\selectfont\strut DDDrop\par\vspace{1pt}\setlength{\unitlength}{\dimexpr\linewidth/1000\relax}\begin{picture}(1000,750.000)\put(0,0){\includegraphics[width=\linewidth,bb=0 0 1344 768,viewport=344 264 664 504,clip]{figures/small_motion/dinos/rgb.jpg}}\put(351.562,343.750){\color{suppcyan}\linethickness{0.45pt}\framebox(218.750,218.750){} }\put(656.250,328.125){\color{supporange}\linethickness{0.45pt}\framebox(187.500,187.500){} }\end{picture}\par\vspace{2pt}\hbox to\linewidth{{\setlength{\fboxsep}{0pt}\color{suppcyan}\fbox{\includegraphics[width=0.478\linewidth,bb=0 0 960 1200,viewport=488 728 712 952,clip]{figures/small_motion/dinos/rgb_details.jpg}}}\hfill{\setlength{\fboxsep}{0pt}\color{supporange}\fbox{\includegraphics[width=0.478\linewidth,bb=0 0 960 1200,viewport=728 728 952 952,clip]{figures/small_motion/dinos/rgb_details.jpg}}}}\end{minipage}\par\vspace{4pt}
\noindent \begin{minipage}[t]{0.194\textwidth}\centering\fontsize{7.4}{8.5}\selectfont\strut DDGS\par\vspace{1pt}\setlength{\unitlength}{\dimexpr\linewidth/1000\relax}\begin{picture}(1000,750.000)\put(0,0){\includegraphics[width=\linewidth,bb=0 0 1344 768,viewport=680 264 1000 504,clip]{figures/small_motion/dinos/rgb.jpg}}\put(351.562,343.750){\color{suppcyan}\linethickness{0.45pt}\framebox(218.750,218.750){} }\put(656.250,328.125){\color{supporange}\linethickness{0.45pt}\framebox(187.500,187.500){} }\end{picture}\par\vspace{2pt}\hbox to\linewidth{{\setlength{\fboxsep}{0pt}\color{suppcyan}\fbox{\includegraphics[width=0.478\linewidth,bb=0 0 960 1200,viewport=8 488 232 712,clip]{figures/small_motion/dinos/rgb_details.jpg}}}\hfill{\setlength{\fboxsep}{0pt}\color{supporange}\fbox{\includegraphics[width=0.478\linewidth,bb=0 0 960 1200,viewport=248 488 472 712,clip]{figures/small_motion/dinos/rgb_details.jpg}}}}\end{minipage}\hfill\begin{minipage}[t]{0.194\textwidth}\centering\fontsize{7.4}{8.5}\selectfont\strut G4Splat\par\vspace{1pt}\setlength{\unitlength}{\dimexpr\linewidth/1000\relax}\begin{picture}(1000,750.000)\put(0,0){\includegraphics[width=\linewidth,bb=0 0 1344 768,viewport=1016 264 1336 504,clip]{figures/small_motion/dinos/rgb.jpg}}\put(351.562,343.750){\color{suppcyan}\linethickness{0.45pt}\framebox(218.750,218.750){} }\put(656.250,328.125){\color{supporange}\linethickness{0.45pt}\framebox(187.500,187.500){} }\end{picture}\par\vspace{2pt}\hbox to\linewidth{{\setlength{\fboxsep}{0pt}\color{suppcyan}\fbox{\includegraphics[width=0.478\linewidth,bb=0 0 960 1200,viewport=488 488 712 712,clip]{figures/small_motion/dinos/rgb_details.jpg}}}\hfill{\setlength{\fboxsep}{0pt}\color{supporange}\fbox{\includegraphics[width=0.478\linewidth,bb=0 0 960 1200,viewport=728 488 952 712,clip]{figures/small_motion/dinos/rgb_details.jpg}}}}\end{minipage}\hfill\begin{minipage}[t]{0.194\textwidth}\centering\fontsize{7.4}{8.5}\selectfont\strut Vanilla$^{\dagger}$\par\vspace{1pt}\setlength{\unitlength}{\dimexpr\linewidth/1000\relax}\begin{picture}(1000,750.000)\put(0,0){\includegraphics[width=\linewidth,bb=0 0 1344 768,viewport=8 8 328 248,clip]{figures/small_motion/dinos/rgb.jpg}}\put(351.562,343.750){\color{suppcyan}\linethickness{0.45pt}\framebox(218.750,218.750){} }\put(656.250,328.125){\color{supporange}\linethickness{0.45pt}\framebox(187.500,187.500){} }\end{picture}\par\vspace{2pt}\hbox to\linewidth{{\setlength{\fboxsep}{0pt}\color{suppcyan}\fbox{\includegraphics[width=0.478\linewidth,bb=0 0 960 1200,viewport=8 248 232 472,clip]{figures/small_motion/dinos/rgb_details.jpg}}}\hfill{\setlength{\fboxsep}{0pt}\color{supporange}\fbox{\includegraphics[width=0.478\linewidth,bb=0 0 960 1200,viewport=248 248 472 472,clip]{figures/small_motion/dinos/rgb_details.jpg}}}}\end{minipage}\hfill\begin{minipage}[t]{0.194\textwidth}\centering\fontsize{7.4}{8.5}\selectfont\strut Mip-Splatting\par\vspace{1pt}\setlength{\unitlength}{\dimexpr\linewidth/1000\relax}\begin{picture}(1000,750.000)\put(0,0){\includegraphics[width=\linewidth,bb=0 0 1344 768,viewport=344 8 664 248,clip]{figures/small_motion/dinos/rgb.jpg}}\put(351.562,343.750){\color{suppcyan}\linethickness{0.45pt}\framebox(218.750,218.750){} }\put(656.250,328.125){\color{supporange}\linethickness{0.45pt}\framebox(187.500,187.500){} }\end{picture}\par\vspace{2pt}\hbox to\linewidth{{\setlength{\fboxsep}{0pt}\color{suppcyan}\fbox{\includegraphics[width=0.478\linewidth,bb=0 0 960 1200,viewport=488 248 712 472,clip]{figures/small_motion/dinos/rgb_details.jpg}}}\hfill{\setlength{\fboxsep}{0pt}\color{supporange}\fbox{\includegraphics[width=0.478\linewidth,bb=0 0 960 1200,viewport=728 248 952 472,clip]{figures/small_motion/dinos/rgb_details.jpg}}}}\end{minipage}\hfill\begin{minipage}[t]{0.194\textwidth}\centering\fontsize{7.4}{8.5}\selectfont\strut \textbf{NBGS}\par\vspace{1pt}\setlength{\unitlength}{\dimexpr\linewidth/1000\relax}\begin{picture}(1000,750.000)\put(0,0){\includegraphics[width=\linewidth,bb=0 0 1344 768,viewport=680 8 1000 248,clip]{figures/small_motion/dinos/rgb.jpg}}\put(351.562,343.750){\color{suppcyan}\linethickness{0.45pt}\framebox(218.750,218.750){} }\put(656.250,328.125){\color{supporange}\linethickness{0.45pt}\framebox(187.500,187.500){} }\end{picture}\par\vspace{2pt}\hbox to\linewidth{{\setlength{\fboxsep}{0pt}\color{suppcyan}\fbox{\includegraphics[width=0.478\linewidth,bb=0 0 960 1200,viewport=8 8 232 232,clip]{figures/small_motion/dinos/rgb_details.jpg}}}\hfill{\setlength{\fboxsep}{0pt}\color{supporange}\fbox{\includegraphics[width=0.478\linewidth,bb=0 0 960 1200,viewport=248 8 472 232,clip]{figures/small_motion/dinos/rgb_details.jpg}}}}\end{minipage}\par\vspace{4pt}
\noindent{\small\bfseries Relative rendered depth}\par\vspace{2pt}
\noindent \begin{minipage}[t]{0.194\textwidth}\centering\fontsize{7.4}{8.5}\selectfont\strut Relative depth\par\vspace{4pt}\raggedright Per-method p02--p98. Black: invalid. No depth ground truth or metric units.\par\vspace{5pt}\includegraphics[width=\linewidth,height=5pt]{figures/legends/small_motion_depth_scale.png}\par near\hfill far\end{minipage}\hfill\begin{minipage}[t]{0.194\textwidth}\centering\fontsize{7.4}{8.5}\selectfont\strut Vanilla original\par\vspace{1pt}\includegraphics[width=\linewidth,bb=0 0 1344 768,viewport=8 520 328 760,clip]{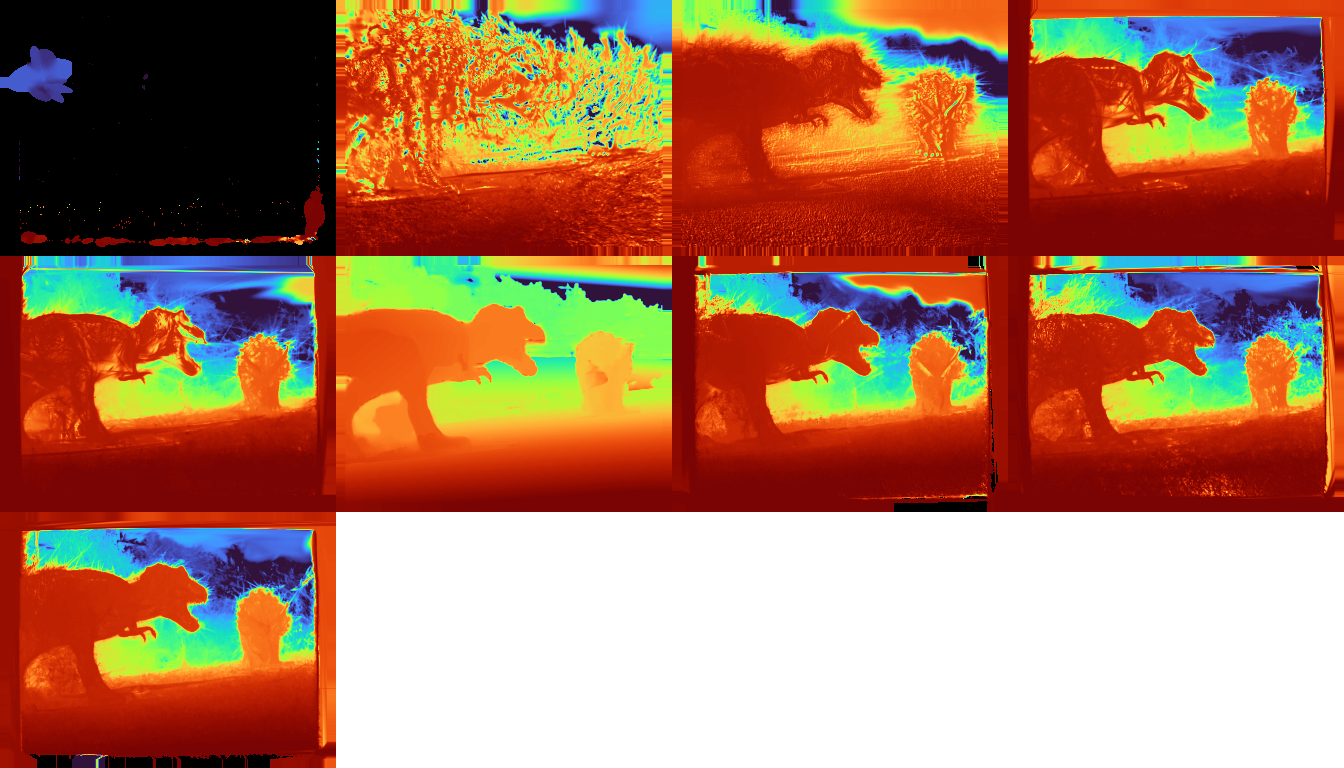}\end{minipage}\hfill\begin{minipage}[t]{0.194\textwidth}\centering\fontsize{7.4}{8.5}\selectfont\strut Small GS\par\vspace{1pt}\includegraphics[width=\linewidth,bb=0 0 1344 768,viewport=344 520 664 760,clip]{figures/small_motion/dinos/depth.png}\end{minipage}\hfill\begin{minipage}[t]{0.194\textwidth}\centering\fontsize{7.4}{8.5}\selectfont\strut InstantSplat\par\vspace{1pt}\includegraphics[width=\linewidth,bb=0 0 1344 768,viewport=680 520 1000 760,clip]{figures/small_motion/dinos/depth.png}\end{minipage}\hfill\begin{minipage}[t]{0.194\textwidth}\centering\fontsize{7.4}{8.5}\selectfont\strut DDDrop\par\vspace{1pt}\includegraphics[width=\linewidth,bb=0 0 1344 768,viewport=1016 520 1336 760,clip]{figures/small_motion/dinos/depth.png}\end{minipage}\par\vspace{4pt}
\noindent \begin{minipage}[t]{0.194\textwidth}\centering\fontsize{7.4}{8.5}\selectfont\strut DDGS\par\vspace{1pt}\includegraphics[width=\linewidth,bb=0 0 1344 768,viewport=8 264 328 504,clip]{figures/small_motion/dinos/depth.png}\end{minipage}\hfill\begin{minipage}[t]{0.194\textwidth}\centering\fontsize{7.4}{8.5}\selectfont\strut G4Splat\par\vspace{1pt}\includegraphics[width=\linewidth,bb=0 0 1344 768,viewport=344 264 664 504,clip]{figures/small_motion/dinos/depth.png}\end{minipage}\hfill\begin{minipage}[t]{0.194\textwidth}\centering\fontsize{7.4}{8.5}\selectfont\strut Vanilla$^{\dagger}$\par\vspace{1pt}\includegraphics[width=\linewidth,bb=0 0 1344 768,viewport=680 264 1000 504,clip]{figures/small_motion/dinos/depth.png}\end{minipage}\hfill\begin{minipage}[t]{0.194\textwidth}\centering\fontsize{7.4}{8.5}\selectfont\strut Mip-Splatting\par\vspace{1pt}\includegraphics[width=\linewidth,bb=0 0 1344 768,viewport=1016 264 1336 504,clip]{figures/small_motion/dinos/depth.png}\end{minipage}\hfill\begin{minipage}[t]{0.194\textwidth}\centering\fontsize{7.4}{8.5}\selectfont\strut \textbf{NBGS}\par\vspace{1pt}\includegraphics[width=\linewidth,bb=0 0 1344 768,viewport=8 8 328 248,clip]{figures/small_motion/dinos/depth.png}\end{minipage}\par\vspace{4pt}
\noindent{\small\bfseries Absolute RGB error}\par\vspace{2pt}
\noindent \begin{minipage}[t]{0.194\textwidth}\centering\fontsize{7.4}{8.5}\selectfont\strut RGB error, $[0,3]$\par\vspace{4pt}\raggedright Channel-summed absolute error. Method-matched RGB reference.\par\vspace{5pt}\includegraphics[width=\linewidth,height=5pt]{figures/legends/small_motion_error_scale.png}\par 0\hfill 3\end{minipage}\hfill\begin{minipage}[t]{0.194\textwidth}\centering\fontsize{7.4}{8.5}\selectfont\strut Vanilla original\par\vspace{1pt}\includegraphics[width=\linewidth,bb=0 0 1344 768,viewport=8 520 328 760,clip]{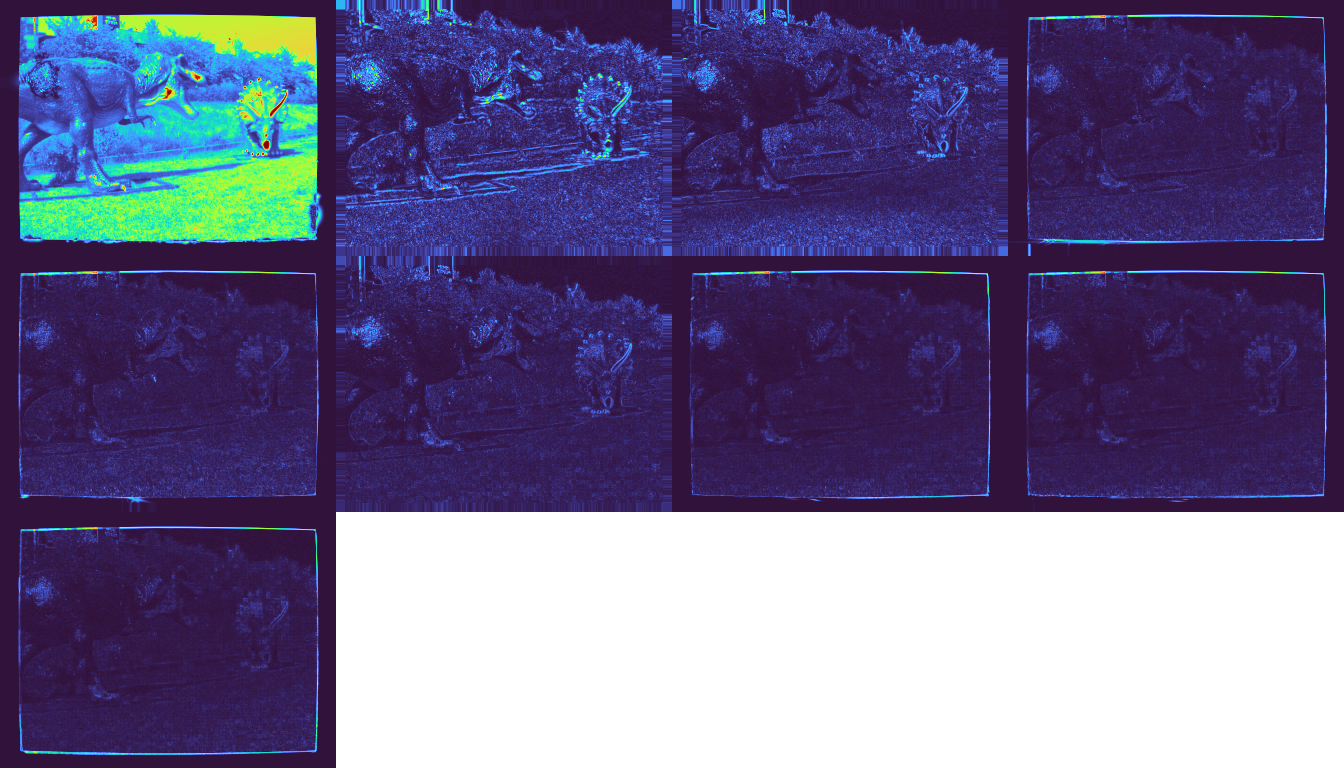}\end{minipage}\hfill\begin{minipage}[t]{0.194\textwidth}\centering\fontsize{7.4}{8.5}\selectfont\strut Small GS\par\vspace{1pt}\includegraphics[width=\linewidth,bb=0 0 1344 768,viewport=344 520 664 760,clip]{figures/small_motion/dinos/rgb_error.png}\end{minipage}\hfill\begin{minipage}[t]{0.194\textwidth}\centering\fontsize{7.4}{8.5}\selectfont\strut InstantSplat\par\vspace{1pt}\includegraphics[width=\linewidth,bb=0 0 1344 768,viewport=680 520 1000 760,clip]{figures/small_motion/dinos/rgb_error.png}\end{minipage}\hfill\begin{minipage}[t]{0.194\textwidth}\centering\fontsize{7.4}{8.5}\selectfont\strut DDDrop\par\vspace{1pt}\includegraphics[width=\linewidth,bb=0 0 1344 768,viewport=1016 520 1336 760,clip]{figures/small_motion/dinos/rgb_error.png}\end{minipage}\par\vspace{4pt}
\noindent \begin{minipage}[t]{0.194\textwidth}\centering\fontsize{7.4}{8.5}\selectfont\strut DDGS\par\vspace{1pt}\includegraphics[width=\linewidth,bb=0 0 1344 768,viewport=8 264 328 504,clip]{figures/small_motion/dinos/rgb_error.png}\end{minipage}\hfill\begin{minipage}[t]{0.194\textwidth}\centering\fontsize{7.4}{8.5}\selectfont\strut G4Splat\par\vspace{1pt}\includegraphics[width=\linewidth,bb=0 0 1344 768,viewport=344 264 664 504,clip]{figures/small_motion/dinos/rgb_error.png}\end{minipage}\hfill\begin{minipage}[t]{0.194\textwidth}\centering\fontsize{7.4}{8.5}\selectfont\strut Vanilla$^{\dagger}$\par\vspace{1pt}\includegraphics[width=\linewidth,bb=0 0 1344 768,viewport=680 264 1000 504,clip]{figures/small_motion/dinos/rgb_error.png}\end{minipage}\hfill\begin{minipage}[t]{0.194\textwidth}\centering\fontsize{7.4}{8.5}\selectfont\strut Mip-Splatting\par\vspace{1pt}\includegraphics[width=\linewidth,bb=0 0 1344 768,viewport=1016 264 1336 504,clip]{figures/small_motion/dinos/rgb_error.png}\end{minipage}\hfill\begin{minipage}[t]{0.194\textwidth}\centering\fontsize{7.4}{8.5}\selectfont\strut \textbf{NBGS}\par\vspace{1pt}\includegraphics[width=\linewidth,bb=0 0 1344 768,viewport=8 8 328 248,clip]{figures/small_motion/dinos/rgb_error.png}\end{minipage}\par\vspace{4pt}
\end{minipage}\end{lrbox}
\PlaceSuppScene
\clearpage
\pdfbookmark[3]{Flowers}{supp.scene.flowers}
\begin{lrbox}{\SuppSceneBox}\begin{minipage}{\textwidth}
\noindent{\large\bfseries Flowers}\hfill{\small\color{black!65}Small-Motion dataset}\par\vspace{3pt}
\noindent{\small\bfseries Rendered RGB}\par\vspace{2pt}
\noindent \begin{minipage}[t]{0.194\textwidth}\centering\fontsize{7.4}{8.5}\selectfont\strut Original RGB\par\vspace{1pt}\includegraphics[width=\linewidth,bb=0 0 1344 624,viewport=8 436 328 616,clip]{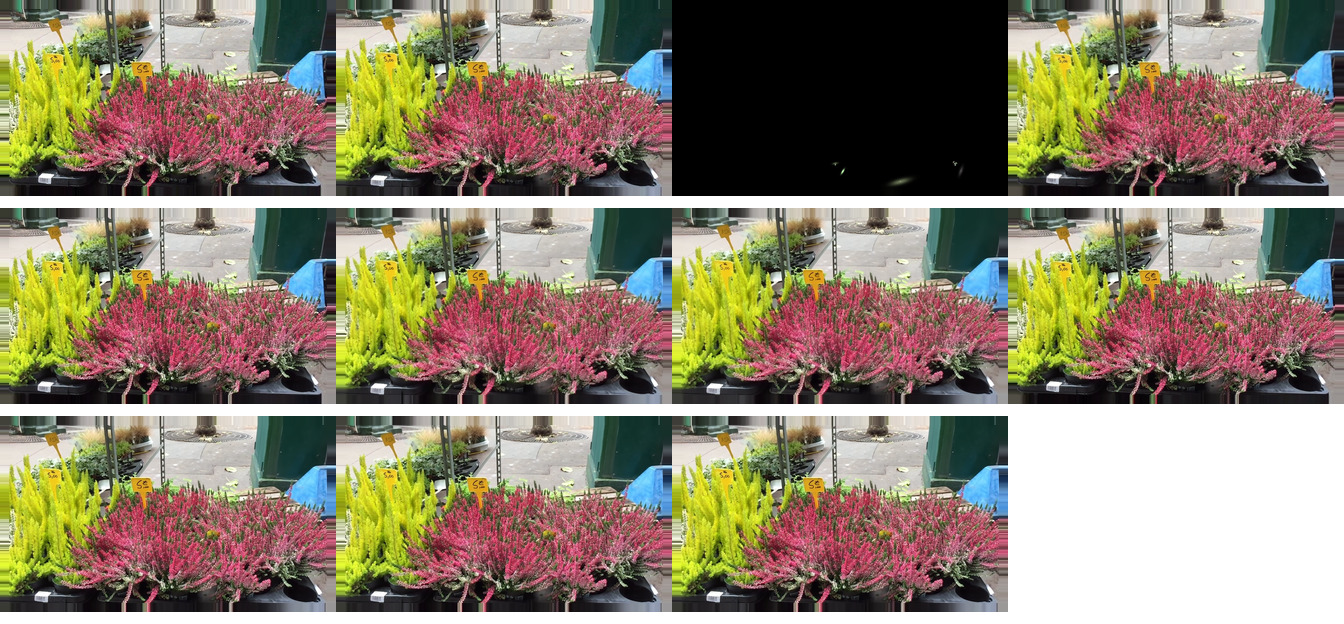}\par\vspace{2pt}\strut SfSM-undistorted RGB\par\vspace{1pt}\includegraphics[width=\linewidth,bb=0 0 1344 624,viewport=344 436 664 616,clip]{figures/small_motion/flowers/rgb.jpg}\end{minipage}\hfill\begin{minipage}[t]{0.194\textwidth}\centering\fontsize{7.4}{8.5}\selectfont\strut Vanilla original\par\vspace{1pt}\setlength{\unitlength}{\dimexpr\linewidth/1000\relax}\begin{picture}(1000,562.500)\put(0,0){\includegraphics[width=\linewidth,bb=0 0 1344 624,viewport=680 436 1000 616,clip]{figures/small_motion/flowers/rgb.jpg}}\put(78.125,203.125){\color{suppcyan}\linethickness{0.45pt}\framebox(187.500,187.500){} }\put(640.625,101.562){\color{supporange}\linethickness{0.45pt}\framebox(187.500,187.500){} }\end{picture}\par\vspace{2pt}\hbox to\linewidth{{\setlength{\fboxsep}{0pt}\color{suppcyan}\fbox{\includegraphics[width=0.478\linewidth,bb=0 0 960 1200,viewport=8 968 232 1192,clip]{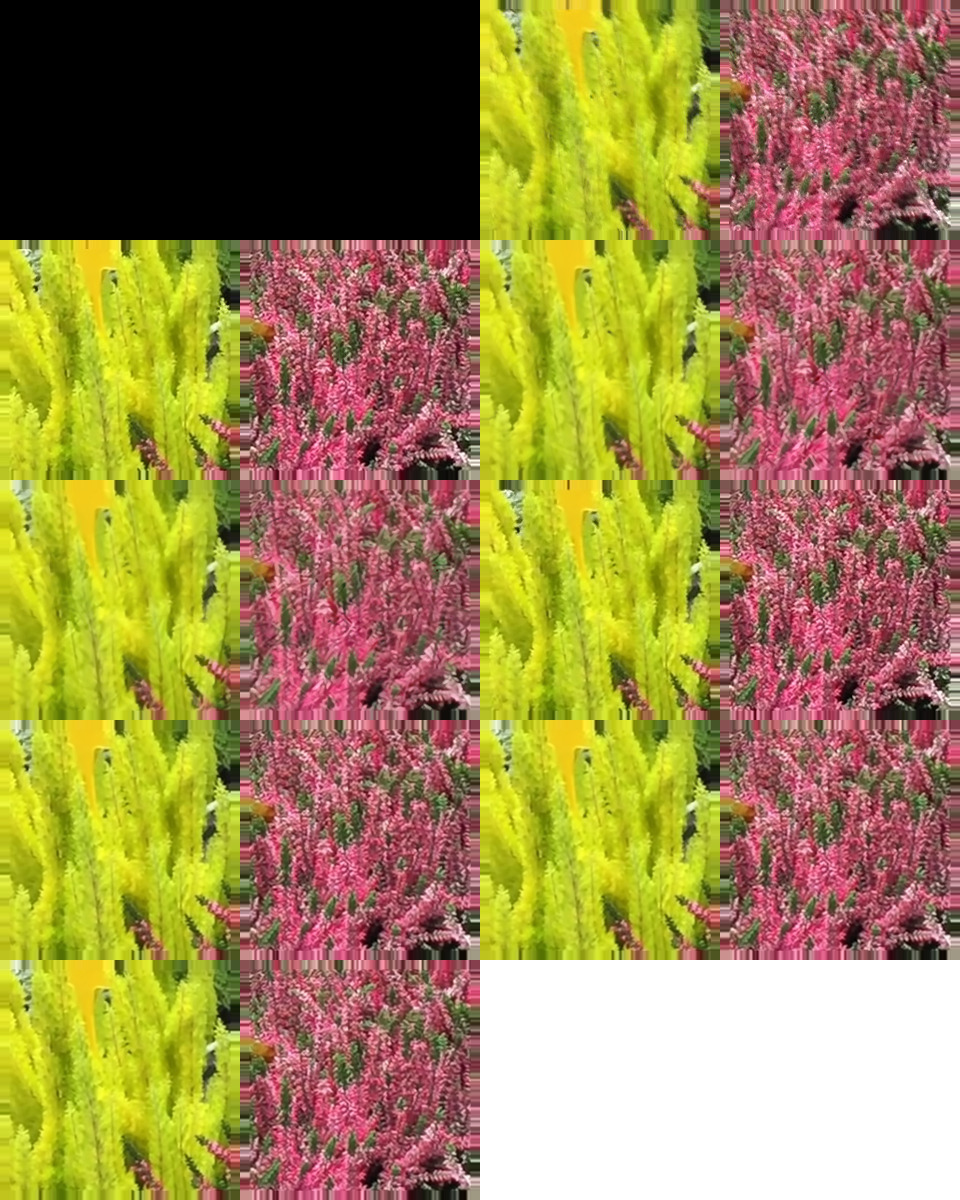}}}\hfill{\setlength{\fboxsep}{0pt}\color{supporange}\fbox{\includegraphics[width=0.478\linewidth,bb=0 0 960 1200,viewport=248 968 472 1192,clip]{figures/small_motion/flowers/rgb_details.jpg}}}}\end{minipage}\hfill\begin{minipage}[t]{0.194\textwidth}\centering\fontsize{7.4}{8.5}\selectfont\strut Small GS\par\vspace{1pt}\setlength{\unitlength}{\dimexpr\linewidth/1000\relax}\begin{picture}(1000,562.500)\put(0,0){\includegraphics[width=\linewidth,bb=0 0 1344 624,viewport=1016 436 1336 616,clip]{figures/small_motion/flowers/rgb.jpg}}\put(78.125,203.125){\color{suppcyan}\linethickness{0.45pt}\framebox(187.500,187.500){} }\put(640.625,101.562){\color{supporange}\linethickness{0.45pt}\framebox(187.500,187.500){} }\end{picture}\par\vspace{2pt}\hbox to\linewidth{{\setlength{\fboxsep}{0pt}\color{suppcyan}\fbox{\includegraphics[width=0.478\linewidth,bb=0 0 960 1200,viewport=488 968 712 1192,clip]{figures/small_motion/flowers/rgb_details.jpg}}}\hfill{\setlength{\fboxsep}{0pt}\color{supporange}\fbox{\includegraphics[width=0.478\linewidth,bb=0 0 960 1200,viewport=728 968 952 1192,clip]{figures/small_motion/flowers/rgb_details.jpg}}}}\end{minipage}\hfill\begin{minipage}[t]{0.194\textwidth}\centering\fontsize{7.4}{8.5}\selectfont\strut InstantSplat\par\vspace{1pt}\setlength{\unitlength}{\dimexpr\linewidth/1000\relax}\begin{picture}(1000,562.500)\put(0,0){\includegraphics[width=\linewidth,bb=0 0 1344 624,viewport=8 228 328 408,clip]{figures/small_motion/flowers/rgb.jpg}}\put(78.125,203.125){\color{suppcyan}\linethickness{0.45pt}\framebox(187.500,187.500){} }\put(640.625,101.562){\color{supporange}\linethickness{0.45pt}\framebox(187.500,187.500){} }\end{picture}\par\vspace{2pt}\hbox to\linewidth{{\setlength{\fboxsep}{0pt}\color{suppcyan}\fbox{\includegraphics[width=0.478\linewidth,bb=0 0 960 1200,viewport=8 728 232 952,clip]{figures/small_motion/flowers/rgb_details.jpg}}}\hfill{\setlength{\fboxsep}{0pt}\color{supporange}\fbox{\includegraphics[width=0.478\linewidth,bb=0 0 960 1200,viewport=248 728 472 952,clip]{figures/small_motion/flowers/rgb_details.jpg}}}}\end{minipage}\hfill\begin{minipage}[t]{0.194\textwidth}\centering\fontsize{7.4}{8.5}\selectfont\strut DDDrop\par\vspace{1pt}\setlength{\unitlength}{\dimexpr\linewidth/1000\relax}\begin{picture}(1000,562.500)\put(0,0){\includegraphics[width=\linewidth,bb=0 0 1344 624,viewport=344 228 664 408,clip]{figures/small_motion/flowers/rgb.jpg}}\put(78.125,203.125){\color{suppcyan}\linethickness{0.45pt}\framebox(187.500,187.500){} }\put(640.625,101.562){\color{supporange}\linethickness{0.45pt}\framebox(187.500,187.500){} }\end{picture}\par\vspace{2pt}\hbox to\linewidth{{\setlength{\fboxsep}{0pt}\color{suppcyan}\fbox{\includegraphics[width=0.478\linewidth,bb=0 0 960 1200,viewport=488 728 712 952,clip]{figures/small_motion/flowers/rgb_details.jpg}}}\hfill{\setlength{\fboxsep}{0pt}\color{supporange}\fbox{\includegraphics[width=0.478\linewidth,bb=0 0 960 1200,viewport=728 728 952 952,clip]{figures/small_motion/flowers/rgb_details.jpg}}}}\end{minipage}\par\vspace{4pt}
\noindent \begin{minipage}[t]{0.194\textwidth}\centering\fontsize{7.4}{8.5}\selectfont\strut DDGS\par\vspace{1pt}\setlength{\unitlength}{\dimexpr\linewidth/1000\relax}\begin{picture}(1000,562.500)\put(0,0){\includegraphics[width=\linewidth,bb=0 0 1344 624,viewport=680 228 1000 408,clip]{figures/small_motion/flowers/rgb.jpg}}\put(78.125,203.125){\color{suppcyan}\linethickness{0.45pt}\framebox(187.500,187.500){} }\put(640.625,101.562){\color{supporange}\linethickness{0.45pt}\framebox(187.500,187.500){} }\end{picture}\par\vspace{2pt}\hbox to\linewidth{{\setlength{\fboxsep}{0pt}\color{suppcyan}\fbox{\includegraphics[width=0.478\linewidth,bb=0 0 960 1200,viewport=8 488 232 712,clip]{figures/small_motion/flowers/rgb_details.jpg}}}\hfill{\setlength{\fboxsep}{0pt}\color{supporange}\fbox{\includegraphics[width=0.478\linewidth,bb=0 0 960 1200,viewport=248 488 472 712,clip]{figures/small_motion/flowers/rgb_details.jpg}}}}\end{minipage}\hfill\begin{minipage}[t]{0.194\textwidth}\centering\fontsize{7.4}{8.5}\selectfont\strut G4Splat\par\vspace{1pt}\setlength{\unitlength}{\dimexpr\linewidth/1000\relax}\begin{picture}(1000,562.500)\put(0,0){\includegraphics[width=\linewidth,bb=0 0 1344 624,viewport=1016 228 1336 408,clip]{figures/small_motion/flowers/rgb.jpg}}\put(78.125,203.125){\color{suppcyan}\linethickness{0.45pt}\framebox(187.500,187.500){} }\put(640.625,101.562){\color{supporange}\linethickness{0.45pt}\framebox(187.500,187.500){} }\end{picture}\par\vspace{2pt}\hbox to\linewidth{{\setlength{\fboxsep}{0pt}\color{suppcyan}\fbox{\includegraphics[width=0.478\linewidth,bb=0 0 960 1200,viewport=488 488 712 712,clip]{figures/small_motion/flowers/rgb_details.jpg}}}\hfill{\setlength{\fboxsep}{0pt}\color{supporange}\fbox{\includegraphics[width=0.478\linewidth,bb=0 0 960 1200,viewport=728 488 952 712,clip]{figures/small_motion/flowers/rgb_details.jpg}}}}\end{minipage}\hfill\begin{minipage}[t]{0.194\textwidth}\centering\fontsize{7.4}{8.5}\selectfont\strut Vanilla$^{\dagger}$\par\vspace{1pt}\setlength{\unitlength}{\dimexpr\linewidth/1000\relax}\begin{picture}(1000,562.500)\put(0,0){\includegraphics[width=\linewidth,bb=0 0 1344 624,viewport=8 20 328 200,clip]{figures/small_motion/flowers/rgb.jpg}}\put(78.125,203.125){\color{suppcyan}\linethickness{0.45pt}\framebox(187.500,187.500){} }\put(640.625,101.562){\color{supporange}\linethickness{0.45pt}\framebox(187.500,187.500){} }\end{picture}\par\vspace{2pt}\hbox to\linewidth{{\setlength{\fboxsep}{0pt}\color{suppcyan}\fbox{\includegraphics[width=0.478\linewidth,bb=0 0 960 1200,viewport=8 248 232 472,clip]{figures/small_motion/flowers/rgb_details.jpg}}}\hfill{\setlength{\fboxsep}{0pt}\color{supporange}\fbox{\includegraphics[width=0.478\linewidth,bb=0 0 960 1200,viewport=248 248 472 472,clip]{figures/small_motion/flowers/rgb_details.jpg}}}}\end{minipage}\hfill\begin{minipage}[t]{0.194\textwidth}\centering\fontsize{7.4}{8.5}\selectfont\strut Mip-Splatting\par\vspace{1pt}\setlength{\unitlength}{\dimexpr\linewidth/1000\relax}\begin{picture}(1000,562.500)\put(0,0){\includegraphics[width=\linewidth,bb=0 0 1344 624,viewport=344 20 664 200,clip]{figures/small_motion/flowers/rgb.jpg}}\put(78.125,203.125){\color{suppcyan}\linethickness{0.45pt}\framebox(187.500,187.500){} }\put(640.625,101.562){\color{supporange}\linethickness{0.45pt}\framebox(187.500,187.500){} }\end{picture}\par\vspace{2pt}\hbox to\linewidth{{\setlength{\fboxsep}{0pt}\color{suppcyan}\fbox{\includegraphics[width=0.478\linewidth,bb=0 0 960 1200,viewport=488 248 712 472,clip]{figures/small_motion/flowers/rgb_details.jpg}}}\hfill{\setlength{\fboxsep}{0pt}\color{supporange}\fbox{\includegraphics[width=0.478\linewidth,bb=0 0 960 1200,viewport=728 248 952 472,clip]{figures/small_motion/flowers/rgb_details.jpg}}}}\end{minipage}\hfill\begin{minipage}[t]{0.194\textwidth}\centering\fontsize{7.4}{8.5}\selectfont\strut \textbf{NBGS}\par\vspace{1pt}\setlength{\unitlength}{\dimexpr\linewidth/1000\relax}\begin{picture}(1000,562.500)\put(0,0){\includegraphics[width=\linewidth,bb=0 0 1344 624,viewport=680 20 1000 200,clip]{figures/small_motion/flowers/rgb.jpg}}\put(78.125,203.125){\color{suppcyan}\linethickness{0.45pt}\framebox(187.500,187.500){} }\put(640.625,101.562){\color{supporange}\linethickness{0.45pt}\framebox(187.500,187.500){} }\end{picture}\par\vspace{2pt}\hbox to\linewidth{{\setlength{\fboxsep}{0pt}\color{suppcyan}\fbox{\includegraphics[width=0.478\linewidth,bb=0 0 960 1200,viewport=8 8 232 232,clip]{figures/small_motion/flowers/rgb_details.jpg}}}\hfill{\setlength{\fboxsep}{0pt}\color{supporange}\fbox{\includegraphics[width=0.478\linewidth,bb=0 0 960 1200,viewport=248 8 472 232,clip]{figures/small_motion/flowers/rgb_details.jpg}}}}\end{minipage}\par\vspace{4pt}
\noindent{\small\bfseries Relative rendered depth}\par\vspace{2pt}
\noindent \begin{minipage}[t]{0.194\textwidth}\centering\fontsize{7.4}{8.5}\selectfont\strut Relative depth\par\vspace{4pt}\raggedright Per-method p02--p98. Black: invalid. No depth ground truth or metric units.\par\vspace{5pt}\includegraphics[width=\linewidth,height=5pt]{figures/legends/small_motion_depth_scale.png}\par near\hfill far\end{minipage}\hfill\begin{minipage}[t]{0.194\textwidth}\centering\fontsize{7.4}{8.5}\selectfont\strut Vanilla original\par\vspace{1pt}\includegraphics[width=\linewidth,bb=0 0 1344 624,viewport=8 436 328 616,clip]{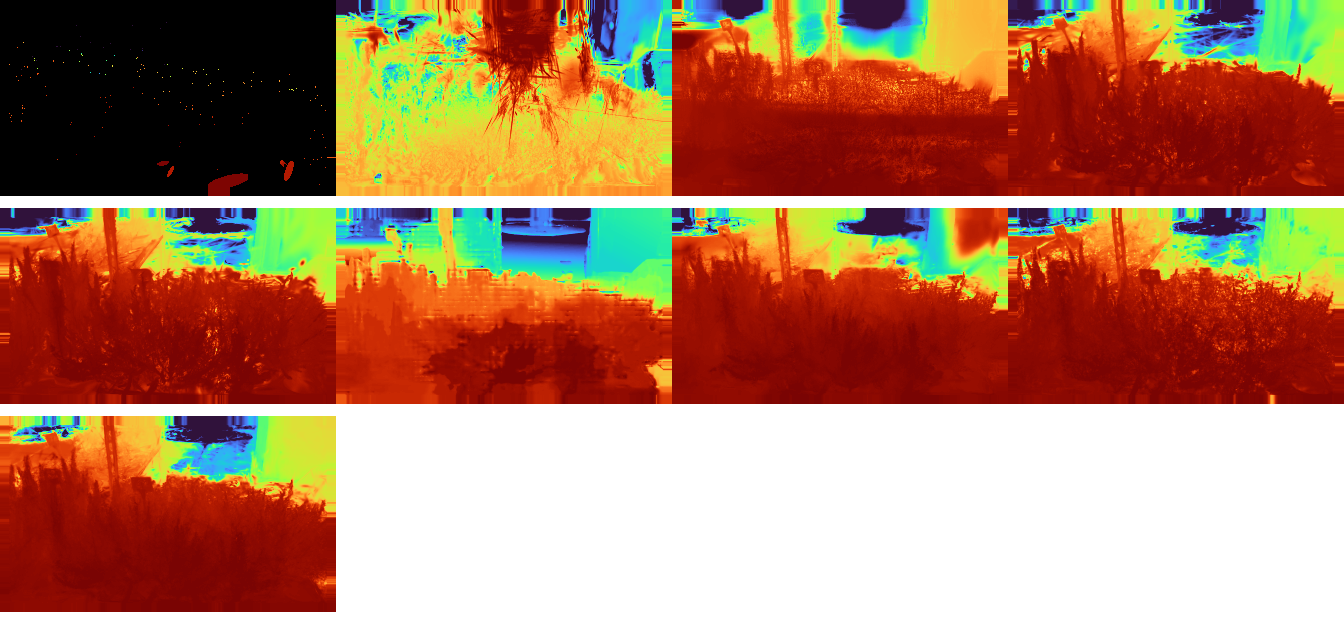}\end{minipage}\hfill\begin{minipage}[t]{0.194\textwidth}\centering\fontsize{7.4}{8.5}\selectfont\strut Small GS\par\vspace{1pt}\includegraphics[width=\linewidth,bb=0 0 1344 624,viewport=344 436 664 616,clip]{figures/small_motion/flowers/depth.png}\end{minipage}\hfill\begin{minipage}[t]{0.194\textwidth}\centering\fontsize{7.4}{8.5}\selectfont\strut InstantSplat\par\vspace{1pt}\includegraphics[width=\linewidth,bb=0 0 1344 624,viewport=680 436 1000 616,clip]{figures/small_motion/flowers/depth.png}\end{minipage}\hfill\begin{minipage}[t]{0.194\textwidth}\centering\fontsize{7.4}{8.5}\selectfont\strut DDDrop\par\vspace{1pt}\includegraphics[width=\linewidth,bb=0 0 1344 624,viewport=1016 436 1336 616,clip]{figures/small_motion/flowers/depth.png}\end{minipage}\par\vspace{4pt}
\noindent \begin{minipage}[t]{0.194\textwidth}\centering\fontsize{7.4}{8.5}\selectfont\strut DDGS\par\vspace{1pt}\includegraphics[width=\linewidth,bb=0 0 1344 624,viewport=8 228 328 408,clip]{figures/small_motion/flowers/depth.png}\end{minipage}\hfill\begin{minipage}[t]{0.194\textwidth}\centering\fontsize{7.4}{8.5}\selectfont\strut G4Splat\par\vspace{1pt}\includegraphics[width=\linewidth,bb=0 0 1344 624,viewport=344 228 664 408,clip]{figures/small_motion/flowers/depth.png}\end{minipage}\hfill\begin{minipage}[t]{0.194\textwidth}\centering\fontsize{7.4}{8.5}\selectfont\strut Vanilla$^{\dagger}$\par\vspace{1pt}\includegraphics[width=\linewidth,bb=0 0 1344 624,viewport=680 228 1000 408,clip]{figures/small_motion/flowers/depth.png}\end{minipage}\hfill\begin{minipage}[t]{0.194\textwidth}\centering\fontsize{7.4}{8.5}\selectfont\strut Mip-Splatting\par\vspace{1pt}\includegraphics[width=\linewidth,bb=0 0 1344 624,viewport=1016 228 1336 408,clip]{figures/small_motion/flowers/depth.png}\end{minipage}\hfill\begin{minipage}[t]{0.194\textwidth}\centering\fontsize{7.4}{8.5}\selectfont\strut \textbf{NBGS}\par\vspace{1pt}\includegraphics[width=\linewidth,bb=0 0 1344 624,viewport=8 20 328 200,clip]{figures/small_motion/flowers/depth.png}\end{minipage}\par\vspace{4pt}
\noindent{\small\bfseries Absolute RGB error}\par\vspace{2pt}
\noindent \begin{minipage}[t]{0.194\textwidth}\centering\fontsize{7.4}{8.5}\selectfont\strut RGB error, $[0,3]$\par\vspace{4pt}\raggedright Channel-summed absolute error. Method-matched RGB reference.\par\vspace{5pt}\includegraphics[width=\linewidth,height=5pt]{figures/legends/small_motion_error_scale.png}\par 0\hfill 3\end{minipage}\hfill\begin{minipage}[t]{0.194\textwidth}\centering\fontsize{7.4}{8.5}\selectfont\strut Vanilla original\par\vspace{1pt}\includegraphics[width=\linewidth,bb=0 0 1344 624,viewport=8 436 328 616,clip]{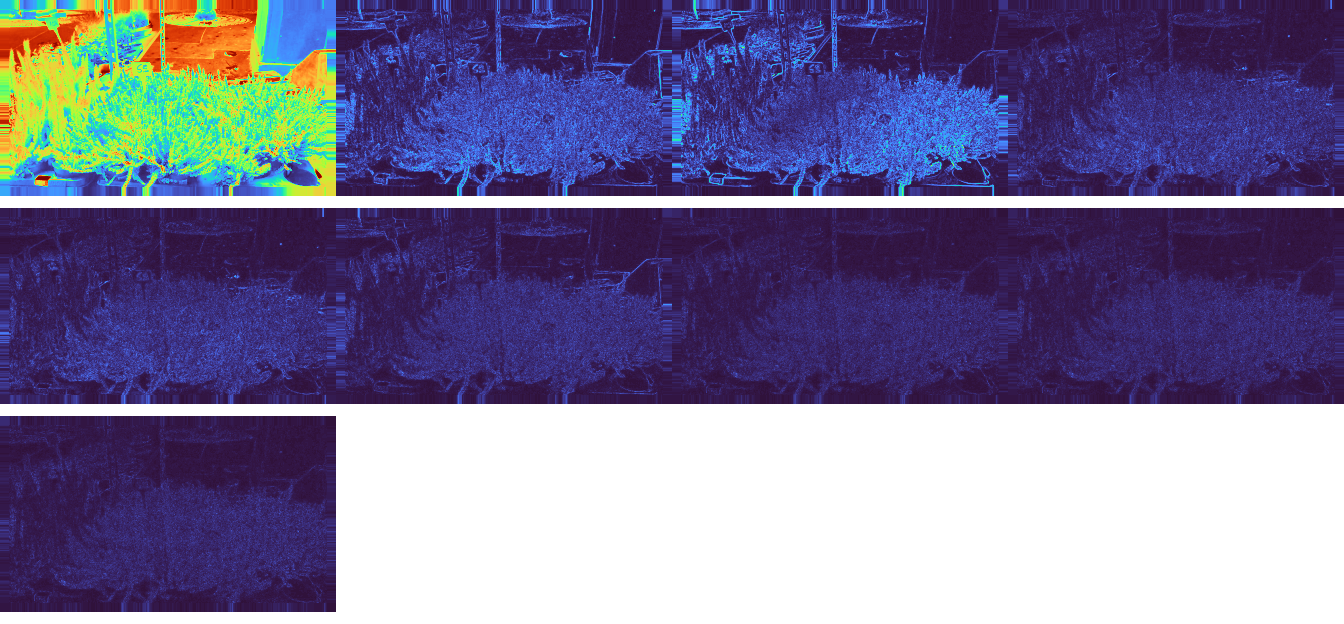}\end{minipage}\hfill\begin{minipage}[t]{0.194\textwidth}\centering\fontsize{7.4}{8.5}\selectfont\strut Small GS\par\vspace{1pt}\includegraphics[width=\linewidth,bb=0 0 1344 624,viewport=344 436 664 616,clip]{figures/small_motion/flowers/rgb_error.png}\end{minipage}\hfill\begin{minipage}[t]{0.194\textwidth}\centering\fontsize{7.4}{8.5}\selectfont\strut InstantSplat\par\vspace{1pt}\includegraphics[width=\linewidth,bb=0 0 1344 624,viewport=680 436 1000 616,clip]{figures/small_motion/flowers/rgb_error.png}\end{minipage}\hfill\begin{minipage}[t]{0.194\textwidth}\centering\fontsize{7.4}{8.5}\selectfont\strut DDDrop\par\vspace{1pt}\includegraphics[width=\linewidth,bb=0 0 1344 624,viewport=1016 436 1336 616,clip]{figures/small_motion/flowers/rgb_error.png}\end{minipage}\par\vspace{4pt}
\noindent \begin{minipage}[t]{0.194\textwidth}\centering\fontsize{7.4}{8.5}\selectfont\strut DDGS\par\vspace{1pt}\includegraphics[width=\linewidth,bb=0 0 1344 624,viewport=8 228 328 408,clip]{figures/small_motion/flowers/rgb_error.png}\end{minipage}\hfill\begin{minipage}[t]{0.194\textwidth}\centering\fontsize{7.4}{8.5}\selectfont\strut G4Splat\par\vspace{1pt}\includegraphics[width=\linewidth,bb=0 0 1344 624,viewport=344 228 664 408,clip]{figures/small_motion/flowers/rgb_error.png}\end{minipage}\hfill\begin{minipage}[t]{0.194\textwidth}\centering\fontsize{7.4}{8.5}\selectfont\strut Vanilla$^{\dagger}$\par\vspace{1pt}\includegraphics[width=\linewidth,bb=0 0 1344 624,viewport=680 228 1000 408,clip]{figures/small_motion/flowers/rgb_error.png}\end{minipage}\hfill\begin{minipage}[t]{0.194\textwidth}\centering\fontsize{7.4}{8.5}\selectfont\strut Mip-Splatting\par\vspace{1pt}\includegraphics[width=\linewidth,bb=0 0 1344 624,viewport=1016 228 1336 408,clip]{figures/small_motion/flowers/rgb_error.png}\end{minipage}\hfill\begin{minipage}[t]{0.194\textwidth}\centering\fontsize{7.4}{8.5}\selectfont\strut \textbf{NBGS}\par\vspace{1pt}\includegraphics[width=\linewidth,bb=0 0 1344 624,viewport=8 20 328 200,clip]{figures/small_motion/flowers/rgb_error.png}\end{minipage}\par\vspace{4pt}
\end{minipage}\end{lrbox}
\PlaceSuppScene
\clearpage
\pdfbookmark[3]{Frogs}{supp.scene.frogs}
\begin{lrbox}{\SuppSceneBox}\begin{minipage}{\textwidth}
\noindent{\large\bfseries Frogs}\hfill{\small\color{black!65}Small-Motion dataset}\par\vspace{3pt}
\noindent{\small\bfseries Rendered RGB}\par\vspace{2pt}
\noindent \begin{minipage}[t]{0.194\textwidth}\centering\fontsize{7.4}{8.5}\selectfont\strut Original RGB\par\vspace{1pt}\includegraphics[width=\linewidth,bb=0 0 1344 624,viewport=8 436 328 616,clip]{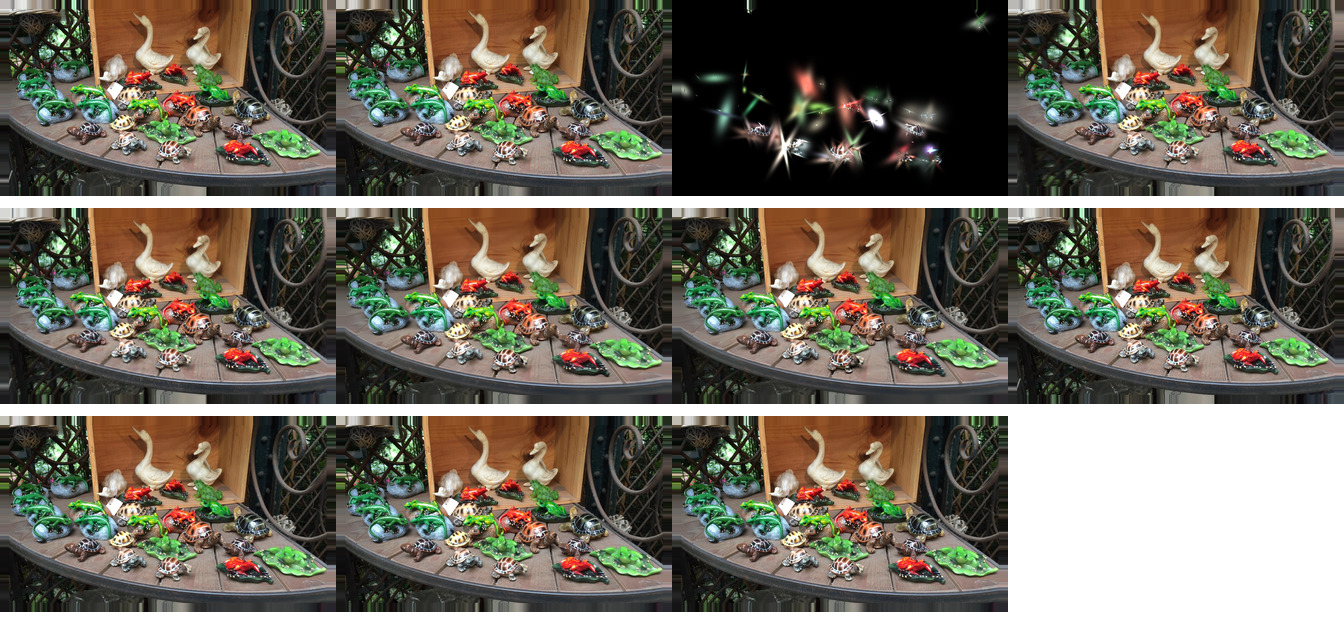}\par\vspace{2pt}\strut SfSM-undistorted RGB\par\vspace{1pt}\includegraphics[width=\linewidth,bb=0 0 1344 624,viewport=344 436 664 616,clip]{figures/small_motion/frogs/rgb.jpg}\end{minipage}\hfill\begin{minipage}[t]{0.194\textwidth}\centering\fontsize{7.4}{8.5}\selectfont\strut Vanilla original\par\vspace{1pt}\setlength{\unitlength}{\dimexpr\linewidth/1000\relax}\begin{picture}(1000,562.500)\put(0,0){\includegraphics[width=\linewidth,bb=0 0 1344 624,viewport=680 436 1000 616,clip]{figures/small_motion/frogs/rgb.jpg}}\put(338.542,93.750){\color{suppcyan}\linethickness{0.45pt}\framebox(145.833,145.833){} }\put(619.792,119.792){\color{supporange}\linethickness{0.45pt}\framebox(145.833,145.833){} }\end{picture}\par\vspace{2pt}\hbox to\linewidth{{\setlength{\fboxsep}{0pt}\color{suppcyan}\fbox{\includegraphics[width=0.478\linewidth,bb=0 0 960 1200,viewport=8 968 232 1192,clip]{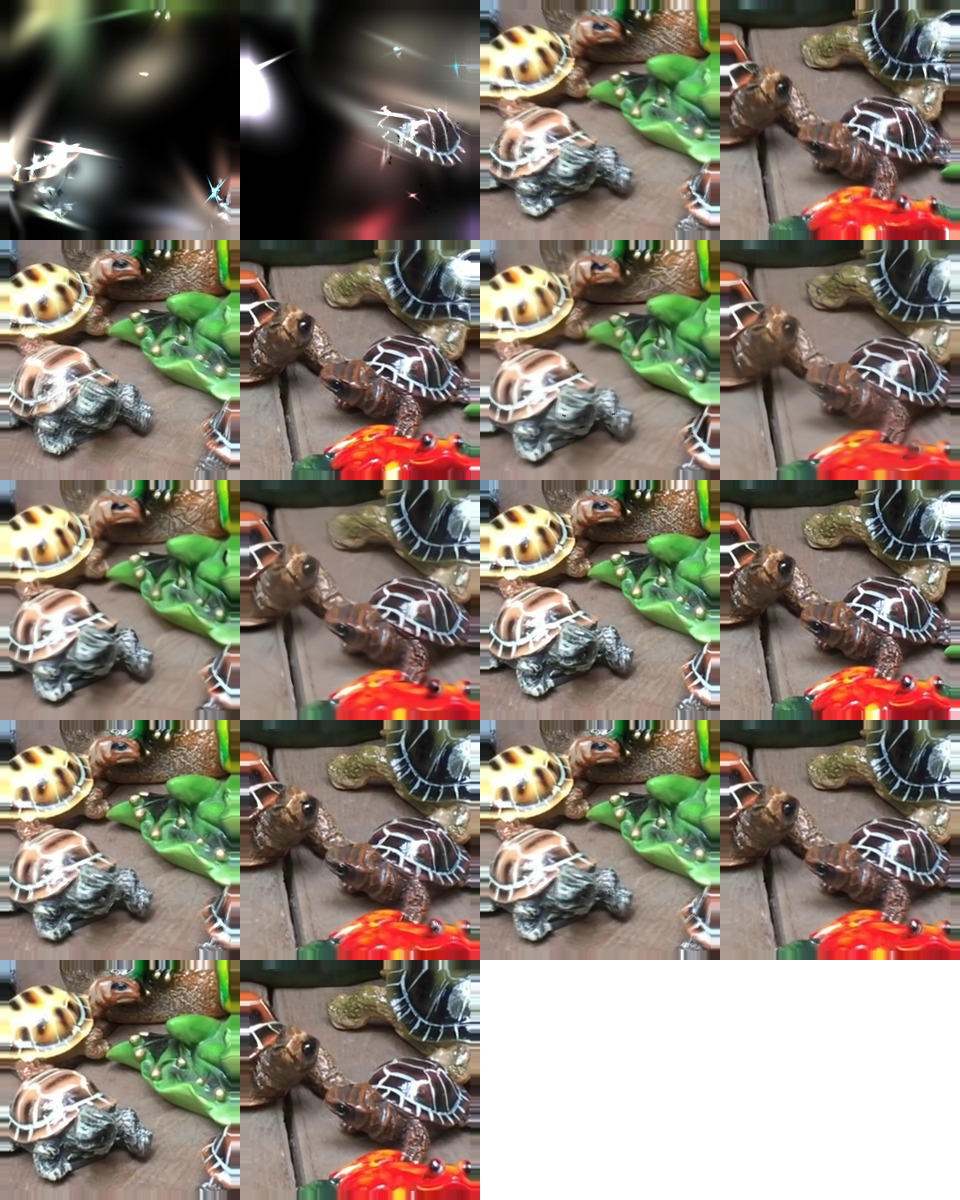}}}\hfill{\setlength{\fboxsep}{0pt}\color{supporange}\fbox{\includegraphics[width=0.478\linewidth,bb=0 0 960 1200,viewport=248 968 472 1192,clip]{figures/small_motion/frogs/rgb_details.jpg}}}}\end{minipage}\hfill\begin{minipage}[t]{0.194\textwidth}\centering\fontsize{7.4}{8.5}\selectfont\strut Small GS\par\vspace{1pt}\setlength{\unitlength}{\dimexpr\linewidth/1000\relax}\begin{picture}(1000,562.500)\put(0,0){\includegraphics[width=\linewidth,bb=0 0 1344 624,viewport=1016 436 1336 616,clip]{figures/small_motion/frogs/rgb.jpg}}\put(338.542,93.750){\color{suppcyan}\linethickness{0.45pt}\framebox(145.833,145.833){} }\put(619.792,119.792){\color{supporange}\linethickness{0.45pt}\framebox(145.833,145.833){} }\end{picture}\par\vspace{2pt}\hbox to\linewidth{{\setlength{\fboxsep}{0pt}\color{suppcyan}\fbox{\includegraphics[width=0.478\linewidth,bb=0 0 960 1200,viewport=488 968 712 1192,clip]{figures/small_motion/frogs/rgb_details.jpg}}}\hfill{\setlength{\fboxsep}{0pt}\color{supporange}\fbox{\includegraphics[width=0.478\linewidth,bb=0 0 960 1200,viewport=728 968 952 1192,clip]{figures/small_motion/frogs/rgb_details.jpg}}}}\end{minipage}\hfill\begin{minipage}[t]{0.194\textwidth}\centering\fontsize{7.4}{8.5}\selectfont\strut InstantSplat\par\vspace{1pt}\setlength{\unitlength}{\dimexpr\linewidth/1000\relax}\begin{picture}(1000,562.500)\put(0,0){\includegraphics[width=\linewidth,bb=0 0 1344 624,viewport=8 228 328 408,clip]{figures/small_motion/frogs/rgb.jpg}}\put(338.542,93.750){\color{suppcyan}\linethickness{0.45pt}\framebox(145.833,145.833){} }\put(619.792,119.792){\color{supporange}\linethickness{0.45pt}\framebox(145.833,145.833){} }\end{picture}\par\vspace{2pt}\hbox to\linewidth{{\setlength{\fboxsep}{0pt}\color{suppcyan}\fbox{\includegraphics[width=0.478\linewidth,bb=0 0 960 1200,viewport=8 728 232 952,clip]{figures/small_motion/frogs/rgb_details.jpg}}}\hfill{\setlength{\fboxsep}{0pt}\color{supporange}\fbox{\includegraphics[width=0.478\linewidth,bb=0 0 960 1200,viewport=248 728 472 952,clip]{figures/small_motion/frogs/rgb_details.jpg}}}}\end{minipage}\hfill\begin{minipage}[t]{0.194\textwidth}\centering\fontsize{7.4}{8.5}\selectfont\strut DDDrop\par\vspace{1pt}\setlength{\unitlength}{\dimexpr\linewidth/1000\relax}\begin{picture}(1000,562.500)\put(0,0){\includegraphics[width=\linewidth,bb=0 0 1344 624,viewport=344 228 664 408,clip]{figures/small_motion/frogs/rgb.jpg}}\put(338.542,93.750){\color{suppcyan}\linethickness{0.45pt}\framebox(145.833,145.833){} }\put(619.792,119.792){\color{supporange}\linethickness{0.45pt}\framebox(145.833,145.833){} }\end{picture}\par\vspace{2pt}\hbox to\linewidth{{\setlength{\fboxsep}{0pt}\color{suppcyan}\fbox{\includegraphics[width=0.478\linewidth,bb=0 0 960 1200,viewport=488 728 712 952,clip]{figures/small_motion/frogs/rgb_details.jpg}}}\hfill{\setlength{\fboxsep}{0pt}\color{supporange}\fbox{\includegraphics[width=0.478\linewidth,bb=0 0 960 1200,viewport=728 728 952 952,clip]{figures/small_motion/frogs/rgb_details.jpg}}}}\end{minipage}\par\vspace{4pt}
\noindent \begin{minipage}[t]{0.194\textwidth}\centering\fontsize{7.4}{8.5}\selectfont\strut DDGS\par\vspace{1pt}\setlength{\unitlength}{\dimexpr\linewidth/1000\relax}\begin{picture}(1000,562.500)\put(0,0){\includegraphics[width=\linewidth,bb=0 0 1344 624,viewport=680 228 1000 408,clip]{figures/small_motion/frogs/rgb.jpg}}\put(338.542,93.750){\color{suppcyan}\linethickness{0.45pt}\framebox(145.833,145.833){} }\put(619.792,119.792){\color{supporange}\linethickness{0.45pt}\framebox(145.833,145.833){} }\end{picture}\par\vspace{2pt}\hbox to\linewidth{{\setlength{\fboxsep}{0pt}\color{suppcyan}\fbox{\includegraphics[width=0.478\linewidth,bb=0 0 960 1200,viewport=8 488 232 712,clip]{figures/small_motion/frogs/rgb_details.jpg}}}\hfill{\setlength{\fboxsep}{0pt}\color{supporange}\fbox{\includegraphics[width=0.478\linewidth,bb=0 0 960 1200,viewport=248 488 472 712,clip]{figures/small_motion/frogs/rgb_details.jpg}}}}\end{minipage}\hfill\begin{minipage}[t]{0.194\textwidth}\centering\fontsize{7.4}{8.5}\selectfont\strut G4Splat\par\vspace{1pt}\setlength{\unitlength}{\dimexpr\linewidth/1000\relax}\begin{picture}(1000,562.500)\put(0,0){\includegraphics[width=\linewidth,bb=0 0 1344 624,viewport=1016 228 1336 408,clip]{figures/small_motion/frogs/rgb.jpg}}\put(338.542,93.750){\color{suppcyan}\linethickness{0.45pt}\framebox(145.833,145.833){} }\put(619.792,119.792){\color{supporange}\linethickness{0.45pt}\framebox(145.833,145.833){} }\end{picture}\par\vspace{2pt}\hbox to\linewidth{{\setlength{\fboxsep}{0pt}\color{suppcyan}\fbox{\includegraphics[width=0.478\linewidth,bb=0 0 960 1200,viewport=488 488 712 712,clip]{figures/small_motion/frogs/rgb_details.jpg}}}\hfill{\setlength{\fboxsep}{0pt}\color{supporange}\fbox{\includegraphics[width=0.478\linewidth,bb=0 0 960 1200,viewport=728 488 952 712,clip]{figures/small_motion/frogs/rgb_details.jpg}}}}\end{minipage}\hfill\begin{minipage}[t]{0.194\textwidth}\centering\fontsize{7.4}{8.5}\selectfont\strut Vanilla$^{\dagger}$\par\vspace{1pt}\setlength{\unitlength}{\dimexpr\linewidth/1000\relax}\begin{picture}(1000,562.500)\put(0,0){\includegraphics[width=\linewidth,bb=0 0 1344 624,viewport=8 20 328 200,clip]{figures/small_motion/frogs/rgb.jpg}}\put(338.542,93.750){\color{suppcyan}\linethickness{0.45pt}\framebox(145.833,145.833){} }\put(619.792,119.792){\color{supporange}\linethickness{0.45pt}\framebox(145.833,145.833){} }\end{picture}\par\vspace{2pt}\hbox to\linewidth{{\setlength{\fboxsep}{0pt}\color{suppcyan}\fbox{\includegraphics[width=0.478\linewidth,bb=0 0 960 1200,viewport=8 248 232 472,clip]{figures/small_motion/frogs/rgb_details.jpg}}}\hfill{\setlength{\fboxsep}{0pt}\color{supporange}\fbox{\includegraphics[width=0.478\linewidth,bb=0 0 960 1200,viewport=248 248 472 472,clip]{figures/small_motion/frogs/rgb_details.jpg}}}}\end{minipage}\hfill\begin{minipage}[t]{0.194\textwidth}\centering\fontsize{7.4}{8.5}\selectfont\strut Mip-Splatting\par\vspace{1pt}\setlength{\unitlength}{\dimexpr\linewidth/1000\relax}\begin{picture}(1000,562.500)\put(0,0){\includegraphics[width=\linewidth,bb=0 0 1344 624,viewport=344 20 664 200,clip]{figures/small_motion/frogs/rgb.jpg}}\put(338.542,93.750){\color{suppcyan}\linethickness{0.45pt}\framebox(145.833,145.833){} }\put(619.792,119.792){\color{supporange}\linethickness{0.45pt}\framebox(145.833,145.833){} }\end{picture}\par\vspace{2pt}\hbox to\linewidth{{\setlength{\fboxsep}{0pt}\color{suppcyan}\fbox{\includegraphics[width=0.478\linewidth,bb=0 0 960 1200,viewport=488 248 712 472,clip]{figures/small_motion/frogs/rgb_details.jpg}}}\hfill{\setlength{\fboxsep}{0pt}\color{supporange}\fbox{\includegraphics[width=0.478\linewidth,bb=0 0 960 1200,viewport=728 248 952 472,clip]{figures/small_motion/frogs/rgb_details.jpg}}}}\end{minipage}\hfill\begin{minipage}[t]{0.194\textwidth}\centering\fontsize{7.4}{8.5}\selectfont\strut \textbf{NBGS}\par\vspace{1pt}\setlength{\unitlength}{\dimexpr\linewidth/1000\relax}\begin{picture}(1000,562.500)\put(0,0){\includegraphics[width=\linewidth,bb=0 0 1344 624,viewport=680 20 1000 200,clip]{figures/small_motion/frogs/rgb.jpg}}\put(338.542,93.750){\color{suppcyan}\linethickness{0.45pt}\framebox(145.833,145.833){} }\put(619.792,119.792){\color{supporange}\linethickness{0.45pt}\framebox(145.833,145.833){} }\end{picture}\par\vspace{2pt}\hbox to\linewidth{{\setlength{\fboxsep}{0pt}\color{suppcyan}\fbox{\includegraphics[width=0.478\linewidth,bb=0 0 960 1200,viewport=8 8 232 232,clip]{figures/small_motion/frogs/rgb_details.jpg}}}\hfill{\setlength{\fboxsep}{0pt}\color{supporange}\fbox{\includegraphics[width=0.478\linewidth,bb=0 0 960 1200,viewport=248 8 472 232,clip]{figures/small_motion/frogs/rgb_details.jpg}}}}\end{minipage}\par\vspace{4pt}
\noindent{\small\bfseries Relative rendered depth}\par\vspace{2pt}
\noindent \begin{minipage}[t]{0.194\textwidth}\centering\fontsize{7.4}{8.5}\selectfont\strut Relative depth\par\vspace{4pt}\raggedright Per-method p02--p98. Black: invalid. No depth ground truth or metric units.\par\vspace{5pt}\includegraphics[width=\linewidth,height=5pt]{figures/legends/small_motion_depth_scale.png}\par near\hfill far\end{minipage}\hfill\begin{minipage}[t]{0.194\textwidth}\centering\fontsize{7.4}{8.5}\selectfont\strut Vanilla original\par\vspace{1pt}\includegraphics[width=\linewidth,bb=0 0 1344 624,viewport=8 436 328 616,clip]{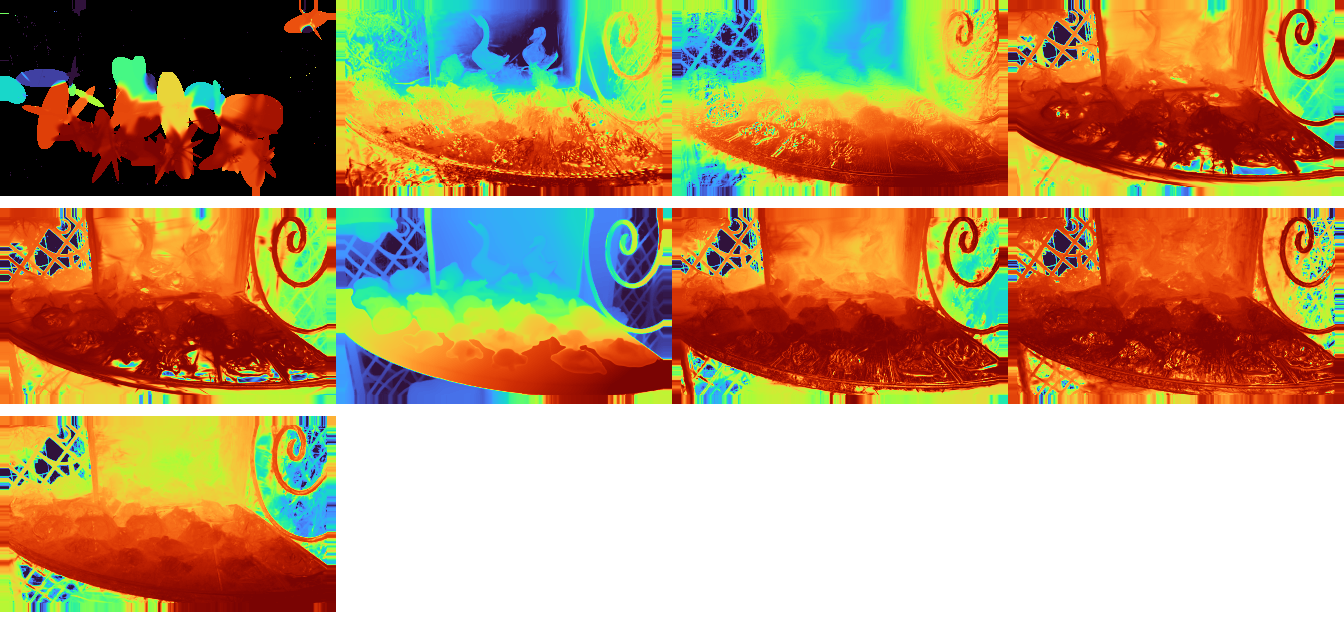}\end{minipage}\hfill\begin{minipage}[t]{0.194\textwidth}\centering\fontsize{7.4}{8.5}\selectfont\strut Small GS\par\vspace{1pt}\includegraphics[width=\linewidth,bb=0 0 1344 624,viewport=344 436 664 616,clip]{figures/small_motion/frogs/depth.png}\end{minipage}\hfill\begin{minipage}[t]{0.194\textwidth}\centering\fontsize{7.4}{8.5}\selectfont\strut InstantSplat\par\vspace{1pt}\includegraphics[width=\linewidth,bb=0 0 1344 624,viewport=680 436 1000 616,clip]{figures/small_motion/frogs/depth.png}\end{minipage}\hfill\begin{minipage}[t]{0.194\textwidth}\centering\fontsize{7.4}{8.5}\selectfont\strut DDDrop\par\vspace{1pt}\includegraphics[width=\linewidth,bb=0 0 1344 624,viewport=1016 436 1336 616,clip]{figures/small_motion/frogs/depth.png}\end{minipage}\par\vspace{4pt}
\noindent \begin{minipage}[t]{0.194\textwidth}\centering\fontsize{7.4}{8.5}\selectfont\strut DDGS\par\vspace{1pt}\includegraphics[width=\linewidth,bb=0 0 1344 624,viewport=8 228 328 408,clip]{figures/small_motion/frogs/depth.png}\end{minipage}\hfill\begin{minipage}[t]{0.194\textwidth}\centering\fontsize{7.4}{8.5}\selectfont\strut G4Splat\par\vspace{1pt}\includegraphics[width=\linewidth,bb=0 0 1344 624,viewport=344 228 664 408,clip]{figures/small_motion/frogs/depth.png}\end{minipage}\hfill\begin{minipage}[t]{0.194\textwidth}\centering\fontsize{7.4}{8.5}\selectfont\strut Vanilla$^{\dagger}$\par\vspace{1pt}\includegraphics[width=\linewidth,bb=0 0 1344 624,viewport=680 228 1000 408,clip]{figures/small_motion/frogs/depth.png}\end{minipage}\hfill\begin{minipage}[t]{0.194\textwidth}\centering\fontsize{7.4}{8.5}\selectfont\strut Mip-Splatting\par\vspace{1pt}\includegraphics[width=\linewidth,bb=0 0 1344 624,viewport=1016 228 1336 408,clip]{figures/small_motion/frogs/depth.png}\end{minipage}\hfill\begin{minipage}[t]{0.194\textwidth}\centering\fontsize{7.4}{8.5}\selectfont\strut \textbf{NBGS}\par\vspace{1pt}\includegraphics[width=\linewidth,bb=0 0 1344 624,viewport=8 20 328 200,clip]{figures/small_motion/frogs/depth.png}\end{minipage}\par\vspace{4pt}
\noindent{\small\bfseries Absolute RGB error}\par\vspace{2pt}
\noindent \begin{minipage}[t]{0.194\textwidth}\centering\fontsize{7.4}{8.5}\selectfont\strut RGB error, $[0,3]$\par\vspace{4pt}\raggedright Channel-summed absolute error. Method-matched RGB reference.\par\vspace{5pt}\includegraphics[width=\linewidth,height=5pt]{figures/legends/small_motion_error_scale.png}\par 0\hfill 3\end{minipage}\hfill\begin{minipage}[t]{0.194\textwidth}\centering\fontsize{7.4}{8.5}\selectfont\strut Vanilla original\par\vspace{1pt}\includegraphics[width=\linewidth,bb=0 0 1344 624,viewport=8 436 328 616,clip]{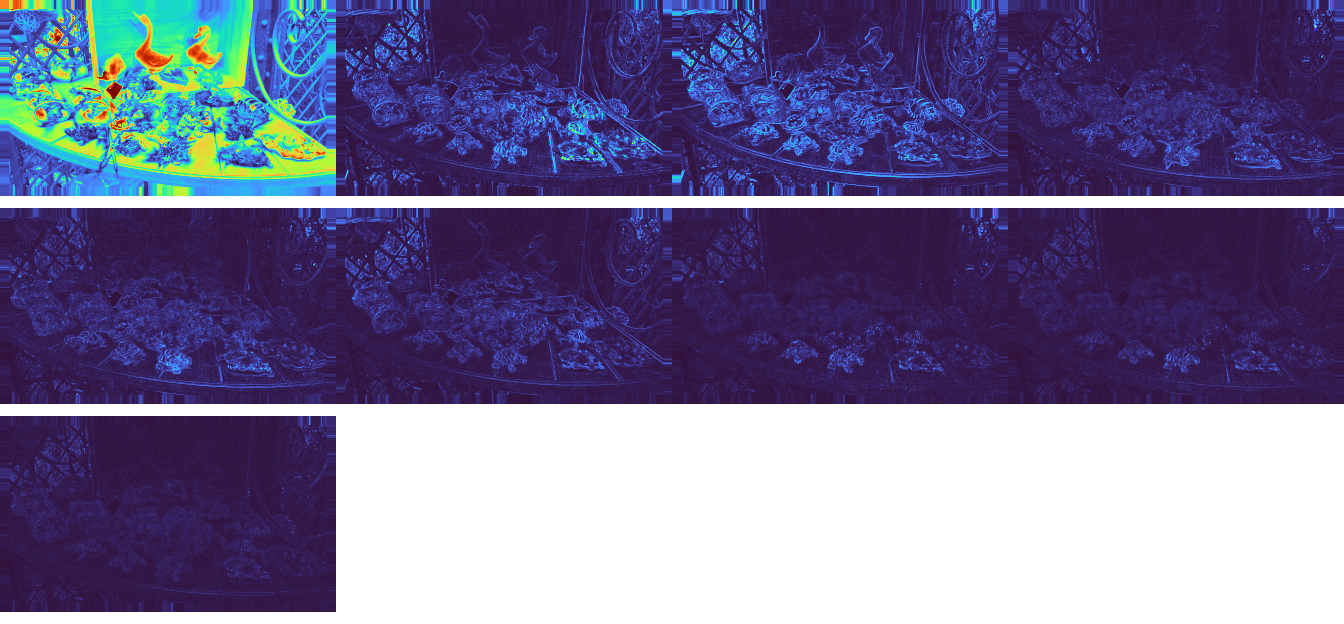}\end{minipage}\hfill\begin{minipage}[t]{0.194\textwidth}\centering\fontsize{7.4}{8.5}\selectfont\strut Small GS\par\vspace{1pt}\includegraphics[width=\linewidth,bb=0 0 1344 624,viewport=344 436 664 616,clip]{figures/small_motion/frogs/rgb_error.png}\end{minipage}\hfill\begin{minipage}[t]{0.194\textwidth}\centering\fontsize{7.4}{8.5}\selectfont\strut InstantSplat\par\vspace{1pt}\includegraphics[width=\linewidth,bb=0 0 1344 624,viewport=680 436 1000 616,clip]{figures/small_motion/frogs/rgb_error.png}\end{minipage}\hfill\begin{minipage}[t]{0.194\textwidth}\centering\fontsize{7.4}{8.5}\selectfont\strut DDDrop\par\vspace{1pt}\includegraphics[width=\linewidth,bb=0 0 1344 624,viewport=1016 436 1336 616,clip]{figures/small_motion/frogs/rgb_error.png}\end{minipage}\par\vspace{4pt}
\noindent \begin{minipage}[t]{0.194\textwidth}\centering\fontsize{7.4}{8.5}\selectfont\strut DDGS\par\vspace{1pt}\includegraphics[width=\linewidth,bb=0 0 1344 624,viewport=8 228 328 408,clip]{figures/small_motion/frogs/rgb_error.png}\end{minipage}\hfill\begin{minipage}[t]{0.194\textwidth}\centering\fontsize{7.4}{8.5}\selectfont\strut G4Splat\par\vspace{1pt}\includegraphics[width=\linewidth,bb=0 0 1344 624,viewport=344 228 664 408,clip]{figures/small_motion/frogs/rgb_error.png}\end{minipage}\hfill\begin{minipage}[t]{0.194\textwidth}\centering\fontsize{7.4}{8.5}\selectfont\strut Vanilla$^{\dagger}$\par\vspace{1pt}\includegraphics[width=\linewidth,bb=0 0 1344 624,viewport=680 228 1000 408,clip]{figures/small_motion/frogs/rgb_error.png}\end{minipage}\hfill\begin{minipage}[t]{0.194\textwidth}\centering\fontsize{7.4}{8.5}\selectfont\strut Mip-Splatting\par\vspace{1pt}\includegraphics[width=\linewidth,bb=0 0 1344 624,viewport=1016 228 1336 408,clip]{figures/small_motion/frogs/rgb_error.png}\end{minipage}\hfill\begin{minipage}[t]{0.194\textwidth}\centering\fontsize{7.4}{8.5}\selectfont\strut \textbf{NBGS}\par\vspace{1pt}\includegraphics[width=\linewidth,bb=0 0 1344 624,viewport=8 20 328 200,clip]{figures/small_motion/frogs/rgb_error.png}\end{minipage}\par\vspace{4pt}
\end{minipage}\end{lrbox}
\PlaceSuppScene
\clearpage
\pdfbookmark[3]{Louvre}{supp.scene.louvre}
\begin{lrbox}{\SuppSceneBox}\begin{minipage}{\textwidth}
\noindent{\large\bfseries Louvre}\hfill{\small\color{black!65}Small-Motion dataset}\par\vspace{3pt}
\noindent{\small\bfseries Rendered RGB}\par\vspace{2pt}
\noindent \begin{minipage}[t]{0.194\textwidth}\centering\fontsize{7.4}{8.5}\selectfont\strut Original RGB\par\vspace{1pt}\includegraphics[width=\linewidth,bb=0 0 1344 624,viewport=8 436 328 616,clip]{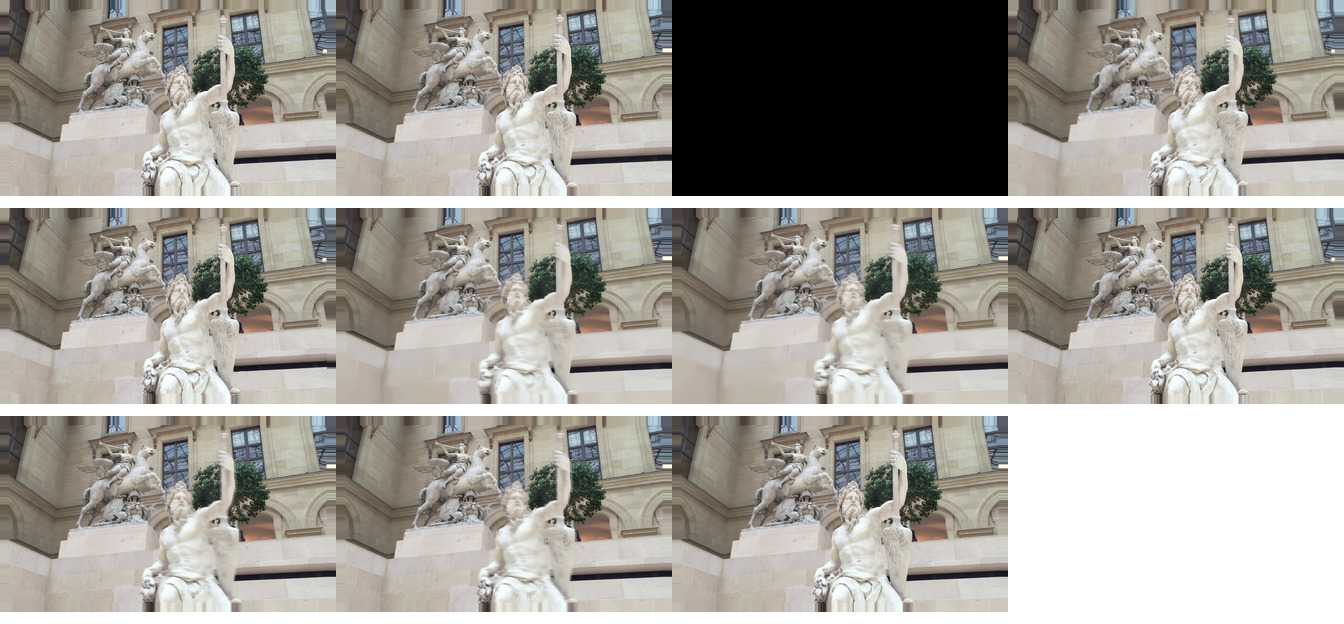}\par\vspace{2pt}\strut SfSM-undistorted RGB\par\vspace{1pt}\includegraphics[width=\linewidth,bb=0 0 1344 624,viewport=344 436 664 616,clip]{figures/small_motion/louvre/rgb.jpg}\end{minipage}\hfill\begin{minipage}[t]{0.194\textwidth}\centering\fontsize{7.4}{8.5}\selectfont\strut Vanilla original\par\vspace{1pt}\setlength{\unitlength}{\dimexpr\linewidth/1000\relax}\begin{picture}(1000,562.500)\put(0,0){\includegraphics[width=\linewidth,bb=0 0 1344 624,viewport=680 436 1000 616,clip]{figures/small_motion/louvre/rgb.jpg}}\put(515.625,270.833){\color{suppcyan}\linethickness{0.45pt}\framebox(83.333,83.333){} }\put(638.021,403.646){\color{supporange}\linethickness{0.45pt}\framebox(72.917,72.917){} }\end{picture}\par\vspace{2pt}\hbox to\linewidth{{\setlength{\fboxsep}{0pt}\color{suppcyan}\fbox{\includegraphics[width=0.478\linewidth,bb=0 0 704 880,viewport=8 712 168 872,clip]{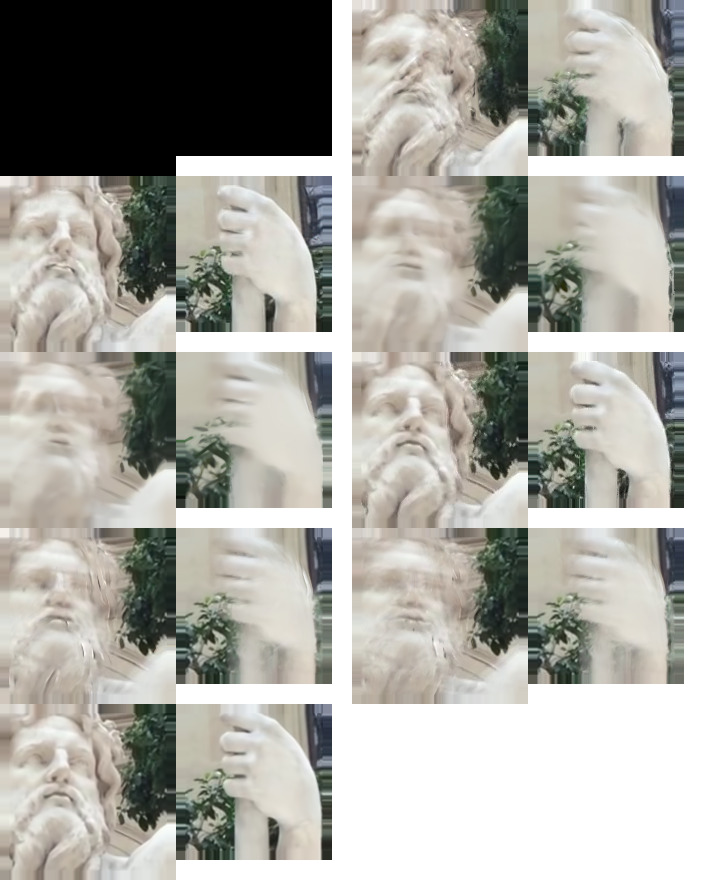}}}\hfill{\setlength{\fboxsep}{0pt}\color{supporange}\fbox{\includegraphics[width=0.478\linewidth,bb=0 0 704 880,viewport=184 732 324 872,clip]{figures/small_motion/louvre/rgb_details.jpg}}}}\end{minipage}\hfill\begin{minipage}[t]{0.194\textwidth}\centering\fontsize{7.4}{8.5}\selectfont\strut Small GS\par\vspace{1pt}\setlength{\unitlength}{\dimexpr\linewidth/1000\relax}\begin{picture}(1000,562.500)\put(0,0){\includegraphics[width=\linewidth,bb=0 0 1344 624,viewport=1016 436 1336 616,clip]{figures/small_motion/louvre/rgb.jpg}}\put(515.625,270.833){\color{suppcyan}\linethickness{0.45pt}\framebox(83.333,83.333){} }\put(638.021,403.646){\color{supporange}\linethickness{0.45pt}\framebox(72.917,72.917){} }\end{picture}\par\vspace{2pt}\hbox to\linewidth{{\setlength{\fboxsep}{0pt}\color{suppcyan}\fbox{\includegraphics[width=0.478\linewidth,bb=0 0 704 880,viewport=360 712 520 872,clip]{figures/small_motion/louvre/rgb_details.jpg}}}\hfill{\setlength{\fboxsep}{0pt}\color{supporange}\fbox{\includegraphics[width=0.478\linewidth,bb=0 0 704 880,viewport=536 732 676 872,clip]{figures/small_motion/louvre/rgb_details.jpg}}}}\end{minipage}\hfill\begin{minipage}[t]{0.194\textwidth}\centering\fontsize{7.4}{8.5}\selectfont\strut InstantSplat\par\vspace{1pt}\setlength{\unitlength}{\dimexpr\linewidth/1000\relax}\begin{picture}(1000,562.500)\put(0,0){\includegraphics[width=\linewidth,bb=0 0 1344 624,viewport=8 228 328 408,clip]{figures/small_motion/louvre/rgb.jpg}}\put(515.625,270.833){\color{suppcyan}\linethickness{0.45pt}\framebox(83.333,83.333){} }\put(638.021,403.646){\color{supporange}\linethickness{0.45pt}\framebox(72.917,72.917){} }\end{picture}\par\vspace{2pt}\hbox to\linewidth{{\setlength{\fboxsep}{0pt}\color{suppcyan}\fbox{\includegraphics[width=0.478\linewidth,bb=0 0 704 880,viewport=8 536 168 696,clip]{figures/small_motion/louvre/rgb_details.jpg}}}\hfill{\setlength{\fboxsep}{0pt}\color{supporange}\fbox{\includegraphics[width=0.478\linewidth,bb=0 0 704 880,viewport=184 556 324 696,clip]{figures/small_motion/louvre/rgb_details.jpg}}}}\end{minipage}\hfill\begin{minipage}[t]{0.194\textwidth}\centering\fontsize{7.4}{8.5}\selectfont\strut DDDrop\par\vspace{1pt}\setlength{\unitlength}{\dimexpr\linewidth/1000\relax}\begin{picture}(1000,562.500)\put(0,0){\includegraphics[width=\linewidth,bb=0 0 1344 624,viewport=344 228 664 408,clip]{figures/small_motion/louvre/rgb.jpg}}\put(515.625,270.833){\color{suppcyan}\linethickness{0.45pt}\framebox(83.333,83.333){} }\put(638.021,403.646){\color{supporange}\linethickness{0.45pt}\framebox(72.917,72.917){} }\end{picture}\par\vspace{2pt}\hbox to\linewidth{{\setlength{\fboxsep}{0pt}\color{suppcyan}\fbox{\includegraphics[width=0.478\linewidth,bb=0 0 704 880,viewport=360 536 520 696,clip]{figures/small_motion/louvre/rgb_details.jpg}}}\hfill{\setlength{\fboxsep}{0pt}\color{supporange}\fbox{\includegraphics[width=0.478\linewidth,bb=0 0 704 880,viewport=536 556 676 696,clip]{figures/small_motion/louvre/rgb_details.jpg}}}}\end{minipage}\par\vspace{4pt}
\noindent \begin{minipage}[t]{0.194\textwidth}\centering\fontsize{7.4}{8.5}\selectfont\strut DDGS\par\vspace{1pt}\setlength{\unitlength}{\dimexpr\linewidth/1000\relax}\begin{picture}(1000,562.500)\put(0,0){\includegraphics[width=\linewidth,bb=0 0 1344 624,viewport=680 228 1000 408,clip]{figures/small_motion/louvre/rgb.jpg}}\put(515.625,270.833){\color{suppcyan}\linethickness{0.45pt}\framebox(83.333,83.333){} }\put(638.021,403.646){\color{supporange}\linethickness{0.45pt}\framebox(72.917,72.917){} }\end{picture}\par\vspace{2pt}\hbox to\linewidth{{\setlength{\fboxsep}{0pt}\color{suppcyan}\fbox{\includegraphics[width=0.478\linewidth,bb=0 0 704 880,viewport=8 360 168 520,clip]{figures/small_motion/louvre/rgb_details.jpg}}}\hfill{\setlength{\fboxsep}{0pt}\color{supporange}\fbox{\includegraphics[width=0.478\linewidth,bb=0 0 704 880,viewport=184 380 324 520,clip]{figures/small_motion/louvre/rgb_details.jpg}}}}\end{minipage}\hfill\begin{minipage}[t]{0.194\textwidth}\centering\fontsize{7.4}{8.5}\selectfont\strut G4Splat\par\vspace{1pt}\setlength{\unitlength}{\dimexpr\linewidth/1000\relax}\begin{picture}(1000,562.500)\put(0,0){\includegraphics[width=\linewidth,bb=0 0 1344 624,viewport=1016 228 1336 408,clip]{figures/small_motion/louvre/rgb.jpg}}\put(515.625,270.833){\color{suppcyan}\linethickness{0.45pt}\framebox(83.333,83.333){} }\put(638.021,403.646){\color{supporange}\linethickness{0.45pt}\framebox(72.917,72.917){} }\end{picture}\par\vspace{2pt}\hbox to\linewidth{{\setlength{\fboxsep}{0pt}\color{suppcyan}\fbox{\includegraphics[width=0.478\linewidth,bb=0 0 704 880,viewport=360 360 520 520,clip]{figures/small_motion/louvre/rgb_details.jpg}}}\hfill{\setlength{\fboxsep}{0pt}\color{supporange}\fbox{\includegraphics[width=0.478\linewidth,bb=0 0 704 880,viewport=536 380 676 520,clip]{figures/small_motion/louvre/rgb_details.jpg}}}}\end{minipage}\hfill\begin{minipage}[t]{0.194\textwidth}\centering\fontsize{7.4}{8.5}\selectfont\strut Vanilla$^{\dagger}$\par\vspace{1pt}\setlength{\unitlength}{\dimexpr\linewidth/1000\relax}\begin{picture}(1000,562.500)\put(0,0){\includegraphics[width=\linewidth,bb=0 0 1344 624,viewport=8 20 328 200,clip]{figures/small_motion/louvre/rgb.jpg}}\put(515.625,270.833){\color{suppcyan}\linethickness{0.45pt}\framebox(83.333,83.333){} }\put(638.021,403.646){\color{supporange}\linethickness{0.45pt}\framebox(72.917,72.917){} }\end{picture}\par\vspace{2pt}\hbox to\linewidth{{\setlength{\fboxsep}{0pt}\color{suppcyan}\fbox{\includegraphics[width=0.478\linewidth,bb=0 0 704 880,viewport=8 184 168 344,clip]{figures/small_motion/louvre/rgb_details.jpg}}}\hfill{\setlength{\fboxsep}{0pt}\color{supporange}\fbox{\includegraphics[width=0.478\linewidth,bb=0 0 704 880,viewport=184 204 324 344,clip]{figures/small_motion/louvre/rgb_details.jpg}}}}\end{minipage}\hfill\begin{minipage}[t]{0.194\textwidth}\centering\fontsize{7.4}{8.5}\selectfont\strut Mip-Splatting\par\vspace{1pt}\setlength{\unitlength}{\dimexpr\linewidth/1000\relax}\begin{picture}(1000,562.500)\put(0,0){\includegraphics[width=\linewidth,bb=0 0 1344 624,viewport=344 20 664 200,clip]{figures/small_motion/louvre/rgb.jpg}}\put(515.625,270.833){\color{suppcyan}\linethickness{0.45pt}\framebox(83.333,83.333){} }\put(638.021,403.646){\color{supporange}\linethickness{0.45pt}\framebox(72.917,72.917){} }\end{picture}\par\vspace{2pt}\hbox to\linewidth{{\setlength{\fboxsep}{0pt}\color{suppcyan}\fbox{\includegraphics[width=0.478\linewidth,bb=0 0 704 880,viewport=360 184 520 344,clip]{figures/small_motion/louvre/rgb_details.jpg}}}\hfill{\setlength{\fboxsep}{0pt}\color{supporange}\fbox{\includegraphics[width=0.478\linewidth,bb=0 0 704 880,viewport=536 204 676 344,clip]{figures/small_motion/louvre/rgb_details.jpg}}}}\end{minipage}\hfill\begin{minipage}[t]{0.194\textwidth}\centering\fontsize{7.4}{8.5}\selectfont\strut \textbf{NBGS}\par\vspace{1pt}\setlength{\unitlength}{\dimexpr\linewidth/1000\relax}\begin{picture}(1000,562.500)\put(0,0){\includegraphics[width=\linewidth,bb=0 0 1344 624,viewport=680 20 1000 200,clip]{figures/small_motion/louvre/rgb.jpg}}\put(515.625,270.833){\color{suppcyan}\linethickness{0.45pt}\framebox(83.333,83.333){} }\put(638.021,403.646){\color{supporange}\linethickness{0.45pt}\framebox(72.917,72.917){} }\end{picture}\par\vspace{2pt}\hbox to\linewidth{{\setlength{\fboxsep}{0pt}\color{suppcyan}\fbox{\includegraphics[width=0.478\linewidth,bb=0 0 704 880,viewport=8 8 168 168,clip]{figures/small_motion/louvre/rgb_details.jpg}}}\hfill{\setlength{\fboxsep}{0pt}\color{supporange}\fbox{\includegraphics[width=0.478\linewidth,bb=0 0 704 880,viewport=184 28 324 168,clip]{figures/small_motion/louvre/rgb_details.jpg}}}}\end{minipage}\par\vspace{4pt}
\noindent{\small\bfseries Relative rendered depth}\par\vspace{2pt}
\noindent \begin{minipage}[t]{0.194\textwidth}\centering\fontsize{7.4}{8.5}\selectfont\strut Relative depth\par\vspace{4pt}\raggedright Per-method p02--p98. Black: invalid. No depth ground truth or metric units.\par\vspace{5pt}\includegraphics[width=\linewidth,height=5pt]{figures/legends/small_motion_depth_scale.png}\par near\hfill far\end{minipage}\hfill\begin{minipage}[t]{0.194\textwidth}\centering\fontsize{7.4}{8.5}\selectfont\strut Vanilla original\par\vspace{1pt}\includegraphics[width=\linewidth,bb=0 0 1344 624,viewport=8 436 328 616,clip]{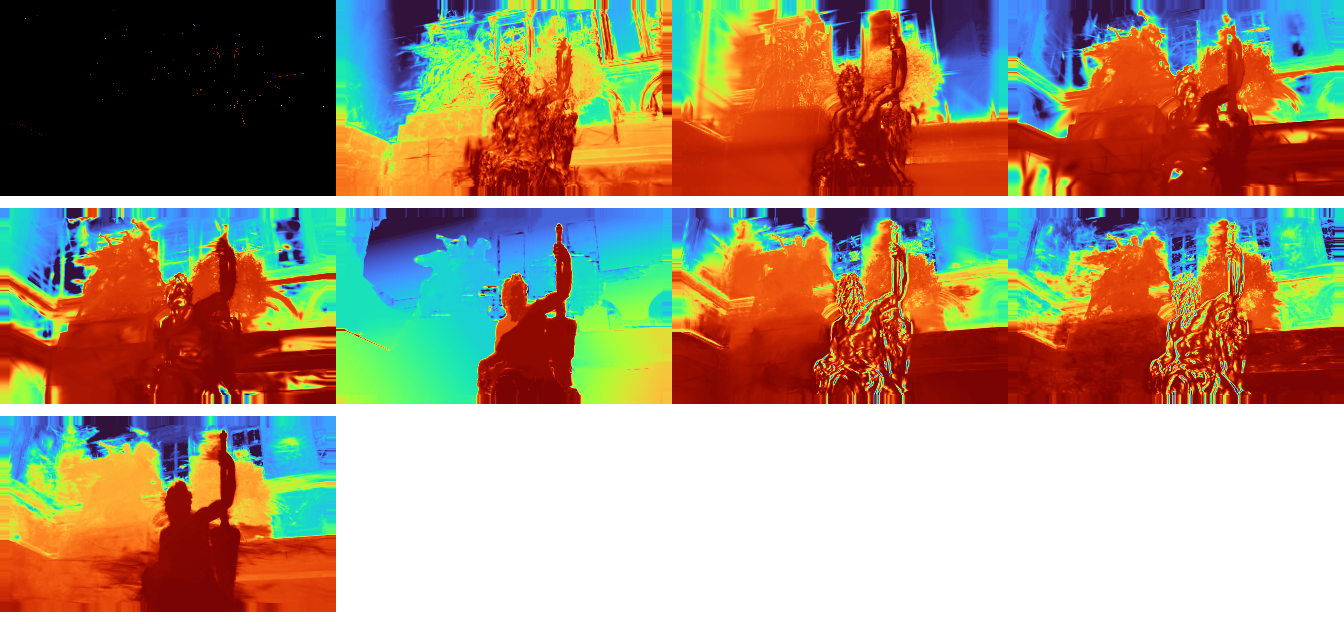}\end{minipage}\hfill\begin{minipage}[t]{0.194\textwidth}\centering\fontsize{7.4}{8.5}\selectfont\strut Small GS\par\vspace{1pt}\includegraphics[width=\linewidth,bb=0 0 1344 624,viewport=344 436 664 616,clip]{figures/small_motion/louvre/depth.png}\end{minipage}\hfill\begin{minipage}[t]{0.194\textwidth}\centering\fontsize{7.4}{8.5}\selectfont\strut InstantSplat\par\vspace{1pt}\includegraphics[width=\linewidth,bb=0 0 1344 624,viewport=680 436 1000 616,clip]{figures/small_motion/louvre/depth.png}\end{minipage}\hfill\begin{minipage}[t]{0.194\textwidth}\centering\fontsize{7.4}{8.5}\selectfont\strut DDDrop\par\vspace{1pt}\includegraphics[width=\linewidth,bb=0 0 1344 624,viewport=1016 436 1336 616,clip]{figures/small_motion/louvre/depth.png}\end{minipage}\par\vspace{4pt}
\noindent \begin{minipage}[t]{0.194\textwidth}\centering\fontsize{7.4}{8.5}\selectfont\strut DDGS\par\vspace{1pt}\includegraphics[width=\linewidth,bb=0 0 1344 624,viewport=8 228 328 408,clip]{figures/small_motion/louvre/depth.png}\end{minipage}\hfill\begin{minipage}[t]{0.194\textwidth}\centering\fontsize{7.4}{8.5}\selectfont\strut G4Splat\par\vspace{1pt}\includegraphics[width=\linewidth,bb=0 0 1344 624,viewport=344 228 664 408,clip]{figures/small_motion/louvre/depth.png}\end{minipage}\hfill\begin{minipage}[t]{0.194\textwidth}\centering\fontsize{7.4}{8.5}\selectfont\strut Vanilla$^{\dagger}$\par\vspace{1pt}\includegraphics[width=\linewidth,bb=0 0 1344 624,viewport=680 228 1000 408,clip]{figures/small_motion/louvre/depth.png}\end{minipage}\hfill\begin{minipage}[t]{0.194\textwidth}\centering\fontsize{7.4}{8.5}\selectfont\strut Mip-Splatting\par\vspace{1pt}\includegraphics[width=\linewidth,bb=0 0 1344 624,viewport=1016 228 1336 408,clip]{figures/small_motion/louvre/depth.png}\end{minipage}\hfill\begin{minipage}[t]{0.194\textwidth}\centering\fontsize{7.4}{8.5}\selectfont\strut \textbf{NBGS}\par\vspace{1pt}\includegraphics[width=\linewidth,bb=0 0 1344 624,viewport=8 20 328 200,clip]{figures/small_motion/louvre/depth.png}\end{minipage}\par\vspace{4pt}
\noindent{\small\bfseries Absolute RGB error}\par\vspace{2pt}
\noindent \begin{minipage}[t]{0.194\textwidth}\centering\fontsize{7.4}{8.5}\selectfont\strut RGB error, $[0,3]$\par\vspace{4pt}\raggedright Channel-summed absolute error. Method-matched RGB reference.\par\vspace{5pt}\includegraphics[width=\linewidth,height=5pt]{figures/legends/small_motion_error_scale.png}\par 0\hfill 3\end{minipage}\hfill\begin{minipage}[t]{0.194\textwidth}\centering\fontsize{7.4}{8.5}\selectfont\strut Vanilla original\par\vspace{1pt}\includegraphics[width=\linewidth,bb=0 0 1344 624,viewport=8 436 328 616,clip]{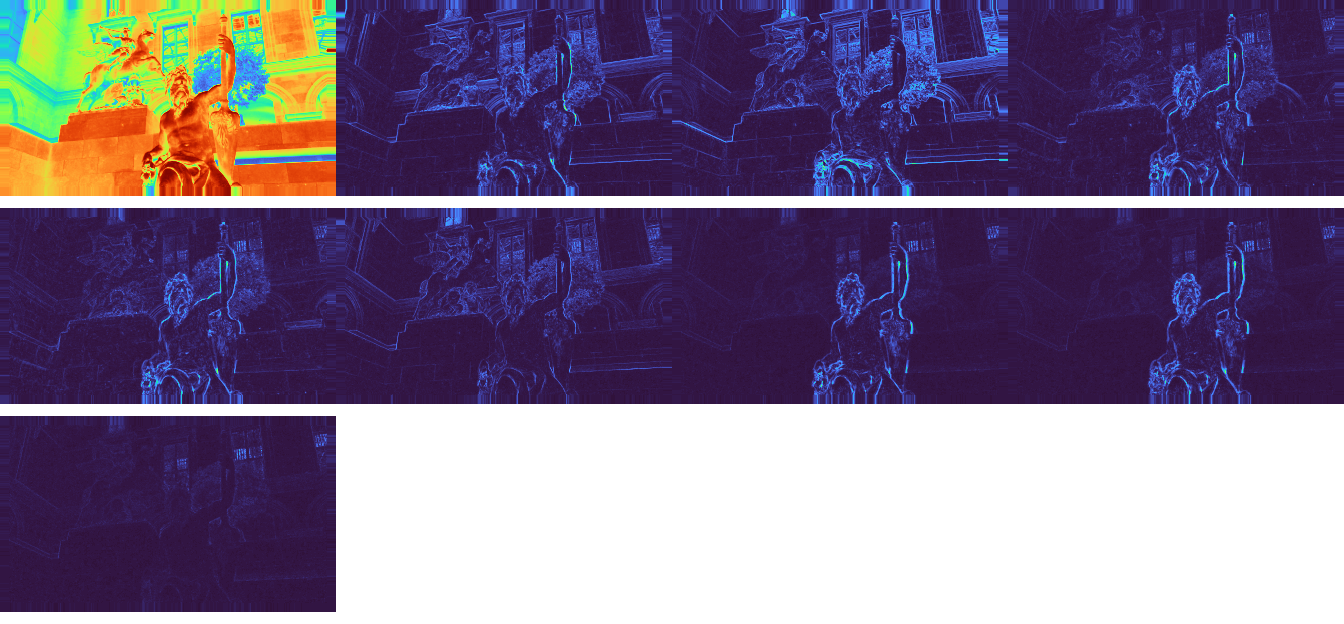}\end{minipage}\hfill\begin{minipage}[t]{0.194\textwidth}\centering\fontsize{7.4}{8.5}\selectfont\strut Small GS\par\vspace{1pt}\includegraphics[width=\linewidth,bb=0 0 1344 624,viewport=344 436 664 616,clip]{figures/small_motion/louvre/rgb_error.png}\end{minipage}\hfill\begin{minipage}[t]{0.194\textwidth}\centering\fontsize{7.4}{8.5}\selectfont\strut InstantSplat\par\vspace{1pt}\includegraphics[width=\linewidth,bb=0 0 1344 624,viewport=680 436 1000 616,clip]{figures/small_motion/louvre/rgb_error.png}\end{minipage}\hfill\begin{minipage}[t]{0.194\textwidth}\centering\fontsize{7.4}{8.5}\selectfont\strut DDDrop\par\vspace{1pt}\includegraphics[width=\linewidth,bb=0 0 1344 624,viewport=1016 436 1336 616,clip]{figures/small_motion/louvre/rgb_error.png}\end{minipage}\par\vspace{4pt}
\noindent \begin{minipage}[t]{0.194\textwidth}\centering\fontsize{7.4}{8.5}\selectfont\strut DDGS\par\vspace{1pt}\includegraphics[width=\linewidth,bb=0 0 1344 624,viewport=8 228 328 408,clip]{figures/small_motion/louvre/rgb_error.png}\end{minipage}\hfill\begin{minipage}[t]{0.194\textwidth}\centering\fontsize{7.4}{8.5}\selectfont\strut G4Splat\par\vspace{1pt}\includegraphics[width=\linewidth,bb=0 0 1344 624,viewport=344 228 664 408,clip]{figures/small_motion/louvre/rgb_error.png}\end{minipage}\hfill\begin{minipage}[t]{0.194\textwidth}\centering\fontsize{7.4}{8.5}\selectfont\strut Vanilla$^{\dagger}$\par\vspace{1pt}\includegraphics[width=\linewidth,bb=0 0 1344 624,viewport=680 228 1000 408,clip]{figures/small_motion/louvre/rgb_error.png}\end{minipage}\hfill\begin{minipage}[t]{0.194\textwidth}\centering\fontsize{7.4}{8.5}\selectfont\strut Mip-Splatting\par\vspace{1pt}\includegraphics[width=\linewidth,bb=0 0 1344 624,viewport=1016 228 1336 408,clip]{figures/small_motion/louvre/rgb_error.png}\end{minipage}\hfill\begin{minipage}[t]{0.194\textwidth}\centering\fontsize{7.4}{8.5}\selectfont\strut \textbf{NBGS}\par\vspace{1pt}\includegraphics[width=\linewidth,bb=0 0 1344 624,viewport=8 20 328 200,clip]{figures/small_motion/louvre/rgb_error.png}\end{minipage}\par\vspace{4pt}
\end{minipage}\end{lrbox}
\PlaceSuppScene
\clearpage
\pdfbookmark[3]{Notre-Dame}{supp.scene.notre_dame}
\begin{lrbox}{\SuppSceneBox}\begin{minipage}{\textwidth}
\noindent{\large\bfseries Notre-Dame}\hfill{\small\color{black!65}Small-Motion dataset}\par\vspace{3pt}
\noindent{\small\bfseries Rendered RGB}\par\vspace{2pt}
\noindent \begin{minipage}[t]{0.194\textwidth}\centering\fontsize{7.4}{8.5}\selectfont\strut Original RGB\par\vspace{1pt}\includegraphics[width=\linewidth,bb=0 0 1344 624,viewport=8 436 328 616,clip]{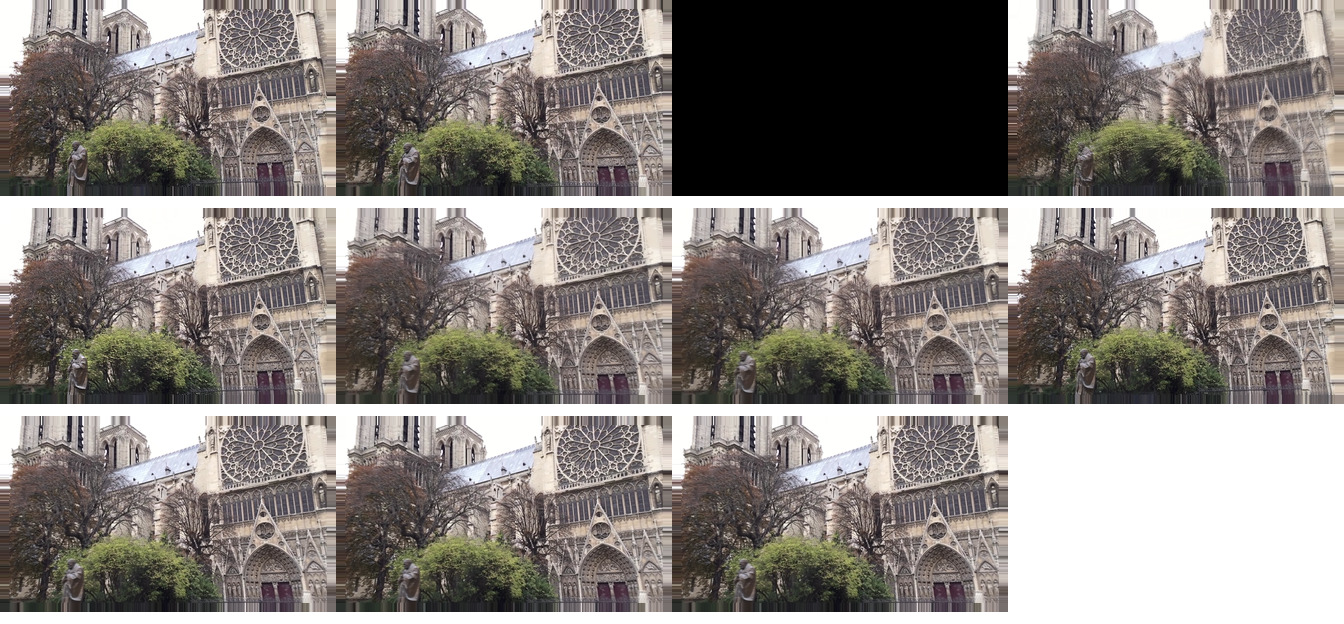}\par\vspace{2pt}\strut SfSM-undistorted RGB\par\vspace{1pt}\includegraphics[width=\linewidth,bb=0 0 1344 624,viewport=344 436 664 616,clip]{figures/small_motion/notre_dame/rgb.jpg}\end{minipage}\hfill\begin{minipage}[t]{0.194\textwidth}\centering\fontsize{7.4}{8.5}\selectfont\strut Vanilla original\par\vspace{1pt}\setlength{\unitlength}{\dimexpr\linewidth/1000\relax}\begin{picture}(1000,562.500)\put(0,0){\includegraphics[width=\linewidth,bb=0 0 1344 624,viewport=680 436 1000 616,clip]{figures/small_motion/notre_dame/rgb.jpg}}\put(679.688,343.750){\color{suppcyan}\linethickness{0.45pt}\framebox(203.125,203.125){} }\put(671.875,39.062){\color{supporange}\linethickness{0.45pt}\framebox(203.125,203.125){} }\end{picture}\par\vspace{2pt}\hbox to\linewidth{{\setlength{\fboxsep}{0pt}\color{suppcyan}\fbox{\includegraphics[width=0.478\linewidth,bb=0 0 960 1200,viewport=8 968 232 1192,clip]{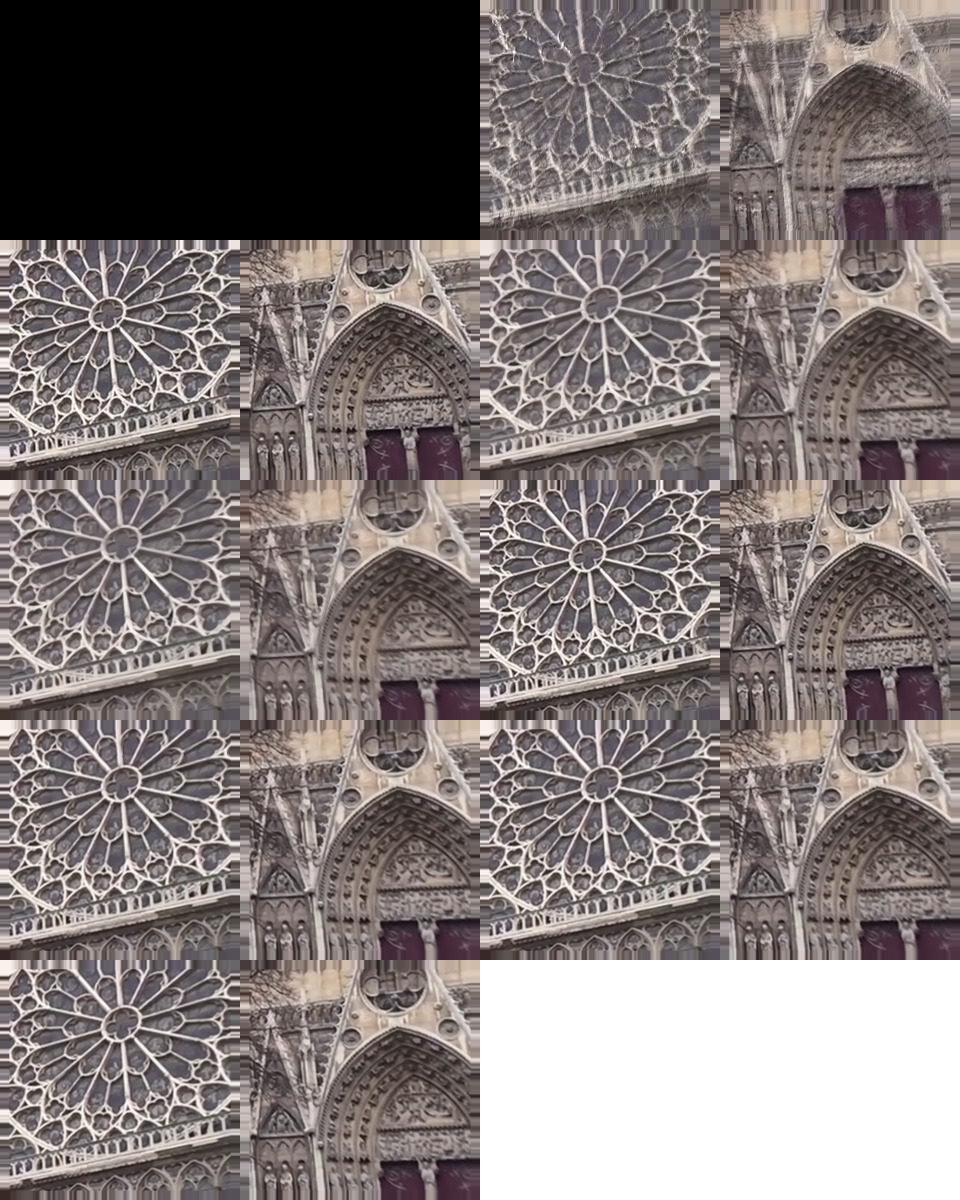}}}\hfill{\setlength{\fboxsep}{0pt}\color{supporange}\fbox{\includegraphics[width=0.478\linewidth,bb=0 0 960 1200,viewport=248 968 472 1192,clip]{figures/small_motion/notre_dame/rgb_details.jpg}}}}\end{minipage}\hfill\begin{minipage}[t]{0.194\textwidth}\centering\fontsize{7.4}{8.5}\selectfont\strut Small GS\par\vspace{1pt}\setlength{\unitlength}{\dimexpr\linewidth/1000\relax}\begin{picture}(1000,562.500)\put(0,0){\includegraphics[width=\linewidth,bb=0 0 1344 624,viewport=1016 436 1336 616,clip]{figures/small_motion/notre_dame/rgb.jpg}}\put(679.688,343.750){\color{suppcyan}\linethickness{0.45pt}\framebox(203.125,203.125){} }\put(671.875,39.062){\color{supporange}\linethickness{0.45pt}\framebox(203.125,203.125){} }\end{picture}\par\vspace{2pt}\hbox to\linewidth{{\setlength{\fboxsep}{0pt}\color{suppcyan}\fbox{\includegraphics[width=0.478\linewidth,bb=0 0 960 1200,viewport=488 968 712 1192,clip]{figures/small_motion/notre_dame/rgb_details.jpg}}}\hfill{\setlength{\fboxsep}{0pt}\color{supporange}\fbox{\includegraphics[width=0.478\linewidth,bb=0 0 960 1200,viewport=728 968 952 1192,clip]{figures/small_motion/notre_dame/rgb_details.jpg}}}}\end{minipage}\hfill\begin{minipage}[t]{0.194\textwidth}\centering\fontsize{7.4}{8.5}\selectfont\strut InstantSplat\par\vspace{1pt}\setlength{\unitlength}{\dimexpr\linewidth/1000\relax}\begin{picture}(1000,562.500)\put(0,0){\includegraphics[width=\linewidth,bb=0 0 1344 624,viewport=8 228 328 408,clip]{figures/small_motion/notre_dame/rgb.jpg}}\put(679.688,343.750){\color{suppcyan}\linethickness{0.45pt}\framebox(203.125,203.125){} }\put(671.875,39.062){\color{supporange}\linethickness{0.45pt}\framebox(203.125,203.125){} }\end{picture}\par\vspace{2pt}\hbox to\linewidth{{\setlength{\fboxsep}{0pt}\color{suppcyan}\fbox{\includegraphics[width=0.478\linewidth,bb=0 0 960 1200,viewport=8 728 232 952,clip]{figures/small_motion/notre_dame/rgb_details.jpg}}}\hfill{\setlength{\fboxsep}{0pt}\color{supporange}\fbox{\includegraphics[width=0.478\linewidth,bb=0 0 960 1200,viewport=248 728 472 952,clip]{figures/small_motion/notre_dame/rgb_details.jpg}}}}\end{minipage}\hfill\begin{minipage}[t]{0.194\textwidth}\centering\fontsize{7.4}{8.5}\selectfont\strut DDDrop\par\vspace{1pt}\setlength{\unitlength}{\dimexpr\linewidth/1000\relax}\begin{picture}(1000,562.500)\put(0,0){\includegraphics[width=\linewidth,bb=0 0 1344 624,viewport=344 228 664 408,clip]{figures/small_motion/notre_dame/rgb.jpg}}\put(679.688,343.750){\color{suppcyan}\linethickness{0.45pt}\framebox(203.125,203.125){} }\put(671.875,39.062){\color{supporange}\linethickness{0.45pt}\framebox(203.125,203.125){} }\end{picture}\par\vspace{2pt}\hbox to\linewidth{{\setlength{\fboxsep}{0pt}\color{suppcyan}\fbox{\includegraphics[width=0.478\linewidth,bb=0 0 960 1200,viewport=488 728 712 952,clip]{figures/small_motion/notre_dame/rgb_details.jpg}}}\hfill{\setlength{\fboxsep}{0pt}\color{supporange}\fbox{\includegraphics[width=0.478\linewidth,bb=0 0 960 1200,viewport=728 728 952 952,clip]{figures/small_motion/notre_dame/rgb_details.jpg}}}}\end{minipage}\par\vspace{4pt}
\noindent \begin{minipage}[t]{0.194\textwidth}\centering\fontsize{7.4}{8.5}\selectfont\strut DDGS\par\vspace{1pt}\setlength{\unitlength}{\dimexpr\linewidth/1000\relax}\begin{picture}(1000,562.500)\put(0,0){\includegraphics[width=\linewidth,bb=0 0 1344 624,viewport=680 228 1000 408,clip]{figures/small_motion/notre_dame/rgb.jpg}}\put(679.688,343.750){\color{suppcyan}\linethickness{0.45pt}\framebox(203.125,203.125){} }\put(671.875,39.062){\color{supporange}\linethickness{0.45pt}\framebox(203.125,203.125){} }\end{picture}\par\vspace{2pt}\hbox to\linewidth{{\setlength{\fboxsep}{0pt}\color{suppcyan}\fbox{\includegraphics[width=0.478\linewidth,bb=0 0 960 1200,viewport=8 488 232 712,clip]{figures/small_motion/notre_dame/rgb_details.jpg}}}\hfill{\setlength{\fboxsep}{0pt}\color{supporange}\fbox{\includegraphics[width=0.478\linewidth,bb=0 0 960 1200,viewport=248 488 472 712,clip]{figures/small_motion/notre_dame/rgb_details.jpg}}}}\end{minipage}\hfill\begin{minipage}[t]{0.194\textwidth}\centering\fontsize{7.4}{8.5}\selectfont\strut G4Splat\par\vspace{1pt}\setlength{\unitlength}{\dimexpr\linewidth/1000\relax}\begin{picture}(1000,562.500)\put(0,0){\includegraphics[width=\linewidth,bb=0 0 1344 624,viewport=1016 228 1336 408,clip]{figures/small_motion/notre_dame/rgb.jpg}}\put(679.688,343.750){\color{suppcyan}\linethickness{0.45pt}\framebox(203.125,203.125){} }\put(671.875,39.062){\color{supporange}\linethickness{0.45pt}\framebox(203.125,203.125){} }\end{picture}\par\vspace{2pt}\hbox to\linewidth{{\setlength{\fboxsep}{0pt}\color{suppcyan}\fbox{\includegraphics[width=0.478\linewidth,bb=0 0 960 1200,viewport=488 488 712 712,clip]{figures/small_motion/notre_dame/rgb_details.jpg}}}\hfill{\setlength{\fboxsep}{0pt}\color{supporange}\fbox{\includegraphics[width=0.478\linewidth,bb=0 0 960 1200,viewport=728 488 952 712,clip]{figures/small_motion/notre_dame/rgb_details.jpg}}}}\end{minipage}\hfill\begin{minipage}[t]{0.194\textwidth}\centering\fontsize{7.4}{8.5}\selectfont\strut Vanilla$^{\dagger}$\par\vspace{1pt}\setlength{\unitlength}{\dimexpr\linewidth/1000\relax}\begin{picture}(1000,562.500)\put(0,0){\includegraphics[width=\linewidth,bb=0 0 1344 624,viewport=8 20 328 200,clip]{figures/small_motion/notre_dame/rgb.jpg}}\put(679.688,343.750){\color{suppcyan}\linethickness{0.45pt}\framebox(203.125,203.125){} }\put(671.875,39.062){\color{supporange}\linethickness{0.45pt}\framebox(203.125,203.125){} }\end{picture}\par\vspace{2pt}\hbox to\linewidth{{\setlength{\fboxsep}{0pt}\color{suppcyan}\fbox{\includegraphics[width=0.478\linewidth,bb=0 0 960 1200,viewport=8 248 232 472,clip]{figures/small_motion/notre_dame/rgb_details.jpg}}}\hfill{\setlength{\fboxsep}{0pt}\color{supporange}\fbox{\includegraphics[width=0.478\linewidth,bb=0 0 960 1200,viewport=248 248 472 472,clip]{figures/small_motion/notre_dame/rgb_details.jpg}}}}\end{minipage}\hfill\begin{minipage}[t]{0.194\textwidth}\centering\fontsize{7.4}{8.5}\selectfont\strut Mip-Splatting\par\vspace{1pt}\setlength{\unitlength}{\dimexpr\linewidth/1000\relax}\begin{picture}(1000,562.500)\put(0,0){\includegraphics[width=\linewidth,bb=0 0 1344 624,viewport=344 20 664 200,clip]{figures/small_motion/notre_dame/rgb.jpg}}\put(679.688,343.750){\color{suppcyan}\linethickness{0.45pt}\framebox(203.125,203.125){} }\put(671.875,39.062){\color{supporange}\linethickness{0.45pt}\framebox(203.125,203.125){} }\end{picture}\par\vspace{2pt}\hbox to\linewidth{{\setlength{\fboxsep}{0pt}\color{suppcyan}\fbox{\includegraphics[width=0.478\linewidth,bb=0 0 960 1200,viewport=488 248 712 472,clip]{figures/small_motion/notre_dame/rgb_details.jpg}}}\hfill{\setlength{\fboxsep}{0pt}\color{supporange}\fbox{\includegraphics[width=0.478\linewidth,bb=0 0 960 1200,viewport=728 248 952 472,clip]{figures/small_motion/notre_dame/rgb_details.jpg}}}}\end{minipage}\hfill\begin{minipage}[t]{0.194\textwidth}\centering\fontsize{7.4}{8.5}\selectfont\strut \textbf{NBGS}\par\vspace{1pt}\setlength{\unitlength}{\dimexpr\linewidth/1000\relax}\begin{picture}(1000,562.500)\put(0,0){\includegraphics[width=\linewidth,bb=0 0 1344 624,viewport=680 20 1000 200,clip]{figures/small_motion/notre_dame/rgb.jpg}}\put(679.688,343.750){\color{suppcyan}\linethickness{0.45pt}\framebox(203.125,203.125){} }\put(671.875,39.062){\color{supporange}\linethickness{0.45pt}\framebox(203.125,203.125){} }\end{picture}\par\vspace{2pt}\hbox to\linewidth{{\setlength{\fboxsep}{0pt}\color{suppcyan}\fbox{\includegraphics[width=0.478\linewidth,bb=0 0 960 1200,viewport=8 8 232 232,clip]{figures/small_motion/notre_dame/rgb_details.jpg}}}\hfill{\setlength{\fboxsep}{0pt}\color{supporange}\fbox{\includegraphics[width=0.478\linewidth,bb=0 0 960 1200,viewport=248 8 472 232,clip]{figures/small_motion/notre_dame/rgb_details.jpg}}}}\end{minipage}\par\vspace{4pt}
\noindent{\small\bfseries Relative rendered depth}\par\vspace{2pt}
\noindent \begin{minipage}[t]{0.194\textwidth}\centering\fontsize{7.4}{8.5}\selectfont\strut Relative depth\par\vspace{4pt}\raggedright Per-method p02--p98. Black: invalid. No depth ground truth or metric units.\par\vspace{5pt}\includegraphics[width=\linewidth,height=5pt]{figures/legends/small_motion_depth_scale.png}\par near\hfill far\end{minipage}\hfill\begin{minipage}[t]{0.194\textwidth}\centering\fontsize{7.4}{8.5}\selectfont\strut Vanilla original\par\vspace{1pt}\includegraphics[width=\linewidth,bb=0 0 1344 624,viewport=8 436 328 616,clip]{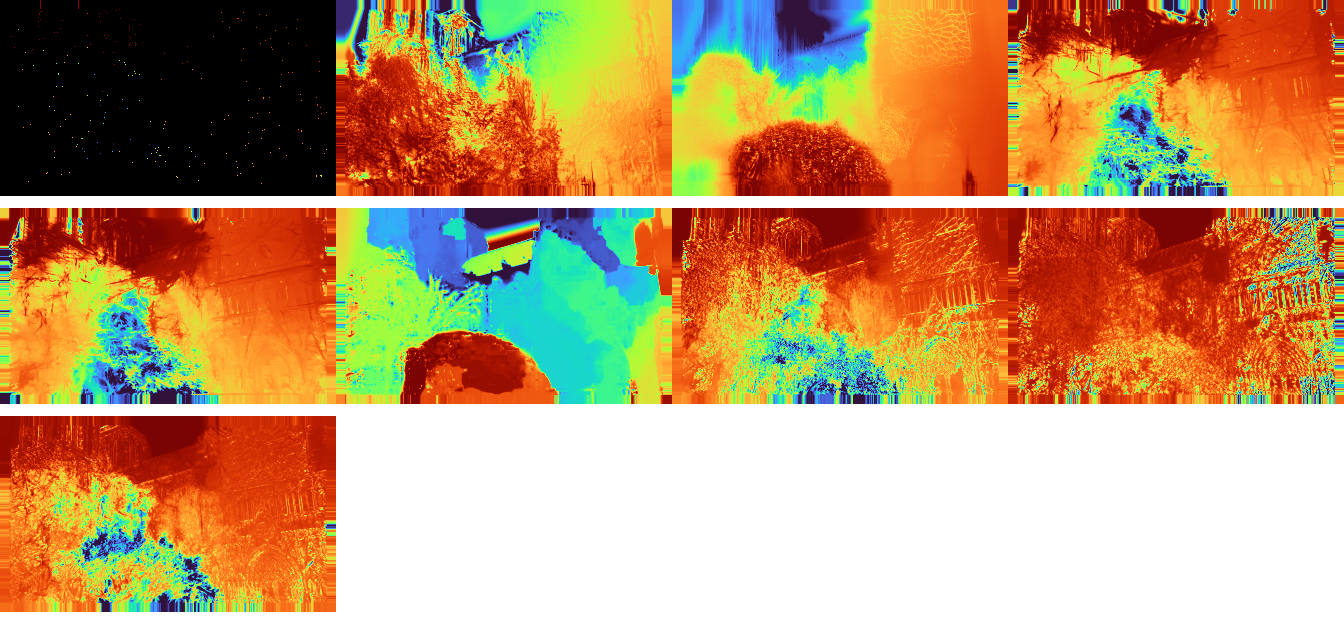}\end{minipage}\hfill\begin{minipage}[t]{0.194\textwidth}\centering\fontsize{7.4}{8.5}\selectfont\strut Small GS\par\vspace{1pt}\includegraphics[width=\linewidth,bb=0 0 1344 624,viewport=344 436 664 616,clip]{figures/small_motion/notre_dame/depth.png}\end{minipage}\hfill\begin{minipage}[t]{0.194\textwidth}\centering\fontsize{7.4}{8.5}\selectfont\strut InstantSplat\par\vspace{1pt}\includegraphics[width=\linewidth,bb=0 0 1344 624,viewport=680 436 1000 616,clip]{figures/small_motion/notre_dame/depth.png}\end{minipage}\hfill\begin{minipage}[t]{0.194\textwidth}\centering\fontsize{7.4}{8.5}\selectfont\strut DDDrop\par\vspace{1pt}\includegraphics[width=\linewidth,bb=0 0 1344 624,viewport=1016 436 1336 616,clip]{figures/small_motion/notre_dame/depth.png}\end{minipage}\par\vspace{4pt}
\noindent \begin{minipage}[t]{0.194\textwidth}\centering\fontsize{7.4}{8.5}\selectfont\strut DDGS\par\vspace{1pt}\includegraphics[width=\linewidth,bb=0 0 1344 624,viewport=8 228 328 408,clip]{figures/small_motion/notre_dame/depth.png}\end{minipage}\hfill\begin{minipage}[t]{0.194\textwidth}\centering\fontsize{7.4}{8.5}\selectfont\strut G4Splat\par\vspace{1pt}\includegraphics[width=\linewidth,bb=0 0 1344 624,viewport=344 228 664 408,clip]{figures/small_motion/notre_dame/depth.png}\end{minipage}\hfill\begin{minipage}[t]{0.194\textwidth}\centering\fontsize{7.4}{8.5}\selectfont\strut Vanilla$^{\dagger}$\par\vspace{1pt}\includegraphics[width=\linewidth,bb=0 0 1344 624,viewport=680 228 1000 408,clip]{figures/small_motion/notre_dame/depth.png}\end{minipage}\hfill\begin{minipage}[t]{0.194\textwidth}\centering\fontsize{7.4}{8.5}\selectfont\strut Mip-Splatting\par\vspace{1pt}\includegraphics[width=\linewidth,bb=0 0 1344 624,viewport=1016 228 1336 408,clip]{figures/small_motion/notre_dame/depth.png}\end{minipage}\hfill\begin{minipage}[t]{0.194\textwidth}\centering\fontsize{7.4}{8.5}\selectfont\strut \textbf{NBGS}\par\vspace{1pt}\includegraphics[width=\linewidth,bb=0 0 1344 624,viewport=8 20 328 200,clip]{figures/small_motion/notre_dame/depth.png}\end{minipage}\par\vspace{4pt}
\noindent{\small\bfseries Absolute RGB error}\par\vspace{2pt}
\noindent \begin{minipage}[t]{0.194\textwidth}\centering\fontsize{7.4}{8.5}\selectfont\strut RGB error, $[0,3]$\par\vspace{4pt}\raggedright Channel-summed absolute error. Method-matched RGB reference.\par\vspace{5pt}\includegraphics[width=\linewidth,height=5pt]{figures/legends/small_motion_error_scale.png}\par 0\hfill 3\end{minipage}\hfill\begin{minipage}[t]{0.194\textwidth}\centering\fontsize{7.4}{8.5}\selectfont\strut Vanilla original\par\vspace{1pt}\includegraphics[width=\linewidth,bb=0 0 1344 624,viewport=8 436 328 616,clip]{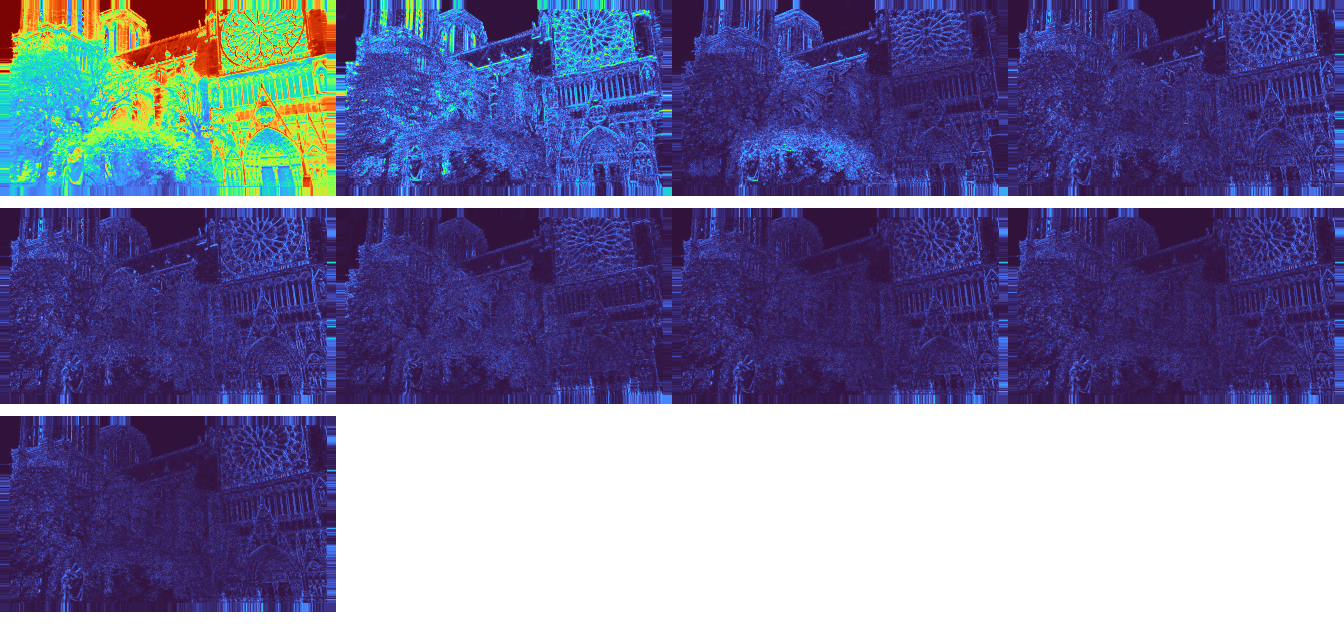}\end{minipage}\hfill\begin{minipage}[t]{0.194\textwidth}\centering\fontsize{7.4}{8.5}\selectfont\strut Small GS\par\vspace{1pt}\includegraphics[width=\linewidth,bb=0 0 1344 624,viewport=344 436 664 616,clip]{figures/small_motion/notre_dame/rgb_error.png}\end{minipage}\hfill\begin{minipage}[t]{0.194\textwidth}\centering\fontsize{7.4}{8.5}\selectfont\strut InstantSplat\par\vspace{1pt}\includegraphics[width=\linewidth,bb=0 0 1344 624,viewport=680 436 1000 616,clip]{figures/small_motion/notre_dame/rgb_error.png}\end{minipage}\hfill\begin{minipage}[t]{0.194\textwidth}\centering\fontsize{7.4}{8.5}\selectfont\strut DDDrop\par\vspace{1pt}\includegraphics[width=\linewidth,bb=0 0 1344 624,viewport=1016 436 1336 616,clip]{figures/small_motion/notre_dame/rgb_error.png}\end{minipage}\par\vspace{4pt}
\noindent \begin{minipage}[t]{0.194\textwidth}\centering\fontsize{7.4}{8.5}\selectfont\strut DDGS\par\vspace{1pt}\includegraphics[width=\linewidth,bb=0 0 1344 624,viewport=8 228 328 408,clip]{figures/small_motion/notre_dame/rgb_error.png}\end{minipage}\hfill\begin{minipage}[t]{0.194\textwidth}\centering\fontsize{7.4}{8.5}\selectfont\strut G4Splat\par\vspace{1pt}\includegraphics[width=\linewidth,bb=0 0 1344 624,viewport=344 228 664 408,clip]{figures/small_motion/notre_dame/rgb_error.png}\end{minipage}\hfill\begin{minipage}[t]{0.194\textwidth}\centering\fontsize{7.4}{8.5}\selectfont\strut Vanilla$^{\dagger}$\par\vspace{1pt}\includegraphics[width=\linewidth,bb=0 0 1344 624,viewport=680 228 1000 408,clip]{figures/small_motion/notre_dame/rgb_error.png}\end{minipage}\hfill\begin{minipage}[t]{0.194\textwidth}\centering\fontsize{7.4}{8.5}\selectfont\strut Mip-Splatting\par\vspace{1pt}\includegraphics[width=\linewidth,bb=0 0 1344 624,viewport=1016 228 1336 408,clip]{figures/small_motion/notre_dame/rgb_error.png}\end{minipage}\hfill\begin{minipage}[t]{0.194\textwidth}\centering\fontsize{7.4}{8.5}\selectfont\strut \textbf{NBGS}\par\vspace{1pt}\includegraphics[width=\linewidth,bb=0 0 1344 624,viewport=8 20 328 200,clip]{figures/small_motion/notre_dame/rgb_error.png}\end{minipage}\par\vspace{4pt}
\end{minipage}\end{lrbox}
\PlaceSuppScene
\clearpage
\pdfbookmark[3]{Plants}{supp.scene.plants}
\begin{lrbox}{\SuppSceneBox}\begin{minipage}{\textwidth}
\noindent{\large\bfseries Plants}\hfill{\small\color{black!65}Small-Motion dataset}\par\vspace{3pt}
\noindent{\small\bfseries Rendered RGB}\par\vspace{2pt}
\noindent \begin{minipage}[t]{0.194\textwidth}\centering\fontsize{7.4}{8.5}\selectfont\strut Original RGB\par\vspace{1pt}\includegraphics[width=\linewidth,bb=0 0 1344 624,viewport=8 436 328 616,clip]{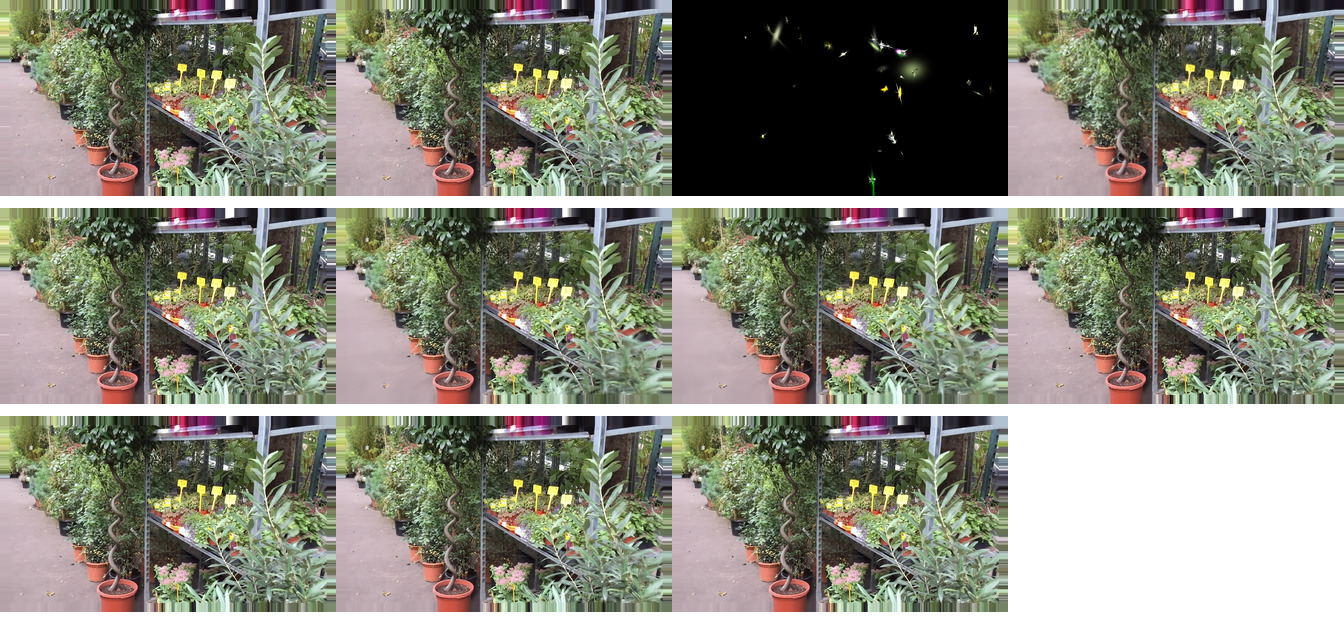}\par\vspace{2pt}\strut SfSM-undistorted RGB\par\vspace{1pt}\includegraphics[width=\linewidth,bb=0 0 1344 624,viewport=344 436 664 616,clip]{figures/small_motion/plants/rgb.jpg}\end{minipage}\hfill\begin{minipage}[t]{0.194\textwidth}\centering\fontsize{7.4}{8.5}\selectfont\strut Vanilla original\par\vspace{1pt}\setlength{\unitlength}{\dimexpr\linewidth/1000\relax}\begin{picture}(1000,562.500)\put(0,0){\includegraphics[width=\linewidth,bb=0 0 1344 624,viewport=680 436 1000 616,clip]{figures/small_motion/plants/rgb.jpg}}\put(550.000,0.000){\color{suppcyan}\linethickness{0.45pt}\framebox(125.000,125.000){} }\put(700.000,0.000){\color{supporange}\linethickness{0.45pt}\framebox(125.000,125.000){} }\end{picture}\par\vspace{2pt}\hbox to\linewidth{{\setlength{\fboxsep}{0pt}\color{suppcyan}\fbox{\includegraphics[width=0.478\linewidth,bb=0 0 704 880,viewport=8 712 168 872,clip]{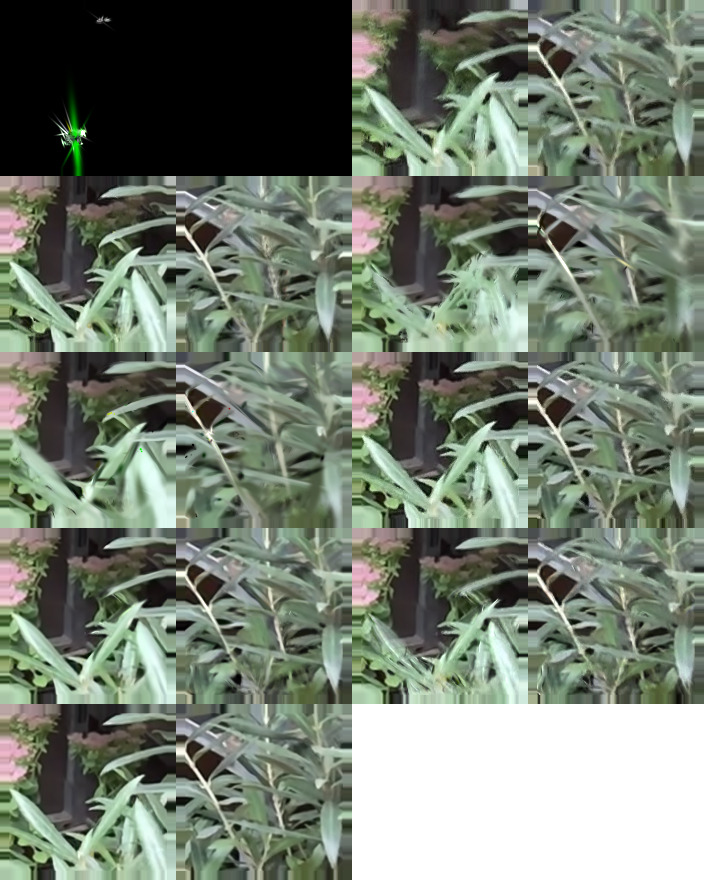}}}\hfill{\setlength{\fboxsep}{0pt}\color{supporange}\fbox{\includegraphics[width=0.478\linewidth,bb=0 0 704 880,viewport=184 712 344 872,clip]{figures/small_motion/plants/rgb_details.jpg}}}}\end{minipage}\hfill\begin{minipage}[t]{0.194\textwidth}\centering\fontsize{7.4}{8.5}\selectfont\strut Small GS\par\vspace{1pt}\setlength{\unitlength}{\dimexpr\linewidth/1000\relax}\begin{picture}(1000,562.500)\put(0,0){\includegraphics[width=\linewidth,bb=0 0 1344 624,viewport=1016 436 1336 616,clip]{figures/small_motion/plants/rgb.jpg}}\put(550.000,0.000){\color{suppcyan}\linethickness{0.45pt}\framebox(125.000,125.000){} }\put(700.000,0.000){\color{supporange}\linethickness{0.45pt}\framebox(125.000,125.000){} }\end{picture}\par\vspace{2pt}\hbox to\linewidth{{\setlength{\fboxsep}{0pt}\color{suppcyan}\fbox{\includegraphics[width=0.478\linewidth,bb=0 0 704 880,viewport=360 712 520 872,clip]{figures/small_motion/plants/rgb_details.jpg}}}\hfill{\setlength{\fboxsep}{0pt}\color{supporange}\fbox{\includegraphics[width=0.478\linewidth,bb=0 0 704 880,viewport=536 712 696 872,clip]{figures/small_motion/plants/rgb_details.jpg}}}}\end{minipage}\hfill\begin{minipage}[t]{0.194\textwidth}\centering\fontsize{7.4}{8.5}\selectfont\strut InstantSplat\par\vspace{1pt}\setlength{\unitlength}{\dimexpr\linewidth/1000\relax}\begin{picture}(1000,562.500)\put(0,0){\includegraphics[width=\linewidth,bb=0 0 1344 624,viewport=8 228 328 408,clip]{figures/small_motion/plants/rgb.jpg}}\put(550.000,0.000){\color{suppcyan}\linethickness{0.45pt}\framebox(125.000,125.000){} }\put(700.000,0.000){\color{supporange}\linethickness{0.45pt}\framebox(125.000,125.000){} }\end{picture}\par\vspace{2pt}\hbox to\linewidth{{\setlength{\fboxsep}{0pt}\color{suppcyan}\fbox{\includegraphics[width=0.478\linewidth,bb=0 0 704 880,viewport=8 536 168 696,clip]{figures/small_motion/plants/rgb_details.jpg}}}\hfill{\setlength{\fboxsep}{0pt}\color{supporange}\fbox{\includegraphics[width=0.478\linewidth,bb=0 0 704 880,viewport=184 536 344 696,clip]{figures/small_motion/plants/rgb_details.jpg}}}}\end{minipage}\hfill\begin{minipage}[t]{0.194\textwidth}\centering\fontsize{7.4}{8.5}\selectfont\strut DDDrop\par\vspace{1pt}\setlength{\unitlength}{\dimexpr\linewidth/1000\relax}\begin{picture}(1000,562.500)\put(0,0){\includegraphics[width=\linewidth,bb=0 0 1344 624,viewport=344 228 664 408,clip]{figures/small_motion/plants/rgb.jpg}}\put(550.000,0.000){\color{suppcyan}\linethickness{0.45pt}\framebox(125.000,125.000){} }\put(700.000,0.000){\color{supporange}\linethickness{0.45pt}\framebox(125.000,125.000){} }\end{picture}\par\vspace{2pt}\hbox to\linewidth{{\setlength{\fboxsep}{0pt}\color{suppcyan}\fbox{\includegraphics[width=0.478\linewidth,bb=0 0 704 880,viewport=360 536 520 696,clip]{figures/small_motion/plants/rgb_details.jpg}}}\hfill{\setlength{\fboxsep}{0pt}\color{supporange}\fbox{\includegraphics[width=0.478\linewidth,bb=0 0 704 880,viewport=536 536 696 696,clip]{figures/small_motion/plants/rgb_details.jpg}}}}\end{minipage}\par\vspace{4pt}
\noindent \begin{minipage}[t]{0.194\textwidth}\centering\fontsize{7.4}{8.5}\selectfont\strut DDGS\par\vspace{1pt}\setlength{\unitlength}{\dimexpr\linewidth/1000\relax}\begin{picture}(1000,562.500)\put(0,0){\includegraphics[width=\linewidth,bb=0 0 1344 624,viewport=680 228 1000 408,clip]{figures/small_motion/plants/rgb.jpg}}\put(550.000,0.000){\color{suppcyan}\linethickness{0.45pt}\framebox(125.000,125.000){} }\put(700.000,0.000){\color{supporange}\linethickness{0.45pt}\framebox(125.000,125.000){} }\end{picture}\par\vspace{2pt}\hbox to\linewidth{{\setlength{\fboxsep}{0pt}\color{suppcyan}\fbox{\includegraphics[width=0.478\linewidth,bb=0 0 704 880,viewport=8 360 168 520,clip]{figures/small_motion/plants/rgb_details.jpg}}}\hfill{\setlength{\fboxsep}{0pt}\color{supporange}\fbox{\includegraphics[width=0.478\linewidth,bb=0 0 704 880,viewport=184 360 344 520,clip]{figures/small_motion/plants/rgb_details.jpg}}}}\end{minipage}\hfill\begin{minipage}[t]{0.194\textwidth}\centering\fontsize{7.4}{8.5}\selectfont\strut G4Splat\par\vspace{1pt}\setlength{\unitlength}{\dimexpr\linewidth/1000\relax}\begin{picture}(1000,562.500)\put(0,0){\includegraphics[width=\linewidth,bb=0 0 1344 624,viewport=1016 228 1336 408,clip]{figures/small_motion/plants/rgb.jpg}}\put(550.000,0.000){\color{suppcyan}\linethickness{0.45pt}\framebox(125.000,125.000){} }\put(700.000,0.000){\color{supporange}\linethickness{0.45pt}\framebox(125.000,125.000){} }\end{picture}\par\vspace{2pt}\hbox to\linewidth{{\setlength{\fboxsep}{0pt}\color{suppcyan}\fbox{\includegraphics[width=0.478\linewidth,bb=0 0 704 880,viewport=360 360 520 520,clip]{figures/small_motion/plants/rgb_details.jpg}}}\hfill{\setlength{\fboxsep}{0pt}\color{supporange}\fbox{\includegraphics[width=0.478\linewidth,bb=0 0 704 880,viewport=536 360 696 520,clip]{figures/small_motion/plants/rgb_details.jpg}}}}\end{minipage}\hfill\begin{minipage}[t]{0.194\textwidth}\centering\fontsize{7.4}{8.5}\selectfont\strut Vanilla$^{\dagger}$\par\vspace{1pt}\setlength{\unitlength}{\dimexpr\linewidth/1000\relax}\begin{picture}(1000,562.500)\put(0,0){\includegraphics[width=\linewidth,bb=0 0 1344 624,viewport=8 20 328 200,clip]{figures/small_motion/plants/rgb.jpg}}\put(550.000,0.000){\color{suppcyan}\linethickness{0.45pt}\framebox(125.000,125.000){} }\put(700.000,0.000){\color{supporange}\linethickness{0.45pt}\framebox(125.000,125.000){} }\end{picture}\par\vspace{2pt}\hbox to\linewidth{{\setlength{\fboxsep}{0pt}\color{suppcyan}\fbox{\includegraphics[width=0.478\linewidth,bb=0 0 704 880,viewport=8 184 168 344,clip]{figures/small_motion/plants/rgb_details.jpg}}}\hfill{\setlength{\fboxsep}{0pt}\color{supporange}\fbox{\includegraphics[width=0.478\linewidth,bb=0 0 704 880,viewport=184 184 344 344,clip]{figures/small_motion/plants/rgb_details.jpg}}}}\end{minipage}\hfill\begin{minipage}[t]{0.194\textwidth}\centering\fontsize{7.4}{8.5}\selectfont\strut Mip-Splatting\par\vspace{1pt}\setlength{\unitlength}{\dimexpr\linewidth/1000\relax}\begin{picture}(1000,562.500)\put(0,0){\includegraphics[width=\linewidth,bb=0 0 1344 624,viewport=344 20 664 200,clip]{figures/small_motion/plants/rgb.jpg}}\put(550.000,0.000){\color{suppcyan}\linethickness{0.45pt}\framebox(125.000,125.000){} }\put(700.000,0.000){\color{supporange}\linethickness{0.45pt}\framebox(125.000,125.000){} }\end{picture}\par\vspace{2pt}\hbox to\linewidth{{\setlength{\fboxsep}{0pt}\color{suppcyan}\fbox{\includegraphics[width=0.478\linewidth,bb=0 0 704 880,viewport=360 184 520 344,clip]{figures/small_motion/plants/rgb_details.jpg}}}\hfill{\setlength{\fboxsep}{0pt}\color{supporange}\fbox{\includegraphics[width=0.478\linewidth,bb=0 0 704 880,viewport=536 184 696 344,clip]{figures/small_motion/plants/rgb_details.jpg}}}}\end{minipage}\hfill\begin{minipage}[t]{0.194\textwidth}\centering\fontsize{7.4}{8.5}\selectfont\strut \textbf{NBGS}\par\vspace{1pt}\setlength{\unitlength}{\dimexpr\linewidth/1000\relax}\begin{picture}(1000,562.500)\put(0,0){\includegraphics[width=\linewidth,bb=0 0 1344 624,viewport=680 20 1000 200,clip]{figures/small_motion/plants/rgb.jpg}}\put(550.000,0.000){\color{suppcyan}\linethickness{0.45pt}\framebox(125.000,125.000){} }\put(700.000,0.000){\color{supporange}\linethickness{0.45pt}\framebox(125.000,125.000){} }\end{picture}\par\vspace{2pt}\hbox to\linewidth{{\setlength{\fboxsep}{0pt}\color{suppcyan}\fbox{\includegraphics[width=0.478\linewidth,bb=0 0 704 880,viewport=8 8 168 168,clip]{figures/small_motion/plants/rgb_details.jpg}}}\hfill{\setlength{\fboxsep}{0pt}\color{supporange}\fbox{\includegraphics[width=0.478\linewidth,bb=0 0 704 880,viewport=184 8 344 168,clip]{figures/small_motion/plants/rgb_details.jpg}}}}\end{minipage}\par\vspace{4pt}
\noindent{\small\bfseries Relative rendered depth}\par\vspace{2pt}
\noindent \begin{minipage}[t]{0.194\textwidth}\centering\fontsize{7.4}{8.5}\selectfont\strut Relative depth\par\vspace{4pt}\raggedright Per-method p02--p98. Black: invalid. No depth ground truth or metric units.\par\vspace{5pt}\includegraphics[width=\linewidth,height=5pt]{figures/legends/small_motion_depth_scale.png}\par near\hfill far\end{minipage}\hfill\begin{minipage}[t]{0.194\textwidth}\centering\fontsize{7.4}{8.5}\selectfont\strut Vanilla original\par\vspace{1pt}\includegraphics[width=\linewidth,bb=0 0 1344 624,viewport=8 436 328 616,clip]{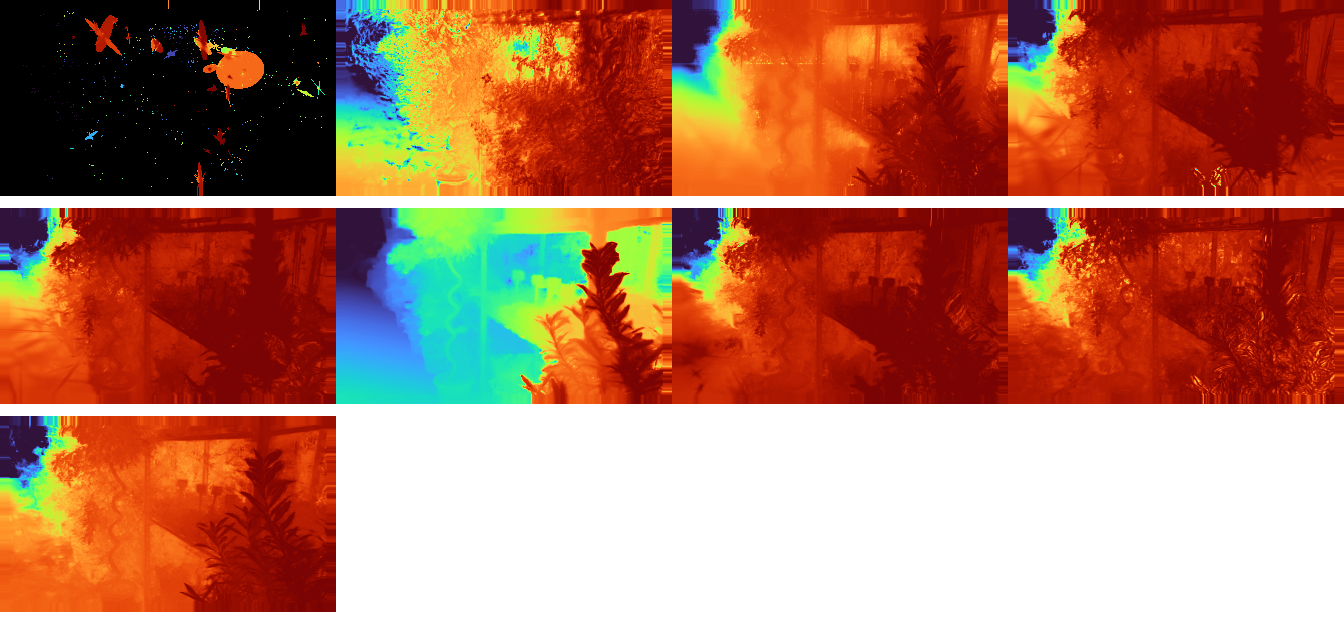}\end{minipage}\hfill\begin{minipage}[t]{0.194\textwidth}\centering\fontsize{7.4}{8.5}\selectfont\strut Small GS\par\vspace{1pt}\includegraphics[width=\linewidth,bb=0 0 1344 624,viewport=344 436 664 616,clip]{figures/small_motion/plants/depth.png}\end{minipage}\hfill\begin{minipage}[t]{0.194\textwidth}\centering\fontsize{7.4}{8.5}\selectfont\strut InstantSplat\par\vspace{1pt}\includegraphics[width=\linewidth,bb=0 0 1344 624,viewport=680 436 1000 616,clip]{figures/small_motion/plants/depth.png}\end{minipage}\hfill\begin{minipage}[t]{0.194\textwidth}\centering\fontsize{7.4}{8.5}\selectfont\strut DDDrop\par\vspace{1pt}\includegraphics[width=\linewidth,bb=0 0 1344 624,viewport=1016 436 1336 616,clip]{figures/small_motion/plants/depth.png}\end{minipage}\par\vspace{4pt}
\noindent \begin{minipage}[t]{0.194\textwidth}\centering\fontsize{7.4}{8.5}\selectfont\strut DDGS\par\vspace{1pt}\includegraphics[width=\linewidth,bb=0 0 1344 624,viewport=8 228 328 408,clip]{figures/small_motion/plants/depth.png}\end{minipage}\hfill\begin{minipage}[t]{0.194\textwidth}\centering\fontsize{7.4}{8.5}\selectfont\strut G4Splat\par\vspace{1pt}\includegraphics[width=\linewidth,bb=0 0 1344 624,viewport=344 228 664 408,clip]{figures/small_motion/plants/depth.png}\end{minipage}\hfill\begin{minipage}[t]{0.194\textwidth}\centering\fontsize{7.4}{8.5}\selectfont\strut Vanilla$^{\dagger}$\par\vspace{1pt}\includegraphics[width=\linewidth,bb=0 0 1344 624,viewport=680 228 1000 408,clip]{figures/small_motion/plants/depth.png}\end{minipage}\hfill\begin{minipage}[t]{0.194\textwidth}\centering\fontsize{7.4}{8.5}\selectfont\strut Mip-Splatting\par\vspace{1pt}\includegraphics[width=\linewidth,bb=0 0 1344 624,viewport=1016 228 1336 408,clip]{figures/small_motion/plants/depth.png}\end{minipage}\hfill\begin{minipage}[t]{0.194\textwidth}\centering\fontsize{7.4}{8.5}\selectfont\strut \textbf{NBGS}\par\vspace{1pt}\includegraphics[width=\linewidth,bb=0 0 1344 624,viewport=8 20 328 200,clip]{figures/small_motion/plants/depth.png}\end{minipage}\par\vspace{4pt}
\noindent{\small\bfseries Absolute RGB error}\par\vspace{2pt}
\noindent \begin{minipage}[t]{0.194\textwidth}\centering\fontsize{7.4}{8.5}\selectfont\strut RGB error, $[0,3]$\par\vspace{4pt}\raggedright Channel-summed absolute error. Method-matched RGB reference.\par\vspace{5pt}\includegraphics[width=\linewidth,height=5pt]{figures/legends/small_motion_error_scale.png}\par 0\hfill 3\end{minipage}\hfill\begin{minipage}[t]{0.194\textwidth}\centering\fontsize{7.4}{8.5}\selectfont\strut Vanilla original\par\vspace{1pt}\includegraphics[width=\linewidth,bb=0 0 1344 624,viewport=8 436 328 616,clip]{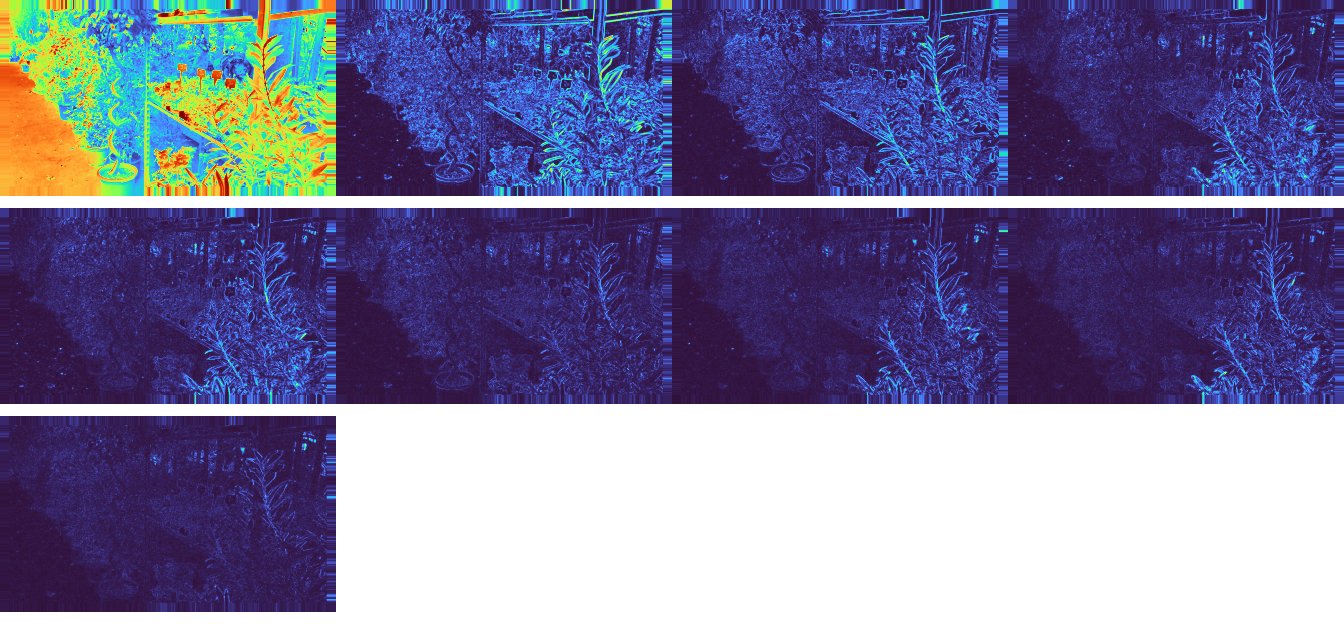}\end{minipage}\hfill\begin{minipage}[t]{0.194\textwidth}\centering\fontsize{7.4}{8.5}\selectfont\strut Small GS\par\vspace{1pt}\includegraphics[width=\linewidth,bb=0 0 1344 624,viewport=344 436 664 616,clip]{figures/small_motion/plants/rgb_error.png}\end{minipage}\hfill\begin{minipage}[t]{0.194\textwidth}\centering\fontsize{7.4}{8.5}\selectfont\strut InstantSplat\par\vspace{1pt}\includegraphics[width=\linewidth,bb=0 0 1344 624,viewport=680 436 1000 616,clip]{figures/small_motion/plants/rgb_error.png}\end{minipage}\hfill\begin{minipage}[t]{0.194\textwidth}\centering\fontsize{7.4}{8.5}\selectfont\strut DDDrop\par\vspace{1pt}\includegraphics[width=\linewidth,bb=0 0 1344 624,viewport=1016 436 1336 616,clip]{figures/small_motion/plants/rgb_error.png}\end{minipage}\par\vspace{4pt}
\noindent \begin{minipage}[t]{0.194\textwidth}\centering\fontsize{7.4}{8.5}\selectfont\strut DDGS\par\vspace{1pt}\includegraphics[width=\linewidth,bb=0 0 1344 624,viewport=8 228 328 408,clip]{figures/small_motion/plants/rgb_error.png}\end{minipage}\hfill\begin{minipage}[t]{0.194\textwidth}\centering\fontsize{7.4}{8.5}\selectfont\strut G4Splat\par\vspace{1pt}\includegraphics[width=\linewidth,bb=0 0 1344 624,viewport=344 228 664 408,clip]{figures/small_motion/plants/rgb_error.png}\end{minipage}\hfill\begin{minipage}[t]{0.194\textwidth}\centering\fontsize{7.4}{8.5}\selectfont\strut Vanilla$^{\dagger}$\par\vspace{1pt}\includegraphics[width=\linewidth,bb=0 0 1344 624,viewport=680 228 1000 408,clip]{figures/small_motion/plants/rgb_error.png}\end{minipage}\hfill\begin{minipage}[t]{0.194\textwidth}\centering\fontsize{7.4}{8.5}\selectfont\strut Mip-Splatting\par\vspace{1pt}\includegraphics[width=\linewidth,bb=0 0 1344 624,viewport=1016 228 1336 408,clip]{figures/small_motion/plants/rgb_error.png}\end{minipage}\hfill\begin{minipage}[t]{0.194\textwidth}\centering\fontsize{7.4}{8.5}\selectfont\strut \textbf{NBGS}\par\vspace{1pt}\includegraphics[width=\linewidth,bb=0 0 1344 624,viewport=8 20 328 200,clip]{figures/small_motion/plants/rgb_error.png}\end{minipage}\par\vspace{4pt}
\end{minipage}\end{lrbox}
\PlaceSuppScene
\clearpage
\pdfbookmark[3]{Stairs}{supp.scene.stairs}
\begin{lrbox}{\SuppSceneBox}\begin{minipage}{\textwidth}
\noindent{\large\bfseries Stairs}\hfill{\small\color{black!65}Small-Motion dataset}\par\vspace{3pt}
\noindent{\small\bfseries Rendered RGB}\par\vspace{2pt}
\noindent \begin{minipage}[t]{0.194\textwidth}\centering\fontsize{7.4}{8.5}\selectfont\strut Original RGB\par\vspace{1pt}\includegraphics[width=\linewidth,bb=0 0 1344 624,viewport=8 436 328 616,clip]{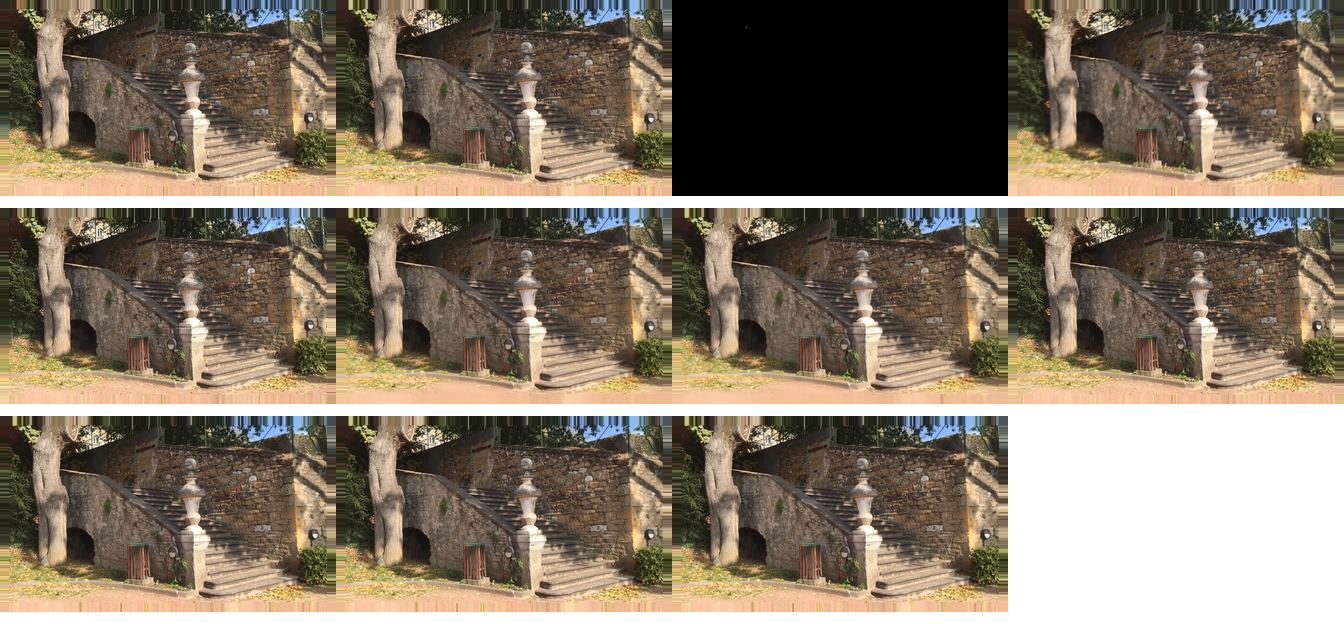}\par\vspace{2pt}\strut SfSM-undistorted RGB\par\vspace{1pt}\includegraphics[width=\linewidth,bb=0 0 1344 624,viewport=344 436 664 616,clip]{figures/small_motion/stairs/rgb.jpg}\end{minipage}\hfill\begin{minipage}[t]{0.194\textwidth}\centering\fontsize{7.4}{8.5}\selectfont\strut Vanilla original\par\vspace{1pt}\setlength{\unitlength}{\dimexpr\linewidth/1000\relax}\begin{picture}(1000,562.500)\put(0,0){\includegraphics[width=\linewidth,bb=0 0 1344 624,viewport=680 436 1000 616,clip]{figures/small_motion/stairs/rgb.jpg}}\put(460.938,140.625){\color{suppcyan}\linethickness{0.45pt}\framebox(187.500,187.500){} }\put(640.625,218.750){\color{supporange}\linethickness{0.45pt}\framebox(187.500,187.500){} }\end{picture}\par\vspace{2pt}\hbox to\linewidth{{\setlength{\fboxsep}{0pt}\color{suppcyan}\fbox{\includegraphics[width=0.478\linewidth,bb=0 0 960 1200,viewport=8 968 232 1192,clip]{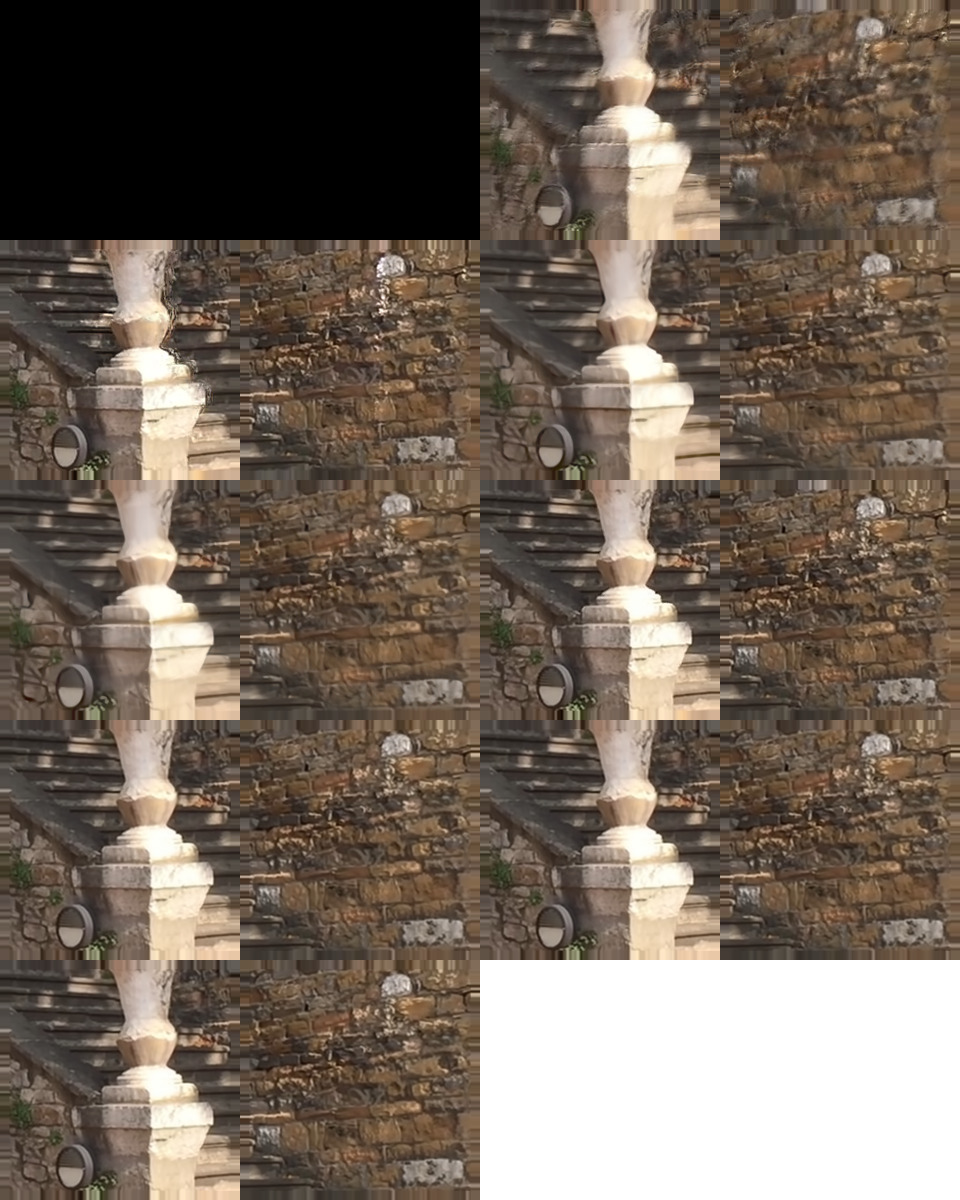}}}\hfill{\setlength{\fboxsep}{0pt}\color{supporange}\fbox{\includegraphics[width=0.478\linewidth,bb=0 0 960 1200,viewport=248 968 472 1192,clip]{figures/small_motion/stairs/rgb_details.jpg}}}}\end{minipage}\hfill\begin{minipage}[t]{0.194\textwidth}\centering\fontsize{7.4}{8.5}\selectfont\strut Small GS\par\vspace{1pt}\setlength{\unitlength}{\dimexpr\linewidth/1000\relax}\begin{picture}(1000,562.500)\put(0,0){\includegraphics[width=\linewidth,bb=0 0 1344 624,viewport=1016 436 1336 616,clip]{figures/small_motion/stairs/rgb.jpg}}\put(460.938,140.625){\color{suppcyan}\linethickness{0.45pt}\framebox(187.500,187.500){} }\put(640.625,218.750){\color{supporange}\linethickness{0.45pt}\framebox(187.500,187.500){} }\end{picture}\par\vspace{2pt}\hbox to\linewidth{{\setlength{\fboxsep}{0pt}\color{suppcyan}\fbox{\includegraphics[width=0.478\linewidth,bb=0 0 960 1200,viewport=488 968 712 1192,clip]{figures/small_motion/stairs/rgb_details.jpg}}}\hfill{\setlength{\fboxsep}{0pt}\color{supporange}\fbox{\includegraphics[width=0.478\linewidth,bb=0 0 960 1200,viewport=728 968 952 1192,clip]{figures/small_motion/stairs/rgb_details.jpg}}}}\end{minipage}\hfill\begin{minipage}[t]{0.194\textwidth}\centering\fontsize{7.4}{8.5}\selectfont\strut InstantSplat\par\vspace{1pt}\setlength{\unitlength}{\dimexpr\linewidth/1000\relax}\begin{picture}(1000,562.500)\put(0,0){\includegraphics[width=\linewidth,bb=0 0 1344 624,viewport=8 228 328 408,clip]{figures/small_motion/stairs/rgb.jpg}}\put(460.938,140.625){\color{suppcyan}\linethickness{0.45pt}\framebox(187.500,187.500){} }\put(640.625,218.750){\color{supporange}\linethickness{0.45pt}\framebox(187.500,187.500){} }\end{picture}\par\vspace{2pt}\hbox to\linewidth{{\setlength{\fboxsep}{0pt}\color{suppcyan}\fbox{\includegraphics[width=0.478\linewidth,bb=0 0 960 1200,viewport=8 728 232 952,clip]{figures/small_motion/stairs/rgb_details.jpg}}}\hfill{\setlength{\fboxsep}{0pt}\color{supporange}\fbox{\includegraphics[width=0.478\linewidth,bb=0 0 960 1200,viewport=248 728 472 952,clip]{figures/small_motion/stairs/rgb_details.jpg}}}}\end{minipage}\hfill\begin{minipage}[t]{0.194\textwidth}\centering\fontsize{7.4}{8.5}\selectfont\strut DDDrop\par\vspace{1pt}\setlength{\unitlength}{\dimexpr\linewidth/1000\relax}\begin{picture}(1000,562.500)\put(0,0){\includegraphics[width=\linewidth,bb=0 0 1344 624,viewport=344 228 664 408,clip]{figures/small_motion/stairs/rgb.jpg}}\put(460.938,140.625){\color{suppcyan}\linethickness{0.45pt}\framebox(187.500,187.500){} }\put(640.625,218.750){\color{supporange}\linethickness{0.45pt}\framebox(187.500,187.500){} }\end{picture}\par\vspace{2pt}\hbox to\linewidth{{\setlength{\fboxsep}{0pt}\color{suppcyan}\fbox{\includegraphics[width=0.478\linewidth,bb=0 0 960 1200,viewport=488 728 712 952,clip]{figures/small_motion/stairs/rgb_details.jpg}}}\hfill{\setlength{\fboxsep}{0pt}\color{supporange}\fbox{\includegraphics[width=0.478\linewidth,bb=0 0 960 1200,viewport=728 728 952 952,clip]{figures/small_motion/stairs/rgb_details.jpg}}}}\end{minipage}\par\vspace{4pt}
\noindent \begin{minipage}[t]{0.194\textwidth}\centering\fontsize{7.4}{8.5}\selectfont\strut DDGS\par\vspace{1pt}\setlength{\unitlength}{\dimexpr\linewidth/1000\relax}\begin{picture}(1000,562.500)\put(0,0){\includegraphics[width=\linewidth,bb=0 0 1344 624,viewport=680 228 1000 408,clip]{figures/small_motion/stairs/rgb.jpg}}\put(460.938,140.625){\color{suppcyan}\linethickness{0.45pt}\framebox(187.500,187.500){} }\put(640.625,218.750){\color{supporange}\linethickness{0.45pt}\framebox(187.500,187.500){} }\end{picture}\par\vspace{2pt}\hbox to\linewidth{{\setlength{\fboxsep}{0pt}\color{suppcyan}\fbox{\includegraphics[width=0.478\linewidth,bb=0 0 960 1200,viewport=8 488 232 712,clip]{figures/small_motion/stairs/rgb_details.jpg}}}\hfill{\setlength{\fboxsep}{0pt}\color{supporange}\fbox{\includegraphics[width=0.478\linewidth,bb=0 0 960 1200,viewport=248 488 472 712,clip]{figures/small_motion/stairs/rgb_details.jpg}}}}\end{minipage}\hfill\begin{minipage}[t]{0.194\textwidth}\centering\fontsize{7.4}{8.5}\selectfont\strut G4Splat\par\vspace{1pt}\setlength{\unitlength}{\dimexpr\linewidth/1000\relax}\begin{picture}(1000,562.500)\put(0,0){\includegraphics[width=\linewidth,bb=0 0 1344 624,viewport=1016 228 1336 408,clip]{figures/small_motion/stairs/rgb.jpg}}\put(460.938,140.625){\color{suppcyan}\linethickness{0.45pt}\framebox(187.500,187.500){} }\put(640.625,218.750){\color{supporange}\linethickness{0.45pt}\framebox(187.500,187.500){} }\end{picture}\par\vspace{2pt}\hbox to\linewidth{{\setlength{\fboxsep}{0pt}\color{suppcyan}\fbox{\includegraphics[width=0.478\linewidth,bb=0 0 960 1200,viewport=488 488 712 712,clip]{figures/small_motion/stairs/rgb_details.jpg}}}\hfill{\setlength{\fboxsep}{0pt}\color{supporange}\fbox{\includegraphics[width=0.478\linewidth,bb=0 0 960 1200,viewport=728 488 952 712,clip]{figures/small_motion/stairs/rgb_details.jpg}}}}\end{minipage}\hfill\begin{minipage}[t]{0.194\textwidth}\centering\fontsize{7.4}{8.5}\selectfont\strut Vanilla$^{\dagger}$\par\vspace{1pt}\setlength{\unitlength}{\dimexpr\linewidth/1000\relax}\begin{picture}(1000,562.500)\put(0,0){\includegraphics[width=\linewidth,bb=0 0 1344 624,viewport=8 20 328 200,clip]{figures/small_motion/stairs/rgb.jpg}}\put(460.938,140.625){\color{suppcyan}\linethickness{0.45pt}\framebox(187.500,187.500){} }\put(640.625,218.750){\color{supporange}\linethickness{0.45pt}\framebox(187.500,187.500){} }\end{picture}\par\vspace{2pt}\hbox to\linewidth{{\setlength{\fboxsep}{0pt}\color{suppcyan}\fbox{\includegraphics[width=0.478\linewidth,bb=0 0 960 1200,viewport=8 248 232 472,clip]{figures/small_motion/stairs/rgb_details.jpg}}}\hfill{\setlength{\fboxsep}{0pt}\color{supporange}\fbox{\includegraphics[width=0.478\linewidth,bb=0 0 960 1200,viewport=248 248 472 472,clip]{figures/small_motion/stairs/rgb_details.jpg}}}}\end{minipage}\hfill\begin{minipage}[t]{0.194\textwidth}\centering\fontsize{7.4}{8.5}\selectfont\strut Mip-Splatting\par\vspace{1pt}\setlength{\unitlength}{\dimexpr\linewidth/1000\relax}\begin{picture}(1000,562.500)\put(0,0){\includegraphics[width=\linewidth,bb=0 0 1344 624,viewport=344 20 664 200,clip]{figures/small_motion/stairs/rgb.jpg}}\put(460.938,140.625){\color{suppcyan}\linethickness{0.45pt}\framebox(187.500,187.500){} }\put(640.625,218.750){\color{supporange}\linethickness{0.45pt}\framebox(187.500,187.500){} }\end{picture}\par\vspace{2pt}\hbox to\linewidth{{\setlength{\fboxsep}{0pt}\color{suppcyan}\fbox{\includegraphics[width=0.478\linewidth,bb=0 0 960 1200,viewport=488 248 712 472,clip]{figures/small_motion/stairs/rgb_details.jpg}}}\hfill{\setlength{\fboxsep}{0pt}\color{supporange}\fbox{\includegraphics[width=0.478\linewidth,bb=0 0 960 1200,viewport=728 248 952 472,clip]{figures/small_motion/stairs/rgb_details.jpg}}}}\end{minipage}\hfill\begin{minipage}[t]{0.194\textwidth}\centering\fontsize{7.4}{8.5}\selectfont\strut \textbf{NBGS}\par\vspace{1pt}\setlength{\unitlength}{\dimexpr\linewidth/1000\relax}\begin{picture}(1000,562.500)\put(0,0){\includegraphics[width=\linewidth,bb=0 0 1344 624,viewport=680 20 1000 200,clip]{figures/small_motion/stairs/rgb.jpg}}\put(460.938,140.625){\color{suppcyan}\linethickness{0.45pt}\framebox(187.500,187.500){} }\put(640.625,218.750){\color{supporange}\linethickness{0.45pt}\framebox(187.500,187.500){} }\end{picture}\par\vspace{2pt}\hbox to\linewidth{{\setlength{\fboxsep}{0pt}\color{suppcyan}\fbox{\includegraphics[width=0.478\linewidth,bb=0 0 960 1200,viewport=8 8 232 232,clip]{figures/small_motion/stairs/rgb_details.jpg}}}\hfill{\setlength{\fboxsep}{0pt}\color{supporange}\fbox{\includegraphics[width=0.478\linewidth,bb=0 0 960 1200,viewport=248 8 472 232,clip]{figures/small_motion/stairs/rgb_details.jpg}}}}\end{minipage}\par\vspace{4pt}
\noindent{\small\bfseries Relative rendered depth}\par\vspace{2pt}
\noindent \begin{minipage}[t]{0.194\textwidth}\centering\fontsize{7.4}{8.5}\selectfont\strut Relative depth\par\vspace{4pt}\raggedright Per-method p02--p98. Black: invalid. No depth ground truth or metric units.\par\vspace{5pt}\includegraphics[width=\linewidth,height=5pt]{figures/legends/small_motion_depth_scale.png}\par near\hfill far\end{minipage}\hfill\begin{minipage}[t]{0.194\textwidth}\centering\fontsize{7.4}{8.5}\selectfont\strut Vanilla original\par\vspace{1pt}\includegraphics[width=\linewidth,bb=0 0 1344 624,viewport=8 436 328 616,clip]{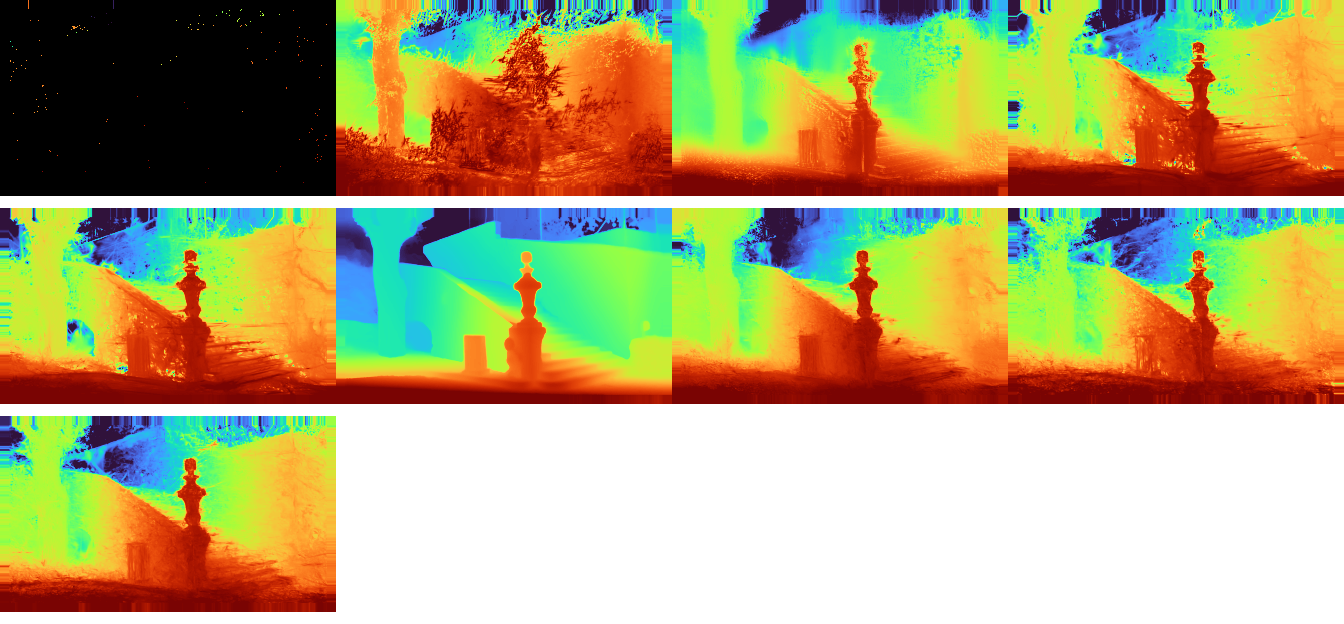}\end{minipage}\hfill\begin{minipage}[t]{0.194\textwidth}\centering\fontsize{7.4}{8.5}\selectfont\strut Small GS\par\vspace{1pt}\includegraphics[width=\linewidth,bb=0 0 1344 624,viewport=344 436 664 616,clip]{figures/small_motion/stairs/depth.png}\end{minipage}\hfill\begin{minipage}[t]{0.194\textwidth}\centering\fontsize{7.4}{8.5}\selectfont\strut InstantSplat\par\vspace{1pt}\includegraphics[width=\linewidth,bb=0 0 1344 624,viewport=680 436 1000 616,clip]{figures/small_motion/stairs/depth.png}\end{minipage}\hfill\begin{minipage}[t]{0.194\textwidth}\centering\fontsize{7.4}{8.5}\selectfont\strut DDDrop\par\vspace{1pt}\includegraphics[width=\linewidth,bb=0 0 1344 624,viewport=1016 436 1336 616,clip]{figures/small_motion/stairs/depth.png}\end{minipage}\par\vspace{4pt}
\noindent \begin{minipage}[t]{0.194\textwidth}\centering\fontsize{7.4}{8.5}\selectfont\strut DDGS\par\vspace{1pt}\includegraphics[width=\linewidth,bb=0 0 1344 624,viewport=8 228 328 408,clip]{figures/small_motion/stairs/depth.png}\end{minipage}\hfill\begin{minipage}[t]{0.194\textwidth}\centering\fontsize{7.4}{8.5}\selectfont\strut G4Splat\par\vspace{1pt}\includegraphics[width=\linewidth,bb=0 0 1344 624,viewport=344 228 664 408,clip]{figures/small_motion/stairs/depth.png}\end{minipage}\hfill\begin{minipage}[t]{0.194\textwidth}\centering\fontsize{7.4}{8.5}\selectfont\strut Vanilla$^{\dagger}$\par\vspace{1pt}\includegraphics[width=\linewidth,bb=0 0 1344 624,viewport=680 228 1000 408,clip]{figures/small_motion/stairs/depth.png}\end{minipage}\hfill\begin{minipage}[t]{0.194\textwidth}\centering\fontsize{7.4}{8.5}\selectfont\strut Mip-Splatting\par\vspace{1pt}\includegraphics[width=\linewidth,bb=0 0 1344 624,viewport=1016 228 1336 408,clip]{figures/small_motion/stairs/depth.png}\end{minipage}\hfill\begin{minipage}[t]{0.194\textwidth}\centering\fontsize{7.4}{8.5}\selectfont\strut \textbf{NBGS}\par\vspace{1pt}\includegraphics[width=\linewidth,bb=0 0 1344 624,viewport=8 20 328 200,clip]{figures/small_motion/stairs/depth.png}\end{minipage}\par\vspace{4pt}
\noindent{\small\bfseries Absolute RGB error}\par\vspace{2pt}
\noindent \begin{minipage}[t]{0.194\textwidth}\centering\fontsize{7.4}{8.5}\selectfont\strut RGB error, $[0,3]$\par\vspace{4pt}\raggedright Channel-summed absolute error. Method-matched RGB reference.\par\vspace{5pt}\includegraphics[width=\linewidth,height=5pt]{figures/legends/small_motion_error_scale.png}\par 0\hfill 3\end{minipage}\hfill\begin{minipage}[t]{0.194\textwidth}\centering\fontsize{7.4}{8.5}\selectfont\strut Vanilla original\par\vspace{1pt}\includegraphics[width=\linewidth,bb=0 0 1344 624,viewport=8 436 328 616,clip]{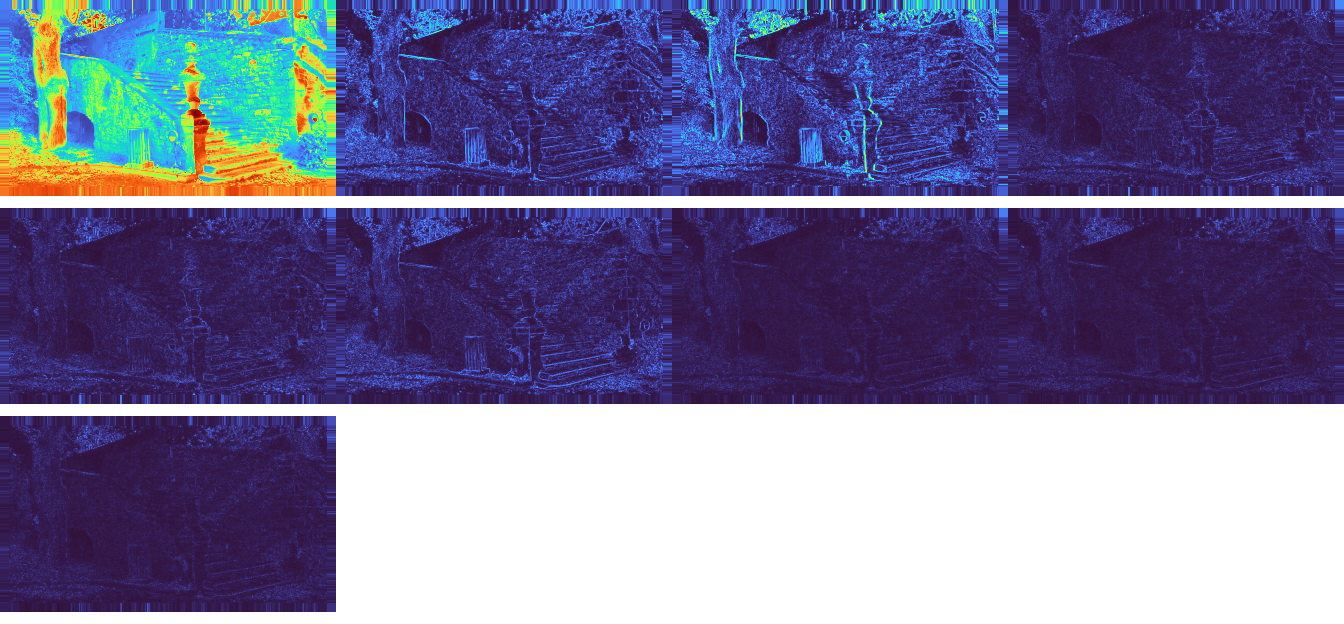}\end{minipage}\hfill\begin{minipage}[t]{0.194\textwidth}\centering\fontsize{7.4}{8.5}\selectfont\strut Small GS\par\vspace{1pt}\includegraphics[width=\linewidth,bb=0 0 1344 624,viewport=344 436 664 616,clip]{figures/small_motion/stairs/rgb_error.png}\end{minipage}\hfill\begin{minipage}[t]{0.194\textwidth}\centering\fontsize{7.4}{8.5}\selectfont\strut InstantSplat\par\vspace{1pt}\includegraphics[width=\linewidth,bb=0 0 1344 624,viewport=680 436 1000 616,clip]{figures/small_motion/stairs/rgb_error.png}\end{minipage}\hfill\begin{minipage}[t]{0.194\textwidth}\centering\fontsize{7.4}{8.5}\selectfont\strut DDDrop\par\vspace{1pt}\includegraphics[width=\linewidth,bb=0 0 1344 624,viewport=1016 436 1336 616,clip]{figures/small_motion/stairs/rgb_error.png}\end{minipage}\par\vspace{4pt}
\noindent \begin{minipage}[t]{0.194\textwidth}\centering\fontsize{7.4}{8.5}\selectfont\strut DDGS\par\vspace{1pt}\includegraphics[width=\linewidth,bb=0 0 1344 624,viewport=8 228 328 408,clip]{figures/small_motion/stairs/rgb_error.png}\end{minipage}\hfill\begin{minipage}[t]{0.194\textwidth}\centering\fontsize{7.4}{8.5}\selectfont\strut G4Splat\par\vspace{1pt}\includegraphics[width=\linewidth,bb=0 0 1344 624,viewport=344 228 664 408,clip]{figures/small_motion/stairs/rgb_error.png}\end{minipage}\hfill\begin{minipage}[t]{0.194\textwidth}\centering\fontsize{7.4}{8.5}\selectfont\strut Vanilla$^{\dagger}$\par\vspace{1pt}\includegraphics[width=\linewidth,bb=0 0 1344 624,viewport=680 228 1000 408,clip]{figures/small_motion/stairs/rgb_error.png}\end{minipage}\hfill\begin{minipage}[t]{0.194\textwidth}\centering\fontsize{7.4}{8.5}\selectfont\strut Mip-Splatting\par\vspace{1pt}\includegraphics[width=\linewidth,bb=0 0 1344 624,viewport=1016 228 1336 408,clip]{figures/small_motion/stairs/rgb_error.png}\end{minipage}\hfill\begin{minipage}[t]{0.194\textwidth}\centering\fontsize{7.4}{8.5}\selectfont\strut \textbf{NBGS}\par\vspace{1pt}\includegraphics[width=\linewidth,bb=0 0 1344 624,viewport=8 20 328 200,clip]{figures/small_motion/stairs/rgb_error.png}\end{minipage}\par\vspace{4pt}
\end{minipage}\end{lrbox}
\PlaceSuppScene
\clearpage
\pdfbookmark[3]{Stones}{supp.scene.stones}
\begin{lrbox}{\SuppSceneBox}\begin{minipage}{\textwidth}
\noindent{\large\bfseries Stones}\hfill{\small\color{black!65}Small-Motion dataset}\par\vspace{3pt}
\noindent{\small\bfseries Rendered RGB}\par\vspace{2pt}
\noindent \begin{minipage}[t]{0.194\textwidth}\centering\fontsize{7.4}{8.5}\selectfont\strut Original RGB\par\vspace{1pt}\includegraphics[width=\linewidth,bb=0 0 1344 624,viewport=8 436 328 616,clip]{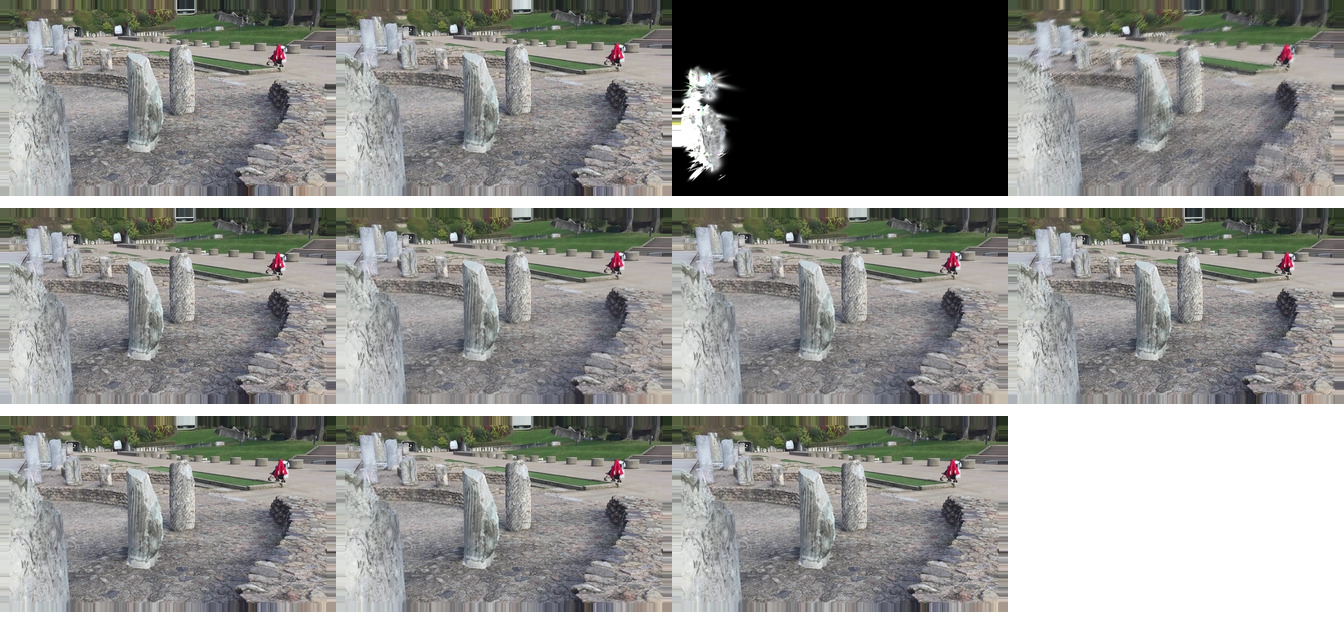}\par\vspace{2pt}\strut SfSM-undistorted RGB\par\vspace{1pt}\includegraphics[width=\linewidth,bb=0 0 1344 624,viewport=344 436 664 616,clip]{figures/small_motion/stones/rgb.jpg}\end{minipage}\hfill\begin{minipage}[t]{0.194\textwidth}\centering\fontsize{7.4}{8.5}\selectfont\strut Vanilla original\par\vspace{1pt}\setlength{\unitlength}{\dimexpr\linewidth/1000\relax}\begin{picture}(1000,562.500)\put(0,0){\includegraphics[width=\linewidth,bb=0 0 1344 624,viewport=680 436 1000 616,clip]{figures/small_motion/stones/rgb.jpg}}\put(328.125,171.875){\color{suppcyan}\linethickness{0.45pt}\framebox(187.500,187.500){} }\put(593.750,62.500){\color{supporange}\linethickness{0.45pt}\framebox(171.875,171.875){} }\end{picture}\par\vspace{2pt}\hbox to\linewidth{{\setlength{\fboxsep}{0pt}\color{suppcyan}\fbox{\includegraphics[width=0.478\linewidth,bb=0 0 960 1200,viewport=8 968 232 1192,clip]{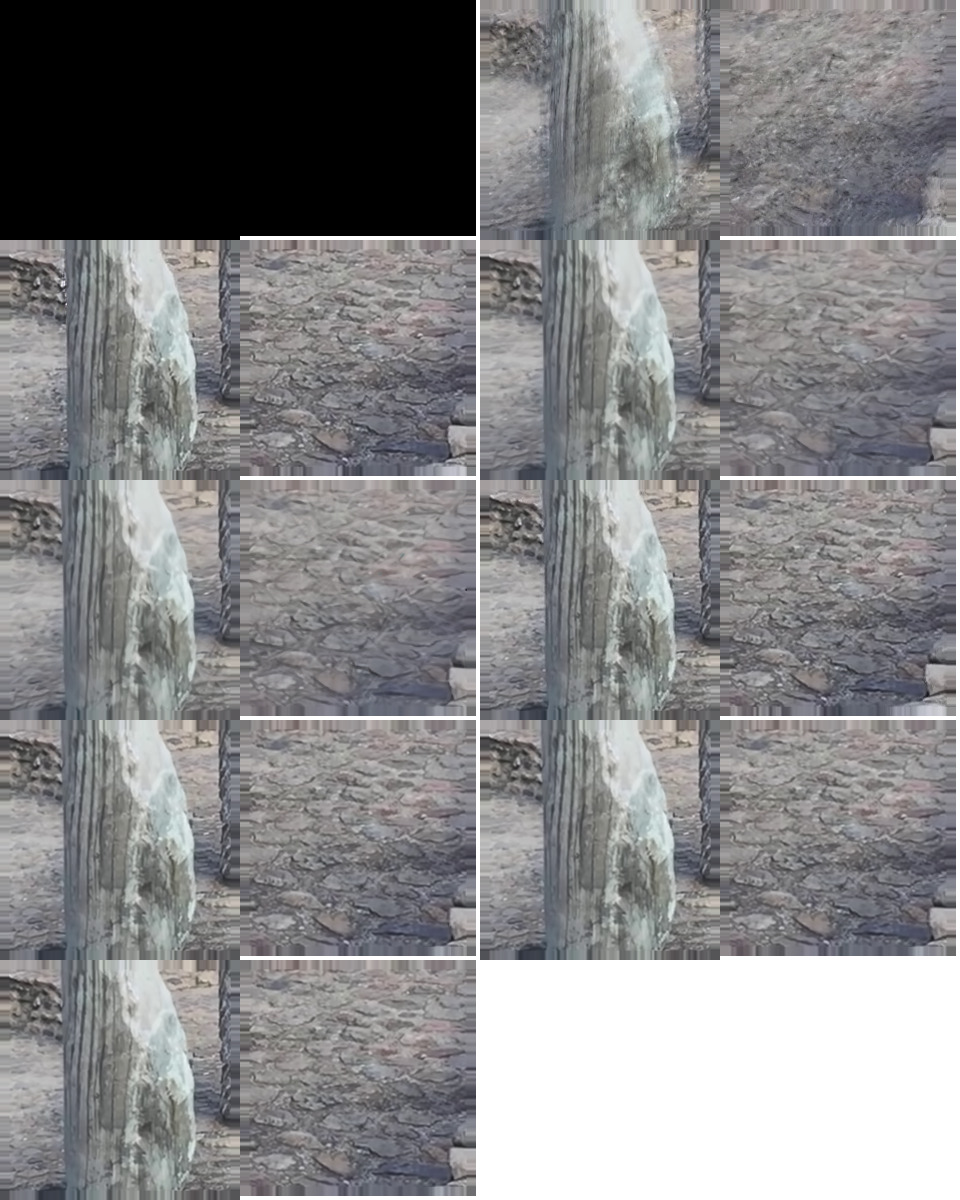}}}\hfill{\setlength{\fboxsep}{0pt}\color{supporange}\fbox{\includegraphics[width=0.478\linewidth,bb=0 0 960 1200,viewport=248 972 468 1192,clip]{figures/small_motion/stones/rgb_details.jpg}}}}\end{minipage}\hfill\begin{minipage}[t]{0.194\textwidth}\centering\fontsize{7.4}{8.5}\selectfont\strut Small GS\par\vspace{1pt}\setlength{\unitlength}{\dimexpr\linewidth/1000\relax}\begin{picture}(1000,562.500)\put(0,0){\includegraphics[width=\linewidth,bb=0 0 1344 624,viewport=1016 436 1336 616,clip]{figures/small_motion/stones/rgb.jpg}}\put(328.125,171.875){\color{suppcyan}\linethickness{0.45pt}\framebox(187.500,187.500){} }\put(593.750,62.500){\color{supporange}\linethickness{0.45pt}\framebox(171.875,171.875){} }\end{picture}\par\vspace{2pt}\hbox to\linewidth{{\setlength{\fboxsep}{0pt}\color{suppcyan}\fbox{\includegraphics[width=0.478\linewidth,bb=0 0 960 1200,viewport=488 968 712 1192,clip]{figures/small_motion/stones/rgb_details.jpg}}}\hfill{\setlength{\fboxsep}{0pt}\color{supporange}\fbox{\includegraphics[width=0.478\linewidth,bb=0 0 960 1200,viewport=728 972 948 1192,clip]{figures/small_motion/stones/rgb_details.jpg}}}}\end{minipage}\hfill\begin{minipage}[t]{0.194\textwidth}\centering\fontsize{7.4}{8.5}\selectfont\strut InstantSplat\par\vspace{1pt}\setlength{\unitlength}{\dimexpr\linewidth/1000\relax}\begin{picture}(1000,562.500)\put(0,0){\includegraphics[width=\linewidth,bb=0 0 1344 624,viewport=8 228 328 408,clip]{figures/small_motion/stones/rgb.jpg}}\put(328.125,171.875){\color{suppcyan}\linethickness{0.45pt}\framebox(187.500,187.500){} }\put(593.750,62.500){\color{supporange}\linethickness{0.45pt}\framebox(171.875,171.875){} }\end{picture}\par\vspace{2pt}\hbox to\linewidth{{\setlength{\fboxsep}{0pt}\color{suppcyan}\fbox{\includegraphics[width=0.478\linewidth,bb=0 0 960 1200,viewport=8 728 232 952,clip]{figures/small_motion/stones/rgb_details.jpg}}}\hfill{\setlength{\fboxsep}{0pt}\color{supporange}\fbox{\includegraphics[width=0.478\linewidth,bb=0 0 960 1200,viewport=248 732 468 952,clip]{figures/small_motion/stones/rgb_details.jpg}}}}\end{minipage}\hfill\begin{minipage}[t]{0.194\textwidth}\centering\fontsize{7.4}{8.5}\selectfont\strut DDDrop\par\vspace{1pt}\setlength{\unitlength}{\dimexpr\linewidth/1000\relax}\begin{picture}(1000,562.500)\put(0,0){\includegraphics[width=\linewidth,bb=0 0 1344 624,viewport=344 228 664 408,clip]{figures/small_motion/stones/rgb.jpg}}\put(328.125,171.875){\color{suppcyan}\linethickness{0.45pt}\framebox(187.500,187.500){} }\put(593.750,62.500){\color{supporange}\linethickness{0.45pt}\framebox(171.875,171.875){} }\end{picture}\par\vspace{2pt}\hbox to\linewidth{{\setlength{\fboxsep}{0pt}\color{suppcyan}\fbox{\includegraphics[width=0.478\linewidth,bb=0 0 960 1200,viewport=488 728 712 952,clip]{figures/small_motion/stones/rgb_details.jpg}}}\hfill{\setlength{\fboxsep}{0pt}\color{supporange}\fbox{\includegraphics[width=0.478\linewidth,bb=0 0 960 1200,viewport=728 732 948 952,clip]{figures/small_motion/stones/rgb_details.jpg}}}}\end{minipage}\par\vspace{4pt}
\noindent \begin{minipage}[t]{0.194\textwidth}\centering\fontsize{7.4}{8.5}\selectfont\strut DDGS\par\vspace{1pt}\setlength{\unitlength}{\dimexpr\linewidth/1000\relax}\begin{picture}(1000,562.500)\put(0,0){\includegraphics[width=\linewidth,bb=0 0 1344 624,viewport=680 228 1000 408,clip]{figures/small_motion/stones/rgb.jpg}}\put(328.125,171.875){\color{suppcyan}\linethickness{0.45pt}\framebox(187.500,187.500){} }\put(593.750,62.500){\color{supporange}\linethickness{0.45pt}\framebox(171.875,171.875){} }\end{picture}\par\vspace{2pt}\hbox to\linewidth{{\setlength{\fboxsep}{0pt}\color{suppcyan}\fbox{\includegraphics[width=0.478\linewidth,bb=0 0 960 1200,viewport=8 488 232 712,clip]{figures/small_motion/stones/rgb_details.jpg}}}\hfill{\setlength{\fboxsep}{0pt}\color{supporange}\fbox{\includegraphics[width=0.478\linewidth,bb=0 0 960 1200,viewport=248 492 468 712,clip]{figures/small_motion/stones/rgb_details.jpg}}}}\end{minipage}\hfill\begin{minipage}[t]{0.194\textwidth}\centering\fontsize{7.4}{8.5}\selectfont\strut G4Splat\par\vspace{1pt}\setlength{\unitlength}{\dimexpr\linewidth/1000\relax}\begin{picture}(1000,562.500)\put(0,0){\includegraphics[width=\linewidth,bb=0 0 1344 624,viewport=1016 228 1336 408,clip]{figures/small_motion/stones/rgb.jpg}}\put(328.125,171.875){\color{suppcyan}\linethickness{0.45pt}\framebox(187.500,187.500){} }\put(593.750,62.500){\color{supporange}\linethickness{0.45pt}\framebox(171.875,171.875){} }\end{picture}\par\vspace{2pt}\hbox to\linewidth{{\setlength{\fboxsep}{0pt}\color{suppcyan}\fbox{\includegraphics[width=0.478\linewidth,bb=0 0 960 1200,viewport=488 488 712 712,clip]{figures/small_motion/stones/rgb_details.jpg}}}\hfill{\setlength{\fboxsep}{0pt}\color{supporange}\fbox{\includegraphics[width=0.478\linewidth,bb=0 0 960 1200,viewport=728 492 948 712,clip]{figures/small_motion/stones/rgb_details.jpg}}}}\end{minipage}\hfill\begin{minipage}[t]{0.194\textwidth}\centering\fontsize{7.4}{8.5}\selectfont\strut Vanilla$^{\dagger}$\par\vspace{1pt}\setlength{\unitlength}{\dimexpr\linewidth/1000\relax}\begin{picture}(1000,562.500)\put(0,0){\includegraphics[width=\linewidth,bb=0 0 1344 624,viewport=8 20 328 200,clip]{figures/small_motion/stones/rgb.jpg}}\put(328.125,171.875){\color{suppcyan}\linethickness{0.45pt}\framebox(187.500,187.500){} }\put(593.750,62.500){\color{supporange}\linethickness{0.45pt}\framebox(171.875,171.875){} }\end{picture}\par\vspace{2pt}\hbox to\linewidth{{\setlength{\fboxsep}{0pt}\color{suppcyan}\fbox{\includegraphics[width=0.478\linewidth,bb=0 0 960 1200,viewport=8 248 232 472,clip]{figures/small_motion/stones/rgb_details.jpg}}}\hfill{\setlength{\fboxsep}{0pt}\color{supporange}\fbox{\includegraphics[width=0.478\linewidth,bb=0 0 960 1200,viewport=248 252 468 472,clip]{figures/small_motion/stones/rgb_details.jpg}}}}\end{minipage}\hfill\begin{minipage}[t]{0.194\textwidth}\centering\fontsize{7.4}{8.5}\selectfont\strut Mip-Splatting\par\vspace{1pt}\setlength{\unitlength}{\dimexpr\linewidth/1000\relax}\begin{picture}(1000,562.500)\put(0,0){\includegraphics[width=\linewidth,bb=0 0 1344 624,viewport=344 20 664 200,clip]{figures/small_motion/stones/rgb.jpg}}\put(328.125,171.875){\color{suppcyan}\linethickness{0.45pt}\framebox(187.500,187.500){} }\put(593.750,62.500){\color{supporange}\linethickness{0.45pt}\framebox(171.875,171.875){} }\end{picture}\par\vspace{2pt}\hbox to\linewidth{{\setlength{\fboxsep}{0pt}\color{suppcyan}\fbox{\includegraphics[width=0.478\linewidth,bb=0 0 960 1200,viewport=488 248 712 472,clip]{figures/small_motion/stones/rgb_details.jpg}}}\hfill{\setlength{\fboxsep}{0pt}\color{supporange}\fbox{\includegraphics[width=0.478\linewidth,bb=0 0 960 1200,viewport=728 252 948 472,clip]{figures/small_motion/stones/rgb_details.jpg}}}}\end{minipage}\hfill\begin{minipage}[t]{0.194\textwidth}\centering\fontsize{7.4}{8.5}\selectfont\strut \textbf{NBGS}\par\vspace{1pt}\setlength{\unitlength}{\dimexpr\linewidth/1000\relax}\begin{picture}(1000,562.500)\put(0,0){\includegraphics[width=\linewidth,bb=0 0 1344 624,viewport=680 20 1000 200,clip]{figures/small_motion/stones/rgb.jpg}}\put(328.125,171.875){\color{suppcyan}\linethickness{0.45pt}\framebox(187.500,187.500){} }\put(593.750,62.500){\color{supporange}\linethickness{0.45pt}\framebox(171.875,171.875){} }\end{picture}\par\vspace{2pt}\hbox to\linewidth{{\setlength{\fboxsep}{0pt}\color{suppcyan}\fbox{\includegraphics[width=0.478\linewidth,bb=0 0 960 1200,viewport=8 8 232 232,clip]{figures/small_motion/stones/rgb_details.jpg}}}\hfill{\setlength{\fboxsep}{0pt}\color{supporange}\fbox{\includegraphics[width=0.478\linewidth,bb=0 0 960 1200,viewport=248 12 468 232,clip]{figures/small_motion/stones/rgb_details.jpg}}}}\end{minipage}\par\vspace{4pt}
\noindent{\small\bfseries Relative rendered depth}\par\vspace{2pt}
\noindent \begin{minipage}[t]{0.194\textwidth}\centering\fontsize{7.4}{8.5}\selectfont\strut Relative depth\par\vspace{4pt}\raggedright Per-method p02--p98. Black: invalid. No depth ground truth or metric units.\par\vspace{5pt}\includegraphics[width=\linewidth,height=5pt]{figures/legends/small_motion_depth_scale.png}\par near\hfill far\end{minipage}\hfill\begin{minipage}[t]{0.194\textwidth}\centering\fontsize{7.4}{8.5}\selectfont\strut Vanilla original\par\vspace{1pt}\includegraphics[width=\linewidth,bb=0 0 1344 624,viewport=8 436 328 616,clip]{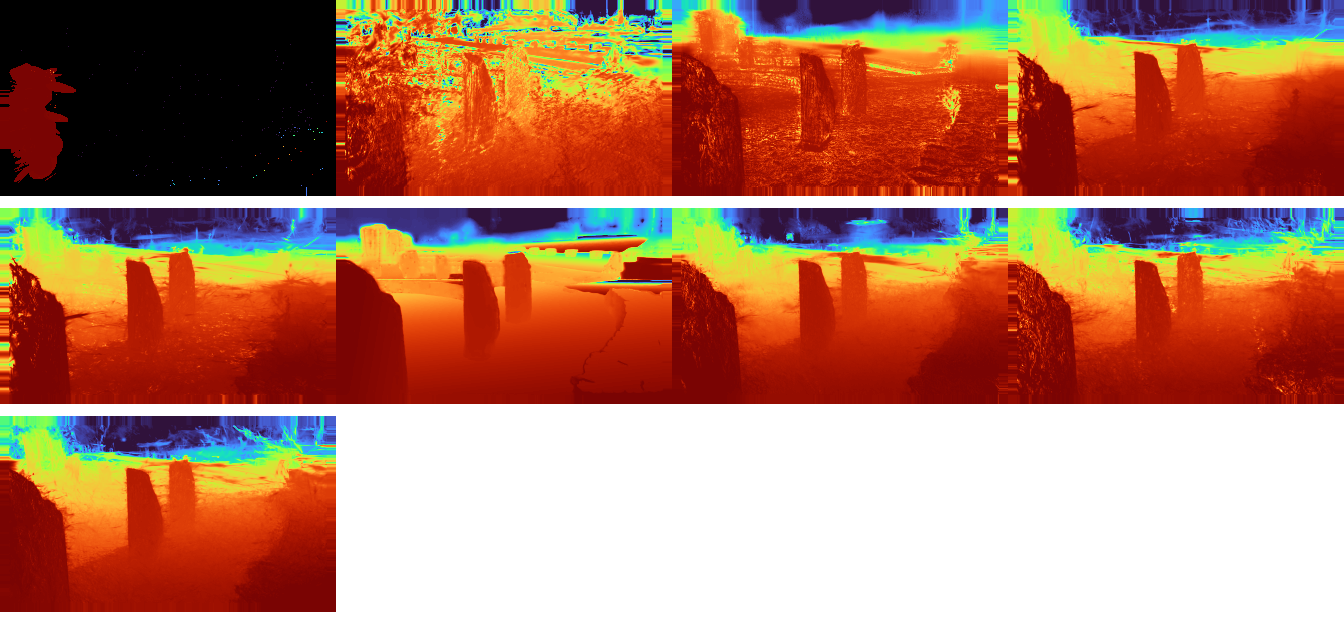}\end{minipage}\hfill\begin{minipage}[t]{0.194\textwidth}\centering\fontsize{7.4}{8.5}\selectfont\strut Small GS\par\vspace{1pt}\includegraphics[width=\linewidth,bb=0 0 1344 624,viewport=344 436 664 616,clip]{figures/small_motion/stones/depth.png}\end{minipage}\hfill\begin{minipage}[t]{0.194\textwidth}\centering\fontsize{7.4}{8.5}\selectfont\strut InstantSplat\par\vspace{1pt}\includegraphics[width=\linewidth,bb=0 0 1344 624,viewport=680 436 1000 616,clip]{figures/small_motion/stones/depth.png}\end{minipage}\hfill\begin{minipage}[t]{0.194\textwidth}\centering\fontsize{7.4}{8.5}\selectfont\strut DDDrop\par\vspace{1pt}\includegraphics[width=\linewidth,bb=0 0 1344 624,viewport=1016 436 1336 616,clip]{figures/small_motion/stones/depth.png}\end{minipage}\par\vspace{4pt}
\noindent \begin{minipage}[t]{0.194\textwidth}\centering\fontsize{7.4}{8.5}\selectfont\strut DDGS\par\vspace{1pt}\includegraphics[width=\linewidth,bb=0 0 1344 624,viewport=8 228 328 408,clip]{figures/small_motion/stones/depth.png}\end{minipage}\hfill\begin{minipage}[t]{0.194\textwidth}\centering\fontsize{7.4}{8.5}\selectfont\strut G4Splat\par\vspace{1pt}\includegraphics[width=\linewidth,bb=0 0 1344 624,viewport=344 228 664 408,clip]{figures/small_motion/stones/depth.png}\end{minipage}\hfill\begin{minipage}[t]{0.194\textwidth}\centering\fontsize{7.4}{8.5}\selectfont\strut Vanilla$^{\dagger}$\par\vspace{1pt}\includegraphics[width=\linewidth,bb=0 0 1344 624,viewport=680 228 1000 408,clip]{figures/small_motion/stones/depth.png}\end{minipage}\hfill\begin{minipage}[t]{0.194\textwidth}\centering\fontsize{7.4}{8.5}\selectfont\strut Mip-Splatting\par\vspace{1pt}\includegraphics[width=\linewidth,bb=0 0 1344 624,viewport=1016 228 1336 408,clip]{figures/small_motion/stones/depth.png}\end{minipage}\hfill\begin{minipage}[t]{0.194\textwidth}\centering\fontsize{7.4}{8.5}\selectfont\strut \textbf{NBGS}\par\vspace{1pt}\includegraphics[width=\linewidth,bb=0 0 1344 624,viewport=8 20 328 200,clip]{figures/small_motion/stones/depth.png}\end{minipage}\par\vspace{4pt}
\noindent{\small\bfseries Absolute RGB error}\par\vspace{2pt}
\noindent \begin{minipage}[t]{0.194\textwidth}\centering\fontsize{7.4}{8.5}\selectfont\strut RGB error, $[0,3]$\par\vspace{4pt}\raggedright Channel-summed absolute error. Method-matched RGB reference.\par\vspace{5pt}\includegraphics[width=\linewidth,height=5pt]{figures/legends/small_motion_error_scale.png}\par 0\hfill 3\end{minipage}\hfill\begin{minipage}[t]{0.194\textwidth}\centering\fontsize{7.4}{8.5}\selectfont\strut Vanilla original\par\vspace{1pt}\includegraphics[width=\linewidth,bb=0 0 1344 624,viewport=8 436 328 616,clip]{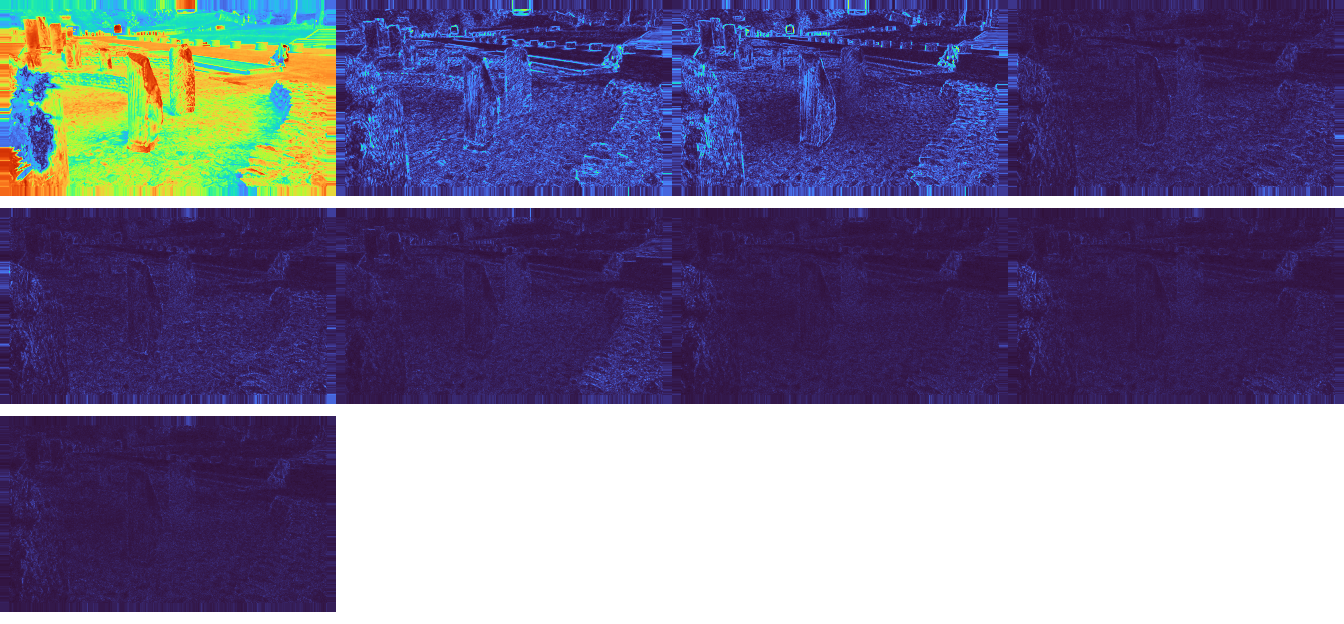}\end{minipage}\hfill\begin{minipage}[t]{0.194\textwidth}\centering\fontsize{7.4}{8.5}\selectfont\strut Small GS\par\vspace{1pt}\includegraphics[width=\linewidth,bb=0 0 1344 624,viewport=344 436 664 616,clip]{figures/small_motion/stones/rgb_error.png}\end{minipage}\hfill\begin{minipage}[t]{0.194\textwidth}\centering\fontsize{7.4}{8.5}\selectfont\strut InstantSplat\par\vspace{1pt}\includegraphics[width=\linewidth,bb=0 0 1344 624,viewport=680 436 1000 616,clip]{figures/small_motion/stones/rgb_error.png}\end{minipage}\hfill\begin{minipage}[t]{0.194\textwidth}\centering\fontsize{7.4}{8.5}\selectfont\strut DDDrop\par\vspace{1pt}\includegraphics[width=\linewidth,bb=0 0 1344 624,viewport=1016 436 1336 616,clip]{figures/small_motion/stones/rgb_error.png}\end{minipage}\par\vspace{4pt}
\noindent \begin{minipage}[t]{0.194\textwidth}\centering\fontsize{7.4}{8.5}\selectfont\strut DDGS\par\vspace{1pt}\includegraphics[width=\linewidth,bb=0 0 1344 624,viewport=8 228 328 408,clip]{figures/small_motion/stones/rgb_error.png}\end{minipage}\hfill\begin{minipage}[t]{0.194\textwidth}\centering\fontsize{7.4}{8.5}\selectfont\strut G4Splat\par\vspace{1pt}\includegraphics[width=\linewidth,bb=0 0 1344 624,viewport=344 228 664 408,clip]{figures/small_motion/stones/rgb_error.png}\end{minipage}\hfill\begin{minipage}[t]{0.194\textwidth}\centering\fontsize{7.4}{8.5}\selectfont\strut Vanilla$^{\dagger}$\par\vspace{1pt}\includegraphics[width=\linewidth,bb=0 0 1344 624,viewport=680 228 1000 408,clip]{figures/small_motion/stones/rgb_error.png}\end{minipage}\hfill\begin{minipage}[t]{0.194\textwidth}\centering\fontsize{7.4}{8.5}\selectfont\strut Mip-Splatting\par\vspace{1pt}\includegraphics[width=\linewidth,bb=0 0 1344 624,viewport=1016 228 1336 408,clip]{figures/small_motion/stones/rgb_error.png}\end{minipage}\hfill\begin{minipage}[t]{0.194\textwidth}\centering\fontsize{7.4}{8.5}\selectfont\strut \textbf{NBGS}\par\vspace{1pt}\includegraphics[width=\linewidth,bb=0 0 1344 624,viewport=8 20 328 200,clip]{figures/small_motion/stones/rgb_error.png}\end{minipage}\par\vspace{4pt}
\end{minipage}\end{lrbox}
\PlaceSuppScene
\clearpage
\pdfbookmark[3]{Trees}{supp.scene.trees}
\begin{lrbox}{\SuppSceneBox}\begin{minipage}{\textwidth}
\noindent{\large\bfseries Trees}\hfill{\small\color{black!65}Small-Motion dataset}\par\vspace{3pt}
\noindent{\small\bfseries Rendered RGB}\par\vspace{2pt}
\noindent \begin{minipage}[t]{0.194\textwidth}\centering\fontsize{7.4}{8.5}\selectfont\strut Original RGB\par\vspace{1pt}\includegraphics[width=\linewidth,bb=0 0 1344 624,viewport=8 436 328 616,clip]{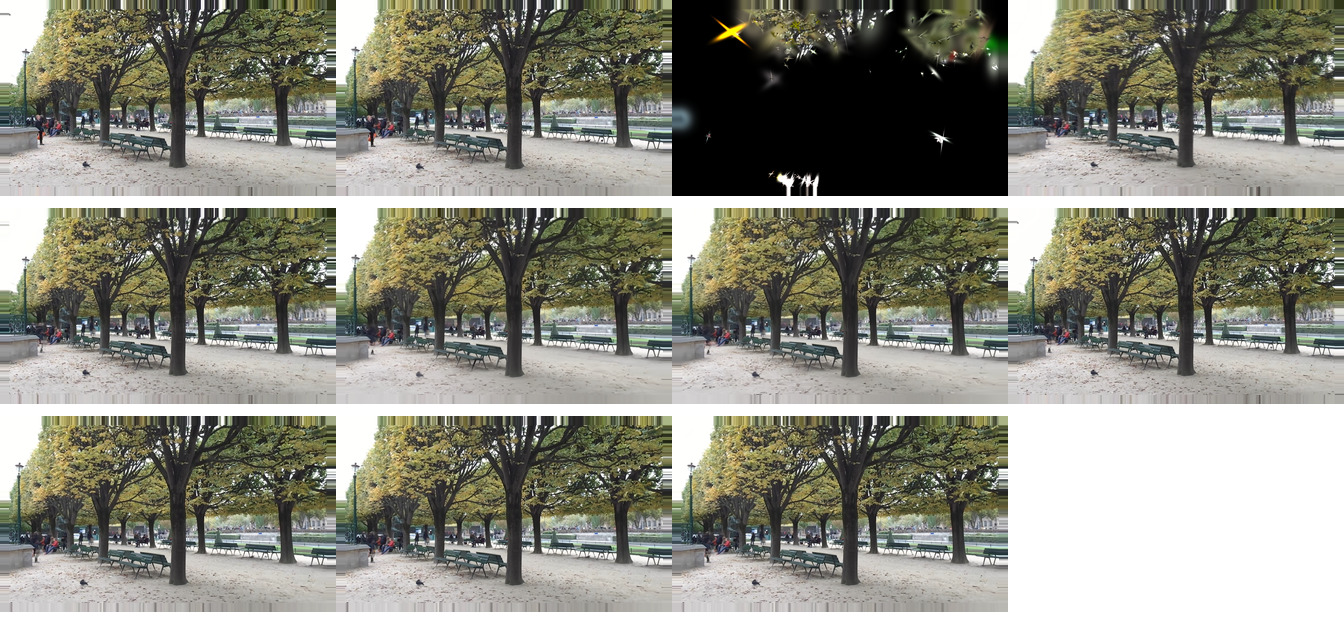}\par\vspace{2pt}\strut SfSM-undistorted RGB\par\vspace{1pt}\includegraphics[width=\linewidth,bb=0 0 1344 624,viewport=344 436 664 616,clip]{figures/small_motion/trees/rgb.jpg}\end{minipage}\hfill\begin{minipage}[t]{0.194\textwidth}\centering\fontsize{7.4}{8.5}\selectfont\strut Vanilla original\par\vspace{1pt}\setlength{\unitlength}{\dimexpr\linewidth/1000\relax}\begin{picture}(1000,562.500)\put(0,0){\includegraphics[width=\linewidth,bb=0 0 1344 624,viewport=680 436 1000 616,clip]{figures/small_motion/trees/rgb.jpg}}\put(421.875,171.875){\color{suppcyan}\linethickness{0.45pt}\framebox(187.500,187.500){} }\put(203.125,54.688){\color{supporange}\linethickness{0.45pt}\framebox(187.500,187.500){} }\end{picture}\par\vspace{2pt}\hbox to\linewidth{{\setlength{\fboxsep}{0pt}\color{suppcyan}\fbox{\includegraphics[width=0.478\linewidth,bb=0 0 960 1200,viewport=8 968 232 1192,clip]{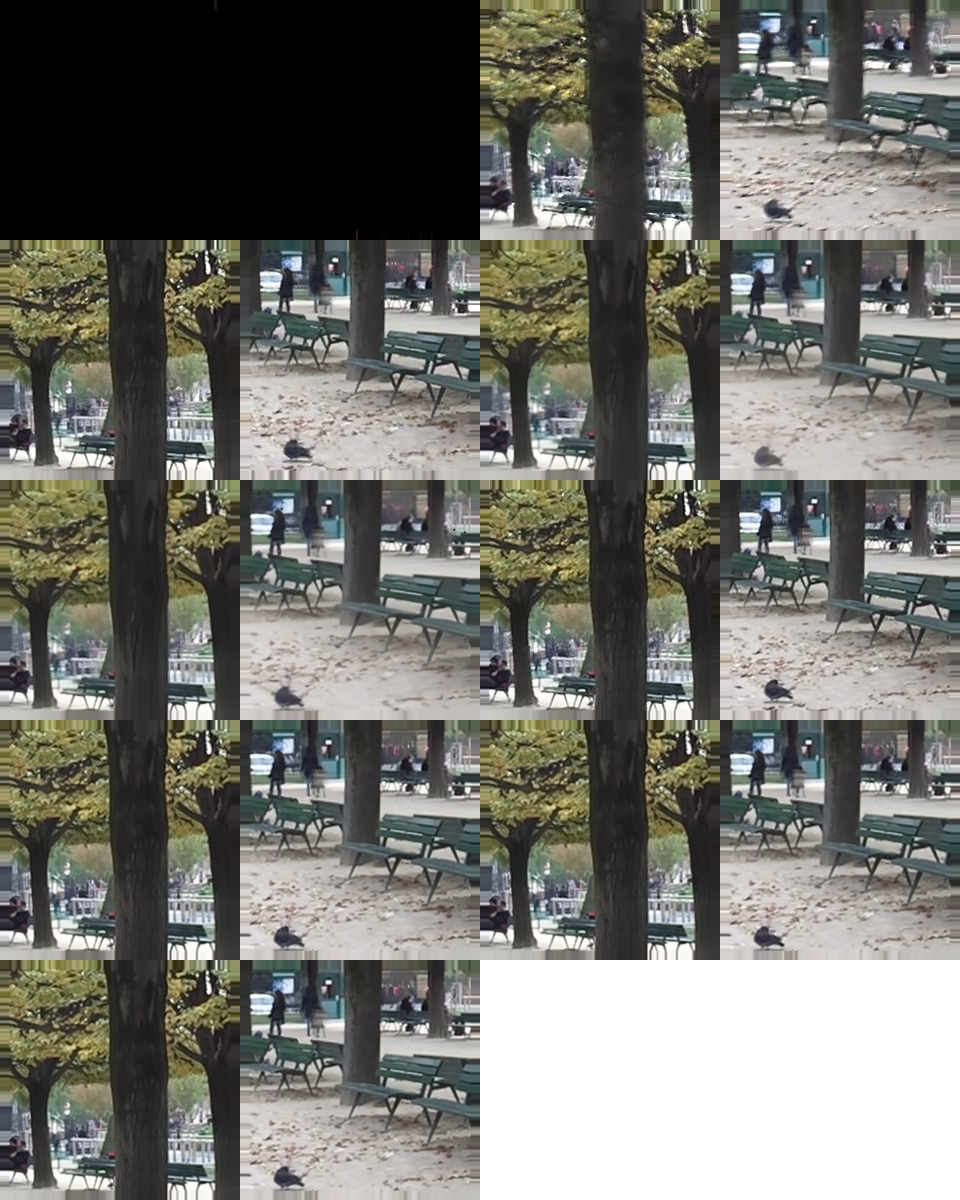}}}\hfill{\setlength{\fboxsep}{0pt}\color{supporange}\fbox{\includegraphics[width=0.478\linewidth,bb=0 0 960 1200,viewport=248 968 472 1192,clip]{figures/small_motion/trees/rgb_details.jpg}}}}\end{minipage}\hfill\begin{minipage}[t]{0.194\textwidth}\centering\fontsize{7.4}{8.5}\selectfont\strut Small GS\par\vspace{1pt}\setlength{\unitlength}{\dimexpr\linewidth/1000\relax}\begin{picture}(1000,562.500)\put(0,0){\includegraphics[width=\linewidth,bb=0 0 1344 624,viewport=1016 436 1336 616,clip]{figures/small_motion/trees/rgb.jpg}}\put(421.875,171.875){\color{suppcyan}\linethickness{0.45pt}\framebox(187.500,187.500){} }\put(203.125,54.688){\color{supporange}\linethickness{0.45pt}\framebox(187.500,187.500){} }\end{picture}\par\vspace{2pt}\hbox to\linewidth{{\setlength{\fboxsep}{0pt}\color{suppcyan}\fbox{\includegraphics[width=0.478\linewidth,bb=0 0 960 1200,viewport=488 968 712 1192,clip]{figures/small_motion/trees/rgb_details.jpg}}}\hfill{\setlength{\fboxsep}{0pt}\color{supporange}\fbox{\includegraphics[width=0.478\linewidth,bb=0 0 960 1200,viewport=728 968 952 1192,clip]{figures/small_motion/trees/rgb_details.jpg}}}}\end{minipage}\hfill\begin{minipage}[t]{0.194\textwidth}\centering\fontsize{7.4}{8.5}\selectfont\strut InstantSplat\par\vspace{1pt}\setlength{\unitlength}{\dimexpr\linewidth/1000\relax}\begin{picture}(1000,562.500)\put(0,0){\includegraphics[width=\linewidth,bb=0 0 1344 624,viewport=8 228 328 408,clip]{figures/small_motion/trees/rgb.jpg}}\put(421.875,171.875){\color{suppcyan}\linethickness{0.45pt}\framebox(187.500,187.500){} }\put(203.125,54.688){\color{supporange}\linethickness{0.45pt}\framebox(187.500,187.500){} }\end{picture}\par\vspace{2pt}\hbox to\linewidth{{\setlength{\fboxsep}{0pt}\color{suppcyan}\fbox{\includegraphics[width=0.478\linewidth,bb=0 0 960 1200,viewport=8 728 232 952,clip]{figures/small_motion/trees/rgb_details.jpg}}}\hfill{\setlength{\fboxsep}{0pt}\color{supporange}\fbox{\includegraphics[width=0.478\linewidth,bb=0 0 960 1200,viewport=248 728 472 952,clip]{figures/small_motion/trees/rgb_details.jpg}}}}\end{minipage}\hfill\begin{minipage}[t]{0.194\textwidth}\centering\fontsize{7.4}{8.5}\selectfont\strut DDDrop\par\vspace{1pt}\setlength{\unitlength}{\dimexpr\linewidth/1000\relax}\begin{picture}(1000,562.500)\put(0,0){\includegraphics[width=\linewidth,bb=0 0 1344 624,viewport=344 228 664 408,clip]{figures/small_motion/trees/rgb.jpg}}\put(421.875,171.875){\color{suppcyan}\linethickness{0.45pt}\framebox(187.500,187.500){} }\put(203.125,54.688){\color{supporange}\linethickness{0.45pt}\framebox(187.500,187.500){} }\end{picture}\par\vspace{2pt}\hbox to\linewidth{{\setlength{\fboxsep}{0pt}\color{suppcyan}\fbox{\includegraphics[width=0.478\linewidth,bb=0 0 960 1200,viewport=488 728 712 952,clip]{figures/small_motion/trees/rgb_details.jpg}}}\hfill{\setlength{\fboxsep}{0pt}\color{supporange}\fbox{\includegraphics[width=0.478\linewidth,bb=0 0 960 1200,viewport=728 728 952 952,clip]{figures/small_motion/trees/rgb_details.jpg}}}}\end{minipage}\par\vspace{4pt}
\noindent \begin{minipage}[t]{0.194\textwidth}\centering\fontsize{7.4}{8.5}\selectfont\strut DDGS\par\vspace{1pt}\setlength{\unitlength}{\dimexpr\linewidth/1000\relax}\begin{picture}(1000,562.500)\put(0,0){\includegraphics[width=\linewidth,bb=0 0 1344 624,viewport=680 228 1000 408,clip]{figures/small_motion/trees/rgb.jpg}}\put(421.875,171.875){\color{suppcyan}\linethickness{0.45pt}\framebox(187.500,187.500){} }\put(203.125,54.688){\color{supporange}\linethickness{0.45pt}\framebox(187.500,187.500){} }\end{picture}\par\vspace{2pt}\hbox to\linewidth{{\setlength{\fboxsep}{0pt}\color{suppcyan}\fbox{\includegraphics[width=0.478\linewidth,bb=0 0 960 1200,viewport=8 488 232 712,clip]{figures/small_motion/trees/rgb_details.jpg}}}\hfill{\setlength{\fboxsep}{0pt}\color{supporange}\fbox{\includegraphics[width=0.478\linewidth,bb=0 0 960 1200,viewport=248 488 472 712,clip]{figures/small_motion/trees/rgb_details.jpg}}}}\end{minipage}\hfill\begin{minipage}[t]{0.194\textwidth}\centering\fontsize{7.4}{8.5}\selectfont\strut G4Splat\par\vspace{1pt}\setlength{\unitlength}{\dimexpr\linewidth/1000\relax}\begin{picture}(1000,562.500)\put(0,0){\includegraphics[width=\linewidth,bb=0 0 1344 624,viewport=1016 228 1336 408,clip]{figures/small_motion/trees/rgb.jpg}}\put(421.875,171.875){\color{suppcyan}\linethickness{0.45pt}\framebox(187.500,187.500){} }\put(203.125,54.688){\color{supporange}\linethickness{0.45pt}\framebox(187.500,187.500){} }\end{picture}\par\vspace{2pt}\hbox to\linewidth{{\setlength{\fboxsep}{0pt}\color{suppcyan}\fbox{\includegraphics[width=0.478\linewidth,bb=0 0 960 1200,viewport=488 488 712 712,clip]{figures/small_motion/trees/rgb_details.jpg}}}\hfill{\setlength{\fboxsep}{0pt}\color{supporange}\fbox{\includegraphics[width=0.478\linewidth,bb=0 0 960 1200,viewport=728 488 952 712,clip]{figures/small_motion/trees/rgb_details.jpg}}}}\end{minipage}\hfill\begin{minipage}[t]{0.194\textwidth}\centering\fontsize{7.4}{8.5}\selectfont\strut Vanilla$^{\dagger}$\par\vspace{1pt}\setlength{\unitlength}{\dimexpr\linewidth/1000\relax}\begin{picture}(1000,562.500)\put(0,0){\includegraphics[width=\linewidth,bb=0 0 1344 624,viewport=8 20 328 200,clip]{figures/small_motion/trees/rgb.jpg}}\put(421.875,171.875){\color{suppcyan}\linethickness{0.45pt}\framebox(187.500,187.500){} }\put(203.125,54.688){\color{supporange}\linethickness{0.45pt}\framebox(187.500,187.500){} }\end{picture}\par\vspace{2pt}\hbox to\linewidth{{\setlength{\fboxsep}{0pt}\color{suppcyan}\fbox{\includegraphics[width=0.478\linewidth,bb=0 0 960 1200,viewport=8 248 232 472,clip]{figures/small_motion/trees/rgb_details.jpg}}}\hfill{\setlength{\fboxsep}{0pt}\color{supporange}\fbox{\includegraphics[width=0.478\linewidth,bb=0 0 960 1200,viewport=248 248 472 472,clip]{figures/small_motion/trees/rgb_details.jpg}}}}\end{minipage}\hfill\begin{minipage}[t]{0.194\textwidth}\centering\fontsize{7.4}{8.5}\selectfont\strut Mip-Splatting\par\vspace{1pt}\setlength{\unitlength}{\dimexpr\linewidth/1000\relax}\begin{picture}(1000,562.500)\put(0,0){\includegraphics[width=\linewidth,bb=0 0 1344 624,viewport=344 20 664 200,clip]{figures/small_motion/trees/rgb.jpg}}\put(421.875,171.875){\color{suppcyan}\linethickness{0.45pt}\framebox(187.500,187.500){} }\put(203.125,54.688){\color{supporange}\linethickness{0.45pt}\framebox(187.500,187.500){} }\end{picture}\par\vspace{2pt}\hbox to\linewidth{{\setlength{\fboxsep}{0pt}\color{suppcyan}\fbox{\includegraphics[width=0.478\linewidth,bb=0 0 960 1200,viewport=488 248 712 472,clip]{figures/small_motion/trees/rgb_details.jpg}}}\hfill{\setlength{\fboxsep}{0pt}\color{supporange}\fbox{\includegraphics[width=0.478\linewidth,bb=0 0 960 1200,viewport=728 248 952 472,clip]{figures/small_motion/trees/rgb_details.jpg}}}}\end{minipage}\hfill\begin{minipage}[t]{0.194\textwidth}\centering\fontsize{7.4}{8.5}\selectfont\strut \textbf{NBGS}\par\vspace{1pt}\setlength{\unitlength}{\dimexpr\linewidth/1000\relax}\begin{picture}(1000,562.500)\put(0,0){\includegraphics[width=\linewidth,bb=0 0 1344 624,viewport=680 20 1000 200,clip]{figures/small_motion/trees/rgb.jpg}}\put(421.875,171.875){\color{suppcyan}\linethickness{0.45pt}\framebox(187.500,187.500){} }\put(203.125,54.688){\color{supporange}\linethickness{0.45pt}\framebox(187.500,187.500){} }\end{picture}\par\vspace{2pt}\hbox to\linewidth{{\setlength{\fboxsep}{0pt}\color{suppcyan}\fbox{\includegraphics[width=0.478\linewidth,bb=0 0 960 1200,viewport=8 8 232 232,clip]{figures/small_motion/trees/rgb_details.jpg}}}\hfill{\setlength{\fboxsep}{0pt}\color{supporange}\fbox{\includegraphics[width=0.478\linewidth,bb=0 0 960 1200,viewport=248 8 472 232,clip]{figures/small_motion/trees/rgb_details.jpg}}}}\end{minipage}\par\vspace{4pt}
\noindent{\small\bfseries Relative rendered depth}\par\vspace{2pt}
\noindent \begin{minipage}[t]{0.194\textwidth}\centering\fontsize{7.4}{8.5}\selectfont\strut Relative depth\par\vspace{4pt}\raggedright Per-method p02--p98. Black: invalid. No depth ground truth or metric units.\par\vspace{5pt}\includegraphics[width=\linewidth,height=5pt]{figures/legends/small_motion_depth_scale.png}\par near\hfill far\end{minipage}\hfill\begin{minipage}[t]{0.194\textwidth}\centering\fontsize{7.4}{8.5}\selectfont\strut Vanilla original\par\vspace{1pt}\includegraphics[width=\linewidth,bb=0 0 1344 624,viewport=8 436 328 616,clip]{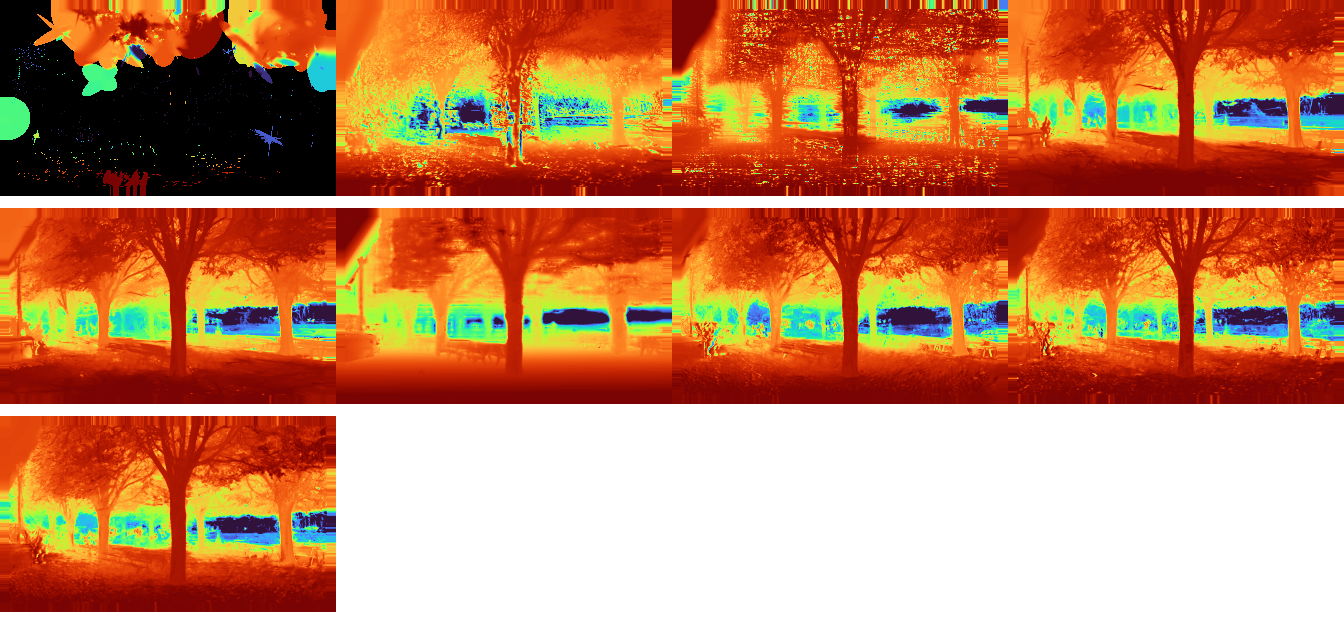}\end{minipage}\hfill\begin{minipage}[t]{0.194\textwidth}\centering\fontsize{7.4}{8.5}\selectfont\strut Small GS\par\vspace{1pt}\includegraphics[width=\linewidth,bb=0 0 1344 624,viewport=344 436 664 616,clip]{figures/small_motion/trees/depth.png}\end{minipage}\hfill\begin{minipage}[t]{0.194\textwidth}\centering\fontsize{7.4}{8.5}\selectfont\strut InstantSplat\par\vspace{1pt}\includegraphics[width=\linewidth,bb=0 0 1344 624,viewport=680 436 1000 616,clip]{figures/small_motion/trees/depth.png}\end{minipage}\hfill\begin{minipage}[t]{0.194\textwidth}\centering\fontsize{7.4}{8.5}\selectfont\strut DDDrop\par\vspace{1pt}\includegraphics[width=\linewidth,bb=0 0 1344 624,viewport=1016 436 1336 616,clip]{figures/small_motion/trees/depth.png}\end{minipage}\par\vspace{4pt}
\noindent \begin{minipage}[t]{0.194\textwidth}\centering\fontsize{7.4}{8.5}\selectfont\strut DDGS\par\vspace{1pt}\includegraphics[width=\linewidth,bb=0 0 1344 624,viewport=8 228 328 408,clip]{figures/small_motion/trees/depth.png}\end{minipage}\hfill\begin{minipage}[t]{0.194\textwidth}\centering\fontsize{7.4}{8.5}\selectfont\strut G4Splat\par\vspace{1pt}\includegraphics[width=\linewidth,bb=0 0 1344 624,viewport=344 228 664 408,clip]{figures/small_motion/trees/depth.png}\end{minipage}\hfill\begin{minipage}[t]{0.194\textwidth}\centering\fontsize{7.4}{8.5}\selectfont\strut Vanilla$^{\dagger}$\par\vspace{1pt}\includegraphics[width=\linewidth,bb=0 0 1344 624,viewport=680 228 1000 408,clip]{figures/small_motion/trees/depth.png}\end{minipage}\hfill\begin{minipage}[t]{0.194\textwidth}\centering\fontsize{7.4}{8.5}\selectfont\strut Mip-Splatting\par\vspace{1pt}\includegraphics[width=\linewidth,bb=0 0 1344 624,viewport=1016 228 1336 408,clip]{figures/small_motion/trees/depth.png}\end{minipage}\hfill\begin{minipage}[t]{0.194\textwidth}\centering\fontsize{7.4}{8.5}\selectfont\strut \textbf{NBGS}\par\vspace{1pt}\includegraphics[width=\linewidth,bb=0 0 1344 624,viewport=8 20 328 200,clip]{figures/small_motion/trees/depth.png}\end{minipage}\par\vspace{4pt}
\noindent{\small\bfseries Absolute RGB error}\par\vspace{2pt}
\noindent \begin{minipage}[t]{0.194\textwidth}\centering\fontsize{7.4}{8.5}\selectfont\strut RGB error, $[0,3]$\par\vspace{4pt}\raggedright Channel-summed absolute error. Method-matched RGB reference.\par\vspace{5pt}\includegraphics[width=\linewidth,height=5pt]{figures/legends/small_motion_error_scale.png}\par 0\hfill 3\end{minipage}\hfill\begin{minipage}[t]{0.194\textwidth}\centering\fontsize{7.4}{8.5}\selectfont\strut Vanilla original\par\vspace{1pt}\includegraphics[width=\linewidth,bb=0 0 1344 624,viewport=8 436 328 616,clip]{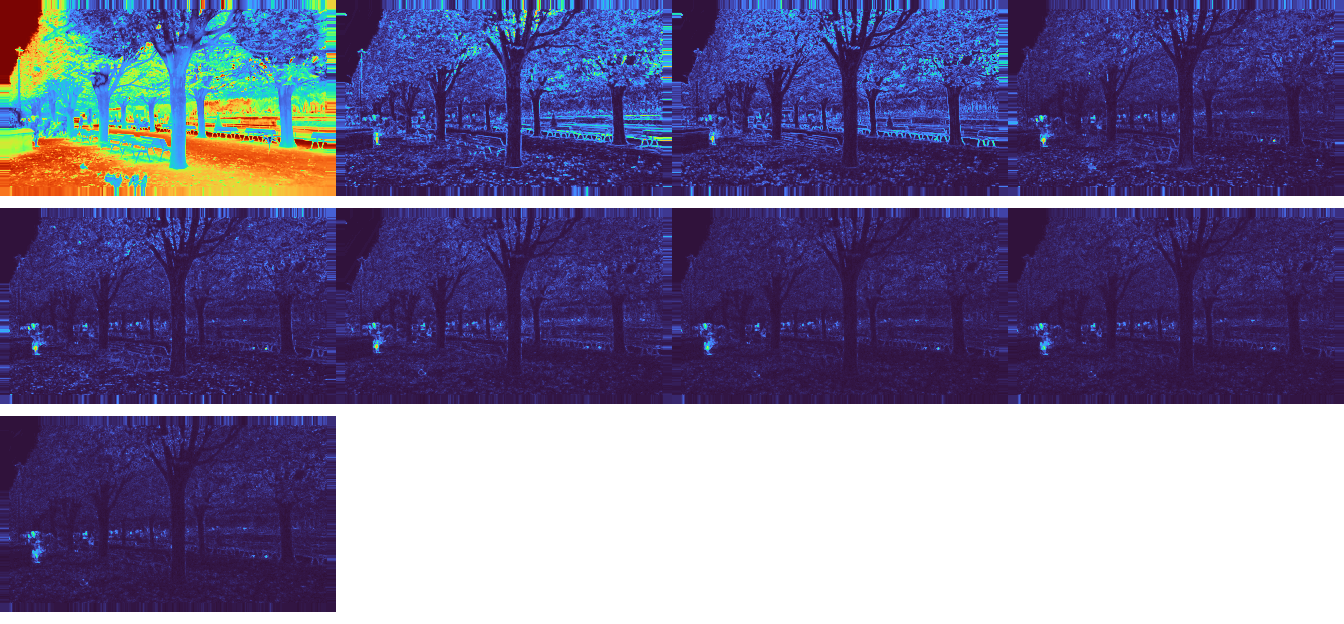}\end{minipage}\hfill\begin{minipage}[t]{0.194\textwidth}\centering\fontsize{7.4}{8.5}\selectfont\strut Small GS\par\vspace{1pt}\includegraphics[width=\linewidth,bb=0 0 1344 624,viewport=344 436 664 616,clip]{figures/small_motion/trees/rgb_error.png}\end{minipage}\hfill\begin{minipage}[t]{0.194\textwidth}\centering\fontsize{7.4}{8.5}\selectfont\strut InstantSplat\par\vspace{1pt}\includegraphics[width=\linewidth,bb=0 0 1344 624,viewport=680 436 1000 616,clip]{figures/small_motion/trees/rgb_error.png}\end{minipage}\hfill\begin{minipage}[t]{0.194\textwidth}\centering\fontsize{7.4}{8.5}\selectfont\strut DDDrop\par\vspace{1pt}\includegraphics[width=\linewidth,bb=0 0 1344 624,viewport=1016 436 1336 616,clip]{figures/small_motion/trees/rgb_error.png}\end{minipage}\par\vspace{4pt}
\noindent \begin{minipage}[t]{0.194\textwidth}\centering\fontsize{7.4}{8.5}\selectfont\strut DDGS\par\vspace{1pt}\includegraphics[width=\linewidth,bb=0 0 1344 624,viewport=8 228 328 408,clip]{figures/small_motion/trees/rgb_error.png}\end{minipage}\hfill\begin{minipage}[t]{0.194\textwidth}\centering\fontsize{7.4}{8.5}\selectfont\strut G4Splat\par\vspace{1pt}\includegraphics[width=\linewidth,bb=0 0 1344 624,viewport=344 228 664 408,clip]{figures/small_motion/trees/rgb_error.png}\end{minipage}\hfill\begin{minipage}[t]{0.194\textwidth}\centering\fontsize{7.4}{8.5}\selectfont\strut Vanilla$^{\dagger}$\par\vspace{1pt}\includegraphics[width=\linewidth,bb=0 0 1344 624,viewport=680 228 1000 408,clip]{figures/small_motion/trees/rgb_error.png}\end{minipage}\hfill\begin{minipage}[t]{0.194\textwidth}\centering\fontsize{7.4}{8.5}\selectfont\strut Mip-Splatting\par\vspace{1pt}\includegraphics[width=\linewidth,bb=0 0 1344 624,viewport=1016 228 1336 408,clip]{figures/small_motion/trees/rgb_error.png}\end{minipage}\hfill\begin{minipage}[t]{0.194\textwidth}\centering\fontsize{7.4}{8.5}\selectfont\strut \textbf{NBGS}\par\vspace{1pt}\includegraphics[width=\linewidth,bb=0 0 1344 624,viewport=8 20 328 200,clip]{figures/small_motion/trees/rgb_error.png}\end{minipage}\par\vspace{4pt}
\end{minipage}\end{lrbox}
\PlaceSuppScene
\clearpage

%% file: sections/phone_comparisons.tex
\clearpage
\subsection{DyCheck iPhone Clips}\label{app:dycheck_visualizations}
\pdfbookmark[3]{apple (iPhone)}{supp.phone.apple}
\phantomsection\label{fig:supp_phone_apple}{\large\bfseries apple (iPhone)}\hfill{\fontsize{8}{9}\selectfont DyCheck dataset~\cite{gao2022dynamic}}\par\vspace{4pt}
\noindent \begin{minipage}[t]{0.18\textwidth}\centering\fontsize{7}{8}\selectfont RGB reference\par\vspace{2pt}\includegraphics[width=\linewidth,bb=0 0 1408 1392,viewport=8 936 344 1384,clip]{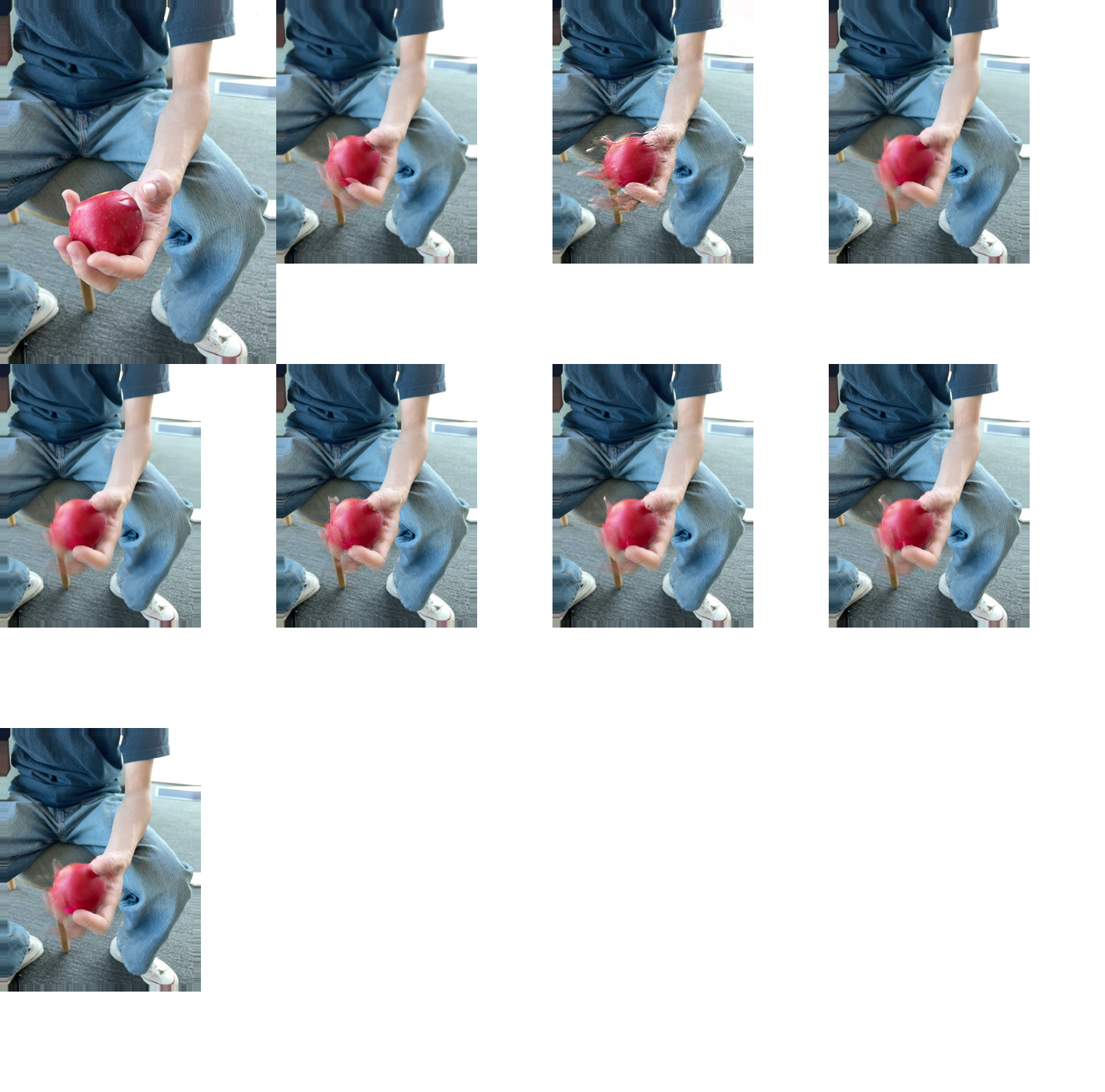}\end{minipage}\hfill\begin{minipage}[t]{0.18\textwidth}\centering\fontsize{7}{8}\selectfont LiDAR reference\par\vspace{2pt}\includegraphics[width=\linewidth,bb=0 0 1408 1392,viewport=8 936 344 1384,clip]{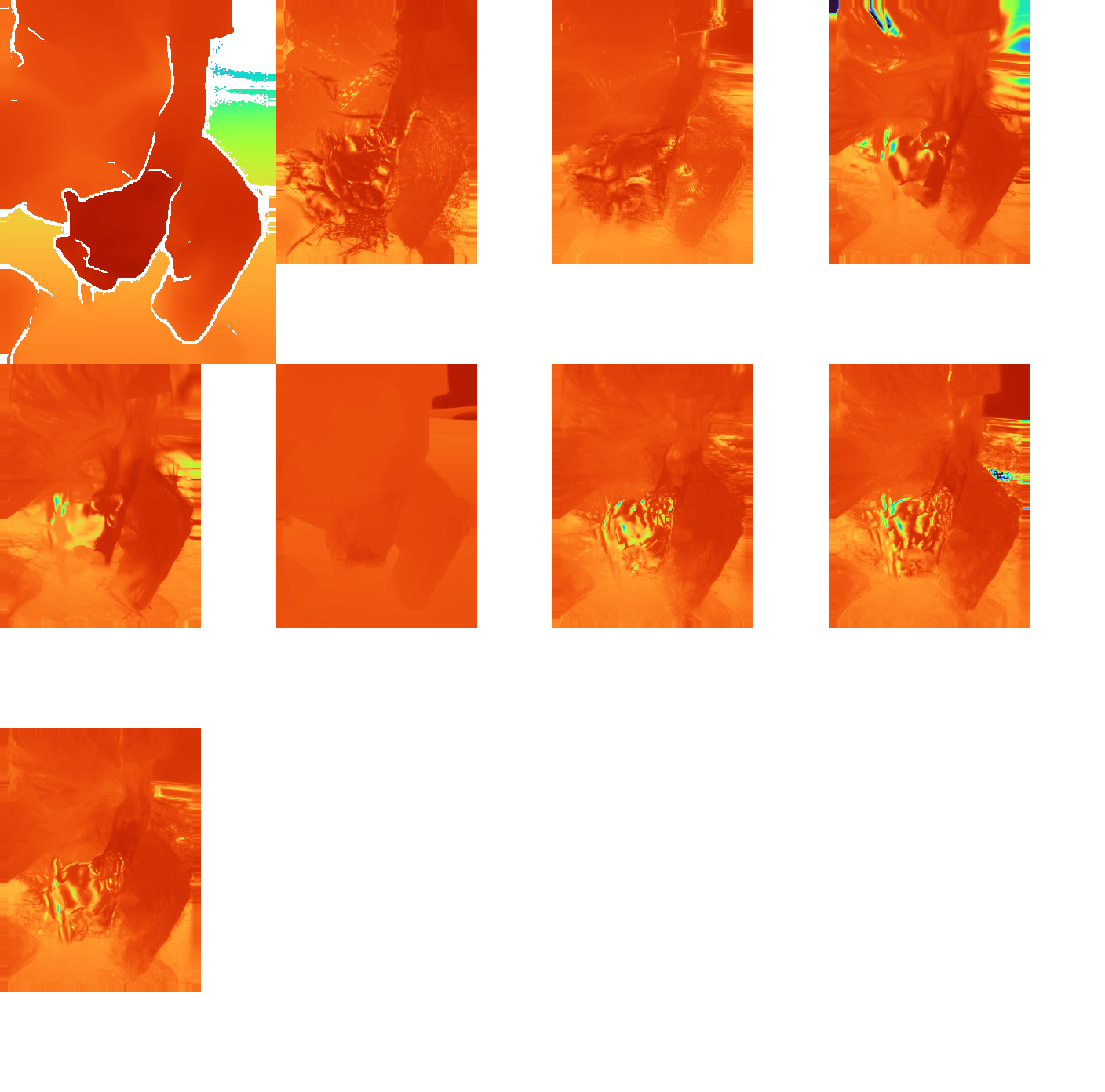}\end{minipage}\hfill\begin{minipage}[t]{0.625\textwidth}\fontsize{8}{9}\selectfont First held-out view, seed 0: \texttt{frame\_006}. Compare each render with the RGB reference, then compare its predicted depth with LiDAR. All depths use the same meter scale. White marks invalid depth, and values above $3$\,m are saturated.\par\vspace{4pt}Depth (m):\par\includegraphics[width=\linewidth,height=5pt]{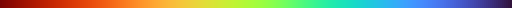}\par $0$\hfill $3$\par\vspace{3pt}RGB error: $\sum_c|I_c-\widehat I_c|$\par\includegraphics[width=\linewidth,height=5pt]{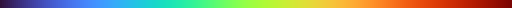}\par $0$\hfill $3$\par\vspace{3pt}Depth error: $|s\widehat Z-Z_{\mathrm{LiDAR}}|$ (m)\par\includegraphics[width=\linewidth,height=5pt]{figures/legends/dycheck_error_scale.png}\par $0$\hfill $\geq1.5$\par\vspace{3pt}{\fontsize{7}{8}\selectfont Depth errors are measured on valid static pixels shared by all methods. White marks excluded pixels. Both displayed phone variants, NBGS-Base and Vanilla (reset off), disable opacity reset and its associated size pruning.}\end{minipage}\par\vspace{5pt}
\noindent \begin{minipage}[t]{0.492\textwidth}\centering{\fontsize{8}{9}\selectfont Small GS}\par\vspace{2pt}\begin{minipage}[t]{0.242\linewidth}\centering\includegraphics[width=\linewidth,bb=0 0 1408 1392,viewport=360 1064 600 1384,clip]{figures/dycheck/apple/rgb.jpg}\par\vspace{1pt}{\fontsize{7}{8}\selectfont RGB}\end{minipage}\hfill\begin{minipage}[t]{0.242\linewidth}\centering\includegraphics[width=\linewidth,bb=0 0 1408 1392,viewport=360 1064 600 1384,clip]{figures/dycheck/apple/depth.png}\par\vspace{1pt}{\fontsize{7}{8}\selectfont Depth}\end{minipage}\hfill\begin{minipage}[t]{0.242\linewidth}\centering\includegraphics[width=\linewidth,bb=0 0 1024 672,viewport=8 344 248 664,clip]{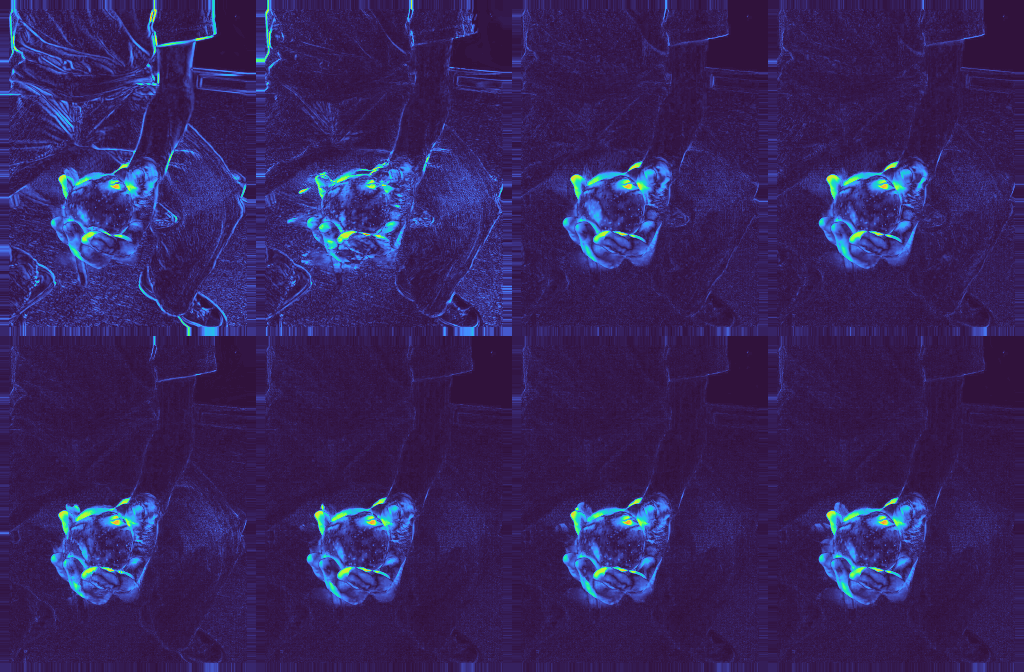}\par\vspace{1pt}{\fontsize{7}{8}\selectfont RGB error}\end{minipage}\hfill\begin{minipage}[t]{0.242\linewidth}\centering\includegraphics[width=\linewidth,bb=0 0 1024 672,viewport=8 344 248 664,clip]{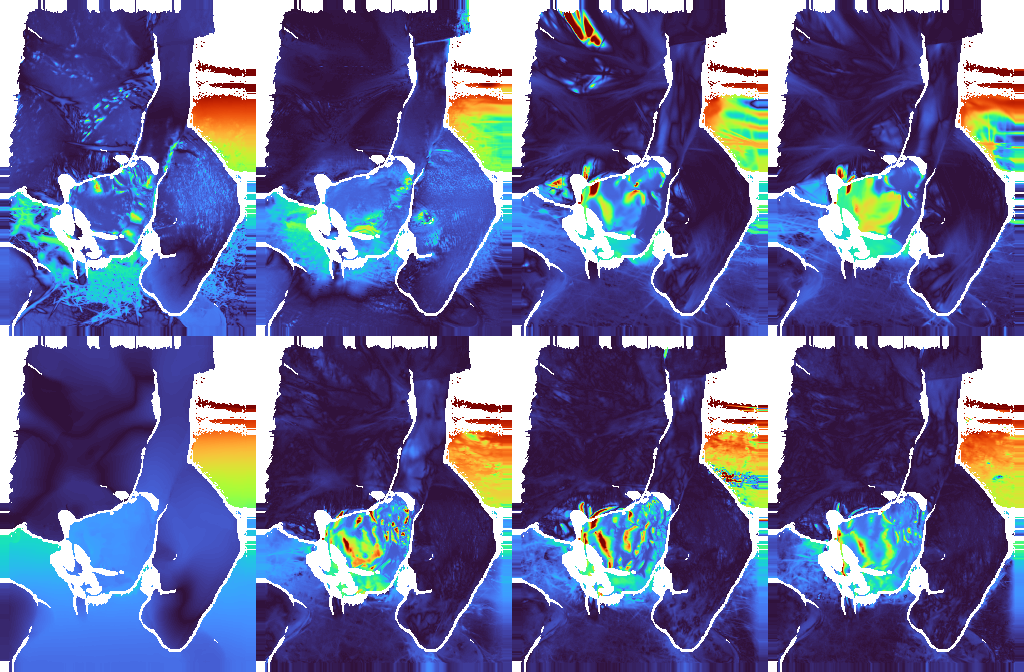}\par\vspace{1pt}{\fontsize{7}{8}\selectfont Depth error}\end{minipage}\end{minipage}\hfill\begin{minipage}[t]{0.492\textwidth}\centering{\fontsize{8}{9}\selectfont InstantSplat}\par\vspace{2pt}\begin{minipage}[t]{0.242\linewidth}\centering\includegraphics[width=\linewidth,bb=0 0 1408 1392,viewport=712 1064 952 1384,clip]{figures/dycheck/apple/rgb.jpg}\par\vspace{1pt}{\fontsize{7}{8}\selectfont RGB}\end{minipage}\hfill\begin{minipage}[t]{0.242\linewidth}\centering\includegraphics[width=\linewidth,bb=0 0 1408 1392,viewport=712 1064 952 1384,clip]{figures/dycheck/apple/depth.png}\par\vspace{1pt}{\fontsize{7}{8}\selectfont Depth}\end{minipage}\hfill\begin{minipage}[t]{0.242\linewidth}\centering\includegraphics[width=\linewidth,bb=0 0 1024 672,viewport=264 344 504 664,clip]{figures/dycheck/apple/rgb_error.png}\par\vspace{1pt}{\fontsize{7}{8}\selectfont RGB error}\end{minipage}\hfill\begin{minipage}[t]{0.242\linewidth}\centering\includegraphics[width=\linewidth,bb=0 0 1024 672,viewport=264 344 504 664,clip]{figures/dycheck/apple/depth_error.png}\par\vspace{1pt}{\fontsize{7}{8}\selectfont Depth error}\end{minipage}\end{minipage}\par\vspace{16pt}
\noindent \begin{minipage}[t]{0.492\textwidth}\centering{\fontsize{8}{9}\selectfont DDDrop + SfSM}\par\vspace{2pt}\begin{minipage}[t]{0.242\linewidth}\centering\includegraphics[width=\linewidth,bb=0 0 1408 1392,viewport=1064 1064 1304 1384,clip]{figures/dycheck/apple/rgb.jpg}\par\vspace{1pt}{\fontsize{7}{8}\selectfont RGB}\end{minipage}\hfill\begin{minipage}[t]{0.242\linewidth}\centering\includegraphics[width=\linewidth,bb=0 0 1408 1392,viewport=1064 1064 1304 1384,clip]{figures/dycheck/apple/depth.png}\par\vspace{1pt}{\fontsize{7}{8}\selectfont Depth}\end{minipage}\hfill\begin{minipage}[t]{0.242\linewidth}\centering\includegraphics[width=\linewidth,bb=0 0 1024 672,viewport=520 344 760 664,clip]{figures/dycheck/apple/rgb_error.png}\par\vspace{1pt}{\fontsize{7}{8}\selectfont RGB error}\end{minipage}\hfill\begin{minipage}[t]{0.242\linewidth}\centering\includegraphics[width=\linewidth,bb=0 0 1024 672,viewport=520 344 760 664,clip]{figures/dycheck/apple/depth_error.png}\par\vspace{1pt}{\fontsize{7}{8}\selectfont Depth error}\end{minipage}\end{minipage}\hfill\begin{minipage}[t]{0.492\textwidth}\centering{\fontsize{8}{9}\selectfont DDGS + SfSM}\par\vspace{2pt}\begin{minipage}[t]{0.242\linewidth}\centering\includegraphics[width=\linewidth,bb=0 0 1408 1392,viewport=8 600 248 920,clip]{figures/dycheck/apple/rgb.jpg}\par\vspace{1pt}{\fontsize{7}{8}\selectfont RGB}\end{minipage}\hfill\begin{minipage}[t]{0.242\linewidth}\centering\includegraphics[width=\linewidth,bb=0 0 1408 1392,viewport=8 600 248 920,clip]{figures/dycheck/apple/depth.png}\par\vspace{1pt}{\fontsize{7}{8}\selectfont Depth}\end{minipage}\hfill\begin{minipage}[t]{0.242\linewidth}\centering\includegraphics[width=\linewidth,bb=0 0 1024 672,viewport=776 344 1016 664,clip]{figures/dycheck/apple/rgb_error.png}\par\vspace{1pt}{\fontsize{7}{8}\selectfont RGB error}\end{minipage}\hfill\begin{minipage}[t]{0.242\linewidth}\centering\includegraphics[width=\linewidth,bb=0 0 1024 672,viewport=776 344 1016 664,clip]{figures/dycheck/apple/depth_error.png}\par\vspace{1pt}{\fontsize{7}{8}\selectfont Depth error}\end{minipage}\end{minipage}\par\vspace{16pt}
\noindent \begin{minipage}[t]{0.492\textwidth}\centering{\fontsize{8}{9}\selectfont G4Splat}\par\vspace{2pt}\begin{minipage}[t]{0.242\linewidth}\centering\includegraphics[width=\linewidth,bb=0 0 1408 1392,viewport=360 600 600 920,clip]{figures/dycheck/apple/rgb.jpg}\par\vspace{1pt}{\fontsize{7}{8}\selectfont RGB}\end{minipage}\hfill\begin{minipage}[t]{0.242\linewidth}\centering\includegraphics[width=\linewidth,bb=0 0 1408 1392,viewport=360 600 600 920,clip]{figures/dycheck/apple/depth.png}\par\vspace{1pt}{\fontsize{7}{8}\selectfont Depth}\end{minipage}\hfill\begin{minipage}[t]{0.242\linewidth}\centering\includegraphics[width=\linewidth,bb=0 0 1024 672,viewport=8 8 248 328,clip]{figures/dycheck/apple/rgb_error.png}\par\vspace{1pt}{\fontsize{7}{8}\selectfont RGB error}\end{minipage}\hfill\begin{minipage}[t]{0.242\linewidth}\centering\includegraphics[width=\linewidth,bb=0 0 1024 672,viewport=8 8 248 328,clip]{figures/dycheck/apple/depth_error.png}\par\vspace{1pt}{\fontsize{7}{8}\selectfont Depth error}\end{minipage}\end{minipage}\hfill\begin{minipage}[t]{0.492\textwidth}\centering{\fontsize{8}{9}\selectfont Vanilla (reset off)}\par\vspace{2pt}\begin{minipage}[t]{0.242\linewidth}\centering\includegraphics[width=\linewidth,bb=0 0 1408 1392,viewport=712 600 952 920,clip]{figures/dycheck/apple/rgb.jpg}\par\vspace{1pt}{\fontsize{7}{8}\selectfont RGB}\end{minipage}\hfill\begin{minipage}[t]{0.242\linewidth}\centering\includegraphics[width=\linewidth,bb=0 0 1408 1392,viewport=712 600 952 920,clip]{figures/dycheck/apple/depth.png}\par\vspace{1pt}{\fontsize{7}{8}\selectfont Depth}\end{minipage}\hfill\begin{minipage}[t]{0.242\linewidth}\centering\includegraphics[width=\linewidth,bb=0 0 1024 672,viewport=264 8 504 328,clip]{figures/dycheck/apple/rgb_error.png}\par\vspace{1pt}{\fontsize{7}{8}\selectfont RGB error}\end{minipage}\hfill\begin{minipage}[t]{0.242\linewidth}\centering\includegraphics[width=\linewidth,bb=0 0 1024 672,viewport=264 8 504 328,clip]{figures/dycheck/apple/depth_error.png}\par\vspace{1pt}{\fontsize{7}{8}\selectfont Depth error}\end{minipage}\end{minipage}\par\vspace{16pt}
\noindent \begin{minipage}[t]{0.492\textwidth}\centering{\fontsize{8}{9}\selectfont Mip-Splatting + SfSM}\par\vspace{2pt}\begin{minipage}[t]{0.242\linewidth}\centering\includegraphics[width=\linewidth,bb=0 0 1408 1392,viewport=1064 600 1304 920,clip]{figures/dycheck/apple/rgb.jpg}\par\vspace{1pt}{\fontsize{7}{8}\selectfont RGB}\end{minipage}\hfill\begin{minipage}[t]{0.242\linewidth}\centering\includegraphics[width=\linewidth,bb=0 0 1408 1392,viewport=1064 600 1304 920,clip]{figures/dycheck/apple/depth.png}\par\vspace{1pt}{\fontsize{7}{8}\selectfont Depth}\end{minipage}\hfill\begin{minipage}[t]{0.242\linewidth}\centering\includegraphics[width=\linewidth,bb=0 0 1024 672,viewport=520 8 760 328,clip]{figures/dycheck/apple/rgb_error.png}\par\vspace{1pt}{\fontsize{7}{8}\selectfont RGB error}\end{minipage}\hfill\begin{minipage}[t]{0.242\linewidth}\centering\includegraphics[width=\linewidth,bb=0 0 1024 672,viewport=520 8 760 328,clip]{figures/dycheck/apple/depth_error.png}\par\vspace{1pt}{\fontsize{7}{8}\selectfont Depth error}\end{minipage}\end{minipage}\hfill\begin{minipage}[t]{0.492\textwidth}\centering{\fontsize{8}{9}\selectfont NBGS-Base}\par\vspace{2pt}\begin{minipage}[t]{0.242\linewidth}\centering\includegraphics[width=\linewidth,bb=0 0 1408 1392,viewport=8 136 248 456,clip]{figures/dycheck/apple/rgb.jpg}\par\vspace{1pt}{\fontsize{7}{8}\selectfont RGB}\end{minipage}\hfill\begin{minipage}[t]{0.242\linewidth}\centering\includegraphics[width=\linewidth,bb=0 0 1408 1392,viewport=8 136 248 456,clip]{figures/dycheck/apple/depth.png}\par\vspace{1pt}{\fontsize{7}{8}\selectfont Depth}\end{minipage}\hfill\begin{minipage}[t]{0.242\linewidth}\centering\includegraphics[width=\linewidth,bb=0 0 1024 672,viewport=776 8 1016 328,clip]{figures/dycheck/apple/rgb_error.png}\par\vspace{1pt}{\fontsize{7}{8}\selectfont RGB error}\end{minipage}\hfill\begin{minipage}[t]{0.242\linewidth}\centering\includegraphics[width=\linewidth,bb=0 0 1024 672,viewport=776 8 1016 328,clip]{figures/dycheck/apple/depth_error.png}\par\vspace{1pt}{\fontsize{7}{8}\selectfont Depth error}\end{minipage}\end{minipage}\par\vspace{16pt}
\clearpage
\pdfbookmark[3]{backpack (iPhone)}{supp.phone.backpack}
\phantomsection\label{fig:supp_phone_backpack}{\large\bfseries backpack (iPhone)}\hfill{\fontsize{8}{9}\selectfont DyCheck dataset}\par\vspace{4pt}
\noindent \begin{minipage}[t]{0.18\textwidth}\centering\fontsize{7}{8}\selectfont RGB reference\par\vspace{2pt}\includegraphics[width=\linewidth,bb=0 0 1408 1392,viewport=8 936 344 1384,clip]{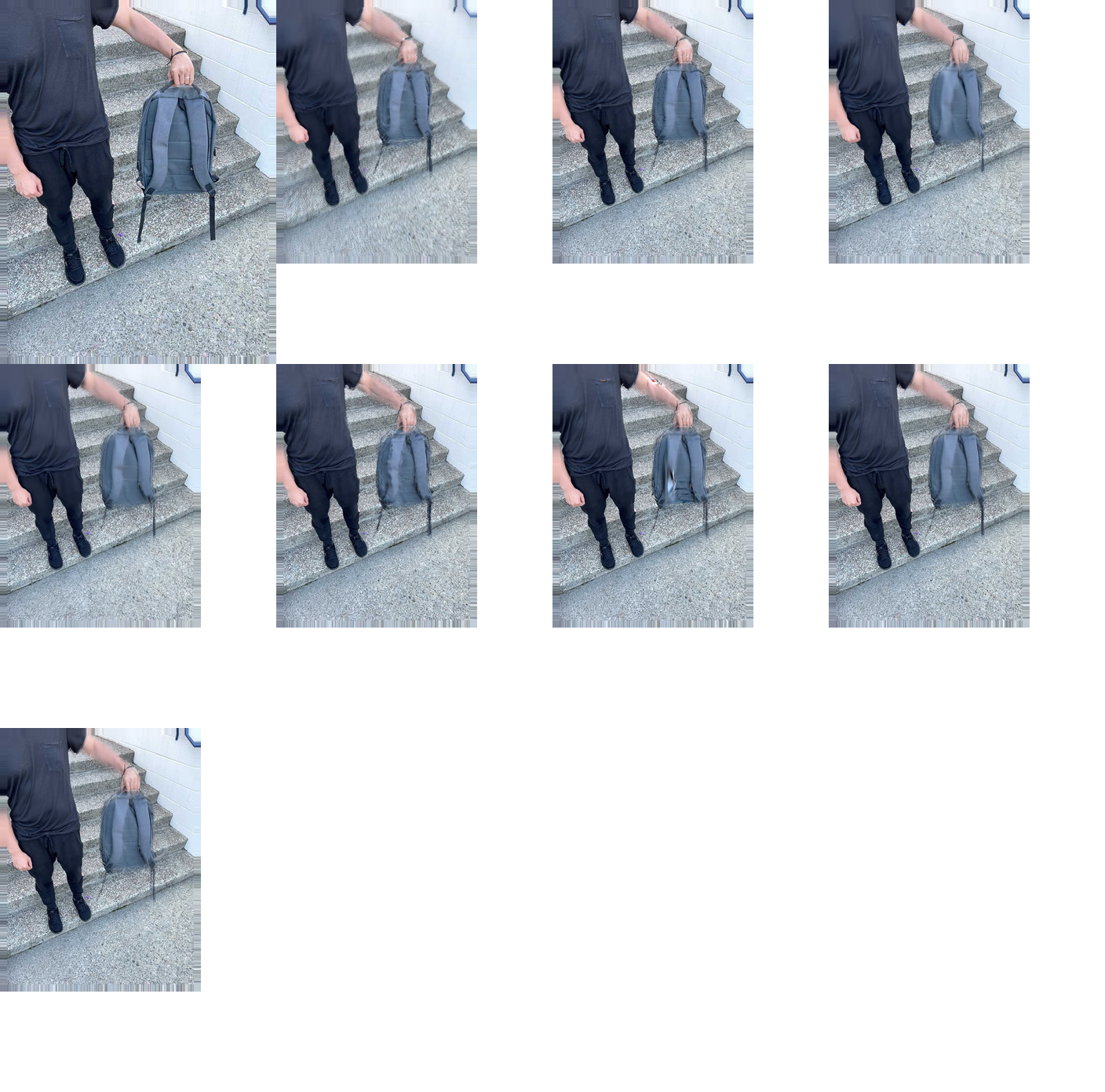}\end{minipage}\hfill\begin{minipage}[t]{0.18\textwidth}\centering\fontsize{7}{8}\selectfont LiDAR reference\par\vspace{2pt}\includegraphics[width=\linewidth,bb=0 0 1408 1392,viewport=8 936 344 1384,clip]{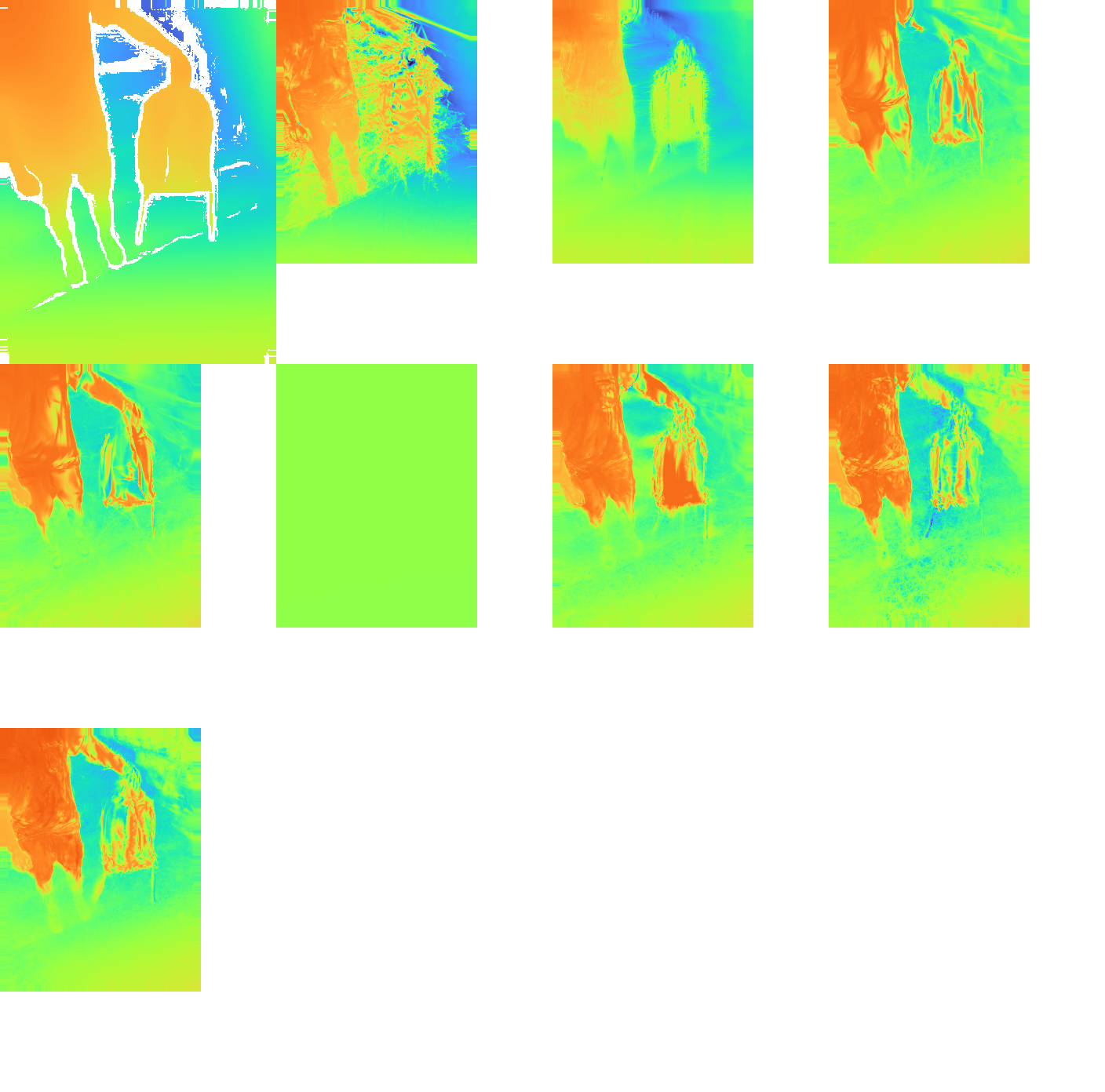}\end{minipage}\hfill\begin{minipage}[t]{0.625\textwidth}\fontsize{8}{9}\selectfont First held-out view, seed 0: \texttt{frame\_009}. Compare each render with the RGB reference, then compare its predicted depth with LiDAR. All depths use the same meter scale. White marks invalid depth, and values above $3$\,m are saturated.\par\vspace{4pt}Depth (m):\par\includegraphics[width=\linewidth,height=5pt]{figures/legends/dycheck_depth_scale.png}\par $0$\hfill $3$\par\vspace{3pt}RGB error: $\sum_c|I_c-\widehat I_c|$\par\includegraphics[width=\linewidth,height=5pt]{figures/legends/dycheck_error_scale.png}\par $0$\hfill $3$\par\vspace{3pt}Depth error: $|s\widehat Z-Z_{\mathrm{LiDAR}}|$ (m)\par\includegraphics[width=\linewidth,height=5pt]{figures/legends/dycheck_error_scale.png}\par $0$\hfill $\geq1.5$\par\vspace{3pt}{\fontsize{7}{8}\selectfont Depth errors are measured on valid static pixels shared by all methods. White marks excluded pixels. Both displayed phone variants, NBGS-Base and Vanilla (reset off), disable opacity reset and its associated size pruning.}\end{minipage}\par\vspace{5pt}
\noindent \begin{minipage}[t]{0.492\textwidth}\centering{\fontsize{8}{9}\selectfont Small GS}\par\vspace{2pt}\begin{minipage}[t]{0.242\linewidth}\centering\includegraphics[width=\linewidth,bb=0 0 1408 1392,viewport=360 1064 600 1384,clip]{figures/dycheck/backpack/rgb.jpg}\par\vspace{1pt}{\fontsize{7}{8}\selectfont RGB}\end{minipage}\hfill\begin{minipage}[t]{0.242\linewidth}\centering\includegraphics[width=\linewidth,bb=0 0 1408 1392,viewport=360 1064 600 1384,clip]{figures/dycheck/backpack/depth.png}\par\vspace{1pt}{\fontsize{7}{8}\selectfont Depth}\end{minipage}\hfill\begin{minipage}[t]{0.242\linewidth}\centering\includegraphics[width=\linewidth,bb=0 0 1024 672,viewport=8 344 248 664,clip]{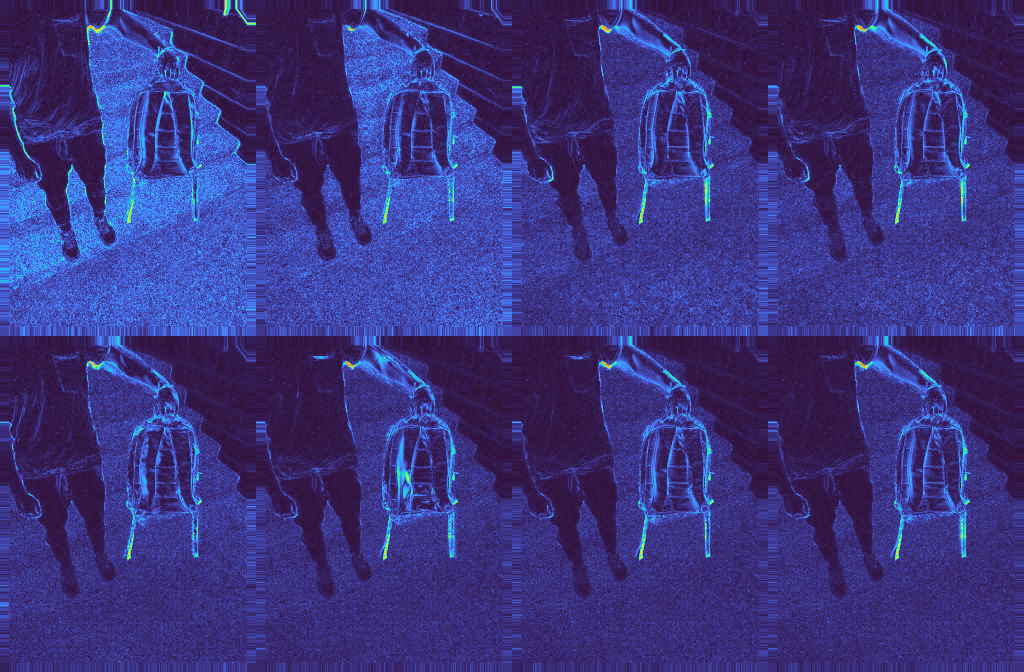}\par\vspace{1pt}{\fontsize{7}{8}\selectfont RGB error}\end{minipage}\hfill\begin{minipage}[t]{0.242\linewidth}\centering\includegraphics[width=\linewidth,bb=0 0 1024 672,viewport=8 344 248 664,clip]{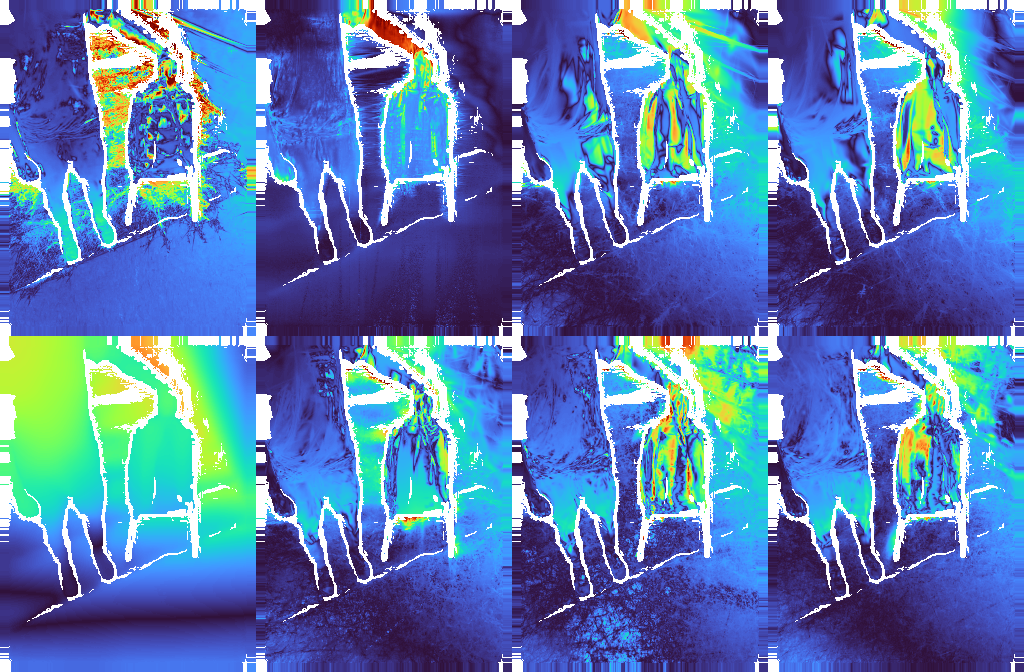}\par\vspace{1pt}{\fontsize{7}{8}\selectfont Depth error}\end{minipage}\end{minipage}\hfill\begin{minipage}[t]{0.492\textwidth}\centering{\fontsize{8}{9}\selectfont InstantSplat}\par\vspace{2pt}\begin{minipage}[t]{0.242\linewidth}\centering\includegraphics[width=\linewidth,bb=0 0 1408 1392,viewport=712 1064 952 1384,clip]{figures/dycheck/backpack/rgb.jpg}\par\vspace{1pt}{\fontsize{7}{8}\selectfont RGB}\end{minipage}\hfill\begin{minipage}[t]{0.242\linewidth}\centering\includegraphics[width=\linewidth,bb=0 0 1408 1392,viewport=712 1064 952 1384,clip]{figures/dycheck/backpack/depth.png}\par\vspace{1pt}{\fontsize{7}{8}\selectfont Depth}\end{minipage}\hfill\begin{minipage}[t]{0.242\linewidth}\centering\includegraphics[width=\linewidth,bb=0 0 1024 672,viewport=264 344 504 664,clip]{figures/dycheck/backpack/rgb_error.png}\par\vspace{1pt}{\fontsize{7}{8}\selectfont RGB error}\end{minipage}\hfill\begin{minipage}[t]{0.242\linewidth}\centering\includegraphics[width=\linewidth,bb=0 0 1024 672,viewport=264 344 504 664,clip]{figures/dycheck/backpack/depth_error.png}\par\vspace{1pt}{\fontsize{7}{8}\selectfont Depth error}\end{minipage}\end{minipage}\par\vspace{16pt}
\noindent \begin{minipage}[t]{0.492\textwidth}\centering{\fontsize{8}{9}\selectfont DDDrop + SfSM}\par\vspace{2pt}\begin{minipage}[t]{0.242\linewidth}\centering\includegraphics[width=\linewidth,bb=0 0 1408 1392,viewport=1064 1064 1304 1384,clip]{figures/dycheck/backpack/rgb.jpg}\par\vspace{1pt}{\fontsize{7}{8}\selectfont RGB}\end{minipage}\hfill\begin{minipage}[t]{0.242\linewidth}\centering\includegraphics[width=\linewidth,bb=0 0 1408 1392,viewport=1064 1064 1304 1384,clip]{figures/dycheck/backpack/depth.png}\par\vspace{1pt}{\fontsize{7}{8}\selectfont Depth}\end{minipage}\hfill\begin{minipage}[t]{0.242\linewidth}\centering\includegraphics[width=\linewidth,bb=0 0 1024 672,viewport=520 344 760 664,clip]{figures/dycheck/backpack/rgb_error.png}\par\vspace{1pt}{\fontsize{7}{8}\selectfont RGB error}\end{minipage}\hfill\begin{minipage}[t]{0.242\linewidth}\centering\includegraphics[width=\linewidth,bb=0 0 1024 672,viewport=520 344 760 664,clip]{figures/dycheck/backpack/depth_error.png}\par\vspace{1pt}{\fontsize{7}{8}\selectfont Depth error}\end{minipage}\end{minipage}\hfill\begin{minipage}[t]{0.492\textwidth}\centering{\fontsize{8}{9}\selectfont DDGS + SfSM}\par\vspace{2pt}\begin{minipage}[t]{0.242\linewidth}\centering\includegraphics[width=\linewidth,bb=0 0 1408 1392,viewport=8 600 248 920,clip]{figures/dycheck/backpack/rgb.jpg}\par\vspace{1pt}{\fontsize{7}{8}\selectfont RGB}\end{minipage}\hfill\begin{minipage}[t]{0.242\linewidth}\centering\includegraphics[width=\linewidth,bb=0 0 1408 1392,viewport=8 600 248 920,clip]{figures/dycheck/backpack/depth.png}\par\vspace{1pt}{\fontsize{7}{8}\selectfont Depth}\end{minipage}\hfill\begin{minipage}[t]{0.242\linewidth}\centering\includegraphics[width=\linewidth,bb=0 0 1024 672,viewport=776 344 1016 664,clip]{figures/dycheck/backpack/rgb_error.png}\par\vspace{1pt}{\fontsize{7}{8}\selectfont RGB error}\end{minipage}\hfill\begin{minipage}[t]{0.242\linewidth}\centering\includegraphics[width=\linewidth,bb=0 0 1024 672,viewport=776 344 1016 664,clip]{figures/dycheck/backpack/depth_error.png}\par\vspace{1pt}{\fontsize{7}{8}\selectfont Depth error}\end{minipage}\end{minipage}\par\vspace{16pt}
\noindent \begin{minipage}[t]{0.492\textwidth}\centering{\fontsize{8}{9}\selectfont G4Splat}\par\vspace{2pt}\begin{minipage}[t]{0.242\linewidth}\centering\includegraphics[width=\linewidth,bb=0 0 1408 1392,viewport=360 600 600 920,clip]{figures/dycheck/backpack/rgb.jpg}\par\vspace{1pt}{\fontsize{7}{8}\selectfont RGB}\end{minipage}\hfill\begin{minipage}[t]{0.242\linewidth}\centering\includegraphics[width=\linewidth,bb=0 0 1408 1392,viewport=360 600 600 920,clip]{figures/dycheck/backpack/depth.png}\par\vspace{1pt}{\fontsize{7}{8}\selectfont Depth}\end{minipage}\hfill\begin{minipage}[t]{0.242\linewidth}\centering\includegraphics[width=\linewidth,bb=0 0 1024 672,viewport=8 8 248 328,clip]{figures/dycheck/backpack/rgb_error.png}\par\vspace{1pt}{\fontsize{7}{8}\selectfont RGB error}\end{minipage}\hfill\begin{minipage}[t]{0.242\linewidth}\centering\includegraphics[width=\linewidth,bb=0 0 1024 672,viewport=8 8 248 328,clip]{figures/dycheck/backpack/depth_error.png}\par\vspace{1pt}{\fontsize{7}{8}\selectfont Depth error}\end{minipage}\end{minipage}\hfill\begin{minipage}[t]{0.492\textwidth}\centering{\fontsize{8}{9}\selectfont Vanilla (reset off)}\par\vspace{2pt}\begin{minipage}[t]{0.242\linewidth}\centering\includegraphics[width=\linewidth,bb=0 0 1408 1392,viewport=712 600 952 920,clip]{figures/dycheck/backpack/rgb.jpg}\par\vspace{1pt}{\fontsize{7}{8}\selectfont RGB}\end{minipage}\hfill\begin{minipage}[t]{0.242\linewidth}\centering\includegraphics[width=\linewidth,bb=0 0 1408 1392,viewport=712 600 952 920,clip]{figures/dycheck/backpack/depth.png}\par\vspace{1pt}{\fontsize{7}{8}\selectfont Depth}\end{minipage}\hfill\begin{minipage}[t]{0.242\linewidth}\centering\includegraphics[width=\linewidth,bb=0 0 1024 672,viewport=264 8 504 328,clip]{figures/dycheck/backpack/rgb_error.png}\par\vspace{1pt}{\fontsize{7}{8}\selectfont RGB error}\end{minipage}\hfill\begin{minipage}[t]{0.242\linewidth}\centering\includegraphics[width=\linewidth,bb=0 0 1024 672,viewport=264 8 504 328,clip]{figures/dycheck/backpack/depth_error.png}\par\vspace{1pt}{\fontsize{7}{8}\selectfont Depth error}\end{minipage}\end{minipage}\par\vspace{16pt}
\noindent \begin{minipage}[t]{0.492\textwidth}\centering{\fontsize{8}{9}\selectfont Mip-Splatting + SfSM}\par\vspace{2pt}\begin{minipage}[t]{0.242\linewidth}\centering\includegraphics[width=\linewidth,bb=0 0 1408 1392,viewport=1064 600 1304 920,clip]{figures/dycheck/backpack/rgb.jpg}\par\vspace{1pt}{\fontsize{7}{8}\selectfont RGB}\end{minipage}\hfill\begin{minipage}[t]{0.242\linewidth}\centering\includegraphics[width=\linewidth,bb=0 0 1408 1392,viewport=1064 600 1304 920,clip]{figures/dycheck/backpack/depth.png}\par\vspace{1pt}{\fontsize{7}{8}\selectfont Depth}\end{minipage}\hfill\begin{minipage}[t]{0.242\linewidth}\centering\includegraphics[width=\linewidth,bb=0 0 1024 672,viewport=520 8 760 328,clip]{figures/dycheck/backpack/rgb_error.png}\par\vspace{1pt}{\fontsize{7}{8}\selectfont RGB error}\end{minipage}\hfill\begin{minipage}[t]{0.242\linewidth}\centering\includegraphics[width=\linewidth,bb=0 0 1024 672,viewport=520 8 760 328,clip]{figures/dycheck/backpack/depth_error.png}\par\vspace{1pt}{\fontsize{7}{8}\selectfont Depth error}\end{minipage}\end{minipage}\hfill\begin{minipage}[t]{0.492\textwidth}\centering{\fontsize{8}{9}\selectfont NBGS-Base}\par\vspace{2pt}\begin{minipage}[t]{0.242\linewidth}\centering\includegraphics[width=\linewidth,bb=0 0 1408 1392,viewport=8 136 248 456,clip]{figures/dycheck/backpack/rgb.jpg}\par\vspace{1pt}{\fontsize{7}{8}\selectfont RGB}\end{minipage}\hfill\begin{minipage}[t]{0.242\linewidth}\centering\includegraphics[width=\linewidth,bb=0 0 1408 1392,viewport=8 136 248 456,clip]{figures/dycheck/backpack/depth.png}\par\vspace{1pt}{\fontsize{7}{8}\selectfont Depth}\end{minipage}\hfill\begin{minipage}[t]{0.242\linewidth}\centering\includegraphics[width=\linewidth,bb=0 0 1024 672,viewport=776 8 1016 328,clip]{figures/dycheck/backpack/rgb_error.png}\par\vspace{1pt}{\fontsize{7}{8}\selectfont RGB error}\end{minipage}\hfill\begin{minipage}[t]{0.242\linewidth}\centering\includegraphics[width=\linewidth,bb=0 0 1024 672,viewport=776 8 1016 328,clip]{figures/dycheck/backpack/depth_error.png}\par\vspace{1pt}{\fontsize{7}{8}\selectfont Depth error}\end{minipage}\end{minipage}\par\vspace{16pt}
\clearpage
\pdfbookmark[3]{block (iPhone)}{supp.phone.block}
\phantomsection\label{fig:supp_phone_block}{\large\bfseries block (iPhone)}\hfill{\fontsize{8}{9}\selectfont DyCheck dataset}\par\vspace{4pt}
\noindent \begin{minipage}[t]{0.18\textwidth}\centering\fontsize{7}{8}\selectfont RGB reference\par\vspace{2pt}\includegraphics[width=\linewidth,bb=0 0 1408 1392,viewport=8 936 344 1384,clip]{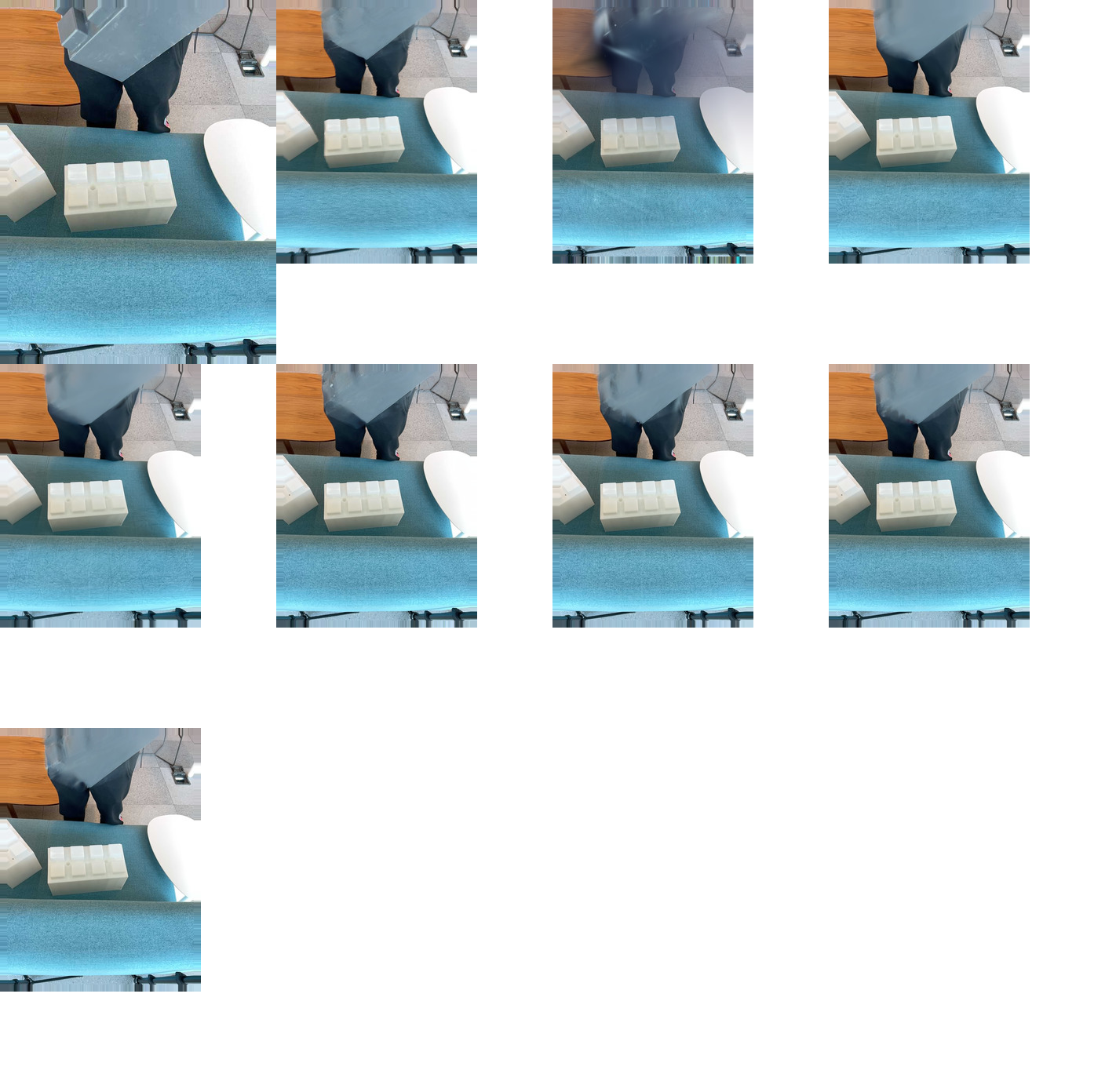}\end{minipage}\hfill\begin{minipage}[t]{0.18\textwidth}\centering\fontsize{7}{8}\selectfont LiDAR reference\par\vspace{2pt}\includegraphics[width=\linewidth,bb=0 0 1408 1392,viewport=8 936 344 1384,clip]{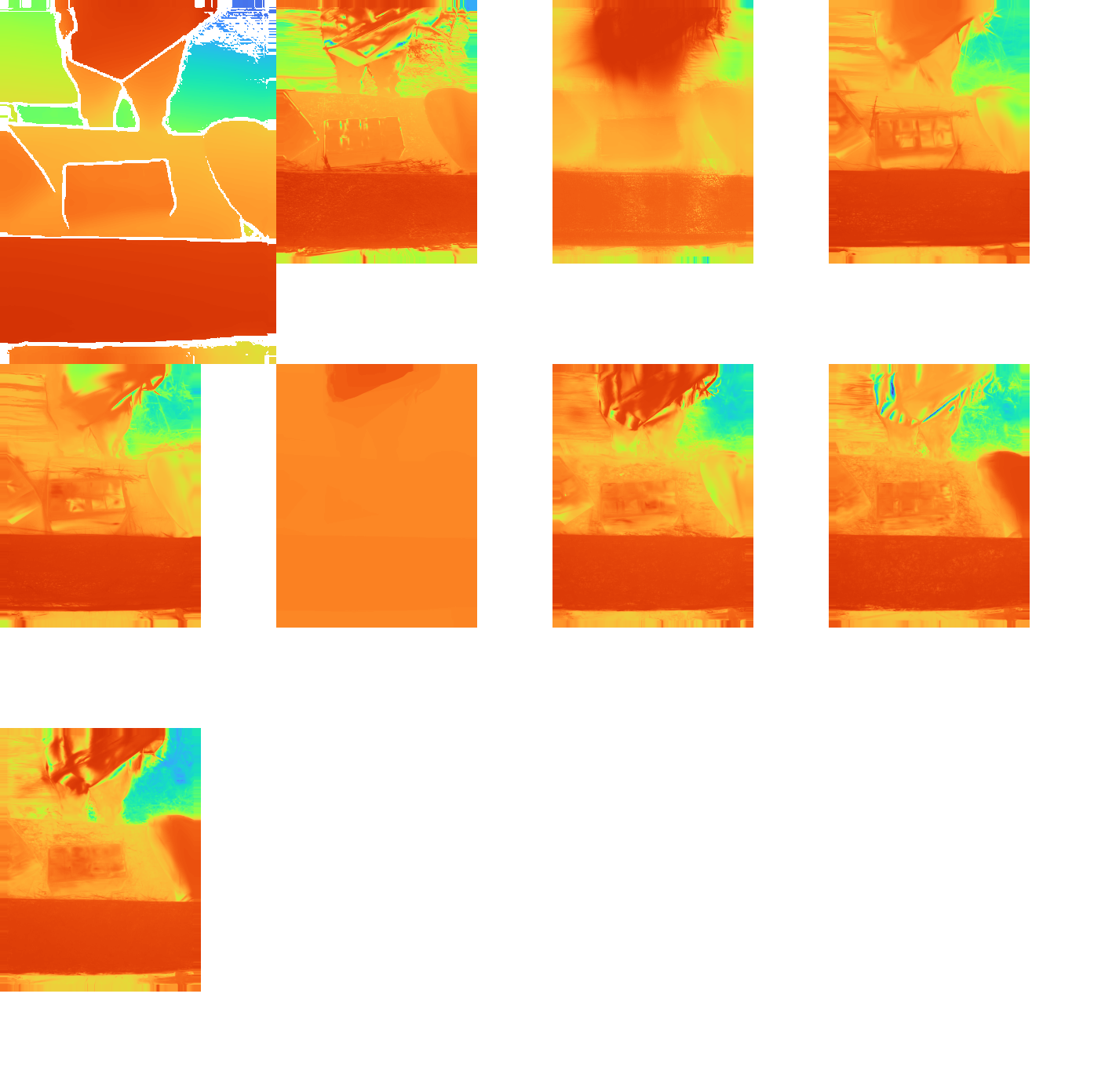}\end{minipage}\hfill\begin{minipage}[t]{0.625\textwidth}\fontsize{8}{9}\selectfont First held-out view, seed 0: \texttt{frame\_009}. Compare each render with the RGB reference, then compare its predicted depth with LiDAR. All depths use the same meter scale. White marks invalid depth, and values above $3$\,m are saturated.\par\vspace{4pt}Depth (m):\par\includegraphics[width=\linewidth,height=5pt]{figures/legends/dycheck_depth_scale.png}\par $0$\hfill $3$\par\vspace{3pt}RGB error: $\sum_c|I_c-\widehat I_c|$\par\includegraphics[width=\linewidth,height=5pt]{figures/legends/dycheck_error_scale.png}\par $0$\hfill $3$\par\vspace{3pt}Depth error: $|s\widehat Z-Z_{\mathrm{LiDAR}}|$ (m)\par\includegraphics[width=\linewidth,height=5pt]{figures/legends/dycheck_error_scale.png}\par $0$\hfill $\geq1.5$\par\vspace{3pt}{\fontsize{7}{8}\selectfont Depth errors are measured on valid static pixels shared by all methods. White marks excluded pixels. Both displayed phone variants, NBGS-Base and Vanilla (reset off), disable opacity reset and its associated size pruning.}\end{minipage}\par\vspace{5pt}
\noindent \begin{minipage}[t]{0.492\textwidth}\centering{\fontsize{8}{9}\selectfont Small GS}\par\vspace{2pt}\begin{minipage}[t]{0.242\linewidth}\centering\includegraphics[width=\linewidth,bb=0 0 1408 1392,viewport=360 1064 600 1384,clip]{figures/dycheck/block/rgb.jpg}\par\vspace{1pt}{\fontsize{7}{8}\selectfont RGB}\end{minipage}\hfill\begin{minipage}[t]{0.242\linewidth}\centering\includegraphics[width=\linewidth,bb=0 0 1408 1392,viewport=360 1064 600 1384,clip]{figures/dycheck/block/depth.png}\par\vspace{1pt}{\fontsize{7}{8}\selectfont Depth}\end{minipage}\hfill\begin{minipage}[t]{0.242\linewidth}\centering\includegraphics[width=\linewidth,bb=0 0 1024 672,viewport=8 344 248 664,clip]{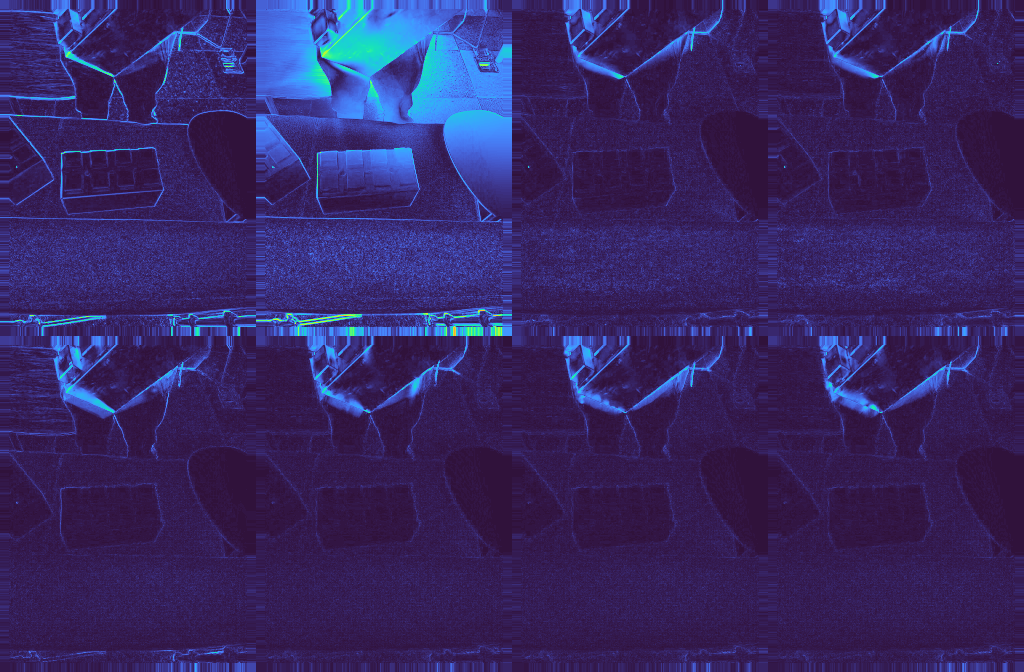}\par\vspace{1pt}{\fontsize{7}{8}\selectfont RGB error}\end{minipage}\hfill\begin{minipage}[t]{0.242\linewidth}\centering\includegraphics[width=\linewidth,bb=0 0 1024 672,viewport=8 344 248 664,clip]{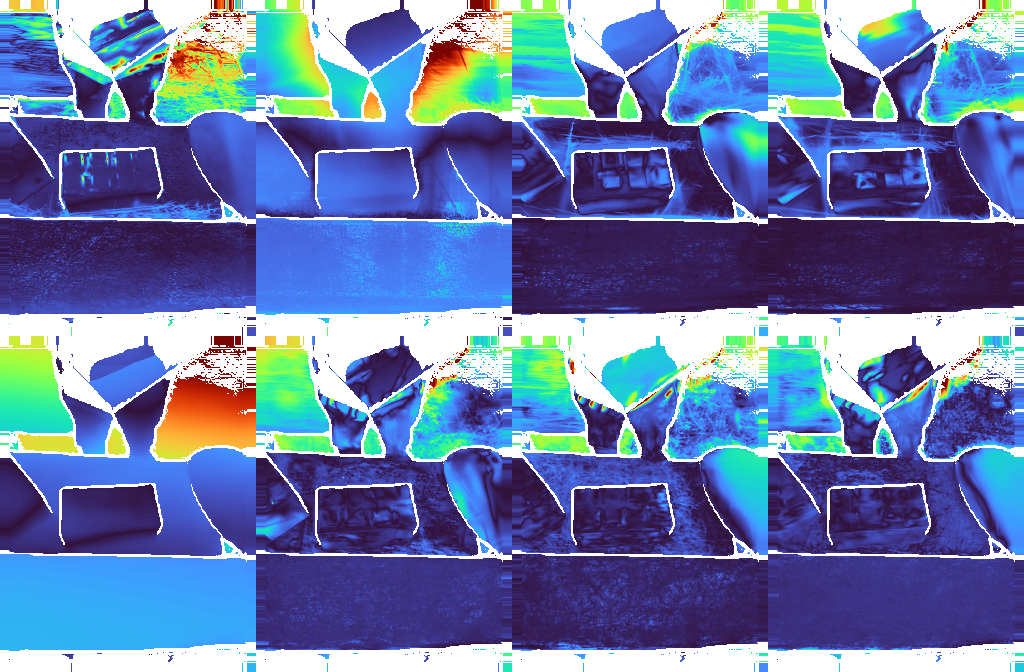}\par\vspace{1pt}{\fontsize{7}{8}\selectfont Depth error}\end{minipage}\end{minipage}\hfill\begin{minipage}[t]{0.492\textwidth}\centering{\fontsize{8}{9}\selectfont InstantSplat}\par\vspace{2pt}\begin{minipage}[t]{0.242\linewidth}\centering\includegraphics[width=\linewidth,bb=0 0 1408 1392,viewport=712 1064 952 1384,clip]{figures/dycheck/block/rgb.jpg}\par\vspace{1pt}{\fontsize{7}{8}\selectfont RGB}\end{minipage}\hfill\begin{minipage}[t]{0.242\linewidth}\centering\includegraphics[width=\linewidth,bb=0 0 1408 1392,viewport=712 1064 952 1384,clip]{figures/dycheck/block/depth.png}\par\vspace{1pt}{\fontsize{7}{8}\selectfont Depth}\end{minipage}\hfill\begin{minipage}[t]{0.242\linewidth}\centering\includegraphics[width=\linewidth,bb=0 0 1024 672,viewport=264 344 504 664,clip]{figures/dycheck/block/rgb_error.png}\par\vspace{1pt}{\fontsize{7}{8}\selectfont RGB error}\end{minipage}\hfill\begin{minipage}[t]{0.242\linewidth}\centering\includegraphics[width=\linewidth,bb=0 0 1024 672,viewport=264 344 504 664,clip]{figures/dycheck/block/depth_error.png}\par\vspace{1pt}{\fontsize{7}{8}\selectfont Depth error}\end{minipage}\end{minipage}\par\vspace{16pt}
\noindent \begin{minipage}[t]{0.492\textwidth}\centering{\fontsize{8}{9}\selectfont DDDrop + SfSM}\par\vspace{2pt}\begin{minipage}[t]{0.242\linewidth}\centering\includegraphics[width=\linewidth,bb=0 0 1408 1392,viewport=1064 1064 1304 1384,clip]{figures/dycheck/block/rgb.jpg}\par\vspace{1pt}{\fontsize{7}{8}\selectfont RGB}\end{minipage}\hfill\begin{minipage}[t]{0.242\linewidth}\centering\includegraphics[width=\linewidth,bb=0 0 1408 1392,viewport=1064 1064 1304 1384,clip]{figures/dycheck/block/depth.png}\par\vspace{1pt}{\fontsize{7}{8}\selectfont Depth}\end{minipage}\hfill\begin{minipage}[t]{0.242\linewidth}\centering\includegraphics[width=\linewidth,bb=0 0 1024 672,viewport=520 344 760 664,clip]{figures/dycheck/block/rgb_error.png}\par\vspace{1pt}{\fontsize{7}{8}\selectfont RGB error}\end{minipage}\hfill\begin{minipage}[t]{0.242\linewidth}\centering\includegraphics[width=\linewidth,bb=0 0 1024 672,viewport=520 344 760 664,clip]{figures/dycheck/block/depth_error.png}\par\vspace{1pt}{\fontsize{7}{8}\selectfont Depth error}\end{minipage}\end{minipage}\hfill\begin{minipage}[t]{0.492\textwidth}\centering{\fontsize{8}{9}\selectfont DDGS + SfSM}\par\vspace{2pt}\begin{minipage}[t]{0.242\linewidth}\centering\includegraphics[width=\linewidth,bb=0 0 1408 1392,viewport=8 600 248 920,clip]{figures/dycheck/block/rgb.jpg}\par\vspace{1pt}{\fontsize{7}{8}\selectfont RGB}\end{minipage}\hfill\begin{minipage}[t]{0.242\linewidth}\centering\includegraphics[width=\linewidth,bb=0 0 1408 1392,viewport=8 600 248 920,clip]{figures/dycheck/block/depth.png}\par\vspace{1pt}{\fontsize{7}{8}\selectfont Depth}\end{minipage}\hfill\begin{minipage}[t]{0.242\linewidth}\centering\includegraphics[width=\linewidth,bb=0 0 1024 672,viewport=776 344 1016 664,clip]{figures/dycheck/block/rgb_error.png}\par\vspace{1pt}{\fontsize{7}{8}\selectfont RGB error}\end{minipage}\hfill\begin{minipage}[t]{0.242\linewidth}\centering\includegraphics[width=\linewidth,bb=0 0 1024 672,viewport=776 344 1016 664,clip]{figures/dycheck/block/depth_error.png}\par\vspace{1pt}{\fontsize{7}{8}\selectfont Depth error}\end{minipage}\end{minipage}\par\vspace{16pt}
\noindent \begin{minipage}[t]{0.492\textwidth}\centering{\fontsize{8}{9}\selectfont G4Splat}\par\vspace{2pt}\begin{minipage}[t]{0.242\linewidth}\centering\includegraphics[width=\linewidth,bb=0 0 1408 1392,viewport=360 600 600 920,clip]{figures/dycheck/block/rgb.jpg}\par\vspace{1pt}{\fontsize{7}{8}\selectfont RGB}\end{minipage}\hfill\begin{minipage}[t]{0.242\linewidth}\centering\includegraphics[width=\linewidth,bb=0 0 1408 1392,viewport=360 600 600 920,clip]{figures/dycheck/block/depth.png}\par\vspace{1pt}{\fontsize{7}{8}\selectfont Depth}\end{minipage}\hfill\begin{minipage}[t]{0.242\linewidth}\centering\includegraphics[width=\linewidth,bb=0 0 1024 672,viewport=8 8 248 328,clip]{figures/dycheck/block/rgb_error.png}\par\vspace{1pt}{\fontsize{7}{8}\selectfont RGB error}\end{minipage}\hfill\begin{minipage}[t]{0.242\linewidth}\centering\includegraphics[width=\linewidth,bb=0 0 1024 672,viewport=8 8 248 328,clip]{figures/dycheck/block/depth_error.png}\par\vspace{1pt}{\fontsize{7}{8}\selectfont Depth error}\end{minipage}\end{minipage}\hfill\begin{minipage}[t]{0.492\textwidth}\centering{\fontsize{8}{9}\selectfont Vanilla (reset off)}\par\vspace{2pt}\begin{minipage}[t]{0.242\linewidth}\centering\includegraphics[width=\linewidth,bb=0 0 1408 1392,viewport=712 600 952 920,clip]{figures/dycheck/block/rgb.jpg}\par\vspace{1pt}{\fontsize{7}{8}\selectfont RGB}\end{minipage}\hfill\begin{minipage}[t]{0.242\linewidth}\centering\includegraphics[width=\linewidth,bb=0 0 1408 1392,viewport=712 600 952 920,clip]{figures/dycheck/block/depth.png}\par\vspace{1pt}{\fontsize{7}{8}\selectfont Depth}\end{minipage}\hfill\begin{minipage}[t]{0.242\linewidth}\centering\includegraphics[width=\linewidth,bb=0 0 1024 672,viewport=264 8 504 328,clip]{figures/dycheck/block/rgb_error.png}\par\vspace{1pt}{\fontsize{7}{8}\selectfont RGB error}\end{minipage}\hfill\begin{minipage}[t]{0.242\linewidth}\centering\includegraphics[width=\linewidth,bb=0 0 1024 672,viewport=264 8 504 328,clip]{figures/dycheck/block/depth_error.png}\par\vspace{1pt}{\fontsize{7}{8}\selectfont Depth error}\end{minipage}\end{minipage}\par\vspace{16pt}
\noindent \begin{minipage}[t]{0.492\textwidth}\centering{\fontsize{8}{9}\selectfont Mip-Splatting + SfSM}\par\vspace{2pt}\begin{minipage}[t]{0.242\linewidth}\centering\includegraphics[width=\linewidth,bb=0 0 1408 1392,viewport=1064 600 1304 920,clip]{figures/dycheck/block/rgb.jpg}\par\vspace{1pt}{\fontsize{7}{8}\selectfont RGB}\end{minipage}\hfill\begin{minipage}[t]{0.242\linewidth}\centering\includegraphics[width=\linewidth,bb=0 0 1408 1392,viewport=1064 600 1304 920,clip]{figures/dycheck/block/depth.png}\par\vspace{1pt}{\fontsize{7}{8}\selectfont Depth}\end{minipage}\hfill\begin{minipage}[t]{0.242\linewidth}\centering\includegraphics[width=\linewidth,bb=0 0 1024 672,viewport=520 8 760 328,clip]{figures/dycheck/block/rgb_error.png}\par\vspace{1pt}{\fontsize{7}{8}\selectfont RGB error}\end{minipage}\hfill\begin{minipage}[t]{0.242\linewidth}\centering\includegraphics[width=\linewidth,bb=0 0 1024 672,viewport=520 8 760 328,clip]{figures/dycheck/block/depth_error.png}\par\vspace{1pt}{\fontsize{7}{8}\selectfont Depth error}\end{minipage}\end{minipage}\hfill\begin{minipage}[t]{0.492\textwidth}\centering{\fontsize{8}{9}\selectfont NBGS-Base}\par\vspace{2pt}\begin{minipage}[t]{0.242\linewidth}\centering\includegraphics[width=\linewidth,bb=0 0 1408 1392,viewport=8 136 248 456,clip]{figures/dycheck/block/rgb.jpg}\par\vspace{1pt}{\fontsize{7}{8}\selectfont RGB}\end{minipage}\hfill\begin{minipage}[t]{0.242\linewidth}\centering\includegraphics[width=\linewidth,bb=0 0 1408 1392,viewport=8 136 248 456,clip]{figures/dycheck/block/depth.png}\par\vspace{1pt}{\fontsize{7}{8}\selectfont Depth}\end{minipage}\hfill\begin{minipage}[t]{0.242\linewidth}\centering\includegraphics[width=\linewidth,bb=0 0 1024 672,viewport=776 8 1016 328,clip]{figures/dycheck/block/rgb_error.png}\par\vspace{1pt}{\fontsize{7}{8}\selectfont RGB error}\end{minipage}\hfill\begin{minipage}[t]{0.242\linewidth}\centering\includegraphics[width=\linewidth,bb=0 0 1024 672,viewport=776 8 1016 328,clip]{figures/dycheck/block/depth_error.png}\par\vspace{1pt}{\fontsize{7}{8}\selectfont Depth error}\end{minipage}\end{minipage}\par\vspace{16pt}
\clearpage
\pdfbookmark[3]{mochi-high-five (iPhone)}{supp.phone.mochi-high-five}
\phantomsection\label{fig:supp_phone_mochi_high_five}{\large\bfseries mochi-high-five (iPhone)}\hfill{\fontsize{8}{9}\selectfont DyCheck dataset}\par\vspace{4pt}
\noindent \begin{minipage}[t]{0.18\textwidth}\centering\fontsize{7}{8}\selectfont RGB reference\par\vspace{2pt}\includegraphics[width=\linewidth,bb=0 0 1408 1392,viewport=8 936 344 1384,clip]{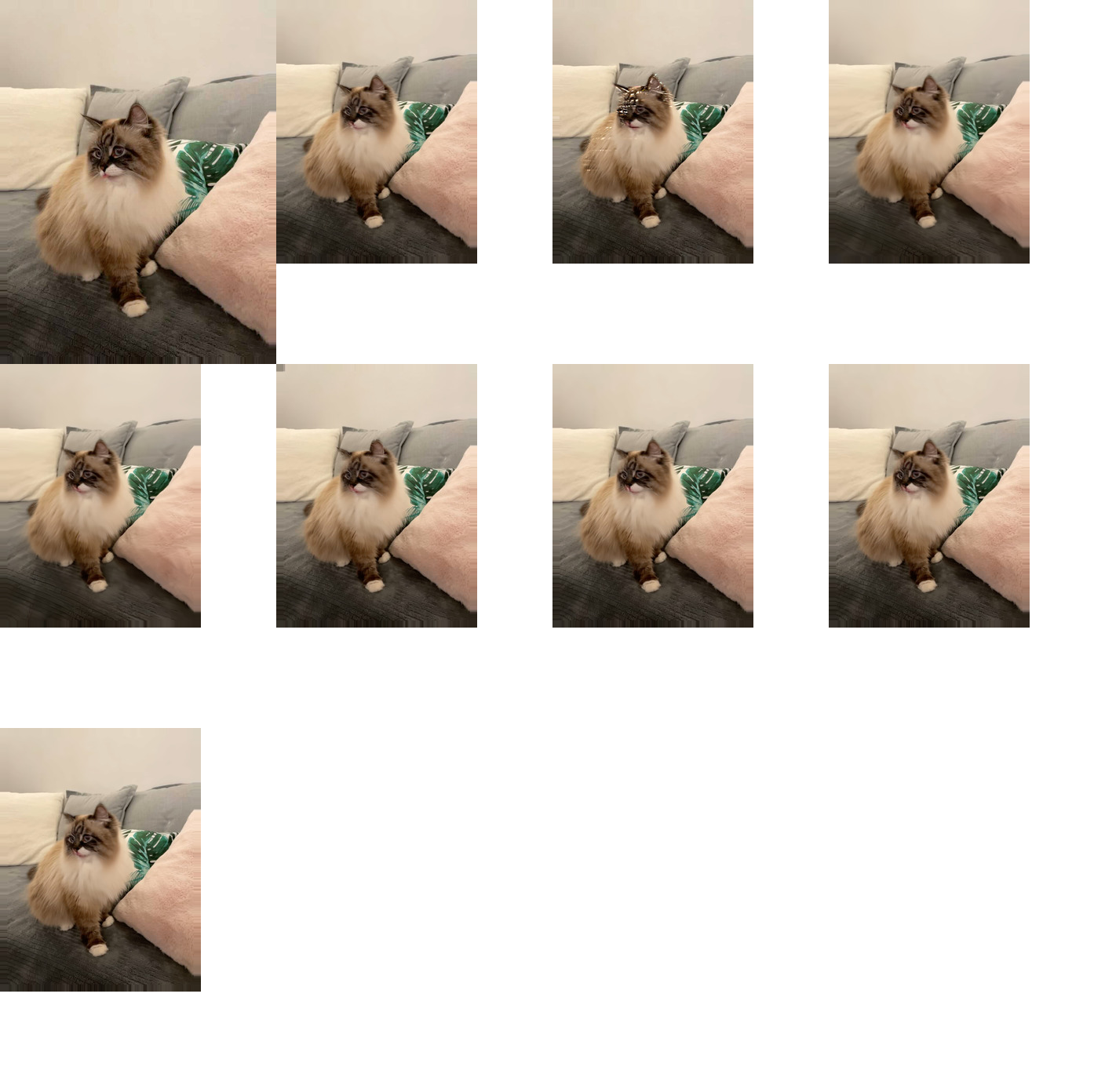}\end{minipage}\hfill\begin{minipage}[t]{0.18\textwidth}\centering\fontsize{7}{8}\selectfont LiDAR reference\par\vspace{2pt}\includegraphics[width=\linewidth,bb=0 0 1408 1392,viewport=8 936 344 1384,clip]{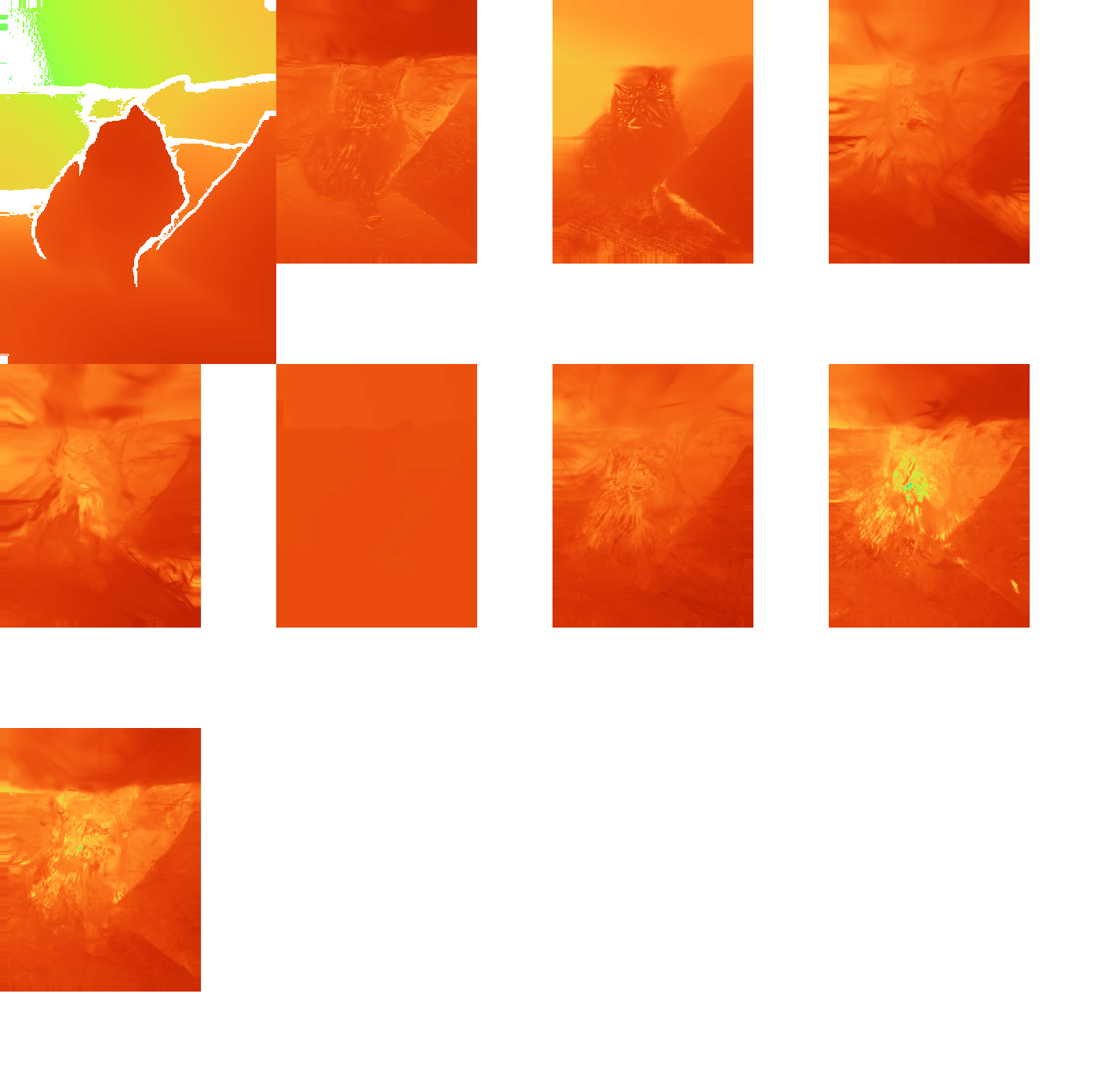}\end{minipage}\hfill\begin{minipage}[t]{0.625\textwidth}\fontsize{8}{9}\selectfont First held-out view, seed 0: \texttt{frame\_009}. Compare each render with the RGB reference, then compare its predicted depth with LiDAR. All depths use the same meter scale. White marks invalid depth, and values above $3$\,m are saturated.\par\vspace{4pt}Depth (m):\par\includegraphics[width=\linewidth,height=5pt]{figures/legends/dycheck_depth_scale.png}\par $0$\hfill $3$\par\vspace{3pt}RGB error: $\sum_c|I_c-\widehat I_c|$\par\includegraphics[width=\linewidth,height=5pt]{figures/legends/dycheck_error_scale.png}\par $0$\hfill $3$\par\vspace{3pt}Depth error: $|s\widehat Z-Z_{\mathrm{LiDAR}}|$ (m)\par\includegraphics[width=\linewidth,height=5pt]{figures/legends/dycheck_error_scale.png}\par $0$\hfill $\geq1.5$\par\vspace{3pt}{\fontsize{7}{8}\selectfont Depth errors are measured on valid static pixels shared by all methods. White marks excluded pixels. Both displayed phone variants, NBGS-Base and Vanilla (reset off), disable opacity reset and its associated size pruning.}\end{minipage}\par\vspace{5pt}
\noindent \begin{minipage}[t]{0.492\textwidth}\centering{\fontsize{8}{9}\selectfont Small GS}\par\vspace{2pt}\begin{minipage}[t]{0.242\linewidth}\centering\includegraphics[width=\linewidth,bb=0 0 1408 1392,viewport=360 1064 600 1384,clip]{figures/dycheck/mochi-high-five/rgb.jpg}\par\vspace{1pt}{\fontsize{7}{8}\selectfont RGB}\end{minipage}\hfill\begin{minipage}[t]{0.242\linewidth}\centering\includegraphics[width=\linewidth,bb=0 0 1408 1392,viewport=360 1064 600 1384,clip]{figures/dycheck/mochi-high-five/depth.png}\par\vspace{1pt}{\fontsize{7}{8}\selectfont Depth}\end{minipage}\hfill\begin{minipage}[t]{0.242\linewidth}\centering\includegraphics[width=\linewidth,bb=0 0 1024 672,viewport=8 344 248 664,clip]{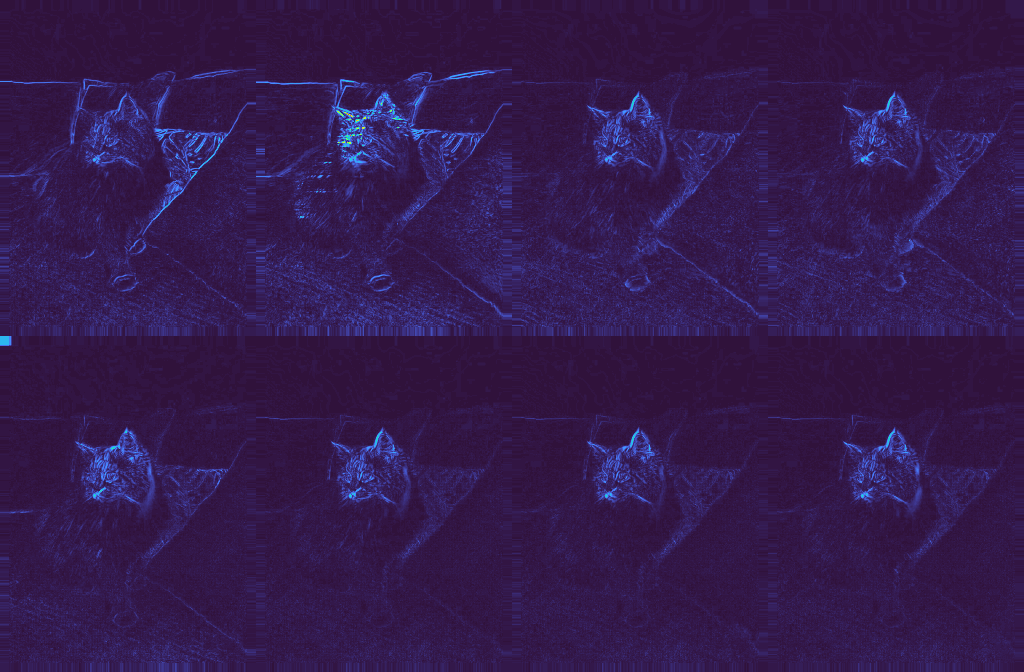}\par\vspace{1pt}{\fontsize{7}{8}\selectfont RGB error}\end{minipage}\hfill\begin{minipage}[t]{0.242\linewidth}\centering\includegraphics[width=\linewidth,bb=0 0 1024 672,viewport=8 344 248 664,clip]{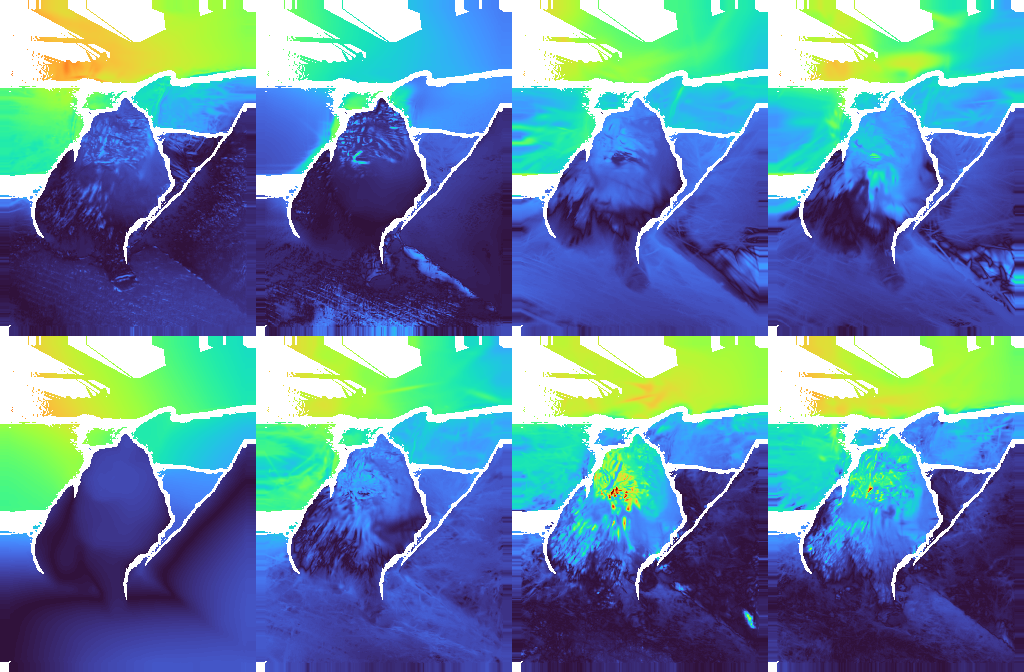}\par\vspace{1pt}{\fontsize{7}{8}\selectfont Depth error}\end{minipage}\end{minipage}\hfill\begin{minipage}[t]{0.492\textwidth}\centering{\fontsize{8}{9}\selectfont InstantSplat}\par\vspace{2pt}\begin{minipage}[t]{0.242\linewidth}\centering\includegraphics[width=\linewidth,bb=0 0 1408 1392,viewport=712 1064 952 1384,clip]{figures/dycheck/mochi-high-five/rgb.jpg}\par\vspace{1pt}{\fontsize{7}{8}\selectfont RGB}\end{minipage}\hfill\begin{minipage}[t]{0.242\linewidth}\centering\includegraphics[width=\linewidth,bb=0 0 1408 1392,viewport=712 1064 952 1384,clip]{figures/dycheck/mochi-high-five/depth.png}\par\vspace{1pt}{\fontsize{7}{8}\selectfont Depth}\end{minipage}\hfill\begin{minipage}[t]{0.242\linewidth}\centering\includegraphics[width=\linewidth,bb=0 0 1024 672,viewport=264 344 504 664,clip]{figures/dycheck/mochi-high-five/rgb_error.png}\par\vspace{1pt}{\fontsize{7}{8}\selectfont RGB error}\end{minipage}\hfill\begin{minipage}[t]{0.242\linewidth}\centering\includegraphics[width=\linewidth,bb=0 0 1024 672,viewport=264 344 504 664,clip]{figures/dycheck/mochi-high-five/depth_error.png}\par\vspace{1pt}{\fontsize{7}{8}\selectfont Depth error}\end{minipage}\end{minipage}\par\vspace{16pt}
\noindent \begin{minipage}[t]{0.492\textwidth}\centering{\fontsize{8}{9}\selectfont DDDrop + SfSM}\par\vspace{2pt}\begin{minipage}[t]{0.242\linewidth}\centering\includegraphics[width=\linewidth,bb=0 0 1408 1392,viewport=1064 1064 1304 1384,clip]{figures/dycheck/mochi-high-five/rgb.jpg}\par\vspace{1pt}{\fontsize{7}{8}\selectfont RGB}\end{minipage}\hfill\begin{minipage}[t]{0.242\linewidth}\centering\includegraphics[width=\linewidth,bb=0 0 1408 1392,viewport=1064 1064 1304 1384,clip]{figures/dycheck/mochi-high-five/depth.png}\par\vspace{1pt}{\fontsize{7}{8}\selectfont Depth}\end{minipage}\hfill\begin{minipage}[t]{0.242\linewidth}\centering\includegraphics[width=\linewidth,bb=0 0 1024 672,viewport=520 344 760 664,clip]{figures/dycheck/mochi-high-five/rgb_error.png}\par\vspace{1pt}{\fontsize{7}{8}\selectfont RGB error}\end{minipage}\hfill\begin{minipage}[t]{0.242\linewidth}\centering\includegraphics[width=\linewidth,bb=0 0 1024 672,viewport=520 344 760 664,clip]{figures/dycheck/mochi-high-five/depth_error.png}\par\vspace{1pt}{\fontsize{7}{8}\selectfont Depth error}\end{minipage}\end{minipage}\hfill\begin{minipage}[t]{0.492\textwidth}\centering{\fontsize{8}{9}\selectfont DDGS + SfSM}\par\vspace{2pt}\begin{minipage}[t]{0.242\linewidth}\centering\includegraphics[width=\linewidth,bb=0 0 1408 1392,viewport=8 600 248 920,clip]{figures/dycheck/mochi-high-five/rgb.jpg}\par\vspace{1pt}{\fontsize{7}{8}\selectfont RGB}\end{minipage}\hfill\begin{minipage}[t]{0.242\linewidth}\centering\includegraphics[width=\linewidth,bb=0 0 1408 1392,viewport=8 600 248 920,clip]{figures/dycheck/mochi-high-five/depth.png}\par\vspace{1pt}{\fontsize{7}{8}\selectfont Depth}\end{minipage}\hfill\begin{minipage}[t]{0.242\linewidth}\centering\includegraphics[width=\linewidth,bb=0 0 1024 672,viewport=776 344 1016 664,clip]{figures/dycheck/mochi-high-five/rgb_error.png}\par\vspace{1pt}{\fontsize{7}{8}\selectfont RGB error}\end{minipage}\hfill\begin{minipage}[t]{0.242\linewidth}\centering\includegraphics[width=\linewidth,bb=0 0 1024 672,viewport=776 344 1016 664,clip]{figures/dycheck/mochi-high-five/depth_error.png}\par\vspace{1pt}{\fontsize{7}{8}\selectfont Depth error}\end{minipage}\end{minipage}\par\vspace{16pt}
\noindent \begin{minipage}[t]{0.492\textwidth}\centering{\fontsize{8}{9}\selectfont G4Splat}\par\vspace{2pt}\begin{minipage}[t]{0.242\linewidth}\centering\includegraphics[width=\linewidth,bb=0 0 1408 1392,viewport=360 600 600 920,clip]{figures/dycheck/mochi-high-five/rgb.jpg}\par\vspace{1pt}{\fontsize{7}{8}\selectfont RGB}\end{minipage}\hfill\begin{minipage}[t]{0.242\linewidth}\centering\includegraphics[width=\linewidth,bb=0 0 1408 1392,viewport=360 600 600 920,clip]{figures/dycheck/mochi-high-five/depth.png}\par\vspace{1pt}{\fontsize{7}{8}\selectfont Depth}\end{minipage}\hfill\begin{minipage}[t]{0.242\linewidth}\centering\includegraphics[width=\linewidth,bb=0 0 1024 672,viewport=8 8 248 328,clip]{figures/dycheck/mochi-high-five/rgb_error.png}\par\vspace{1pt}{\fontsize{7}{8}\selectfont RGB error}\end{minipage}\hfill\begin{minipage}[t]{0.242\linewidth}\centering\includegraphics[width=\linewidth,bb=0 0 1024 672,viewport=8 8 248 328,clip]{figures/dycheck/mochi-high-five/depth_error.png}\par\vspace{1pt}{\fontsize{7}{8}\selectfont Depth error}\end{minipage}\end{minipage}\hfill\begin{minipage}[t]{0.492\textwidth}\centering{\fontsize{8}{9}\selectfont Vanilla (reset off)}\par\vspace{2pt}\begin{minipage}[t]{0.242\linewidth}\centering\includegraphics[width=\linewidth,bb=0 0 1408 1392,viewport=712 600 952 920,clip]{figures/dycheck/mochi-high-five/rgb.jpg}\par\vspace{1pt}{\fontsize{7}{8}\selectfont RGB}\end{minipage}\hfill\begin{minipage}[t]{0.242\linewidth}\centering\includegraphics[width=\linewidth,bb=0 0 1408 1392,viewport=712 600 952 920,clip]{figures/dycheck/mochi-high-five/depth.png}\par\vspace{1pt}{\fontsize{7}{8}\selectfont Depth}\end{minipage}\hfill\begin{minipage}[t]{0.242\linewidth}\centering\includegraphics[width=\linewidth,bb=0 0 1024 672,viewport=264 8 504 328,clip]{figures/dycheck/mochi-high-five/rgb_error.png}\par\vspace{1pt}{\fontsize{7}{8}\selectfont RGB error}\end{minipage}\hfill\begin{minipage}[t]{0.242\linewidth}\centering\includegraphics[width=\linewidth,bb=0 0 1024 672,viewport=264 8 504 328,clip]{figures/dycheck/mochi-high-five/depth_error.png}\par\vspace{1pt}{\fontsize{7}{8}\selectfont Depth error}\end{minipage}\end{minipage}\par\vspace{16pt}
\noindent \begin{minipage}[t]{0.492\textwidth}\centering{\fontsize{8}{9}\selectfont Mip-Splatting + SfSM}\par\vspace{2pt}\begin{minipage}[t]{0.242\linewidth}\centering\includegraphics[width=\linewidth,bb=0 0 1408 1392,viewport=1064 600 1304 920,clip]{figures/dycheck/mochi-high-five/rgb.jpg}\par\vspace{1pt}{\fontsize{7}{8}\selectfont RGB}\end{minipage}\hfill\begin{minipage}[t]{0.242\linewidth}\centering\includegraphics[width=\linewidth,bb=0 0 1408 1392,viewport=1064 600 1304 920,clip]{figures/dycheck/mochi-high-five/depth.png}\par\vspace{1pt}{\fontsize{7}{8}\selectfont Depth}\end{minipage}\hfill\begin{minipage}[t]{0.242\linewidth}\centering\includegraphics[width=\linewidth,bb=0 0 1024 672,viewport=520 8 760 328,clip]{figures/dycheck/mochi-high-five/rgb_error.png}\par\vspace{1pt}{\fontsize{7}{8}\selectfont RGB error}\end{minipage}\hfill\begin{minipage}[t]{0.242\linewidth}\centering\includegraphics[width=\linewidth,bb=0 0 1024 672,viewport=520 8 760 328,clip]{figures/dycheck/mochi-high-five/depth_error.png}\par\vspace{1pt}{\fontsize{7}{8}\selectfont Depth error}\end{minipage}\end{minipage}\hfill\begin{minipage}[t]{0.492\textwidth}\centering{\fontsize{8}{9}\selectfont NBGS-Base}\par\vspace{2pt}\begin{minipage}[t]{0.242\linewidth}\centering\includegraphics[width=\linewidth,bb=0 0 1408 1392,viewport=8 136 248 456,clip]{figures/dycheck/mochi-high-five/rgb.jpg}\par\vspace{1pt}{\fontsize{7}{8}\selectfont RGB}\end{minipage}\hfill\begin{minipage}[t]{0.242\linewidth}\centering\includegraphics[width=\linewidth,bb=0 0 1408 1392,viewport=8 136 248 456,clip]{figures/dycheck/mochi-high-five/depth.png}\par\vspace{1pt}{\fontsize{7}{8}\selectfont Depth}\end{minipage}\hfill\begin{minipage}[t]{0.242\linewidth}\centering\includegraphics[width=\linewidth,bb=0 0 1024 672,viewport=776 8 1016 328,clip]{figures/dycheck/mochi-high-five/rgb_error.png}\par\vspace{1pt}{\fontsize{7}{8}\selectfont RGB error}\end{minipage}\hfill\begin{minipage}[t]{0.242\linewidth}\centering\includegraphics[width=\linewidth,bb=0 0 1024 672,viewport=776 8 1016 328,clip]{figures/dycheck/mochi-high-five/depth_error.png}\par\vspace{1pt}{\fontsize{7}{8}\selectfont Depth error}\end{minipage}\end{minipage}\par\vspace{16pt}
\clearpage

%% file: sections/comparison_protocol.tex
\subsection*{Image references and crop selection}
\paragraph{Small-Motion benchmark (Sec.~\ref{sec:image_quality}).}
For each scene, all methods use the same RGB crop coordinates. We chose the Louvre crops by visual inspection of foreground contours and retained the Plants crops from a search based on local rendering error. The other eight scenes use regions marked in the reference image. Full-image RGB errors show how far the differences extend beyond these crops.

Small GS, InstantSplat and G4Splat use the original RGB reference. The SfSM-based methods use the undistorted reference. RGB error is the sum of absolute differences across the three channels, on a shared $[0,3]$ scale. Each depth map uses its own 2nd and 98th percentiles of positive depth as color limits. Black marks invalid depth.

Vanilla$^{\dagger}$ uses median sparse depth to scale size pruning, the clone/split threshold and the position learning rate. NBGS also uses sparse depth distributions for ray supervision and per-camera near planes. The InstantSplat, G4Splat, DDGS and DDDrop images shown here come from separate runs of these methods. Table~\ref{tab:system_quant} gives the quantitative comparison from the runs reported there.

\newpage
\paragraph{DyCheck iPhone clips (Sec.~\ref{sec:iphone_depth}).}
For each method and scene, we collect the common valid static pixels across the four held-out views. We scale the rendered depths by the median LiDAR depth divided by the median rendered depth over those pixels. Figure~\ref{fig:iphone_depth} illustrates rendered depth alongside the LiDAR reference. The additional error maps here give the absolute difference in meters on valid static pixels shared by all methods. Values above $1.5$\,m take the highest color, and white marks excluded pixels. The labels NBGS-Base and Vanilla (reset off) identify the configurations used for these phone visualizations.